\documentclass{article}
\usepackage{style/foci}
\usepackage{tgpagella}   
\usepackage{tgheros}     

\usepackage[utf8]{inputenc}
\usepackage[T1]{fontenc}
\usepackage{microtype}

\usepackage{amsmath}
\usepackage{amssymb}
\usepackage{amsfonts}
\usepackage{amsthm}
\usepackage{mathtools}
\usepackage{mathrsfs}
\usepackage{physics}
\usepackage{braket}
\usepackage{slashed}
\usepackage{nicefrac}
\usepackage{textcomp}
\usepackage{dsfont}
\usepackage{bbm}
\usepackage{bm}

\usepackage{graphicx}
\usepackage{subcaption}
\usepackage[export]{adjustbox}
\usepackage{float}
\usepackage{booktabs}
\usepackage{dcolumn}
\newcolumntype{d}[1]{D{.}{.}{#1}}
\usepackage{bigstrut, tabularx, multirow, makecell, diagbox}
\usepackage{colortbl}
\usepackage{tabularray}
\UseTblrLibrary{booktabs}
\usepackage{threeparttable}
\usepackage{tablefootnote}
\usepackage{fontawesome5}

\usepackage{placeins}
\usepackage{caption}
\usepackage{footnote}
\usepackage{enumitem}
\usepackage{multicol}
\usepackage{xspace}
\usepackage{titletoc}
\usepackage{titlesec}
\usepackage[bottom]{footmisc}
\usepackage{setspace}

\usepackage{wrapfig}
\usepackage{tikz}
\usepackage{quantikz}
\usepackage{dashbox}
\usepackage{mdframed}
\usepackage{marvosym}
\usepackage{pifont}
\usepackage{CJK}
\usepackage{url}

\usepackage[table,x11names]{xcolor}
\usepackage[most]{tcolorbox}
\tcbuselibrary{breakable}
\usetikzlibrary{decorations.pathreplacing, fit}

\definecolor{primaryblue}{HTML}{0066CC}
\definecolor{accentcyan}{HTML}{00D4AA}
\definecolor{warmorange}{HTML}{FF6B35}
\definecolor{deepgray}{HTML}{2C3E50}
\definecolor{lightgray}{HTML}{F8F9FA}
\definecolor{gradientstart}{HTML}{667eea}
\definecolor{gradientend}{HTML}{764ba2}

\definecolor{citecolor}{HTML}{0071bc}
\definecolor{citeblue}{RGB}{0, 113, 188}
\definecolor{linkcolor}{HTML}{9A4D92}
\definecolor{firebrick}{rgb}{0.698,0.133,0.133}

\definecolor{paleviolet}{HTML}{E1EEFC}
\definecolor{CarolinaUltraLight}{HTML}{E7F4FC}
\definecolor{lightgrey}{RGB}{247, 247, 247}
\definecolor{shadecolor}{HTML}{EFEFEF}
\definecolor{lightyellow}{rgb}{1.0, 0.95, 0.7}
\definecolor{lightblue}{rgb}{0.90, 0.95, 1.0}
\definecolor{light-gray}{gray}{0.95}

\definecolor{darkgrey}{rgb}{0.5, 0.5, 0.5}
\definecolor{darkgreen}{rgb}{0, 0.5, 0}
\definecolor{mydarkblue}{rgb}{0,0.08,0.45}
\definecolor{mydarkblue2}{rgb}{0.133, 0.133, 0.698}
\definecolor{echodrk}{HTML}{0099cc}
\definecolor{mymauve}{rgb}{0.58,0,0.82}
\definecolor{midnightblue}{rgb}{0.1,0.1,0.44}
\definecolor{oxfordblue}{rgb}{0.0,0.13,0.28}
\definecolor{prussianblue}{rgb}{0.0,0.19,0.33}
\definecolor{coolteal}{rgb}{0, 0.45, 0.45}
\definecolor{olive}{rgb}{0.1, 0.3, 0}
\definecolor{mypurple}{rgb}{0.5,0,0.5}
\definecolor{almond}{rgb}{0.94, 0.87, 0.8}

\definecolor{blue_ampEncoding}{HTML}{DAE8FC}
\definecolor{green_encoder}{HTML}{D5E8D4}
\definecolor{purple_decoder}{HTML}{E1D5E7}
\definecolor{yellow_measure}{HTML}{FFF2CC}
\definecolor{gray_block}{HTML}{F5F5F5}
\definecolor{pink_dru}{HTML}{FAD9D5}
\definecolor{orange_v}{HTML}{FAD7AC}

\definecolor{colorA}{rgb}{1,0,0}
\definecolor{colorB}{rgb}{0,0.3,1}
\definecolor{colorC}{rgb}{0.9,0.8,0.2}
\definecolor{colorD}{rgb}{0,0.65,0}
\definecolor{lesslightgray}{rgb}{0.5,0.5,0.5}
\definecolor{fundamental}{RGB}{55, 110, 111}
\definecolor{Gred}{RGB}{219, 50, 54}
\definecolor{ToCgreen}{RGB}{0, 128, 0}
\definecolor{Sepia}{RGB}{112, 66, 20}
\definecolor{Dblue}{rgb}{0,0.08,0.45}
\definecolor{Blue}{rgb}{0, 0, 0.8}
\definecolor{blue}{rgb}{0,0,1}
\definecolor{UNCblue!10}{rgb}{0.84,0.91,0.98}
\definecolor{RowAlt}{rgb}{0.98,0.98,0.99}

\definecolor{CarolinaBlue}{HTML}{7BAFD4}        
\definecolor{CarolinaLightBlue}{HTML}{B3D4E5}   
\definecolor{CarolinaUltraLight}{HTML}{E8F4F8}  
\definecolor{CarolinaText}{HTML}{1C2B33}        

\usepackage[pagebackref=true,breaklinks=true,colorlinks,hyperfootnotes=false]{hyperref}
\hypersetup{
  colorlinks,
  citecolor=citeblue,
  linkcolor=firebrick,
  urlcolor=firebrick
}
\usepackage[nameinlink,capitalize,noabbrev]{cleveref}

\titlespacing\section{0pt}{4pt plus 4pt minus 2pt}{-2pt plus 2pt minus 2pt}
\titlespacing\subsection{0pt}{2pt plus 4pt minus 2pt}{-2pt plus 2pt minus 2pt}
\titlespacing\subsubsection{0pt}{2pt plus 4pt minus 2pt}{-2pt plus 2pt minus 2pt}

\makeatletter
\def\th@remark{%
  \thm@headfont{\bfseries}%
  \normalfont 
  \thm@preskip\topsep \divide\thm@preskip\tw@
  \thm@postskip\thm@preskip
}
\makeatother

\theoremstyle{definition}

\tcbset{focithm/.style={
  enhanced, breakable, frame hidden,
  colback=black!5,
  width=\linewidth,
  enlarge left by=0pt, enlarge right by=0pt,
  boxsep=6pt, boxrule=0pt,
  left=4pt, right=4pt, top=1pt, bottom=1pt,
  arc=8pt,
  before skip=\topsep, after skip=\topsep
}}

\tcolorboxenvironment{theorem}{focithm}

\tcolorboxenvironment{lemma}{focithm}

\tcolorboxenvironment{corollary}{focithm}

\tcolorboxenvironment{proposition}{focithm}

\tcolorboxenvironment{definition}{focithm}

\tcolorboxenvironment{assumption}{focithm}

\tcolorboxenvironment{claim}{focithm}

\tcolorboxenvironment{problem}{focithm}

\tcolorboxenvironment{question}{focithm}

\newtcolorbox{titleblock}{
  enhanced,
  frame hidden,
  colback=CarolinaUltraLight,
  colframe=CarolinaUltraLight,
  boxrule=0pt,
  arc=10pt,
  left=14pt,
  right=14pt,
  top=14pt,
  bottom=14pt,
  width=\linewidth,
  before skip=12pt plus 4pt,
  after skip=12pt plus 4pt,
  grow to left by=1.5pt,
  grow to right by=1.5pt,
  before upper={
    \setlength{\parindent}{0cm}
    \setlength{\parskip}{0.5cm}
  }
}

\crefname{theorem}{Theorem}{Theorems}
\crefname{proposition}{Proposition}{Propositions}
\crefname{lemma}{Lemma}{Lemmas}
\crefname{corollary}{Corollary}{Corollaries}
\crefname{definition}{Definition}{Definitions}
\crefname{assumption}{Assumption}{Assumptions}
\crefname{remark}{Remark}{Remarks}
\crefname{problem}{Problem}{Problems}
\crefname{property}{Property}{property}
\crefname{question}{Question}{Questions}

\numberwithin{equation}{section}
\numberwithin{theorem}{section}
\numberwithin{proposition}{section}
\numberwithin{definition}{section}
\numberwithin{lemma}{section}
\numberwithin{assumption}{section}
\numberwithin{remark}{section}

\def\1{\bm{1}}

\makeatletter
\let\save@mathaccent\mathaccent
\newcommand*\if@single[3]{%
    \setbox0\hbox{${\mathaccent"0362{#1}}^H$}%
    \setbox2\hbox{${\mathaccent"0362{\kern0pt#1}}^H$}%
    \ifdim\ht0=\ht2 #3\else #2\fi
}
\newcommand*\rel@kern[1]{\kern#1\dimexpr\macc@kerna}
\newcommand*\widebar[1]{\@ifnextchar^{{\wide@bar{#1}{0}}}{\wide@bar{#1}{1}}}
\newcommand*\wide@bar[2]{\if@single{#1}{\wide@bar@{#1}{#2}{1}}{\wide@bar@{#1}{#2}{2}}}
\newcommand*\wide@bar@[3]{%
    \begingroup
    \def\mathaccent##1##2{%
        \let\mathaccent\save@mathaccent
        \if#32 \let\macc@nucleus\first@char \fi
        \setbox\z@\hbox{$\macc@style{\macc@nucleus}_{}$}%
        \setbox\tw@\hbox{$\macc@style{\macc@nucleus}{}_{}$}%
        \dimen@\wd\tw@
        \advance\dimen@-\wd\z@
        \divide\dimen@ 3
        \@tempdima\wd\tw@
        \advance\@tempdima-\scriptspace
        \divide\@tempdima 10
        \advance\dimen@-\@tempdima
        \ifdim\dimen@>\z@ \dimen@0pt\fi
        \rel@kern{0.6}\kern-\dimen@
        \if#31
        \overline{\rel@kern{-0.6}\kern\dimen@\macc@nucleus\rel@kern{0.4}\kern\dimen@}%
        \advance\dimen@0.4\dimexpr\macc@kerna
        \let\final@kern#2%
        \ifdim\dimen@<\z@ \let\final@kern1\fi
        \if\final@kern1 \kern-\dimen@\fi
        \else
        \overline{\rel@kern{-0.6}\kern\dimen@#1}%
        \fi
    }%
    \macc@depth\@ne
    \let\math@bgroup\@empty \let\math@egroup\macc@set@skewchar
    \mathsurround\z@ \frozen@everymath{\mathgroup\macc@group\relax}%
    \macc@set@skewchar\relax
    \let\mathaccentV\macc@nested@a
    \if#31
    \macc@nested@a\relax111{#1}%
    \else
    \def\gobble@till@marker##1\endmarker{}%
    \futurelet\first@char\gobble@till@marker#1\endmarker
    \ifcat\noexpand\first@char A\else
    \def\first@char{}%
    \fi
    \macc@nested@a\relax111{\first@char}%
    \fi
    \endgroup
    }
\makeatother

\DeclareMathAlphabet{\mathsfit}{\encodingdefault}{\sfdefault}{m}{sl}
\SetMathAlphabet{\mathsfit}{bold}{\encodingdefault}{\sfdefault}{bx}{n}

\renewcommand{\arraystretch}{1.15}
\makeatletter

\def\foci@authors{}
\def\foci@affil{}
\def\foci@email{}
\def\foci@project{}
\def\foci@date{\today}
\def\foci@authorsep{}

\newcommand{\fociauthor}[2]{%
  \g@addto@macro\foci@authors{\foci@authorsep\mbox{#1\textsuperscript{#2}}\gdef\foci@authorsep{,\ }}}
\newcommand{\fociaffil}[1]{\gdef\foci@affil{#1}}
\newcommand{\fociemail}[1]{\gdef\foci@email{#1}}
\newcommand{\fociproject}[1]{\gdef\foci@project{#1}}
\newcommand{\focidate}[1]{\gdef\foci@date{#1}}
\newcommand{\fociauthorlist}{\gdef\foci@authorsep{}\foci@authors}
\newcommand{\focicorr}{\textrm{\Letter}}
\newcommand{\focimarkeq}{*}

\def\foci@blfootnote{\gdef\@thefnmark{}\@footnotetext}
\newcommand{\focifootnotes}{%
  \foci@blfootnote{\textsuperscript{*}Equal contribution.\quad
    \textsuperscript{\Letter}Corresponding author: \texttt{\foci@email}}}

\newlength{\foci@mw}
\newcommand{\focidrawmark}[3]{%
  \setlength{\foci@mw}{#2}%
  \begin{scope}[shift={#3}, x={\foci@mw/564.03}, y={-\foci@mw/564.03}]
    \foreach \cx/\cy in {282.02/66.03, 282.02/222.08, 215.77/348.8, 348.73/348.8,
                         414.88/474.93, 149.16/474.93, 67.06/222.08, 496.99/222.08}
      \fill[#1] (\cx,\cy) circle[radius=42.04];
  \end{scope}}
\newcommand{\focimark}[2][black]{%
  \setlength{\foci@mw}{#2}%
  \begin{tikzpicture}[x={\foci@mw/564.03}, y={-\foci@mw/564.03}]
    \useasboundingbox (0,0) rectangle (564.03,540.96);
    \focidrawmark{#1}{#2}{(0,0)}
  \end{tikzpicture}}

\fancypagestyle{focititlepage}{\fancyhf{}\renewcommand{\headrulewidth}{0pt}}

\newcommand{\foci@headerline}{%
  \noindent
  \begin{minipage}[c]{0.5\linewidth}
    \includegraphics[height=0.58cm]{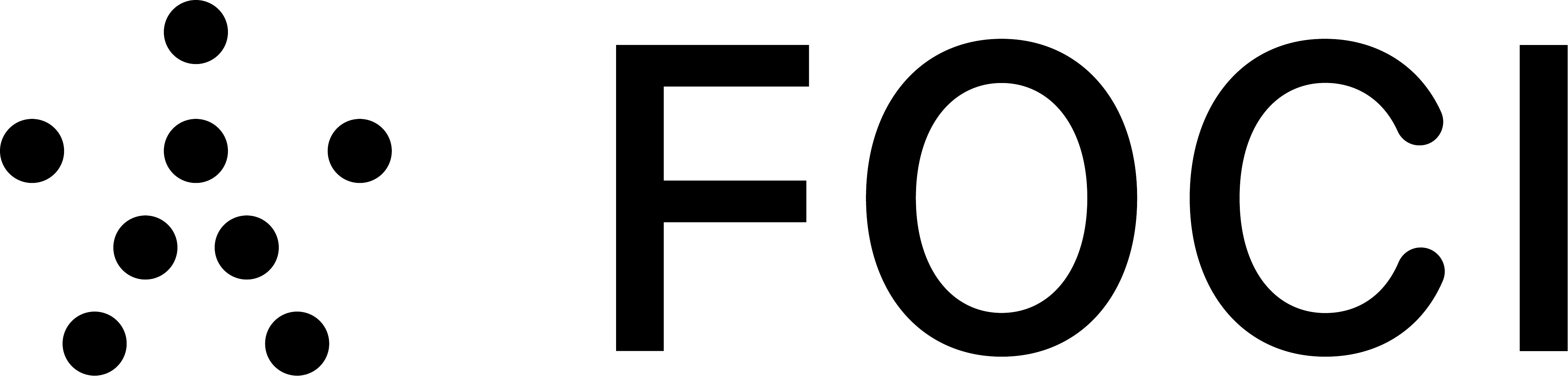}
  \end{minipage}%
  \begin{minipage}[c]{0.5\linewidth}
    \raggedleft\small\itshape\color{black!55}\foci@date
  \end{minipage}\par}

\newcommand{\foci@contactline}{%
  {\footnotesize\color{black!65}%
   \textbf{Email:}~\texttt{\foci@email}\qquad
   \textbf{Project:}~\url{\foci@project}\par}}

\newcommand{\foci@hairline}{\noindent\textcolor{black}{\rule{\linewidth}{0.5pt}}\par}

\newcommand{\focicoverCard}{%
  \thispagestyle{focititlepage}%
  \foci@headerline
  \vskip -0.5em
  \foci@hairline
  \vskip 0.9em
  \begin{tcolorbox}[enhanced, frame hidden, boxrule=0pt, arc=12pt,
      colback=black!4, width=\linewidth,
      left=18pt, right=18pt, top=18pt, bottom=16pt,
      grow to left by=2pt, grow to right by=2pt]
    \noindent
    \begin{minipage}[t]{\linewidth}
      {\setstretch{1.15}\LARGE\bfseries\raggedright\textbf{Actionwave:} A Systematic Benchmark for Waveform-Centric Clinical Prediction Across Care Scenarios\unskip\par}
      \vskip 0.55em
      {\normalsize\fociauthorlist\par}
      \vskip 0.2em
      {\small\color{black!70}\foci@affil\par}
    \end{minipage}\par
    \vskip 0.9em
    {\small\setstretch{1.08}Continuous bedside waveforms record rapid physiological changes that can inform clinical decisions. Existing benchmarks examine selected prediction conditions, but it remains unclear how model advantages depend on signal duration, warning time, and available inputs across care settings.
We introduce \method, a benchmark for waveform-centric clinical prediction across intensive-care, operating-room, and emergency-department settings.
\method~uses a common decision-time prediction interface and groups clinically grounded tasks by what is inferred and when.
It varies conditions relevant to clinical use, including the warning required before an intervention.
We compare diverse architectures and pretrained biosignal encoders under specified model-size and compute budgets. The protocol requires patient-level splits, strict information cutoffs, and comparisons that isolate individual modeling factors. Across 20 tasks and 1.4M task-labelled windows from source datasets, the experimental grids account for over 87,000 model fits.
These comparisons identify task-specific choices and show why matched cohorts, unimodal controls, and explicit timing are necessary when interpreting waveform model performance.\unskip\par}
    \vskip 0.9em
    \foci@contactline
  \end{tcolorbox}
  \vskip 0.4em}

\newcommand{\focicover}{\focicoverCard}

\makeatother

\usepackage{sectsty}
\definecolor{fociblue}{HTML}{0071BC}
\allsectionsfont{\color{fociblue}}
\hypersetup{colorlinks=true,linkcolor=fociblue,citecolor=fociblue,urlcolor=fociblue}

\newcommand{\method}{\textbf{Actionwave}}

\newcommand{\yesicon}{\ding{51}}

\providecommand{\ms}[2]{$#1{\scriptstyle\,\pm\,#2}$}
\newcolumntype{L}[1]{>{\raggedright\arraybackslash}p{#1}}
\newcolumntype{C}[1]{>{\centering\arraybackslash}p{#1}}

\begin{document}

\fociauthor{Jie Peng}{\focimarkeq\textsuperscript{1}}
\fociauthor{Davis Hobley}{\focimarkeq\textsuperscript{1}}
\fociauthor{Song Wang}{\textsuperscript{1}}
\fociauthor{Zachary Plotkin}{\focicorr\textsuperscript{1}}
\fociauthor{Tianlong Chen}{\focicorr\textsuperscript{1},\textsuperscript{2}}

\fociaffil{\textsuperscript{1}Foci, \textsuperscript{2}University of North Carolina at Chapel Hill}
\fociemail{zach@fo.ci, tianlong@fo.ci}
\fociproject{https://fo.ci/}


\pagestyle{fancy}
\fancyhf{}
\renewcommand{\headrulewidth}{0.4pt}
\chead{\small\scshape\textcolor{fociblue}{Actionwave}}
\cfoot{\small\thepage}

\focicover
\focifootnotes

\section{Introduction}

\begin{wrapfigure}{r}{0.25\textwidth}
    \centering
    \vspace{-\intextsep}
    \includegraphics[width=\linewidth]{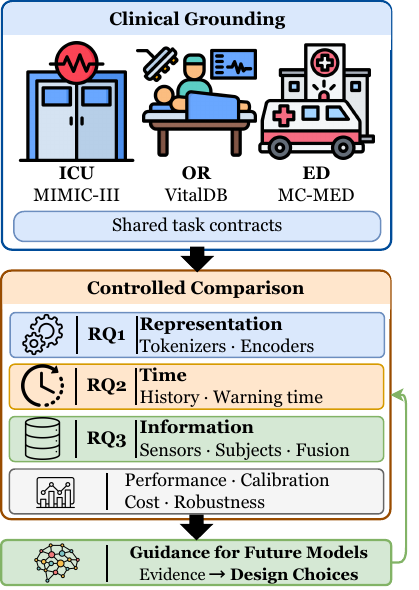}
    \vspace{-6mm}
    \caption{\small \method: clinically grounded tasks and controlled comparisons provide evidence to guide future model development. The comparison axes are summarized in Table~\ref{tab:experiment-count}.}
    \label{fig:design-philosophy-draft}
    \vspace{-4mm}
\end{wrapfigure}
    
Continuous bedside monitoring captures physiological dynamics missed by intermittent clinical measurements~\citep{sundrani2023decompensation}.
Electrocardiography (ECG), arterial blood pressure (ABP), photoplethysmography (PPG), and respiration (RESP) capture changes over seconds to minutes for clinical prediction~\citep{moody2020mimic3waveform,lee2022vitaldb,kansal2025mcmed}.
Their predictive value depends on prediction conditions~\citep{saeb2017usecase,lauritsen2021framing}.
Intensive-care units (ICU), operating rooms (OR), and emergency departments (ED) differ in available sensors, monitoring duration, and the warning time required before an intervention.
Studying these conditions reveals which modeling choices work, where benefits persist, and what needs improvement.

Waveform encoders and biosignal foundation models offer diverse approaches to clinical prediction.
YAIB, MC-BEC, and SignalMC-MED support reproducible evaluation across clinical definitions, modalities, and signal durations~\citep{vandewater2024yaib,chen2023mcbec,gustafsson2026signalmcmed}.
Two issues obscure which raw-waveform modeling choices offer consistent gains across care settings and prediction conditions.
First, encoder comparisons often also change tokenization, sampling, input channels, or computational resources, obscuring architecture's contribution.
Second, changing observation length or forecast gap can alter episode eligibility, follow-up availability, and outcome prevalence.
Performance differences can therefore reflect population shifts or task difficulty alongside modeling choices.
These comparisons require control over model inputs and prediction-task construction.
Table~\ref{tab:benchmark-comparison-draft} compares our raw-waveform task coverage across sources and experimental axes with complementary benchmarks.
\begin{table}[!t]
\centering
\setlength{\tabcolsep}{1.5pt}
\caption{Comparison with clinical time-series and waveform benchmarks, with experimental axes grouped by our three research questions.
Scope columns denote raw-waveform availability, independent source cohorts, and clinical task families. Systolic and diastolic blood pressure count as one family.
RQ columns indicate reported comparisons (\yesicon{}) or none (--). Input panels include feature groups, and gap/horizon coverage includes multiple target horizons.
The \method{} row summarizes the specified benchmark design.}
\vspace{-3mm}
\label{tab:benchmark-comparison-draft}
\label{tab:benchmark-comparison}
\resizebox{\linewidth}{!}{%
\begin{tabular}{@{}l*{9}{c}@{}}
\toprule
& \multicolumn{3}{c}{\textbf{Scope}}
& \multicolumn{1}{c}{\textbf{RQ1}}
& \multicolumn{2}{c}{\textbf{RQ2: Time}}
& \multicolumn{3}{c}{\textbf{RQ3: Information}} \\
\cmidrule(lr){2-4}\cmidrule(lr){5-5}\cmidrule(lr){6-7}\cmidrule(l){8-10}
Benchmark
& Raw waveform
& Multiple sources
& Multiple tasks
& Tokenization
& Observation length
& Gap or horizon
& Input panels
& Missingness
& Data budget \\
\midrule
MIMIC-III~\citep{harutyunyan2019multitask}
& -- & -- & \yesicon
& -- & -- & -- & -- & -- & -- \\
\addlinespace[1pt]
YAIB~\citep{vandewater2024yaib}
& -- & \yesicon & \yesicon
& -- & -- & -- & \yesicon & -- & \yesicon$^{\dagger}$ \\
\addlinespace[1pt]
PulseDB~\citep{wang2023pulsedb}
& \yesicon & \yesicon & --
& -- & -- & -- & -- & -- & -- \\
\addlinespace[1pt]
MC-BEC~\citep{chen2023mcbec}
& \yesicon & -- & \yesicon
& -- & -- & \yesicon & \yesicon & \yesicon & -- \\
\addlinespace[1pt]
MDS-ED~\citep{lopezalcaraz2025enhancing}
& \yesicon & -- & \yesicon
& -- & -- & \yesicon$^{\S}$ & \yesicon & \yesicon$^{\ddagger}$ & -- \\
\addlinespace[1pt]
SignalMC-MED~\citep{gustafsson2026signalmcmed}
& \yesicon & -- & \yesicon
& -- & \yesicon & -- & \yesicon & -- & \yesicon \\
\midrule
\method{} (ours)
& \yesicon & \yesicon & \yesicon
& \yesicon & \yesicon & \yesicon & \yesicon & \yesicon & \yesicon \\
\bottomrule
\end{tabular}%
}
\par\smallskip
\parbox{\linewidth}{\footnotesize
$\dagger$~YAIB varies target-cohort training/fine-tuning sample size.
$\ddagger$~MDS-ED compares including versus omitting missingness masks.
$\S$~MDS-ED evaluates multiple mortality and ICU-admission target horizons.}
\vspace{-3mm}
\end{table}
We ask three questions that inform clinical waveform model development:

\begin{tcolorbox}[
    colback=black!5!white,
    colframe=black!55!white,
    boxrule=0.4pt,
    arc=1.5pt,
    left=5pt,
    right=5pt,
    top=3pt,
    bottom=3pt
]
\textbf{(RQ1) Representation and architecture.}
How do waveform tokenization and encoder choices affect predictive performance and model rankings under controlled comparisons?
\par\smallskip
\textbf{(RQ2) Temporal modeling.}
How do observation history and warning time affect predictability, and how do episode eligibility and outcome prevalence change across these comparisons?
\par\smallskip
\textbf{(RQ3) Information availability.}
When do additional sensors, training subjects, and clinical modalities improve prediction, and how robust are these gains to missing inputs?
\end{tcolorbox}
We introduce \method{} (Figure~\ref{fig:design-philosophy-draft}), a waveform-centric benchmark covering 20 tasks across MIMIC-III intensive-care records, VitalDB surgical records, and MC-MED emergency-department records~\citep{moody2020mimic3waveform,lee2022vitaldb,kansal2025mcmed}.
\method{} represents each prediction as a time-indexed episode with a clinical outcome, observation window, available inputs, and a warning interval where applicable.
This shared interface makes observation history, forecast gap, sensor access, and clinical context explicit experimental factors, providing the following contributions:
\ding{182}~We provide a common framework for task definitions and evaluation across care settings.
\ding{183}~We compare representations, temporal context, and information availability in grids accounting for over 87,000 model fits, including hyperparameter searches and repetitions (Appendix~\ref{app:experiment-count}).
\ding{184}~We identify task-specific trade-offs in discrimination, calibration, efficiency, and robustness to guide model choice under clinical data and warning-time constraints. 
Together, these findings establish a foundation to enable future model development for clinical waveform prediction.

\section{Related Work}
\label{sec:related}

\noindent\textbf{Clinical waveform benchmarks.}
Clinical time-series and waveform benchmarks standardize cohorts, tasks, and input protocols for reproducible comparisons across acute-care settings~\citep{harutyunyan2019multitask,vandewater2024yaib,chen2023mcbec,wang2023pulsedb,lopezalcaraz2025enhancing,gustafsson2026signalmcmed}.
The MIMIC-III clinical time-series benchmark defined mortality, decompensation, length-of-stay, and phenotyping tasks using structured ICU data rather than raw waveforms~\citep{harutyunyan2019multitask}.
The MIMIC-III Waveform Database Matched Subset linked multichannel bedside signals to clinical records without prescribing matched comparisons across temporal or sensing conditions~\citep{moody2020mimic3waveform}.
MC-BEC extended multimodal benchmarking to emergency care with decompensation, disposition, and revisit prediction, including missing-modality tests~\citep{chen2023mcbec}.
MDS-ED evaluated raw and engineered ECG alongside routine clinical data for diagnosis and deterioration prediction in a common emergency-department cohort, including multiple mortality and ICU-admission horizons~\citep{lopezalcaraz2025enhancing}.
SignalMC-MED compared biosignal foundation models on synchronized ECG and PPG while varying modality, fusion, signal duration, and model scale~\citep{gustafsson2026signalmcmed}.
\method{} treats tokenization, imputation, temporal geometry, sensing, data budget, architecture, fusion, and source domain as explicit experimental factors across ICU, operating-room, and emergency-department settings. Its protocol calls for varying one factor at a time under matched episodes, splits, and budgets (Table~\ref{tab:benchmark-comparison}).

\noindent\textbf{Waveform representation learning.}
Time-series models depend on both the representation of samples, channels, and irregular observations and the modeling of temporal and cross-channel dependencies~\citep{nie2023time,zeng2023transformers,zhang2023crossformer,liu2024itransformer,horn2020seft,shukla2021mtand}.
PatchTST segments each channel into fixed-length patches and applies a channel-independent Transformer~\citep{vaswani2017attention}, showing that patching and channel independence can affect long-horizon forecasting accuracy and efficiency~\citep{nie2023time}.
LTSF-Linear~\citep{zeng2023transformers} shows that simple linear models can outperform several Transformer-based methods on standard long-term forecasting datasets, underscoring the importance of baseline choice and experimental design.
Whether these tokenizer and encoder findings hold for clinical waveforms across tasks, sensing policies, missingness, and source domains remains an open question. \method~examines their applicability through comparisons organized around shared task definitions and resource constraints.

\section{Benchmark and Shared Experimental Protocol}
\label{sec:benchmark-protocol-draft}
\label{sec:data-task-construction}

\begin{figure*}[t]
\centering
\includegraphics[width=\textwidth]{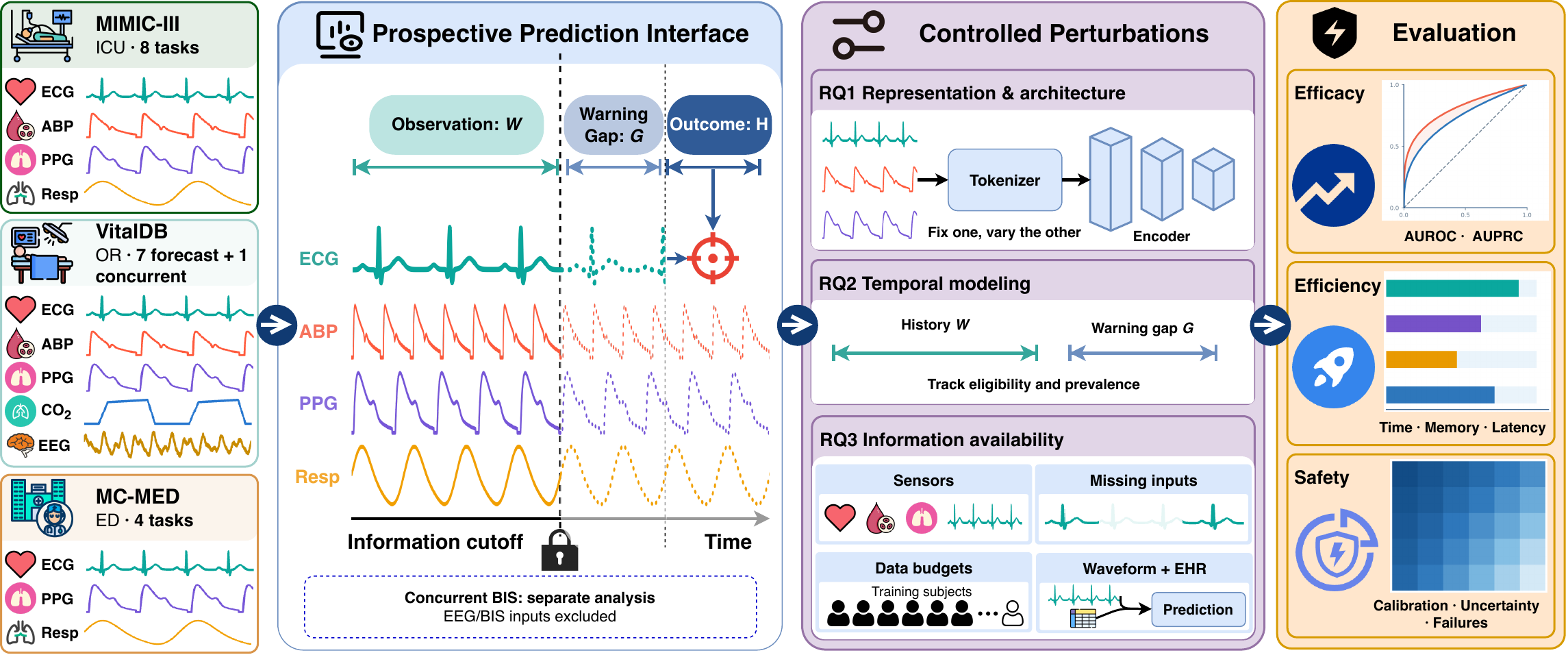}
\vspace{-6mm}
\caption{\method{} links three data sources to a shared decision-time prediction interface and varies representation, time, and information availability to assess efficacy, efficiency, and safety. Task names and cohort support are listed in Tables~\ref{tab:task-name-key} and~\ref{tab:cohorts}.}
\label{fig:overview}
\vspace{-6mm}
\end{figure*}

\method{} covers clinical prediction tasks across intensive-care, operating-room, and emergency-department settings (Sec.~\ref{sec:benchmark-scope-draft}).
A shared episode interface defines outcomes, observation windows, applicable warning intervals, and information cutoffs (Sec.~\ref{sec:benchmark-episodes-draft}).
Each comparison specifies the factor varied, the quantities held fixed, and the model-selection procedure (Sec.~\ref{sec:benchmark-comparisons-draft}).
Figure~\ref{fig:overview} illustrates the workflow and schematic comparisons of efficacy, efficiency, and safety (Sec.~\ref{sec:benchmark-evaluation-draft}).

\subsection{Clinical sources and task coverage}
\label{sec:benchmark-scope-draft}

\begin{wraptable}{r}{0.32\textwidth}
\centering
\scriptsize
\vspace{-\intextsep}
\caption{\footnotesize Eligible cohort unions.
Patients and clinical episodes are deduplicated across tasks. Windows count task-labelled rows across training, validation, and test.}
\label{tab:benchmark-scope-draft}
\vspace{-3mm}
{ 
\setlength{\tabcolsep}{2.7pt} 
\renewcommand{\arraystretch}{1.12}
\begin{tabular}{lrrr}
\toprule
Dataset & Patients & Episodes & Windows \\
\midrule
MIMIC-III & 8,535 & 9,772 & 979,920 \\
VitalDB   & 5,346 & 5,575 & 385,667 \\
MC-MED    & 15,852 & 18,536 & 34,379 \\
In total  & 29,733 & 33,883 & 1,399,966 \\
\bottomrule
\end{tabular}
\vspace{-6mm}
}
\end{wraptable}

\textbf{Clinical Sources.}
\method{} spans three care settings: the intensive care unit (MIMIC-III), operating room (VitalDB), and emergency department (MC-MED)~\citep{moody2020mimic3waveform,lee2022vitaldb,kansal2025mcmed}.
Waveform inputs include ICU monitoring channels, twelve native intraoperative streams, and ED ECG-II and respiration, respectively.
Table~\ref{tab:benchmark-scope-draft} summarizes unique patients and clinical episodes after task-specific eligibility filtering.
\emph{Clinical episodes} are ICU stays, surgical cases, or ED visits. \emph{Windows} are task-labelled prediction examples. One episode can supply several windows across tasks or time points.
\par

\textbf{Task Coverage.}
Full task definitions appear in Appendix~\ref{app:dataset-details}.
\textit{MIMIC-III} includes 8 tasks: acute deterioration, dynamic mortality, respiratory escalation, circulatory deterioration, fluid response, KDIGO stage 2/3 AKI, ventilator liberation, and Sepsis-3.
\textit{VitalDB} includes 8 tasks: intraoperative hypotension, accelerating MAP decline, oxygen desaturation, core hypothermia, $\mathrm{PCO_2}$--$\mathrm{ETCO_2}$ discordance, abrupt $\mathrm{ETCO_2}$ fall, coupled hemodynamic response, and concurrent light BIS state.
We define 4 tasks in \textit{MC-MED}: cardiorespiratory decompensation (60 min), pending hyperkalemia, lactate non-clearance, and pending severe anemia.

These sources provide around 1.4M task-labelled windows across care settings and endpoints. Concurrent BIS is analyzed separately from forecasting.
Table~\ref{tab:cohorts} gives per-task cohort counts.

\subsection{Episode construction and information boundaries}
\label{sec:benchmark-episodes-draft}
\label{sec:episode-data-controls}

\textbf{Episode Construction.}
Using the interface in Figure~\ref{fig:overview}, we construct windows within eligible clinical episodes and represent each example as
\begin{equation}
\begin{aligned}
\mathcal{E} &= (d,s,i,T,W,C,G,H,Y,M),\\
\mathcal{O} &= [T-W,T),\qquad
\mathcal{F}=[T+G,T+G+H),
\end{aligned}
\label{eq:benchmark-episode-draft}
\end{equation}
where $d$, $s$, and $i$ identify the dataset, care setting, and subject. $T$ is the prediction anchor, $W$ is the observation duration, and $C$ denotes permitted channels. $G$ and $H$ specify the forecast gap and target horizon.
$Y$ is the outcome, and $M$ records input validity and outcome ascertainment separately.
Forecasting tasks use observations within $\mathcal{O}$ to predict an endpoint ascertained within $\mathcal{F}$.
The concurrent VitalDB probe targets light BIS state within the observation interval and is analyzed separately.
Candidate anchors follow prespecified surveillance, event, or daily schedules.
Incident-event eligibility excludes prevalent events, handles events in the forecast gap according to the task contract, and requires sufficient follow-up to assign a negative label.
Incompletely ascertained outcomes remain masked.
Waveform views end at the common prediction anchor: from the extracted MIMIC-III history $[T-6\,\mathrm{h},T)$, a five-minute input uses $[T-5\,\mathrm{min},T)$ and a one-hour input uses $[T-1\,\mathrm{h},T)$.
Raw waveform samples from the selected interval enter the model for prediction.

\textbf{Information Boundaries.}
Waveforms and EHR values must precede the applicable decision-time cutoff. Future test results, label and censoring fields, and direct or derived target features are excluded.
The concurrent VitalDB probe withholds EEG, BIS, and other label-defining inputs.
Only input-availability masks enter the model. Outcome-ascertainment masks select labels for training and evaluation.
Subject-level partitions keep repeated visits, surgical cases, and prediction windows within one split or fold.
Dataset-specific preprocessing preserves physical time and input-presence masks. Normalization and learned preprocessing use training subjects only.
Windows with unavailable channels remain eligible.
Appendix~\ref{app:episode-data-controls} details these controls.

\subsection{Controlled comparisons and model selection}
\label{sec:benchmark-comparisons-draft}

Each RQ1--RQ3 comparison in Figure~\ref{fig:overview} specifies one experimental change and quantities held fixed.
Competing configurations should share eligible windows, labels, patient partitions, permitted information, and evaluation procedures within each task, except for the factor varied.
We report changes in eligibility or available training subjects.

\ding{182}~\textbf{Representation Choices (RQ1).}
Encoder comparisons require a common tokenizer\footnote{A tokenizer maps raw waveform samples to a sequence of feature vectors, each called a token. The encoder processes this token sequence.} to isolate encoder effects. Tokenizer comparisons require a common encoder. Selected crossed comparisons test whether tokenizer changes alter encoder rankings.
The reported architecture comparisons use approximately 500k trainable parameters.
When tokenizers are selected for each encoder, we identify the comparison as one of encoder--tokenizer combinations.
Parameter matching does not equate token counts, training cost, or hyperparameter-search effort. These require separate resource specifications.
\ding{183}~\textbf{Physiological Context and Warning Time (RQ2).}
The observation-length comparisons use nested intervals $[T-W,T)$ on the same anchors, holding the prediction cutoff, labels, and patient partitions fixed.
For warning-time comparisons, patient assignments to training, validation, and test partitions remain fixed across gaps, as do the observation view and target horizon.
Each gap defines its own outcome window. Eligibility, follow-up support, and prevalence can therefore vary within these fixed patient partitions.
These controls compare additional history at the same decision time and longer intervention lead times without reassigning patients between partitions.
\ding{184}~\textbf{Information Availability (RQ3).}
Sensor comparisons preserve the evaluation cohort when selected inputs are unavailable, and channel selectors or imputers use training subjects only.
Inference-time missing-input stresses use paired masks without changing training or model selection.
Data-budget comparisons fix validation and test cohorts. The reported MIMIC-III comparison varies training-window quotas, while VitalDB and MC-MED vary unique training subjects and retain their windows.
Fusion comparisons align windows and pre-cutoff information across waveform-only, EHR-only, and fusion models. Gains are assessed against both unimodal references, including the stronger one.

Across five seeds, validation selects checkpoint epochs and hyperparameters, including learning rates and tokenizers where applicable. Selected checkpoints are evaluated on the held-out test set.
Appendices~\ref{app:architecture-tokenizer-protocol} and~\ref{app:evaluation-details} detail the searches, configuration selection, and test-score aggregation.

\subsection{Evaluation: efficacy, efficiency, and safety}
\label{sec:benchmark-evaluation-draft}

The main comparisons and companion figures report predictive performance, training cost, and reliability.
\ding{182}~\textbf{Efficacy.}
AUROC is the main discrimination metric. AUPRC characterizes precision--recall performance. Higher values are better for both.
Task-level scores accompany equal-task macro summaries where recorded.
VitalDB forecast summaries weight gaps equally within each task, then average the seven tasks. Concurrent BIS remains separate.
Prevalence, outcome support, and the scoring estimator are specified in the corresponding study details.
\ding{183}~\textbf{Efficiency.}
We report training time per optimizer step for observation-length comparisons and total training wall time for other experiments, together with accelerator memory and time per training sample. MC-MED temporal panels measure shared-device memory in MiB.
Wall time is measured per run on MIMIC-III and MC-MED and per validation training cell on VitalDB.
The figures label $1000/\text{training throughput}$ as latency, but it measures training cost, not inference or alert-delivery latency.
Hardware, batch size, precision, and schedule constrain comparisons across studies.
\ding{184}~\textbf{Safety.}
Recorded reliability metrics include expected calibration error (ECE, 15 bins), area under the risk--coverage curve (AURC), and failure rates from nonfinite-event, training-attempt, or prediction-file audits, as specified per study.
Lower ECE indicates better calibration, lower AURC better selective prediction, and lower failure rates greater numerical stability.
Separate VitalDB analyses also report Brier score, log loss, and paired sensor/modality-dropout effects.
Seed variation describes training variability. Whole-subject bootstrap intervals estimate uncertainty conditional on fitted models.
Metric definitions, aggregation, and selection qualifications are in Appendix~\ref{app:evaluation-details}. Shaded placeholder panels are excluded from the evidence.

\section{Representation Choices and Model Rankings}
\label{sec:representation-choices-draft}

\begin{figure*}[!t]
\centering
\includegraphics[page=4,width=\textwidth]{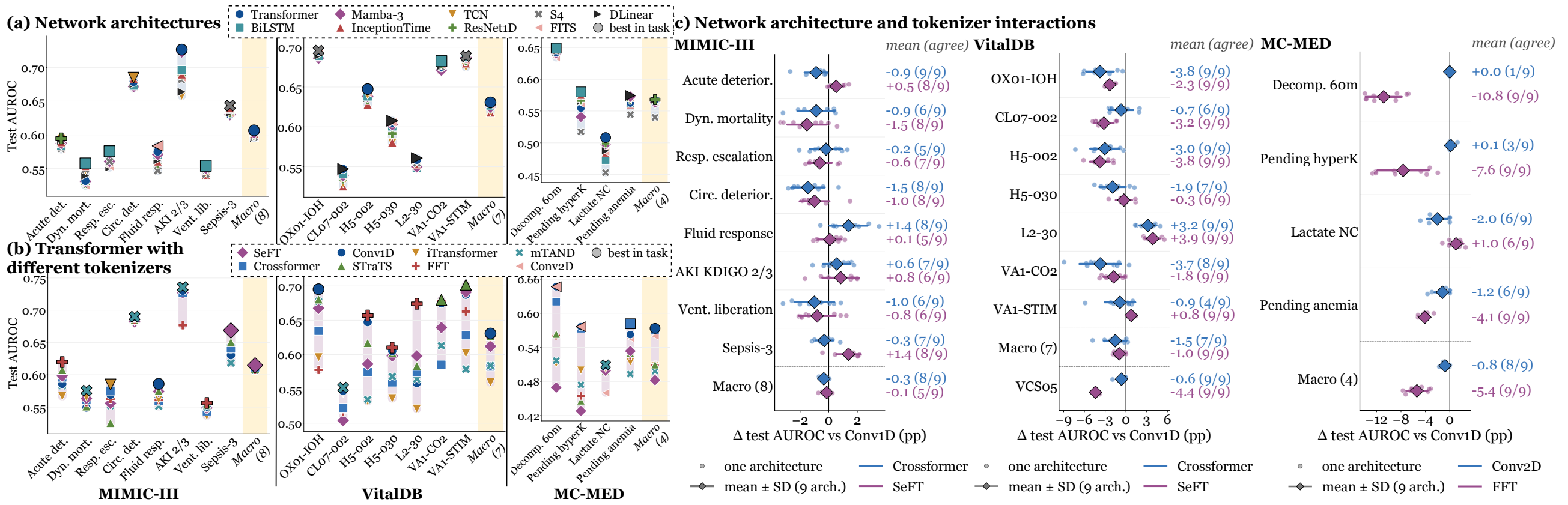}
\vspace{-6mm}
\caption{\textbf{Representation choices and model rankings.}
(a) Encoders with selected tokenizers.
(b) Tokenizers with a fixed Transformer.
(c) Tokenizer--architecture interactions relative to Conv1D, including Conv2D and FFT on MC-MED. Numerical results are in Tables~\ref{tab:main_auroc} and~\ref{tab:representation-test-interactions}.}
\label{fig:representation-results}
\vspace{-6mm}
\end{figure*}

Representation performance depends on task and care setting.
Figure~\ref{fig:representation-results} compares nine encoders (a), seven tokenizers on MIMIC-III and VitalDB and eight on MC-MED with a fixed Transformer (b), and tokenizer changes within each encoder (c).
Encoders are Transformer~\citep{vaswani2017attention}, BiLSTM~\citep{graves2005bilstm}, Mamba-3~\citep{lahoti2026mamba3}, InceptionTime~\citep{fawaz2020inceptiontime}, TCN~\citep{bai2018tcn}, ResNet1D~\citep{wang2017resnet}, S4~\citep{gu2022s4}, FITS~\citep{xu2024fits}, and DLinear~\citep{zeng2023transformers}.
Conv1D projects fixed-length waveform patches following PatchTST~\citep{nie2023time}.
Other tokenizers are SeFT~\citep{horn2020seft}, Crossformer~\citep{zhang2023crossformer}, STraTS~\citep{tipirneni2022strats}, iTransformer~\citep{liu2024itransformer}, FFT spectral projection~\citep{yi2023frets}, and mTAND~\citep{shukla2021mtand}. We adapt tokenizers to the shared encoder budget without reproducing complete systems.
MC-MED additionally compares the local Conv2D tokenizer.

Outlined markers in panels~(a,b) identify each task's highest displayed mean.
Panel~(c) shows $100\times\Delta\mathrm{AUROC}$ from Conv1D for nine encoders (dots), their mean and SD (diamonds and bars), and counts agreeing with the unrounded mean's sign.
Macros average eight MIMIC-III, four MC-MED, and seven VitalDB forecast tasks.
Appendices~\ref{app:task-name-key}, \ref{app:architecture-tokenizer-protocol}, and~\ref{app:representation-results} give task names, protocols, and full scores, respectively.
Restricted waveform inputs and required warning intervals make task AUROCs lower than the 0.8--0.9 reported (Appendix~\ref{app:absolute-performance}).

\noindent\textbf{Tokenization.}\label{sec:tokenizer-effects-draft}
Conv1D alone achieves top-three macro AUROC across all datasets, leading VitalDB (0.631) and MC-MED (0.574) and supporting local patches as a shared baseline (Figure~\ref{fig:representation-results}(b)).
SeFT leads on MIMIC-III (0.615 versus 0.611), driven by Sepsis-3 (0.669 versus 0.630). The other seven tasks collectively favor Conv1D, including fluid response ($-0.012$).
SeFT handles irregular, unaligned timestamped observations~\citep{horn2020seft}, fitting MIMIC-III's sparse channel coverage.
Aggregation outperforms local patches for sepsis, a syndrome of infection-related organ dysfunction~\citep{singer2016sepsis}, but offers less for intervention-linked blood-pressure changes.
See Appendix~\ref{app:representation-test-results} for comparisons.

\noindent\textbf{Architecture.}\label{sec:encoder-effects-draft}
Transformer leads MIMIC-III (0.607) and VitalDB (0.631). On MC-MED, ResNet1D (0.568), DLinear (0.567), and Transformer (0.567) have closely ranked macros.
On MC-MED, BiLSTM leads decompensation (0.649) and hyperkalemia (0.580), DLinear leads anemia (0.574), and Transformer leads lactate nonclearance (0.508).
DLinear's linear temporal projections~\citep{zeng2023transformers} remain competitive for anemia, while the recurrent encoder leads decompensation and hyperkalemia.
Lactate AUROC is 0.453--0.508 across all nine encoders despite decompensation reaching 0.649, consistent with lactate reflecting production, clearance, and treatment over hours beyond the current cardiorespiratory state~\citep{vincent2016lactate}.
Table~\ref{tab:main_auroc} reports the selected encoder--tokenizer combinations and fixed-Transformer tokenizer comparisons.

\noindent\textbf{Tokenizer--architecture interactions.}\label{sec:representation-interactions-draft}
On MC-MED, FFT lowers mean macro AUROC by 0.054 across nine encoders, with losses of 0.108 for decompensation and 0.076 for hyperkalemia in all nine backbones.
Its lactate gain is only 0.010 and occurs in six.
FFT exposes within-patch frequency structure~\citep{yi2023frets}, but local patches are stronger for these ED tasks.
Despite ECGs carrying potassium-related information~\citep{galloway2019hyperkalemia}, repeated FFT losses show that this signal alone does not favor spectral tokenization.
VitalDB shows a reversal: with a fixed Transformer, FFT gains 0.115 on gas discordance but loses 0.118 on hypotension, yielding a macro change of $-0.006$.
Gas exchange reflects ventilation--perfusion physiology, while hypotension prediction draws directly on arterial-waveform changes~\citep{nassar2016capnography,hatib2018machine}.
These results favor spectral projection for gas discordance and local patches for pressure transitions.
Full contrasts and task-level MIMIC-III interaction curves appear in Appendix~\ref{app:representation-test-interactions}.



\section{Physiological Context and Advance Warning}
\label{sec:physiological-context-draft}

\begin{figure*}[!t]
\centering
\vspace{-2mm}
\includegraphics[width=\textwidth]{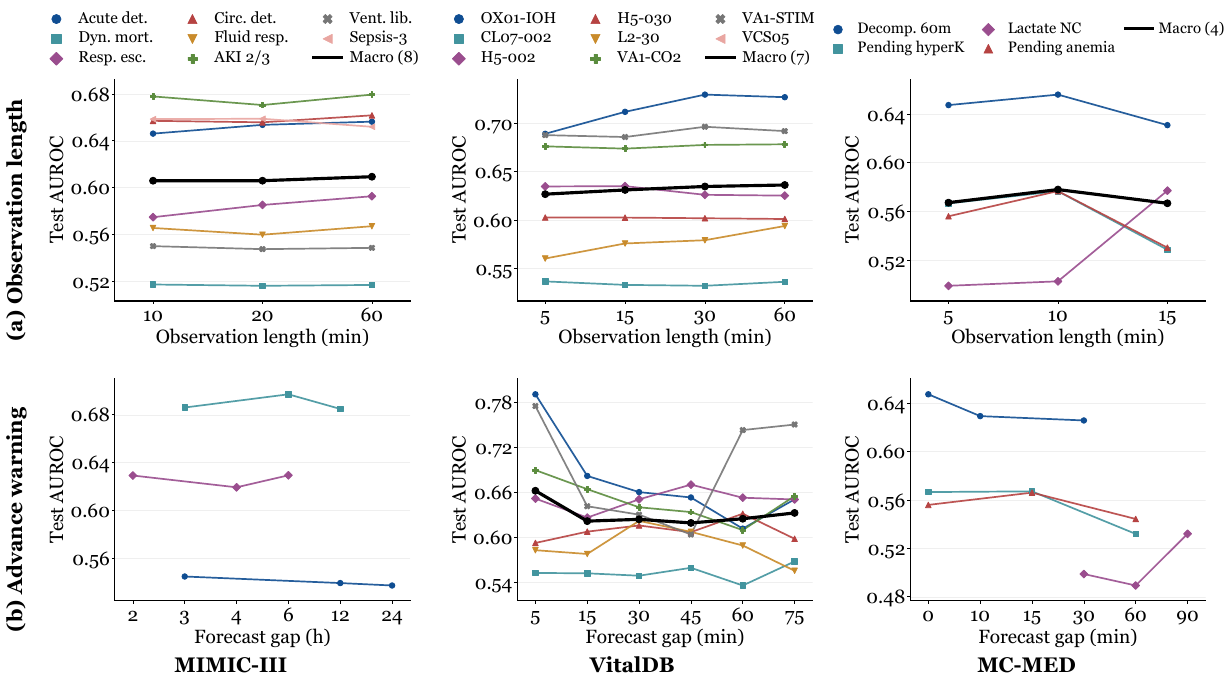}
\vspace{-6mm}
\caption{\textbf{Physiological context and advance warning.}
(a) Nested observation intervals ending at matched anchors.
(b) Task-specific forecast gaps with fixed patient partitions.
In the MIMIC-III warning panel, blue circles denote mortality, teal squares AKI, and purple diamonds Sepsis-3. Task-level values are in Tables~\ref{tab:tasks-observation-length-mimic}--\ref{tab:tasks-forecast-gap-mcmed}.}
\label{fig:context-warning}
\vspace{-6mm}
\end{figure*}

Figure~\ref{fig:context-warning} compares observation lengths and blank forecast gaps across the three datasets (RQ2).
All panels report test AUROC. VitalDB macros exclude the concurrent BIS probe.
Protocols and separate validation results appear in Appendices~\ref{app:temporal-comparisons} and~\ref{app:vitaldb-gaps}.

\noindent\textbf{Observation length.}
Extending MIMIC-III history from ten to 60 minutes changes macro AUROC from 0.606 to 0.609: respiratory escalation improves from 0.575 to 0.593, while mortality remains near 0.517 (Figure~\ref{fig:context-warning}(a)).
On VitalDB, extending five minutes to 60 improves hypotension (0.689 to 0.727) and gas discordance (0.560 to 0.594), but lowers oxygen-desaturation AUROC (0.635 to 0.625).
The pressure-specific gain aligns with predictive pre-event arterial-waveform patterns~\citep{hatib2018machine}. Longer history does not benefit all transitions equally.
MC-MED's macro peaks at ten minutes (0.578), then falls at 15 minutes (0.567): lactate nonclearance improves from 0.502 to 0.577, while the other three tasks decline.
The lactate-specific gain accords with its serial-measurement endpoint and evolving production, clearance, and treatment history~\citep{vincent2016lactate}, whereas pending hyperkalemia and anemia concern tests ordered before prediction.
These matched-anchor comparisons favor endpoint-specific history lengths (Appendix~\ref{app:temporal-comparisons}).

\noindent\textbf{Warning time\footnote{Also called forecast gap.}.}
Training, validation, and test patient partitions remain fixed across warning gaps.
For MIMIC-III, mortality AUROC decreases from 0.545 at three hours to 0.537 at 24 hours, whereas AKI peaks at six hours (0.697) and Sepsis-3 recovers after a four-hour dip.
AKI's intermediate peak accords with renal criteria assessed over time~\citep{kdigo2012aki}: shifting outcome windows changes diagnostic timing alongside warning distance.
For VitalDB, macro AUROC falls from 0.662 at five minutes to 0.621 at 15 minutes, then ranges from 0.619 to 0.632.
This range conceals coupled-response recovery from 0.603 at 45 minutes to 0.750 at 75 minutes, while gas discordance declines from 0.622 at 30 minutes to 0.555 at 75 minutes.
We interpret the coupled-response rebound and 60-minute hypothermia peak as reflecting changing surgical stimulation and heat balance~\citep{ledowski2019nociception,sessler2016thermoregulation}.
For MC-MED, decompensation falls from 0.647 at zero gap to 0.626 at 30 minutes.
Hyperkalemia and anemia retain stronger discrimination at zero or 15 minutes than at 60 minutes.
Useful warning ranges differ between imminent vital-sign transitions and laboratory-defined outcomes. Comparisons must preserve required intervention intervals and account for shifted outcome windows.


\section{Information Availability and Multimodal Value}
\label{sec:information-availability-draft}

\begin{figure*}[!t]
\centering
\includegraphics[page=6,width=\textwidth]{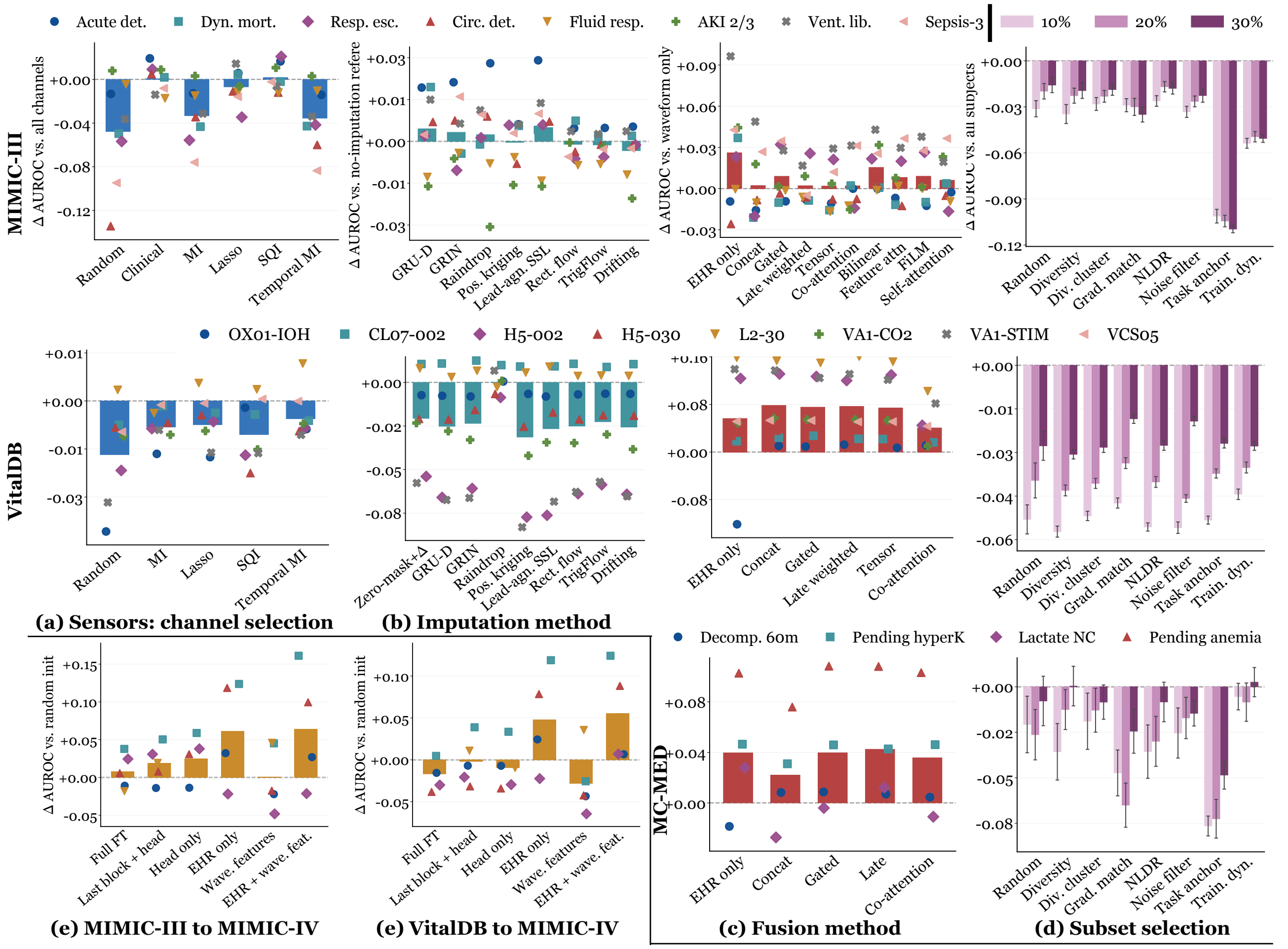}
\vspace{-6mm}
\caption{\textbf{Information availability and multimodal value.}
AUROC changes for (a) sensor selection, (b) missing inputs, (c) waveform--EHR fusion, (d) training-data budgets, and (e) transfer to MIMIC-IV in the two lower-left panels. Light-to-dark shades order window budgets for MIMIC-III and subject budgets for VitalDB and MC-MED (Appendix~\ref{app:information-subjects}).
Bars summarize task averages. Numerical results are in Tables~\ref{tab:tasks-channel-selection-mimic}--\ref{tab:tasks-training-subjects-mcmed} and~\ref{tab:transfer-auroc}.}
\label{fig:information-multimodal}
\vspace{-10mm}
\end{figure*}

Figure~\ref{fig:information-multimodal} compares channel selection, missing-input handling, waveform--EHR fusion, training subsets, and MIMIC-IV transfer~\citep{johnson2023mimiciv} (RQ3).
AUROC references are panel-specific (Appendix~\ref{app:information-comparisons}).
Channel selectors include mutual information (MI)~\citep{peng2005mi}, Lasso~\citep{tibshirani1996lasso}, and signal quality index (SQI)~\citep{clifford2012sqi}, plus random, clinical-priority, and temporal-MI variants.
Missing-input methods include zero-mask plus time-gap encoding~\citep{che2018missing}, GRU-D~\citep{che2018missing}, GRIN~\citep{cini2022grin}, Raindrop~\citep{zhang2022raindrop}, positional kriging adapted from IGNNK~\citep{wu2021ignnk}, lead-agnostic SSL~\citep{oh2022leadagnostic}, and adaptations of Rectified Flow~\citep{liu2023rectifiedflow}, TrigFlow~\citep{lu2025trigflow}, and Drifting~\citep{deng2026drifting}.
Subset selectors use diversity coverage~\citep{sener2018coreset}, diversity clustering~\citep{mcinnes2018umap,sener2018coreset}, gradient matching~\citep{mirzasoleiman2020craig,killamsetty2021gradmatch}, an NLDR-inspired policy~\citep{vazifeh2026nldr}, and training dynamics~\citep{swayamdipta2020cartography,paul2021datadiet}, plus benchmark-specific noise filtering and task-anchor heuristics.
Appendix~\ref{app:method-adaptations} details benchmark adaptations.

\ding{182}~\textbf{Sensors and missingness.}
Clinical/SQI selection preserves MIMIC-III macro AUROC with ten channels. All VitalDB selectors lower it, with temporal MI closest to the all-channel reference (Figure~\ref{fig:information-multimodal}(a,b)).
For MIMIC-III missing inputs, lead-agnostic SSL and GRU-D lead the macro, but all eight alternatives to zero-mask plus $\Delta$ lower AKI and fluid-response AUROC.
On VitalDB, Raindrop approximately matches the native-handling macro, eight alternatives lower it, and all nine reduce desaturation AUROC.
GRIN's cross-sensor reconstruction~\citep{cini2022grin} improves gas discordance by 0.007 but lowers desaturation and coupled-response AUROC by 0.061 and 0.066. Gains for the ventilation--perfusion relation~\citep{nassar2016capnography} do not extend to these transitions.
MC-MED omits these comparisons with only two waveform channels.
In a separate ensemble test, 50\% sensor dropout lowers AUROC from 0.619 to 0.583: handling natural missingness does not ensure resilience to new sensor loss (Appendix~\ref{app:information-sensors}).

\ding{183}~\textbf{Multimodal fusion.}
Bilinear fusion raises MIMIC-III's waveform-only macro most (0.606 to 0.622), below EHR-only (0.632). VitalDB concatenation leads both unimodal models (0.712 versus waveform-only 0.634 and EHR-only 0.690, Figure~\ref{fig:information-multimodal}(c)).
All VitalDB fusion methods improve hypotension, supporting concatenation or prediction weighting without cross-modal attention.
The waveform branch retains value for MIMIC-III circulatory deterioration and VitalDB hypotension, consistent with hemodynamic changes preceding circulatory events~\citep{hyland2020circulatory,hatib2018machine}.
On MC-MED, late fusion leads the macro (0.609), but its lactate AUROC remains below EHR-only (0.520 versus 0.536), consistent with the metabolic and treatment history governing lactate kinetics~\citep{vincent2016lactate}.
Practical fusion is bilinear fusion on MIMIC-III, concatenation or weighted late fusion on VitalDB, and late fusion on MC-MED.
EHR-only controls assess the waveform contribution beyond clinical context (Appendix~\ref{app:information-fusion}).

\ding{184}~\textbf{Training-data budgets.}
Full training data give the highest MIMIC-III macro and VitalDB forecast-macro AUROC. At the largest tested subsets, random selection leads MIMIC-III (30\% of windows), while gradient matching and noise filtering lead VitalDB (20\% of subjects).
MIMIC-III task-anchor selection worsens with larger quotas: prioritizing transitions and hard negatives remains weaker than random sampling (Figure~\ref{fig:information-multimodal}(d)).
On MC-MED, training dynamics and diversity at 50\% of subjects slightly exceed the full-cohort mean (0.5695 and 0.5673 versus 0.5669).
In a separate fixed-recipe VitalDB test, 20\% gradient-matched subjects preserve concurrent BIS AUROC (0.884 versus 0.883) but reduce forecast AUROC from 0.630 to 0.578.
Approximating full-data gradients through selection~\citep{killamsetty2021gradmatch} better preserves current anesthetic-state ranking than pre-event warning information. BIS AUPRC nevertheless falls from 0.606 to 0.592 (Appendix~\ref{app:training-budget-test}).

\ding{185}~\textbf{Transfer to MIMIC-IV.}
For supervised Transformer/Conv1D transfer to five MIMIC-IV targets, head-only adaptation gives the highest tuning macro with MIMIC-III initialization. VitalDB initialization offers no consistent macro benefit (Figure~\ref{fig:information-multimodal}(e)).
Learning a target decision boundary from fixed source features gives the largest aggregate gain without encoder updates.
On circulatory deterioration, head-only adaptation beats random initialization, while logistic regression on the same frozen features falls below it.
Waveform adaptation outperforms EHR alone for circulatory deterioration, whereas EHR accounts for most combined-feature gains on AKI and delirium.
Hemodynamic monitoring captures circulatory instability, while AKI is defined by renal measurements~\citep{hyland2020circulatory,kdigo2012aki} (Appendix~\ref{app:information-transfer}).

\noindent\textbf{Efficacy, efficiency, and safety.}
MIMIC-III clinical channel selection retains AUPRC (0.096 versus 0.098) and reduces peak memory (2.353\% versus 3.395\%), but increases ECE (0.228 versus 0.178).
On VitalDB, weighted late and gated fusion have similar AUPRC (0.1255 and 0.1256) but different ECE (0.0251 and 0.0306).
Selecting configurations by AUROC rewards correct ordering without directly optimizing calibration. Sensor and fusion choices can therefore preserve discrimination while changing probability accuracy.
For risk-based bedside alerts, ranking patients correctly and assigning reliable probabilities for clinical review are distinct requirements~\citep{vancalster2019calibration}.
Companion results appear in Appendices~\ref{app:ees-representation}, \ref{app:ees-temporal}, and~\ref{app:ees-information}. Definitions are in Appendix~\ref{app:evaluation-details}.

\textbf{(RQ1) Representation and architecture:} Tokenizer and encoder choices change model rankings by endpoint. Conv1D leads the VitalDB and MC-MED tokenizer macros, while Transformer leads the MIMIC-III and VitalDB architecture macros. MC-MED favors BiLSTM for decompensation and hyperkalemia and DLinear for anemia.
\textbf{(RQ2) Temporal modeling:} Longer histories offer task-dependent gains that often plateau. Longer warning gaps reduce discrimination for some imminent events but produce nonmonotonic effects elsewhere. Patient partitions stay fixed while shifted outcome windows can change eligibility and prevalence.
\textbf{(RQ3) Information availability:} Full training sets lead MIMIC-III macro and VitalDB forecast-macro AUROC. Clinical-priority and SQI selection preserve MIMIC-III performance with fewer sensors, whereas VitalDB favors all channels. Missing-input handling remains task-dependent: Raindrop matches VitalDB's native-handling macro, but all added methods lower desaturation AUROC.
Detailed comparisons appear in Appendices~\ref{app:representation-results}--\ref{app:information-comparisons}. Appendix~\ref{app:model-development-recommendations} gives task-specific recommendations.




\begin{wrapfigure}{r}{0.3\textwidth}
    \centering
    \vspace{-4mm}
    \includegraphics[width=\linewidth]{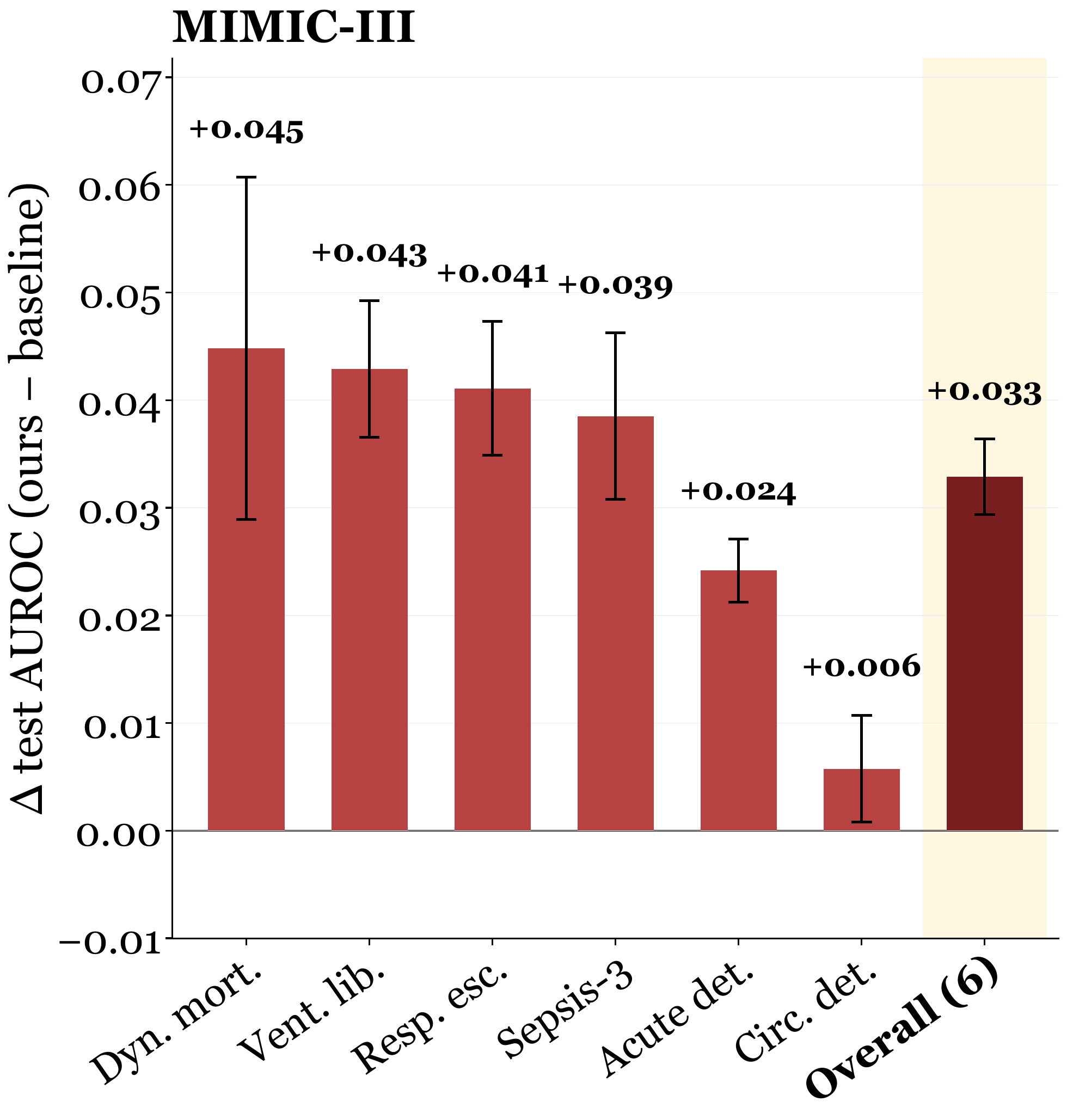}
    \vspace{-7mm}
    \caption{\small AUROC gains over the parameter-matched baseline: six MIMIC-III tasks and macro, five-seed means.}
    \vspace{-14mm}
    \label{fig:moe-ssl-improvement}
\end{wrapfigure}

\noindent\textbf{MoE with self-supervised pretraining.}
To evaluate our benchmark as a foundation for model development, we combine mixture-of-experts (MoE) routing~\citep{fedus2022switch} with self-supervised learning (SSL)~\citep{oh2022leadagnostic} in \textbf{MoE+SSL}.
SSL learns representations from unlabeled waveforms, while an input-dependent router combines expert outputs for clinical prediction.
Pretraining uses training patients only, retaining the benchmark's splits, information cutoffs, and validation-based selection.
We evaluate five-minute waveform inputs on six surveillance-style MIMIC-III tasks: acute and circulatory deterioration, respiratory escalation, sepsis onset, ventilator liberation, and dynamic mortality.
Figure~\ref{fig:moe-ssl-improvement} shows test AUROC gains on all six tasks over the parameter-matched baseline, ranging from $+0.006$ to $+0.045$, with a macro gain of $+0.033$.

\section{Conclusion}
\method{} benchmarks representation, temporal context, and information availability across 20 clinical waveform classification tasks.
Endpoint differences persist across model families: sepsis drives SeFT's MIMIC-III lead, while FFT lowers MC-MED decompensation and hyperkalemia AUROC across all nine encoders.
History and warning time produce opposing task trends. Fusion and transfer reveal complementary EHR information for hypotension and predominantly EHR-driven gains for AKI and delirium.
These task-level comparisons align waveform representations and training configurations with each endpoint's clinical information and intervention time.
The clinically defined events and states support bedside screening, warning decisions, and comparisons across care settings.
Future work will extend to regression tasks that estimate physiological trajectories and quantify the magnitude of clinical change.


\subsection*{AI use statement}
Generative AI tools were used only to correct typographical errors in the manuscript.
The authors take responsibility for the final text, scientific claims, and reported results.

\subsection*{Ethics statement}
\method{} is a retrospective benchmark using existing de-identified clinical datasets: MIMIC-III, VitalDB, and MC-MED, with additional transfer experiments on MIMIC-IV~\citep{moody2020mimic3waveform,lee2022vitaldb,kansal2025mcmed,johnson2023mimiciv}.
The study analyzes previously collected records and involves no new patient recruitment or clinical intervention.
Access to and reuse of the source data remain subject to each provider's access conditions and data-use terms. Reproducing the benchmark requires obtaining the relevant data from those providers.

\subsection*{Reproducibility statement}
Section~\ref{sec:benchmark-protocol-draft} specifies the prediction interface, information cutoffs, patient-level partitioning, and rules for controlled comparisons.
Appendix~\ref{app:dataset-details} details task definitions, eligibility, preprocessing, and cohort support. Appendix~\ref{app:architecture-tokenizer-protocol} documents training settings, search grids, random seeds, and validation-based model selection, with method adaptations in Appendix~\ref{app:method-adaptations}.
Appendix~\ref{app:evaluation-details} defines metrics, aggregation, uncertainty estimates, and resource measurements.
Appendices~\ref{app:temporal-comparisons} and~\ref{app:information-comparisons} describe the temporal and information-availability comparisons, including the MIMIC-IV transfer protocol in Appendix~\ref{app:information-transfer}.
Appendix~\ref{app:numerical-tables} provides the numerical results underlying the reported comparisons.



\section*{Acknowledgements}
In conclusion I wish to say that in working at the problem here dealt with I
have had the loyal assistance of my friend and colleague M.~Besso, and that I am
indebted to him for several valuable suggestions.

\bibliographystyle{style/icml2025}
\bibliography{999_reference}

@article{harutyunyan2019multitask,
  author = {Harutyunyan, Hrayr and Khachatrian, Hrant and Kale, David C. and Ver Steeg, Greg and Galstyan, Aram},
  title = {Multitask Learning and Benchmarking with Clinical Time Series Data},
  journal = {Scientific Data},
  volume = {6},
  pages = {96},
  year = {2019},
  doi = {10.1038/s41597-019-0103-9}
}

@inproceedings{vandewater2024yaib,
  title = {Yet Another {ICU} Benchmark: A Flexible Multi-Center Framework for Clinical {ML}},
  author = {van de Water, Robin and Schmidt, Hendrik and Elbers, Paul and Thoral, Patrick and Arnrich, Bert and Rockenschaub, Patrick},
  booktitle = {International Conference on Learning Representations},
  year = {2024},
  url = {https://proceedings.iclr.cc/paper_files/paper/2024/hash/c26a89073d972f2e6643617b0f3a9e8a-Abstract-Conference.html}
}

@inproceedings{chen2023mcbec,
  author = {Chen, Emma and Kansal, Aman and Chen, Julie and Jin, Boyang Tom and Reisler, Julia Rachel and Kim, David A. and Rajpurkar, Pranav},
  title = {Multimodal Clinical Benchmark for Emergency Care ({MC-BEC}): A Comprehensive Benchmark for Evaluating Foundation Models in Emergency Medicine},
  booktitle = {Advances in Neural Information Processing Systems},
  volume = {36},
  pages = {45794--45811},
  year = {2023}
}

@article{wang2023pulsedb,
  author = {Wang, Weinan and Mohseni, Pedram and Kilgore, Kevin L. and Najafizadeh, Laleh},
  title = {{PulseDB}: A Large, Cleaned Dataset Based on {MIMIC-III} and {VitalDB} for Benchmarking Cuff-less Blood Pressure Estimation Methods},
  journal = {Frontiers in Digital Health},
  volume = {4},
  pages = {1090854},
  year = {2023},
  doi = {10.3389/fdgth.2022.1090854}
}

@article{lopezalcaraz2025enhancing,
  author = {Lopez Alcaraz, Juan Miguel and Bouma, Hjalmar and Strodthoff, Nils},
  title = {Enhancing Clinical Decision Support with Physiological Waveforms---A Multimodal Benchmark in Emergency Care},
  journal = {Computers in Biology and Medicine},
  volume = {192},
  pages = {110196},
  year = {2025},
  doi = {10.1016/j.compbiomed.2025.110196}
}

@article{gustafsson2026signalmcmed,
  author = {Gustafsson, Fredrik K. and Gu, Xiao and Carletti, Mattia and Palo, Patitapaban and Eyre, David W. and Clifton, David A.},
  title = {{SignalMC-MED}: A Multimodal Benchmark for Evaluating Biosignal Foundation Models on Single-Lead {ECG} and {PPG}},
  journal = {arXiv preprint arXiv:2603.09940},
  year = {2026}
}

@article{moody2020mimic3waveform,
  author = {Moody, Benjamin and Moody, George and Villarroel, Mauricio and Clifford, Gari D. and Silva, Ikaro},
  title = {{MIMIC-III} Waveform Database Matched Subset},
  journal = {PhysioNet},
  year = {2020},
  note = {Version 1.0},
  doi = {10.13026/c2294b}
}

@article{lee2022vitaldb,
  author = {Lee, Hyung-Chul and Park, Yoonsang and Yoon, Soo Bin and Yang, Seong Mi and Park, Dongnyeok and Jung, Chul-Woo},
  title = {{VitalDB}, a High-Fidelity Multi-Parameter Vital Signs Database in Surgical Patients},
  journal = {Scientific Data},
  volume = {9},
  pages = {279},
  year = {2022},
  doi = {10.1038/s41597-022-01411-5}
}

@article{kansal2025mcmed,
  author = {Kansal, Aman and Chen, Emma and Jin, Boyang Tom and Rajpurkar, Pranav and Kim, David A.},
  title = {{MC-MED}, Multimodal Clinical Monitoring in the Emergency Department},
  journal = {Scientific Data},
  volume = {12},
  pages = {1094},
  year = {2025},
  doi = {10.1038/s41597-025-05419-5}
}

@article{hatib2018machine,
  title = {Machine-Learning {{Algorithm}} to {{Predict Hypotension Based}} on {{High-fidelity Arterial Pressure Waveform Analysis}}},
  author = {Hatib, Feras and Jian, Zhongping and Buddi, Sai and Lee, Christine and Settels, Jos and Sibert, Karen and Rinehart, Joseph and Cannesson, Maxime},
  date = {2018-10},
  journal = {Anesthesiology},
  volume = {129},
  number = {4},
  eprint = {29894315},
  eprinttype = {pubmed},
  pages = {663--674},
  issn = {1528-1175},
  doi = {10.1097/ALN.0000000000002300},
  year = {2018},
  langid = {english}
}

@article{galloway2019hyperkalemia,
  title = {Development and Validation of a Deep-Learning Model to Screen for Hyperkalemia From the Electrocardiogram},
  author = {Galloway, Conner D. and Valys, Alexander V. and Shreibati, Jacqueline B. and Treiman, Daniel L. and Petterson, Frank L. and Gundotra, Vivek P. and Albert, David E. and Attia, Zachi I. and Carter, Rickey E. and Asirvatham, Samuel J. and Ackerman, Michael J. and Noseworthy, Peter A. and Dillon, John J. and Friedman, Paul A.},
  journal = {JAMA Cardiology},
  volume = {4},
  number = {5},
  pages = {428--436},
  year = {2019},
  doi = {10.1001/jamacardio.2019.0640}
}

@article{lauritsen2021framing,
  title = {The {{Framing}} of Machine Learning Risk Prediction Models Illustrated by Evaluation of Sepsis in General Wards},
  author = {Lauritsen, Simon Meyer and Thiesson, Bo and Jørgensen, Marianne Johansson and Riis, Anders Hammerich and Espelund, Ulrick Skipper and Weile, Jesper Bo and Lange, Jeppe},
  date = {2021-11-15},
  journal = {npj Digital Medicine},
  shortjournal = {Npj Digit. Med.},
  volume = {4},
  number = {1},
  pages = {158},
  publisher = {Nature Publishing Group},
  issn = {2398-6352},
  year = {2021},
  doi = {10.1038/s41746-021-00529-x},
  langid = {english}
}

@inproceedings{nie2023time,
  title = {A {{Time Series}} Is {{Worth}} 64 {{Words}}: {{Long-term Forecasting}} with {{Transformers}}},
  shorttitle = {A {{Time Series}} Is {{Worth}} 64 {{Words}}},
  author = {Nie, Yuqi and Nguyen, Nam H. and Sinthong, Phanwadee and Kalagnanam, Jayant},
  date = {2022-09-29},
  url = {https://openreview.net/forum?id=Jbdc0vTOcol},
  urldate = {2026-08-28},
  booktitle = {The {{Eleventh International Conference}} on {{Learning Representations}}},
  year = {2023},
  langid = {english}
}

@inproceedings{zeng2023transformers,
  title = {Are Transformers Effective for Time Series Forecasting?},
  booktitle = {Proceedings of the {{Thirty-Seventh AAAI Conference}} on {{Artificial Intelligence}} and {{Thirty-Fifth Conference}} on {{Innovative Applications}} of {{Artificial Intelligence}} and {{Thirteenth Symposium}} on {{Educational Advances}} in {{Artificial Intelligence}}},
  author = {Zeng, Ailing and Chen, Muxi and Zhang, Lei and Xu, Qiang},
  date = {2023-02-07},
  series = {{{AAAI}}'23/{{IAAI}}'23/{{EAAI}}'23},
  volume = {37},
  year = {2023},
  pages = {11121--11128},
  publisher = {AAAI Press},
  doi = {10.1609/aaai.v37i9.26317},
  isbn = {978-1-57735-880-0}
}

@article{agniel2018biases,
  title = {{Biases in electronic health record data due to processes within the healthcare system: retrospective observational study}},
  author = {Agniel, Denis and Kohane, Isaac S and Weber, Griffin M},
  journal = {BMJ},
  volume = {361},
  pages = {k1479},
  year = {2018},
  doi = {10.1136/bmj.k1479}
}

@article{beaulieujones2021clinicians,
  title = {{Machine learning for patient risk stratification: standing on, or looking over, the shoulders of clinicians?}},
  author = {Beaulieu-Jones, Brett K and Yuan, William and Brat, Gabriel A and Beam, Andrew L and Weber, Griffin and Ruffin, Marshall and Kohane, Isaac S},
  journal = {NPJ digital medicine},
  volume = {4},
  number = {1},
  pages = {62},
  year = {2021},
  doi = {10.1038/s41746-021-00426-3}
}

@article{che2018missing,
  title = {{Recurrent Neural Networks for Multivariate Time Series with Missing Values}},
  author = {Che, Zhengping and Purushotham, Sanjay and Cho, Kyunghyun and Sontag, David and Liu, Yan},
  journal = {Scientific reports},
  volume = {8},
  number = {1},
  pages = {6085},
  year = {2018},
  doi = {10.1038/s41598-018-24271-9}
}

@article{enevoldsen2022selection,
  title = {{Performance of the Hypotension Prediction Index May Be Overestimated Due to Selection Bias}},
  author = {Enevoldsen, Johannes and Vistisen, Simon T},
  journal = {Anesthesiology},
  volume = {137},
  number = {3},
  pages = {283--289},
  year = {2022},
  doi = {10.1097/ALN.0000000000004320}
}

@article{hanley1982roc,
  title = {{The meaning and use of the area under a receiver operating characteristic (ROC) curve}},
  author = {Hanley, J A and McNeil, B J},
  journal = {Radiology},
  volume = {143},
  number = {1},
  pages = {29--36},
  year = {1982},
  doi = {10.1148/radiology.143.1.7063747}
}

@article{hyland2020circulatory,
  title = {{Early prediction of circulatory failure in the intensive care unit using machine learning}},
  author = {Hyland, Stephanie L and Faltys, Martin and Hüser, Matthias and Lyu, Xinrui and Gumbsch, Thomas and Esteban, Cristóbal and Bock, Christian and Horn, Max and Moor, Michael and Rieck, Bastian and Zimmermann, Marc and Bodenham, Dean and Borgwardt, Karsten and Rätsch, Gunnar and Merz, Tobias M},
  journal = {Nature medicine},
  volume = {26},
  number = {3},
  pages = {364--373},
  year = {2020},
  doi = {10.1038/s41591-020-0789-4}
}

@article{singer2016sepsis,
  title = {The Third International Consensus Definitions for Sepsis and Septic Shock ({Sepsis-3})},
  author = {Singer, Mervyn and Deutschman, Clifford S. and Seymour, Christopher Warren and others},
  journal = {JAMA},
  year = {2016},
  volume = {315},
  number = {8},
  pages = {801--810},
  doi = {10.1001/jama.2016.0287}
}

@article{johnson2018sepsis,
  title = {{A Comparative Analysis of Sepsis Identification Methods in an Electronic Database}},
  author = {Johnson, Alistair E W and Aboab, Jerome and Raffa, Jesse D and Pollard, Tom J and Deliberato, Rodrigo O and Celi, Leo A and Stone, David J},
  journal = {Critical care medicine},
  volume = {46},
  number = {4},
  pages = {494--499},
  year = {2018},
  doi = {10.1097/CCM.0000000000002965}
}

@article{kwon2020anaemia,
  title = {{A deep learning algorithm to detect anaemia with ECGs: a retrospective, multicentre study}},
  author = {Kwon, Joon-Myoung and Cho, Younghoon and Jeon, Ki-Hyun and Cho, Soohyun and Kim, Kyung-Hee and Baek, Seung Don and Jeung, Soomin and Park, Jinsik and Oh, Byung-Hee},
  journal = {The Lancet Digital Health},
  volume = {2},
  number = {7},
  pages = {e358--e367},
  year = {2020},
  doi = {10.1016/S2589-7500(20)30108-4}
}

@article{ledowski2019nociception,
  title = {{Objective monitoring of nociception: a review of current commercial solutions}},
  author = {Ledowski, Thomas},
  journal = {British journal of anaesthesia},
  volume = {123},
  number = {2},
  pages = {e312--e321},
  year = {2019},
  doi = {10.1016/j.bja.2019.03.024}
}

@article{nassar2016capnography,
  title = {{Capnography During Critical Illness}},
  author = {Nassar, Boulos S and Schmidt, Gregory A},
  journal = {Chest},
  volume = {149},
  number = {2},
  pages = {576--585},
  year = {2016},
  doi = {10.1378/chest.15-1369}
}

@article{pierrakos2012pressure,
  title = {{Can changes in arterial pressure be used to detect changes in cardiac index during fluid challenge in patients with septic shock?}},
  author = {Pierrakos, Charalampos and Velissaris, Dimitrios and Scolletta, Sabino and Heenen, Sarah and De Backer, Daniel and Vincent, Jean-Louis},
  journal = {Intensive care medicine},
  volume = {38},
  number = {3},
  pages = {422--428},
  year = {2012},
  doi = {10.1007/s00134-011-2457-0}
}

@article{riley2020sample,
  title = {{Calculating the sample size required for developing a clinical prediction model}},
  author = {Riley, Richard D and Ensor, Joie and Snell, Kym I E and Harrell, Jr., Frank E and Martin, Glen P and Reitsma, Johannes B and Moons, Karel G M and Collins, Gary and van Smeden, Maarten},
  journal = {BMJ},
  volume = {368},
  pages = {m441},
  year = {2020},
  doi = {10.1136/bmj.m441}
}

@article{saeb2017usecase,
  title = {{The need to approximate the use-case in clinical machine learning}},
  author = {Saeb, Sohrab and Lonini, Luca and Jayaraman, Arun and Mohr, David C and Kording, Konrad P},
  journal = {GigaScience},
  volume = {6},
  number = {5},
  pages = {1--9},
  year = {2017},
  doi = {10.1093/gigascience/gix019}
}

@article{saito2015precision,
  title = {{The precision-recall plot is more informative than the ROC plot when evaluating binary classifiers on imbalanced datasets}},
  author = {Saito, Takaya and Rehmsmeier, Marc},
  journal = {PloS one},
  volume = {10},
  number = {3},
  pages = {e0118432},
  year = {2015},
  doi = {10.1371/journal.pone.0118432}
}

@article{sessler2016thermoregulation,
  title = {{Perioperative thermoregulation and heat balance}},
  author = {Sessler, Daniel I},
  journal = {The Lancet},
  volume = {387},
  number = {10038},
  pages = {2655--2664},
  year = {2016},
  doi = {10.1016/S0140-6736(15)00981-2}
}

@article{sundrani2023decompensation,
  title = {{Predicting patient decompensation from continuous physiologic monitoring in the emergency department}},
  author = {Sundrani, Sameer and Chen, Julie and Jin, Boyang Tom and Abad, Zahra Shakeri Hossein and Rajpurkar, Pranav and Kim, David},
  journal = {NPJ digital medicine},
  volume = {6},
  number = {1},
  pages = {60},
  year = {2023},
  doi = {10.1038/s41746-023-00803-0}
}

@article{tomasev2019aki,
  title = {{A clinically applicable approach to continuous prediction of future acute kidney injury}},
  author = {Tomašev, Nenad and Glorot, Xavier and Rae, Jack W and Zielinski, Michal and Askham, Harry and Saraiva, Andre and Mottram, Anne and Meyer, Clemens and Ravuri, Suman and Protsyuk, Ivan and Connell, Alistair and Hughes, Cían O and Karthikesalingam, Alan and Cornebise, Julien and Montgomery, Hugh and Rees, Geraint and Laing, Chris and Baker, Clifton R and Peterson, Kelly and Reeves, Ruth and Hassabis, Demis and King, Dominic and Suleyman, Mustafa and Back, Trevor and Nielson, Christopher and Ledsam, Joseph R and Mohamed, Shakir},
  journal = {Nature},
  volume = {572},
  number = {7767},
  pages = {116--119},
  year = {2019},
  doi = {10.1038/s41586-019-1390-1}
}

@article{torrini2021extubation,
  title = {{Prediction of extubation outcome in critically ill patients: a systematic review and meta-analysis}},
  author = {Torrini, Flavia and Gendreau, Ségolène and Morel, Johanna and Carteaux, Guillaume and Thille, Arnaud W and Antonelli, Massimo and Mekontso Dessap, Armand},
  journal = {Critical care (London, England)},
  volume = {25},
  number = {1},
  pages = {391},
  year = {2021},
  doi = {10.1186/s13054-021-03802-3}
}

@article{vincent2016lactate,
  title = {{The value of blood lactate kinetics in critically ill patients: a systematic review}},
  author = {Vincent, Jean-Louis and Quintairos E Silva, Amanda and Couto, Jr., Lúcio and Taccone, Fabio S},
  journal = {Critical care (London, England)},
  volume = {20},
  number = {1},
  pages = {257},
  year = {2016},
  doi = {10.1186/s13054-016-1403-5}
}

@inproceedings{zhang2023crossformer,
  title = {{Crossformer: Transformer Utilizing Cross-Dimension Dependency for Multivariate Time Series Forecasting}},
  author = {Zhang, Yunhao and Yan, Junchi},
  booktitle = {International Conference on Learning Representations},
  year = {2023},
  url = {https://openreview.net/forum?id=vSVLM2j9eie}
}

@inproceedings{liu2024itransformer,
  title = {{iTransformer: Inverted Transformers Are Effective for Time Series Forecasting}},
  author = {Liu, Yong and Hu, Tengge and Zhang, Haoran and Wu, Haixu and Wang, Shiyu and Ma, Lintao and Long, Mingsheng},
  booktitle = {International Conference on Learning Representations},
  year = {2024},
  url = {https://proceedings.iclr.cc/paper_files/paper/2024/hash/2ea18fdc667e0ef2ad82b2b4d65147ad-Abstract-Conference.html}
}

@inproceedings{horn2020seft,
  title = {{Set Functions for Time Series}},
  author = {Horn, Max and Moor, Michael and Bock, Christian and Rieck, Bastian and Borgwardt, Karsten},
  booktitle = {Proceedings of the 37th International Conference on Machine Learning},
  series = {Proceedings of Machine Learning Research},
  volume = {119},
  pages = {4353--4363},
  publisher = {PMLR},
  year = {2020},
  url = {https://proceedings.mlr.press/v119/horn20a.html}
}

@inproceedings{shukla2021mtand,
  title = {{Multi-Time Attention Networks for Irregularly Sampled Time Series}},
  author = {Shukla, Satya Narayan and Marlin, Benjamin M.},
  booktitle = {International Conference on Learning Representations},
  year = {2021},
  url = {https://arxiv.org/abs/2101.10318}
}

@inproceedings{vaswani2017attention,
  title = {{Attention Is All You Need}},
  author = {Vaswani, Ashish and Shazeer, Noam and Parmar, Niki and Uszkoreit, Jakob and Jones, Llion and Gomez, Aidan N. and Kaiser, Lukasz and Polosukhin, Illia},
  booktitle = {Advances in Neural Information Processing Systems},
  year = {2017},
  url = {https://arxiv.org/abs/1706.03762}
}

@article{graves2005bilstm,
  title = {{Framewise Phoneme Classification with Bidirectional LSTM and Other Neural Network Architectures}},
  author = {Graves, Alex and Schmidhuber, J{\"u}rgen},
  journal = {Neural Networks},
  volume = {18},
  number = {5--6},
  pages = {602--610},
  year = {2005},
  doi = {10.1016/j.neunet.2005.06.042},
  url = {https://www.cs.toronto.edu/~graves/nn_2005.pdf}
}

@inproceedings{lahoti2026mamba3,
  title = {{Mamba-3: Improved Sequence Modeling using State Space Principles}},
  author = {Lahoti, Aakash and Li, Kevin Y. and Chen, Berlin and Wang, Caitlin and Bick, Aviv and Kolter, J. Zico and Dao, Tri and Gu, Albert},
  booktitle = {International Conference on Learning Representations},
  year = {2026},
  url = {https://arxiv.org/abs/2603.15569}
}

@article{fawaz2020inceptiontime,
  title = {{InceptionTime: Finding AlexNet for Time Series Classification}},
  author = {{Ismail Fawaz}, Hassan and Lucas, Benjamin and Forestier, Germain and Pelletier, Charlotte and Schmidt, Daniel F. and Weber, Jonathan and Webb, Geoffrey I. and Idoumghar, Lhassane and Muller, Pierre-Alain and Petitjean, Fran{\c c}ois},
  journal = {Data Mining and Knowledge Discovery},
  year = {2020},
  url = {https://arxiv.org/abs/1909.04939}
}

@article{bai2018tcn,
  title = {{An Empirical Evaluation of Generic Convolutional and Recurrent Networks for Sequence Modeling}},
  author = {Bai, Shaojie and Kolter, J. Zico and Koltun, Vladlen},
  journal = {arXiv preprint arXiv:1803.01271},
  year = {2018},
  url = {https://arxiv.org/abs/1803.01271}
}

@inproceedings{wang2017resnet,
  title = {{Time Series Classification from Scratch with Deep Neural Networks: A Strong Baseline}},
  author = {Wang, Zhiguang and Yan, Weizhong and Oates, Tim},
  booktitle = {International Joint Conference on Neural Networks},
  pages = {1578--1585},
  year = {2017},
  doi = {10.1109/IJCNN.2017.7966039},
  url = {https://arxiv.org/abs/1611.06455}
}

@inproceedings{gu2022s4,
  title = {{Efficiently Modeling Long Sequences with Structured State Spaces}},
  author = {Gu, Albert and Goel, Karan and R{\'e}, Christopher},
  booktitle = {International Conference on Learning Representations},
  year = {2022},
  url = {https://arxiv.org/abs/2111.00396}
}

@inproceedings{xu2024fits,
  title = {{FITS: Modeling Time Series with 10k Parameters}},
  author = {Xu, Zhijian and Zeng, Ailing and Xu, Qiang},
  booktitle = {International Conference on Learning Representations},
  year = {2024},
  url = {https://proceedings.iclr.cc/paper_files/paper/2024/hash/701251e1db4a2e4dd2ef23f5265d5936-Abstract-Conference.html}
}

@article{tipirneni2022strats,
  title = {{Self-Supervised Transformer for Sparse and Irregularly Sampled Multivariate Clinical Time-Series}},
  author = {Tipirneni, Sindhu and Reddy, Chandan K.},
  journal = {ACM Transactions on Knowledge Discovery from Data},
  volume = {16},
  number = {6},
  year = {2022},
  doi = {10.1145/3516367},
  url = {https://arxiv.org/abs/2107.14293}
}

@inproceedings{cini2022grin,
  title = {{Filling the Gaps: Multivariate Time Series Imputation by Graph Neural Networks}},
  author = {Cini, Andrea and Marisca, Ivan and Alippi, Cesare},
  booktitle = {International Conference on Learning Representations},
  year = {2022},
  url = {https://arxiv.org/abs/2108.00298}
}

@inproceedings{zhang2022raindrop,
  title = {{Graph-Guided Network for Irregularly Sampled Multivariate Time Series}},
  author = {Zhang, Xiang and Zeman, Marko and Tsiligkaridis, Theodoros and Zitnik, Marinka},
  booktitle = {International Conference on Learning Representations},
  year = {2022},
  url = {https://arxiv.org/abs/2110.05357}
}

@inproceedings{wu2021ignnk,
  title = {{Inductive Graph Neural Networks for Spatiotemporal Kriging}},
  author = {Wu, Yuankai and Zhuang, Dingyi and Labbe, Aurelie and Sun, Lijun},
  booktitle = {Proceedings of the AAAI Conference on Artificial Intelligence},
  volume = {35},
  pages = {4478--4485},
  year = {2021},
  doi = {10.1609/aaai.v35i5.16575},
  url = {https://ojs.aaai.org/index.php/AAAI/article/view/16575}
}

@inproceedings{oh2022leadagnostic,
  title = {{Lead-agnostic Self-supervised Learning for Local and Global Representations of Electrocardiogram}},
  author = {Oh, Jungwoo and Chung, Hyunseung and Kwon, Joon-myoung and Hong, Dong-gyun and Choi, Edward},
  booktitle = {Proceedings of the Conference on Health, Inference, and Learning},
  volume = {174},
  pages = {338--353},
  year = {2022},
  url = {https://proceedings.mlr.press/v174/oh22a.html}
}

@inproceedings{liu2023rectifiedflow,
  title = {{Flow Straight and Fast: Learning to Generate and Transfer Data with Rectified Flow}},
  author = {Liu, Xingchao and Gong, Chengyue and Liu, Qiang},
  booktitle = {International Conference on Learning Representations},
  year = {2023},
  url = {https://arxiv.org/abs/2209.03003}
}

@inproceedings{lu2025trigflow,
  title = {{Simplifying, Stabilizing and Scaling Continuous-Time Consistency Models}},
  author = {Lu, Cheng and Song, Yang},
  booktitle = {International Conference on Learning Representations},
  year = {2025},
  url = {https://proceedings.iclr.cc/paper_files/paper/2025/hash/7e9c2053258b1bdd32ff2654802cd594-Abstract-Conference.html}
}

@article{deng2026drifting,
  title = {{Generative Modeling via Drifting}},
  author = {Deng, Mingyang and Li, He and Li, Tianhong and Du, Yilun and He, Kaiming},
  journal = {arXiv preprint arXiv:2602.04770},
  year = {2026},
  url = {https://arxiv.org/abs/2602.04770}
}

@inproceedings{sener2018coreset,
  title = {{Active Learning for Convolutional Neural Networks: A Core-Set Approach}},
  author = {Sener, Ozan and Savarese, Silvio},
  booktitle = {International Conference on Learning Representations},
  year = {2018},
  url = {https://arxiv.org/abs/1708.00489}
}

@article{mcinnes2018umap,
  title = {{UMAP: Uniform Manifold Approximation and Projection for Dimension Reduction}},
  author = {McInnes, Leland and Healy, John and Melville, James},
  journal = {arXiv preprint arXiv:1802.03426},
  year = {2018},
  url = {https://arxiv.org/abs/1802.03426}
}

@article{vazifeh2026nldr,
  title = {{Manifold Learning for Personalized and Label-Free Detection of Cardiac Arrhythmias}},
  author = {Vazifeh, Amir Reza and Fleischer, Jason W.},
  journal = {Informatics in Medicine Unlocked},
  volume = {64},
  pages = {101770},
  year = {2026},
  doi = {10.1016/j.imu.2026.101770},
  url = {https://arxiv.org/abs/2506.16494}
}

@inproceedings{mirzasoleiman2020craig,
  title = {{Coresets for Data-efficient Training of Machine Learning Models}},
  author = {Mirzasoleiman, Baharan and Bilmes, Jeff and Leskovec, Jure},
  booktitle = {Proceedings of the 37th International Conference on Machine Learning},
  volume = {119},
  pages = {6950--6960},
  year = {2020},
  url = {https://proceedings.mlr.press/v119/mirzasoleiman20a.html}
}

@inproceedings{killamsetty2021gradmatch,
  title = {{GRAD-MATCH: Gradient Matching based Data Subset Selection for Efficient Deep Model Training}},
  author = {Killamsetty, Krishnateja and S, Durga and Ramakrishnan, Ganesh and De, Abir and Iyer, Rishabh},
  booktitle = {Proceedings of the 38th International Conference on Machine Learning},
  volume = {139},
  pages = {5464--5474},
  year = {2021},
  url = {https://proceedings.mlr.press/v139/killamsetty21a.html}
}

@inproceedings{swayamdipta2020cartography,
  title = {{Dataset Cartography: Mapping and Diagnosing Datasets with Training Dynamics}},
  author = {Swayamdipta, Swabha and Schwartz, Roy and Lourie, Nicholas and Wang, Yizhong and Hajishirzi, Hannaneh and Smith, Noah A. and Choi, Yejin},
  booktitle = {Proceedings of the 2020 Conference on Empirical Methods in Natural Language Processing},
  year = {2020},
  url = {https://arxiv.org/abs/2009.10795}
}

@inproceedings{paul2021datadiet,
  title = {{Deep Learning on a Data Diet: Finding Important Examples Early in Training}},
  author = {Paul, Mansheej and Ganguli, Surya and Dziugaite, Gintare Karolina},
  booktitle = {Advances in Neural Information Processing Systems},
  volume = {34},
  year = {2021},
  url = {https://arxiv.org/abs/2107.07075}
}

@article{peng2005mi,
  title = {{Feature Selection Based on Mutual Information: Criteria of Max-Dependency, Max-Relevance, and Min-Redundancy}},
  author = {Peng, Hanchuan and Long, Fuhui and Ding, Chris},
  journal = {IEEE Transactions on Pattern Analysis and Machine Intelligence},
  volume = {27},
  number = {8},
  pages = {1226--1238},
  year = {2005},
  doi = {10.1109/TPAMI.2005.159},
  url = {https://home.penglab.com/papersall/docpdf/2005_TPAMI_FeaSel.pdf}
}

@article{tibshirani1996lasso,
  title = {{Regression Shrinkage and Selection via the Lasso}},
  author = {Tibshirani, Robert},
  journal = {Journal of the Royal Statistical Society: Series B},
  volume = {58},
  number = {1},
  pages = {267--288},
  year = {1996},
  doi = {10.1111/j.2517-6161.1996.tb02080.x}
}

@article{clifford2012sqi,
  title = {{Signal Quality Indices and Data Fusion for Determining Clinical Acceptability of Electrocardiograms}},
  author = {Clifford, G. D. and Behar, J. and Li, Q. and Rezek, I.},
  journal = {Physiological Measurement},
  volume = {33},
  number = {9},
  pages = {1419--1433},
  year = {2012},
  doi = {10.1088/0967-3334/33/9/1419}
}

@article{johnson2023mimiciv,
  author = {Johnson, Alistair E. W. and Bulgarelli, Lucas and Shen, Lu and Gayles, Alvin and Shammout, Ayad and Horng, Steven and Pollard, Tom J. and Hao, Sicheng and Moody, Benjamin and Gow, Brian and Lehman, Li-Wei H. and Celi, Leo A. and Mark, Roger G.},
  title = {{MIMIC-IV}, a Freely Accessible Electronic Health Record Dataset},
  journal = {Scientific Data},
  volume = {10},
  pages = {1},
  year = {2023},
  doi = {10.1038/s41597-022-01899-x}
}

@article{vancalster2019calibration,
  author = {Van Calster, Ben and McLernon, David J. and van Smeden, Maarten and Wynants, Laure and Steyerberg, Ewout W.},
  title = {Calibration: The Achilles Heel of Predictive Analytics},
  journal = {BMC Medicine},
  volume = {17},
  pages = {230},
  year = {2019},
  doi = {10.1186/s12916-019-1466-7}
}

@article{lundberg2018hypoxaemia,
  author = {Lundberg, Scott M. and Nair, Bala and Vavilala, Monica S. and Horibe, Mayumi and Eisses, Michael J. and Adams, Trevor and Liston, David E. and Low, Daniel King-Wai and Newman, Shu-Fang and Kim, Jerry and Lee, Su-In},
  title = {Explainable Machine-Learning Predictions for the Prevention of Hypoxaemia during Surgery},
  journal = {Nature Biomedical Engineering},
  volume = {2},
  number = {10},
  pages = {749--760},
  year = {2018},
  doi = {10.1038/s41551-018-0304-0}
}

@article{kdigo2012aki,
  author = {{KDIGO Acute Kidney Injury Work Group}},
  title = {{KDIGO} Clinical Practice Guideline for Acute Kidney Injury},
  journal = {Kidney International Supplements},
  volume = {2},
  number = {1},
  pages = {1--138},
  year = {2012},
  url = {https://kdigo.org/wp-content/uploads/2016/10/KDIGO-2012-AKI-Guideline-English.pdf}
}

@inproceedings{yi2023frets,
  author = {Yi, Kun and Zhang, Qi and Fan, Wei and Wang, Shoujin and Wang, Pengyang and He, Hui and Lian, Defu and An, Ning and Cao, Longbing and Niu, Zhendong},
  title = {Frequency-Domain {MLPs} Are More Effective Learners in Time Series Forecasting},
  booktitle = {Advances in Neural Information Processing Systems},
  volume = {36},
  year = {2023},
  url = {https://proceedings.neurips.cc/paper_files/paper/2023/hash/f1d16af76939f476b5f040fd1398c0a3-Abstract-Conference.html}
}

@article{fedus2022switch,
  author = {Fedus, William and Zoph, Barret and Shazeer, Noam},
  title = {{Switch Transformers}: Scaling to Trillion Parameter Models with Simple and Efficient Sparsity},
  journal = {Journal of Machine Learning Research},
  volume = {23},
  number = {120},
  pages = {1--39},
  year = {2022},
  url = {https://www.jmlr.org/papers/v23/21-0998.html}
}

\appendix
\clearpage
\section{Benchmark Definition Details}
\label{app:benchmark-definition}

\subsection{Clinical-Inference and Temporal-Geometry Taxonomies}
\label{app:benchmark-taxonomies}

\noindent\textbf{Clinical-inference taxonomy.}
We group tasks by their primary clinical inference target, rather than by
dataset, input modality, or anatomical system alone. Acute homeostatic failure
and safety-boundary transitions concern impending changes from relative
stability to clinically actionable instability. Pathophysiologic state and
organ dysfunction concern the onset, persistence, or presence of a defined
biological or organ-level abnormality. Functional reserve and intervention
response concern responses to treatment, physiological challenge, or
withdrawal of organ support. Patient trajectory and care-disposition outcomes
cover downstream survival and healthcare utilization. Organ systems provide
a secondary grouping; dataset, prediction horizon, and task formulation
describe other aspects of each task. We assign each task by its primary
endpoint construct. The dataset-specific registries below define each endpoint, including those for the four
selected MC-MED tasks.

\noindent\textbf{Temporal-geometry taxonomy.}
Temporal geometry describes when observations and targets occur relative
to the prediction anchor \(T\).
\ding{182} \textit{Surveillance} issues predictions at repeated eligible
anchors as new observations accumulate.
\ding{183} \textit{Gapped forecasts} use a nonzero \(G\) to make the minimum
warning time explicit.
\ding{184} \textit{Event- and daily-anchored predictions} align \(T\) to a
recorded action, measurement, or recurring care time rather than a uniform
schedule.
\ding{185} \textit{Concurrent held-out-state probes} withhold a target state
or channel within the observation interval and measure shared information
instead of advance warning; they are not pooled with future-event forecasts.
The main comparison uses one primary endpoint and geometry per task.
Alternative heads, gaps, and horizons are evaluated in sensitivity analyses.

\subsection{Evaluation and Aggregation}
\label{app:evaluation-details}
\noindent\textbf{Metrics and uncertainty.}
The numeric supplement reports AUROC, AUPRC, Brier score, log loss, and expected calibration error (ECE) for the completed VitalDB analyses. Forecast summaries give equal weight to gaps within each task and then to the seven forecasting tasks. Concurrent BIS is reported separately.

For validation, we pool predictions from disjoint held folds within each seed, compute a metric for each seed, and average the five values. Fixed-fold test summaries average metrics from the five frozen fold models. The separately labeled ensemble-test stress analysis instead averages model probabilities before computing metrics. The original subject-selection panel averages saved seed--fold validation metrics for both the selected subsets and the full-data comparator. The later gradient-matching test uses individual model metrics and is reported separately in Appendix~\ref{app:training-budget-test}.

Gap and fusion intervals use 2,000 shared whole-subject bootstrap draws, keeping repeated episodes from each subject together. These pointwise intervals are conditional on the fitted models: they do not account for retraining or selection uncertainty and are not adjusted for multiplicity. The 25 seed--fold fits are not 25 independent patient samples. To compare fusion with the stronger unimodal envelope, we select the better unimodal metric separately for each head and draw before aggregation. This envelope is a comparison rule, not a single deployable model.

\noindent\textbf{Runtime and reliability.}
We report training wall time, peak accelerator memory, and time per training sample for the precision, schedule, and workload. Failure summaries use the recorded nonfinite-loss, gradient, and prediction counters; their measurement scope is specified below.

\noindent\textbf{Companion efficacy, efficiency, and safety metrics.}
The companion figures in Appendices~\ref{app:ees-representation}, \ref{app:ees-temporal}, and~\ref{app:ees-information} report AUPRC, training cost, calibration, selective risk, and numerical failures.
AUPRC summarizes precision--recall performance and depends on outcome prevalence.
Observation-length comparisons report mean training time per optimizer step in milliseconds, computed for each fit as $1000\,t_{\mathrm{train}}/N_{\mathrm{updates}}$ before task and seed averaging.
The MC-MED records provide both quantities for all 60 fits at batch size 128; other observation-length timing entries remain unreported until their update counts are verified.
Other experiments retain training wall time per run on MIMIC-III and MC-MED, or per validation training cell on VitalDB.
MIMIC-III memory is peak accelerator memory as a percentage of device capacity; the refreshed MC-MED temporal figures instead report sampled shared-device memory in MiB, which does not isolate a model's memory use.
The plotted latency is $1000/\text{training throughput}$ in milliseconds per training sample, not inference or alert-delivery latency.
ECE uses 15 bins, and AURC integrates selective-prediction risk over coverage; lower values are preferable for both.
MIMIC-III and MC-MED failure rates use the available nonfinite-loss, gradient, or prediction records.
For the refreshed MC-MED temporal figures, the failure rate is failed training attempts divided by started attempts, using the campaign audit (zero of 10,112 attempts); the same audit records zero nonfinite events across 5,320,037 training steps.
For VitalDB, the audited rate of nonfinite or out-of-range prediction files among completed selected models is zero of 78,700 test files; all-attempt training-failure counters were not recorded.

Bars average available task means when conditions cover the same tasks; VitalDB forecast macros exclude concurrent VCS05.
Colored markers show individual tasks; grouped bars distinguish subject budgets from light (smallest) to dark (largest).
Hollow marks identify validation metrics, and filled cost bars show training telemetry.
Task markers replace macro bars when task coverage differs across conditions.
Panels show only recorded metrics and omit missing measurements.
The VitalDB sensor-selection figure pools the available two-, four-, and eight-channel results under each policy.
The source tables propagate task seed SDs as $\sqrt{\sum_t s_t^2}/K$ for $K$ tasks; this summary omits cross-task covariance and is not a patient-level confidence interval.
The per-step timing table instead averages the four task values within each seed and reports the mean and sample SD of the five seed-level macros.

\noindent\textbf{Selection and measurement scope.}
Checkpoint selection and search grids follow Appendix~\ref{app:architecture-tokenizer-protocol}.
VitalDB ECE and AURC come from the September 15 safety supplement, except fusion ECE, which retains the original fusion export used in the main text.
MC-MED representation and fusion safety metrics are recomputed from saved test predictions; earlier-study calibration estimates remain separately identified in Appendix~\ref{app:vitaldb-results}.
VitalDB training telemetry comes from earlier grids, whereas its predictive metrics come from the updated test release; cost comparisons are within device and workload, not across datasets.
The MC-MED representation companion retains the earlier campaign, including its additional FFT sweep; Table~\ref{tab:main_auroc} and the AUROC dot plots use the September 17 update.
Its AUPRC and reliability results therefore describe that earlier campaign, rather than the updated AUROC configurations.
The MIMIC-III companion and updated main AUROC table also use different Conv1D search summaries and some different tokenizer seed statistics.
The temporal companions now use the September 17 collection corresponding to Figure~\ref{fig:context-warning}: duration comparisons average three MIMIC-III and five VitalDB encoder--tokenizer pairs, and MC-MED uses the refreshed ResNet1D/Conv2D campaign.
MC-MED temporal ECE and AURC cover all four endpoints; duration figures report their macro, and gap figures show task-specific values.
Information-availability companions likewise retain earlier exports: their MIMIC-III missingness reference is architecture-only and their subset budgets are reported by subject, whereas the refreshed AUROC study uses zero-mask plus $\Delta$ and training-window quotas, respectively.
MC-MED fusion AUPRC, training time, and failure records cover all four endpoints; its ECE and AURC panels report decompensation.
MC-MED subject-budget results and the full-cohort AUPRC reference use five-seed validation means.

\newcommand{\eesfigure}[3]{%
\begin{figure*}[!t]
\centering
\includegraphics[width=\linewidth]{figures/ees_originals/#1.pdf}
\caption{#3}
\label{#2}
\end{figure*}
}

\subsection{Experiment Count}
\label{app:experiment-count}
We count one model fit for each task/head, experimental condition, hyperparameter setting, training seed, and subject fold.
The reported grids imply a conservative total of 87,825 fits (Table~\ref{tab:experiment-count}): 5,033 task/head-specific conditions, including learning-rate choices, before repetitions across seeds and folds.
MIMIC-III contributes eight task heads and five seeds, MC-MED four heads and five seeds, and the VitalDB core grid 15 heads and five seeds across five subject folds.
Thus, a condition evaluated across a dataset's core heads contributes 40, 20, or 375 fits, respectively.
We include the specified learning-rate searches and count overlapping search conditions and shared reference models only once.
Where the search grid is unspecified, we count one learning-rate setting. We exclude additional unreported tuning trials, restarts, bootstrap draws, and inference-only controls.


MIMIC-III and VitalDB each contain $9\times3+4=31$ encoder--tokenizer configurations, including the four Transformer tokenizers outside their architecture grids.
MC-MED contains $9\times3+6+8=41$ configurations: 27 native architecture-grid pairs, six additional Transformer configurations, and eight additional non-Transformer FFT pairs.
Separately trained configurations within the same tokenizer family count as separate fits.
For MIMIC-III, the architecture search contains $9\times3\times3=81$ pair--learning-rate conditions, and the tokenizer search contains $6\times5+3=33$.
The two searches share eight conditions: two Conv1D learning rates and three each for Crossformer and SeFT with a Transformer.
Their union gives $(81+33-8)\times8\times5=4,240$ fits.
VitalDB searches three rates for each of its 31 pairs, giving $31\times3\times15\times5\times5=34,875$ fits.
MC-MED also searches three learning rates for each of its 41 configurations, giving $41\times3\times4\times5=2,460$ representation fits.

The history count retains the original inventory of MIMIC-III fits at 10, 20, and 60 minutes, two additional MC-MED durations, and three additional durations for each of five VitalDB pairs (Appendices~\ref{app:temporal-comparisons} and~\ref{app:vitaldb-preprocessing}). The refreshed MIMIC-III curve excludes the earlier five-minute point, which came from a separate campaign.
Only the 700 explicitly new VitalDB longer-gap fits are added; the 350 reused fits and the MIMIC-III and MC-MED gap extensions are excluded from this lower bound (Appendix~\ref{app:vitaldb-gaps}).

The documented MIMIC-III and MC-MED grids contribute six channel-selection alternatives and nine missing-input methods per task.
The workload tally retains these original MC-MED grids, although their result panels are omitted because the current input configuration has only two available waveform channels.
The fuller VitalDB grids contribute $48-3=45$ channel conditions and $30-3=27$ missingness conditions after removing the three fixed-pair references (Appendix~\ref{app:vitaldb-preprocessing}); their subsets in the main figure are not counted again.
Subject selection contributes seven methods at three fractions per dataset, excluding the full-data reference.
The refreshed eight-selector grids and added MIMIC-III EHR-only control are not added to this conservative fit inventory.
The counted fusion grids contain nine MIMIC-III methods and five fusion methods plus EHR only on VitalDB and MC-MED; waveform-only references are not added again.
This inventory is broader than the current MC-MED fusion display, which contains four fusion methods plus EHR only (Table~\ref{tab:information-comparison-grids}); transfer-to-MIMIC-IV fits are not added to this lower bound.

\section{Dataset and Task Details}
\label{app:dataset-details}

\subsection{Episode Construction and Data-Control Details}
\label{app:episode-data-controls}

\noindent\textbf{Anchors and temporal windows.}
For each task, we generate candidate decision anchors \(T\), then construct
the observation interval \([T-W,T)\), blank forecast gap \(G\),
and outcome horizon \(H\). MIMIC-III combines sliding-surveillance, daily, and
event anchors; VitalDB separates future-event anchors from the concurrent BIS
probe; and MC-MED applies the task-specific eligibility rules and target windows described in
Appendix~\ref{app:mcmed-details}. Observation intervals are fixed independently of future labels to prevent post-cutoff information from entering the inputs.
For MIMIC-III, we extract the six-hour history \([T-6\,\mathrm{h},T)\) and construct each input as its suffix \([T-W,T)\): five minutes uses \([T-5\,\mathrm{min},T)\), and one hour uses \([T-1\,\mathrm{h},T)\).
Observation-length comparisons retain the same anchors and labels across these nested inputs.

\noindent\textbf{Eligibility and ascertainment.}
Eligibility rules determine whether an anchor can be assigned an interpretable label.
For incident-event tasks, a prevalent event before \(T\) excludes the episode,
an event during \(G\) is excluded or masked according to the task contract,
and only a qualifying event during \(H\) receives a positive label. A negative label requires
sufficient follow-up and target availability to establish that the event did
not occur; missing, stale, or incompletely ascertained outcomes remain masked.
Response tasks instead anchor on a documented action and apply prespecified
baseline, response, and non-overlap rules. Repeated or overlapping episodes
remain valid longitudinal observations but are not treated as independent
patients in aggregation or uncertainty estimation.

\noindent\textbf{Splits and missingness.}
We assign subjects to partitions before constructing model inputs.
Repeated surgical cases from the same VitalDB subject remain in the same
partition, and no subject contributes waveform fragments or prediction anchors
to more than one split. Patient split and fold assignments remain fixed across observation lengths and warning gaps.
Raw waveform samples enter the model tokenizer with channel-presence and sample-validity masks;
missingness is recorded explicitly and is not converted into a
negative observation. We fit normalization statistics, vocabularies, and
learned quality filters on the training partition only.

\noindent\textbf{Leakage controls and auditability.}
Temporal and feature-level controls limit inputs to information available at the decision time. Every
EHR value must be available before the applicable input cutoff, and waveform
samples at or after that cutoff are inaccessible to the model. Outcome-derived
columns, post-anchor results and interventions, final disposition and
diagnoses, realized length of stay, label and censoring fields, and direct or
derived target channels are excluded. Documented pre-anchor treatment or
support is retained when it would have been known at the decision time,
although an already active intervention can make an episode ineligible for an
incident-event task. Each released registry records the dataset version, subject split,
anchor, input and label bounds, channel mask, and preprocessing hash so that
the temporal audit can be reproduced without opening held-out outcomes.

\subsection{Short Task Labels and Full Clinical Names}
\label{app:task-name-key}
The main result tables and figures use the short labels listed in Table~\ref{tab:task-name-key}.
Aliases identify the same endpoint; the dataset-specific definitions in Appendices~\ref{app:mimic-details}, \ref{app:vitaldb-details}, and~\ref{app:mcmed-details} specify its eligibility, timing, and outcome rules.
Macro (8), Macro (7), and Macro (4) denote dataset summaries, not additional tasks; VitalDB's forecast macro excludes VCS05.
The separate five-target MIMIC-IV transfer labels and full names appear in Table~\ref{tab:mimiciv-transfer-cohorts}; their temporal contracts are specific to the adaptation study in Appendix~\ref{app:information-transfer}.
Table~\ref{tab:task-name-key} lists the full names.

\subsection{MIMIC-III Details}
\label{app:mimic-details}
The MIMIC-III Waveform Database Matched Subset contains 10,282 adult ICU patients under the frozen 80/10/10 patient-level split.
Native waveforms are sampled at 125\,Hz and may include ECG, arterial blood pressure, plethysmography, and respiration.
Availability varies across channels and time; channels need not be simultaneous, and the benchmark retains presence and validity masks.
Each ID below counts as one benchmark task.
For tasks with multiple heads, we declare one primary head before evaluation for the main comparison and report all heads in the supplement.

\ding{182}\textbf{Acute deterioration / failure to rescue} predicts near-term deterioration from the preceding six hours. The primary surveillance task uses a zero-gap six-hour horizon, with a three-hour-gap sensitivity analysis for rescue preparation.
\ding{183}\textbf{Dynamic in-hospital mortality} predicts death within 24 hours after a six-hour forecast gap from a six-hour waveform history, serving as a severity and representation benchmark rather than a stand-alone paging rule.
\ding{184}\textbf{Respiratory deterioration and escalation} predicts impending airway or oxygen-support escalation over six hours, with two- and three-hour sensitivities testing whether useful warning persists earlier in the course.
\ding{185}\textbf{Circulatory deterioration and shock} uses the primary \texttt{y\_new\_pressor} head to predict a new vasopressor start within six hours, with zero gap, among patients off vasopressors at the anchor.
Escalation and 24-hour recovery are auxiliary heads from the broader task registry and are excluded from the reported primary-head comparisons.
\ding{186}\textbf{Rapid documented-fluid response} uses the six hours before a fluid event to predict a documented MAP improvement 30--60 minutes afterward. We interpret this endpoint as a hemodynamic-response proxy, not true fluid responsiveness.
\ding{187}\textbf{Incident KDIGO stage 2/3 AKI} predicts new moderate-to-severe kidney injury over the subsequent 23 hours after a one-hour gap, supporting earlier renal-risk monitoring among patients not already meeting the endpoint.
\ding{188}\textbf{Ventilator course and liberation} uses daily and event-anchored predictions to characterize readiness for weaning or extubation and the subsequent need for prolonged ventilation or tracheostomy.
\ding{189}\textbf{Sepsis-3 onset and shock trajectory} predicts sepsis-related organ dysfunction over the next five hours after a one-hour gap and tracks progression toward shock under an explicit suspected-infection rule.

\subsection{VitalDB Details}
\label{app:vitaldb-details}
The waveform-only VitalDB benchmark uses every official waveform stream present in a case and permitted by the task's input rules, drawn from the twelve-stream registry in Table~\ref{tab:vitaldb-channels}.
In waveform-only A--D, missingness, and gap experiments, numeric and laboratory tracks define eligibility and outcomes but are not model inputs. Fusion experiments also use EHR predictors, subject to task-specific pre-cutoff availability and target-source exclusions.
Source sampling rates range from 62.5 to 500\,Hz.
For each episode, the hour ending at the task-specific legal input cutoff is divided into twelve consecutive five-minute fragments.
The model tokenizer receives raw waveform samples from the selected input interval at the sampling rates recorded in the source headers.
The twelve native streams are placed into a fixed 66-slot canonical tensor layout; the remaining slots are structural placeholders rather than additional observed channels.
Unrecorded streams and invalid samples remain NaN and are accompanied by explicit channel-presence and sample-validity masks.
We do not require ECG-II, interpolate across absent streams, or apply complete-case filtering.
Channel-wise median and interquartile-range normalization is fitted independently within the four outer-training folds and clipped to \([-5,5]\).
For models that cannot consume NaNs directly, missing normalized values are replaced by neutral zero only after the observation mask has been constructed, and that mask is provided to the model as a separate input.

VitalDB contributes seven future-event tasks and one concurrent held-out-state probe.
The core future-event panel uses waveform histories ending strictly before the prediction anchor, gaps of 5 or 15 minutes, and task-specific target horizons of 15 or 60 minutes. The separate 30-minute-input gap extension adds 30, 45, 60, and 75 minutes (Appendix~\ref{app:vitaldb-gaps}).
Direct target channels, target-derived features, and post-cutoff samples are excluded.
The concurrent task is analyzed separately and is not pooled with the future-event forecasts.
Missing, stale, or incompletely ascertained outcomes are masked rather than treated as negative.

\ding{182}\textbf{Sustained intraoperative hypotension} predicts a new episode in which invasive MAP remains below 65\,mmHg for at least two continuous minutes after five stable pre-anchor minutes at or above 65\,mmHg.
\ding{183}\textbf{Accelerating MAP decline} predicts an accelerating ten-minute decrease in invasive MAP after a stable pre-anchor period. It captures trajectory shape before a conventional hypotension threshold is necessarily crossed.
\ding{184}\textbf{Incident oxygen desaturation} predicts a sustained \(\mathrm{SpO_2}<92\%\) episode after five stable minutes at or above 94\%, with oxygen-delivery, target-quality, and future-oximetry information excluded from the predictors.
\ding{185}\textbf{Incident core hypothermia} predicts a newly observed core-temperature episode below \(35.5^{\circ}\mathrm{C}\) lasting at least five minutes, subject to explicit temperature freshness and coverage requirements.
\ding{186}\textbf{Abrupt \(\mathrm{ETCO_2}\) fall} predicts a decrease of at least 10\,mmHg from the recent \(\mathrm{ETCO_2}\) baseline that develops within one minute and persists long enough to distinguish it from an isolated sample.
\ding{187}\textbf{\(\mathrm{PCO_2}\)--\(\mathrm{ETCO_2}\) discordance} predicts a newly measured difference of at least 10\,mmHg between blood-gas \(\mathrm{PCO_2}\) and temporally matched \(\mathrm{ETCO_2}\) over a 60-minute target horizon; the label is masked when no qualifying future blood-gas measurement is available.
\ding{188}\textbf{Coupled hemodynamic response} predicts a joint increase of at least 20 beats/min in heart rate and 20\,mmHg in invasive MAP, with the two responses beginning within three minutes and overlapping for at least one minute.
This endpoint represents a coupled hemodynamic pattern and is not interpreted as proof of pain, awareness, or inadequate anesthesia.
\ding{189}\textbf{Concurrent light BIS state} evaluates whether non-BIS physiology identifies a simultaneous, quality-valid BIS value above 60 lasting at least two minutes.
The five-minute label interval is concurrent with the most recent five minutes of the waveform input; longer input views add only earlier physiological context.
EEG1, EEG2, numeric BIS, BIS SQI, and other label-defining BIS inputs are withheld from every model and selector view.
Accordingly, this task is a held-out-state representation probe rather than a forecast and is analyzed separately from the seven future-event tasks.


\subsection{MC-MED Details}
\label{app:mcmed-details}
MC-MED contains 70,545 adult ED patients across 118,385 visits and uses the supplied 78/11/11 chronological, patient-disjoint split; completed representation-comparison test results are reported in Figure~\ref{fig:representation-results}.
Waveform sampling rates are channel-dependent and recorded in each WFDB header.
At least one waveform is available in 83,623 visits (70.6\%), while monitoring interruptions produce segment gaps.
The source provides ECG lead II, plethysmography, and respiration; the native packs used for the reported MC-MED comparisons contain ECG-II and respiration, with channel presence and validity represented explicitly.

The selected MC-MED panel contains four binary prediction tasks, defined relative to prediction time \(T\):
\begin{itemize}
\item \textbf{Physiological decompensation} (\texttt{decompensation\_60m}): predict whether an initially stable patient crosses a heart-rate (HR), mean-arterial-pressure (MAP), or oxygen-saturation (\(\mathrm{SpO_2}\)) deterioration threshold within 60 minutes after \(T\).
\item \textbf{Pending hyperkalemia} (\texttt{pending\_hyperkalemia}): predict whether the first result of a potassium test ordered before \(T\) is \(\geq 5.5\)\,mmol/L during the next 240 minutes.
\item \textbf{Lactate nonclearance} (\texttt{lactate\_nonclearance}): after an initial lactate \(\geq 2\)\,mmol/L, predict whether the minimum repeat lactate measured 30--240 minutes after \(T\) remains \(\geq 2\)\,mmol/L and \(>80\%\) of baseline.
\item \textbf{Pending severe anemia} (\texttt{pending\_severe\_anemia}): predict whether the first result of a hemoglobin test ordered before \(T\) is \(<8\)\,g/dL during the next 240 minutes.
\end{itemize}
Decompensation targets an acute safety-boundary transition, while the laboratory endpoints target the presence or persistence of a pathophysiologic abnormality.
The pending-test tasks condition on an order placed before the prediction time; their future laboratory results define outcomes and are excluded from model inputs.
Lactate nonclearance requires an eligible baseline and repeat measurement; missing target measurements remain unascertained rather than negative.
Table~\ref{tab:cohorts} reports task-specific patients, ED visits, windows, and positive windows across training, validation, and test.

\subsection{Task-level Cohort Support}
\label{app:task-support}

Table~\ref{tab:cohorts} reports the supplied cohort counts for all 20 tasks and all reported development and test partitions.
Clinical episodes are ICU stays for MIMIC-III, surgical cases for VitalDB, and ED visits for MC-MED.
Within each partition, the union rows count each patient and clinical episode once across tasks. Window counts refer to task-labelled prediction examples.
One stay or visit can contribute windows to several tasks; MC-MED has one window per visit \emph{per task}, rather than one across the whole task panel.
Positive-window totals also count task-labelled outcomes, not unique patients with positive outcomes.

Across its patient-disjoint train, validation, and test partitions, MIMIC-III contains 8,535 unique patients, 9,772 ICU stays, and 979,920 task windows.
VitalDB contains 5,346 unique patients, 5,575 surgical cases, and 385,667 task windows across development and test; its five-fold development cohort is counted once.
MC-MED contains 15,852 unique patients, 18,536 ED visits, and 34,379 task windows across all partitions.
Together, the benchmark includes 29,733 patients, 33,883 clinical episodes, and 1,399,966 task-labelled windows.
Patients and episodes are deduplicated across tasks, while windows count each task and forecast-gap head separately without multiplication by model, fold, seed, or observation-length setting.


\subsection{Architecture--Tokenizer Evaluation Protocol}
\label{app:architecture-tokenizer-protocol}

\noindent\textbf{Common protocol.}
The representation comparisons use waveform-only inputs on all three datasets. Unavailable channels are encoded as NaN with a binary mask; we do not apply complete-case filtering. Each cell of Table~\ref{tab:main_auroc} gives the mean and sample SD of test AUROC over five seeds (0--4). For macro rows, we first average tasks within each seed, then summarize the five seed-level macros.

For each candidate configuration, validation performance selects the checkpoint epoch; it also selects hyperparameters, including learning rate and tokenizer where applicable.
The selected checkpoints are then evaluated on the held-out test set to obtain the reported performance.
The dataset-specific search grids are listed below.
Bold and underlining indicate the highest and next-highest unrounded means, respectively; distinct values can round to the same displayed score.

\noindent\textbf{Controlled comparisons and input views.}
Within each comparison, models share the task-specific episodes, patient partitions, permitted channels, and missingness representation.
The MIMIC-III representation input is the five-minute suffix $[T-5\,\mathrm{min},T)$ of the extracted six-hour history; VitalDB uses five-minute inputs ending at its legal cutoff.
The selected tokenizer converts raw waveform samples into the token sequence processed by the encoder.

\noindent\textbf{Part (a): architectures.}
Nine waveform encoders are compared at a matched budget of approximately 500k trainable parameters ($499{,}915--500{,}091$): Transformer, BiLSTM, Mamba-3, InceptionTime, TCN, ResNet1D, S4, FITS and DLinear.
MIMIC-III and VitalDB compare Conv1D, Crossformer, and SeFT under their dataset-specific aggregation rules. MC-MED uses Conv1D across its nine displayed encoders.
\noindent\textbf{Part (b): tokenizers.}
The tokenizer study fixes a Transformer backbone and varies the input tokenizer. Conv1D projects local waveform patches into tokens. Macro means and SDs are recomputed from the selected per-seed task scores. SeFT embeds each observed (time, channel, value) triple as a set element. Crossformer uses dimension-segment-wise embedding with cross-channel attention, and STraTS embeds observation triplets with continuous value embeddings. iTransformer treats each channel as one token; FFT tokenizes each patch's frequency spectrum; mTAND applies multi-time attention over irregular sampling. MC-MED additionally reports a Conv2D tokenizer.

\noindent\textbf{MIMIC-III.}
Eight ICU tasks are derived from the MIMIC-III clinical records and paired with the matched-subset bedside waveforms. Each task extracts a six-hour history ending at the anchor $T$: acute deterioration (major rescue event within $[T, T{+}6\,\mathrm{h})$), dynamic in-hospital mortality (six-hour gap, then death within the following 24\,h), respiratory escalation (new HFNC/NIV/IMV or an upward respiratory-support transition within 6\,h), circulatory deterioration (new vasopressor start within 6\,h; primary head \texttt{y\_new\_pressor}), fluid response (hemodynamic improvement 30--60\,min after a documented rapid crystalloid course; event-anchored), AKI KDIGO stage~2/3 (incident injury within $[T{+}1\,\mathrm{h}, T{+}24\,\mathrm{h})$), ventilator liberation (readiness and durable liberation over 24\,h at daily ventilator checkpoints) and Sepsis-3 (infection-associated organ dysfunction within $[T{+}1\,\mathrm{h}, T{+}6\,\mathrm{h})$).
The representation baseline uses raw samples in $[T-5\,\mathrm{min},T)$ on a 62-channel canonical layout at 125\,Hz, giving 37,500 samples per channel before tokenization. Longer observation views use the corresponding suffix $[T-W,T)$ of the same extracted history. Splits are patient-level with a single held-out test set (854--81{,}359 test windows per task).
Models are trained for 10 epochs with batch size 128, bf16 mixed precision, dropout 0.1 and weight decay $10^{-4}$. The learning-rate grid is $\{10^{-3}, 10^{-4}, 10^{-5}\}$ in part~(a) and $\{10^{-2}, 10^{-3}, 10^{-4}, 10^{-5}, 5{\times}10^{-6}\}$ in part~(b) (the Conv1D baseline searched $\{10^{-2}, 10^{-3}, 10^{-4}\}$). A learning-rate group is eligible only if all five seeds finished without non-finite gradient steps.

\noindent\textbf{VitalDB.}
The VitalDB Core-8 study uses intra-operative waveforms. All 12 available native waveform channels are kept in a canonical 66-slot geometry, and robust normalisation is fitted on the training folds only. Seven forecast tasks (OX01-IOH, CL07-002, H5-002, H5-030, L2-30, VA1-CO2-DROP, VA1-STIM-RESPONSE) are each evaluated at forecast gaps of 5 and 15\,min; the two gap heads are averaged with equal weight into one value per task. The concurrent-state task VCS05 (one head, EEG slots excluded) is reported separately and is not part of the forecast macro.
Training uses five seeds $\times$ five outer whole-subject folds (4{,}596 validation subjects).
In part~(a), the selected tokenizer is Conv1D for all nine encoders. The average test AUROC is got from the five fitted-fold models within each seed on the fixed cohort of 750 test subjects, and the mean and sample SD are reported across five seeds.
The learning rate is selected per architecture/tokenizer condition from $\{10^{-4}, 3{\times}10^{-4}, 10^{-3}\}$. In part~(b), Conv1D, SeFT, and Crossformer use the Transformer conditions of the architecture grid; the reported seven-tokenizer comparison also includes STraTS, iTransformer, FFT, and mTAND (Figure~\ref{fig:representation-results} and Table~\ref{tab:main_auroc}).

\noindent\textbf{MC-MED.}
MC-MED uses the four selected tasks in Appendix~\ref{app:mcmed-details}: physiological decompensation, pending hyperkalemia, lactate nonclearance, and pending severe anemia. It compares nine architectures at the 500k-parameter budget, eight displayed Transformer tokenizers, five seeds, and an equal-task macro over the four endpoints.
Table~\ref{tab:main_auroc} records the updated per-method means and seed SDs. Architecture and tokenizer panels select configurations separately, so their Transformer and Conv1D entries can differ. The rounded interaction annotations from Figure~\ref{fig:representation-results}(c) are recorded in Table~\ref{tab:representation-test-interactions}.

\subsection{Method Attribution and Benchmark Adaptations}
\label{app:method-adaptations}
The citations at first mention in Sections~\ref{sec:representation-choices-draft} and~\ref{sec:information-availability-draft} identify the methodological sources of the displayed configurations.
The benchmark adapts these methods to its waveform inputs, task heads, and parameter budgets. A shared method name does not imply use of the original data, pretraining, ensembling, or optimization schedules.

\noindent\textbf{Encoders and tokenizers.}
The nine encoder families retain their cited architectural basis while using the benchmark's classification heads and matched parameter target.
The Conv1D projection is a local shared baseline inspired by PatchTST's patching design; this attribution does not assign the origin of generic Conv1D tokenization to PatchTST or imply reproduction of its channel-independent architecture.
SeFT, Crossformer, STraTS, iTransformer, and mTAND supply representation mechanisms used with the common Transformer, so their tokenizer results do not evaluate the complete original systems.
The FFT tokenizer applies a Fourier transform and learned spectral projection.
FreTS~\citep{yi2023frets} provides a reference for neural learning after frequency-domain conversion; our local tokenizer feeds the shared encoder and does not reproduce the complete FreTS model.
FITS is a separate encoder family.

\noindent\textbf{Missing-input handling.}
Zero-mask plus $\Delta$ supplies absence and elapsed-time features without reconstructing channels, using the missingness representation formalized in GRU-D~\citep{che2018missing}.
GRIN and Raindrop motivate graph-based sharing across observed channels, while the positional-kriging variant applies the IGNNK idea to a fixed sensor layout~\citep{wu2021ignnk}.
Lead-agnostic SSL uses the random-lead-masking principle of \citet{oh2022leadagnostic}.
Rectified Flow, TrigFlow, and Drifting identify generative objectives adapted for conditional missing-input handling; the cited generative papers do not themselves establish clinical waveform-imputation performance.

\noindent\textbf{Training-subset policies.}
Diversity uses greedy coverage in a patient-representation space; diversity clustering combines manifold projection with proportional coverage within clusters~\citep{sener2018coreset,mcinnes2018umap}.
Gradient matching adapts coreset and matching-pursuit ideas to patient-level proxy gradients~\citep{mirzasoleiman2020craig,killamsetty2021gradmatch}.
Training-dynamics selection combines confidence and variability with an early error score~\citep{swayamdipta2020cartography,paul2021datadiet}.
The NLDR policy draws inspiration from label-free arrhythmia structure in nonlinear embeddings~\citep{vazifeh2026nldr}; allocation across rare patient clusters is a benchmark adaptation, not a subset-selection algorithm evaluated in that paper.
Noise filtering and task anchors are local controls that prioritize recorded signal quality and physiological transitions/hard negatives, respectively.
The current MIMIC-III comparison applies training-window quotas, while VitalDB and MC-MED use subject quotas (Appendix~\ref{app:information-subjects}).

\noindent\textbf{Channel-selection policies.}
MI and Lasso use published feature-selection principles~\citep{peng2005mi,tibshirani1996lasso}; temporal MI is a benchmark variation of the information-based score.
SQI applies quality-based channel ranking, motivated by work on ECG signal acceptability~\citep{clifford2012sqi}, without claiming to reproduce that paper's clinical-quality classifier across every waveform type.
Random selection and clinical-priority channel lists are benchmark controls.

\section{Detailed Representation Results}
\label{app:representation-results}

\subsection{Task-level and Macro Test AUROC}
\label{app:representation-test-results}

Table~\ref{tab:main_auroc} reports the detailed MIMIC-III, VitalDB, and MC-MED architecture and tokenizer results summarized in Figure~\ref{fig:representation-results}(a,b).
The MIMIC-III macro averages eight tasks; VitalDB first weights the two forecast gaps equally within each task and then averages seven forecasting tasks, with VCS05 reported separately.
MC-MED averages its four endpoints with equal weight, retaining the supplied macro means and seed SDs.
Selection and scoring rules are specified in Appendix~\ref{app:architecture-tokenizer-protocol}.

\noindent\textbf{Tokenizer comparisons.}
The tokenizers cover local waveform projections, observation sets or triplets, channel-wise segments, channel tokens, spectral tokens, and multi-time attention; their definitions and search grids appear in Appendix~\ref{app:architecture-tokenizer-protocol}.
Conv1D is the only tokenizer in the macro-AUROC top three on all datasets: it rounds to the same MIMIC-III macro as Crossformer (0.611) and leads VitalDB (0.631) and MC-MED (0.574).
SeFT's MIMIC-III lead (0.615) depends on its Sepsis-3 gain of 0.038 over Conv1D, computed before rounding; the other seven task differences sum to $-0.010$, and their means favor Conv1D (0.6085 versus 0.6070).
SeFT accommodates irregular observations through timestamped-set aggregation~\citep{horn2020seft}, but lowers fluid-response AUROC from 0.586 to 0.574.
On VitalDB, STraTS and FFT approach Conv1D at 0.627 and 0.625, whereas Crossformer falls to 0.583.
On MC-MED, Conv1D and Crossformer lead the macro (0.574 and 0.570), whereas FFT falls to 0.509; Conv1D and Conv2D are nearly tied for decompensation (0.647) and hyperkalemia (0.576), but Crossformer leads anemia (0.582).
Conv1D is a strong shared starting tokenizer, with SeFT retained for MIMIC-III sepsis and Crossformer for MC-MED anemia.

\noindent\textbf{Architecture comparisons.}
Table~\ref{tab:main_auroc}(a) compares selected encoder--tokenizer configurations at the matched parameter budget under the selection rules in Appendix~\ref{app:architecture-tokenizer-protocol}.
Transformer has the highest macro AUROC on MIMIC-III (0.607) and VitalDB (0.631); on MC-MED, ResNet1D (0.568) is closely followed by DLinear, Transformer, and InceptionTime (each 0.567 at the displayed precision).
BiLSTM improves MIMIC-III mortality over Transformer (0.558 versus 0.532) but scores lower on AKI (0.696 versus 0.727).
VitalDB also has endpoint-specific alternatives: DLinear exceeds Transformer on hypothermia (0.608 versus 0.605), and BiLSTM leads abrupt ETCO$_2$ fall (0.683 versus 0.675).
DLinear leads MC-MED anemia (0.574), extending evidence that linear temporal projections can compete with attention models~\citep{zeng2023transformers} to this endpoint; BiLSTM instead leads decompensation (0.649) and hyperkalemia (0.580).
Transformer leads lactate nonclearance (0.508 versus ResNet1D's 0.499), so ResNet1D's small macro lead comes from its gains on hyperkalemia and anemia.
Lactate AUROCs remain near chance across architectures (0.453--0.508), consistent with a five-minute waveform capturing only part of the production, clearance, and treatment history that shapes lactate kinetics over hours~\citep{vincent2016lactate}.
Transformer is a useful initial ICU and operating-room encoder; MC-MED favors a task-specific shortlist including BiLSTM, DLinear, Transformer, and ResNet1D.



\subsection{Task-level AUROC Dot Plots}
\label{app:representation-dotplots}
\begin{figure*}[!t]
\centering
\includegraphics[width=\linewidth]{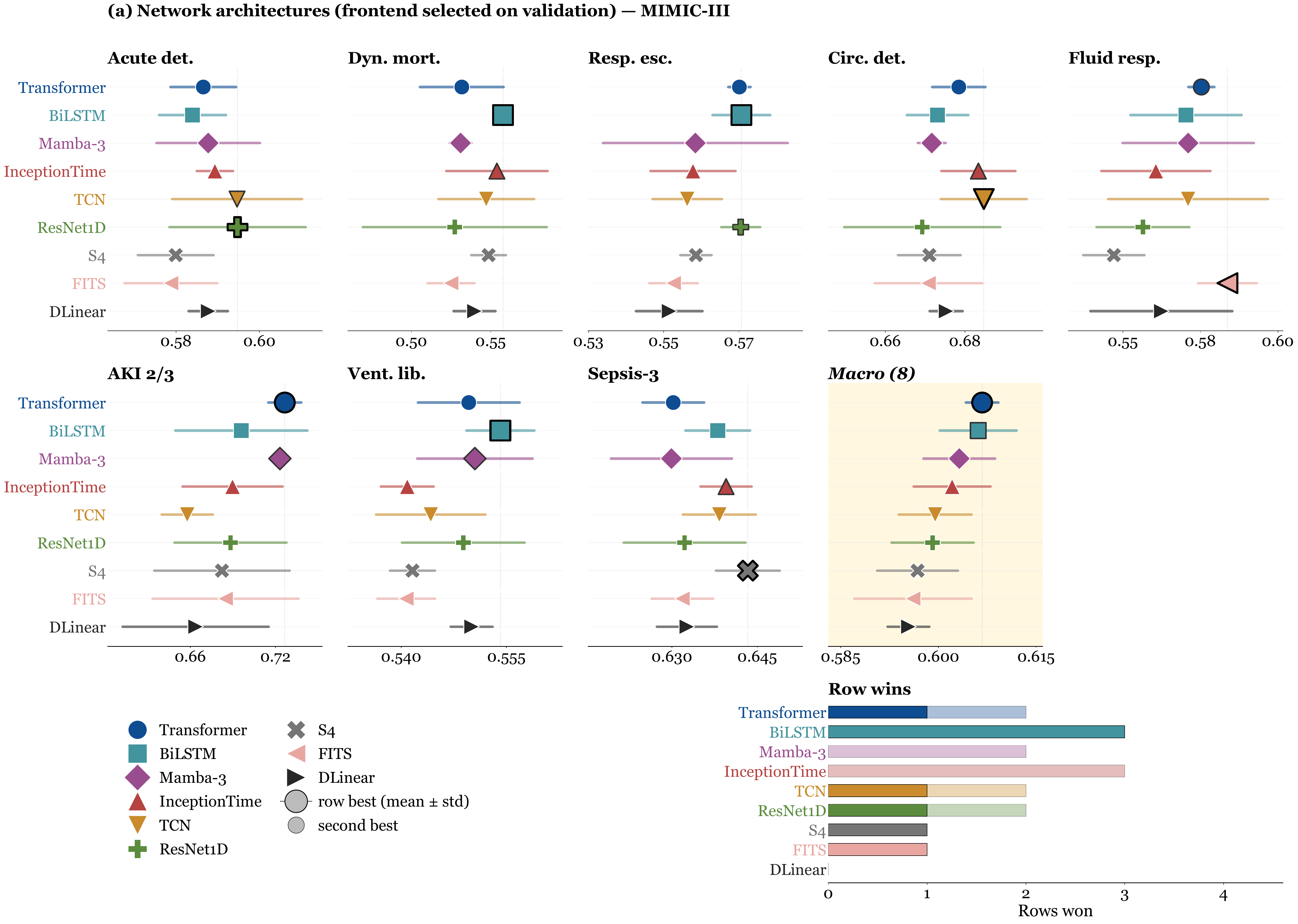}
\caption{\textbf{Architecture AUROC by task --- MIMIC-III.} Nine approximately 500k-parameter encoders with selected tokenizers across eight ICU tasks and their macro. Marker and SD-bar definitions follow Appendix~\ref{app:representation-dotplots}. Numerical values are in Table~\ref{tab:main_auroc}.}
\label{fig:auroc-architectures-mimic}
\end{figure*}

\begin{figure*}[!t]
\centering
\includegraphics[width=\linewidth]{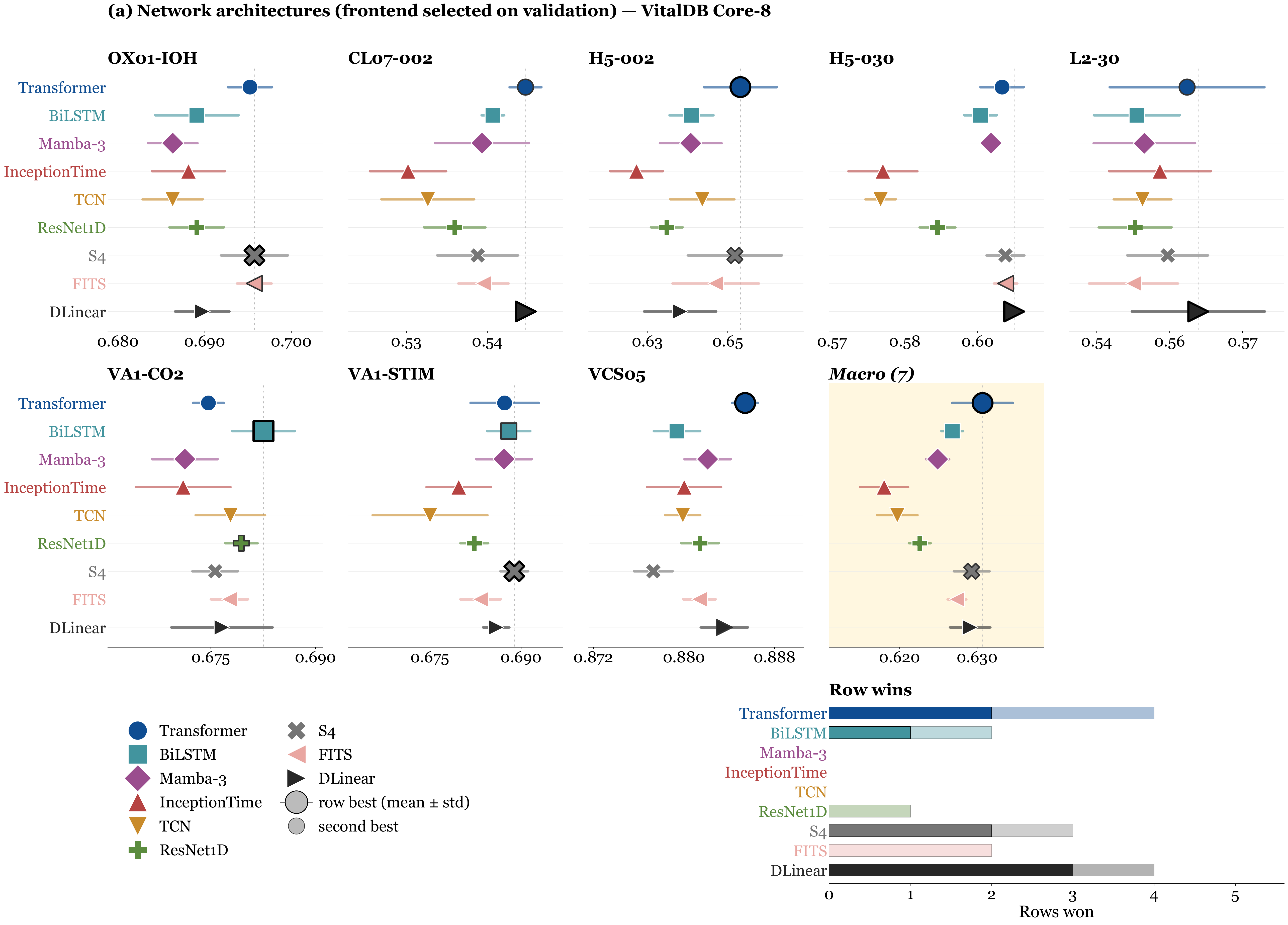}
\caption{\textbf{Architecture AUROC by task --- VitalDB Core-8.} Updated test results for nine encoders with selected tokenizers. Each forecast task gives equal weight to its two gaps; the seven-task macro excludes the concurrent VCS05 probe. Scores average fitted-fold model metrics within each seed, without probability ensembling. Numerical values are in Table~\ref{tab:main_auroc}.}
\label{fig:auroc-architectures-vitaldb}
\end{figure*}

\begin{figure*}[!t]
\centering
\includegraphics[width=\linewidth]{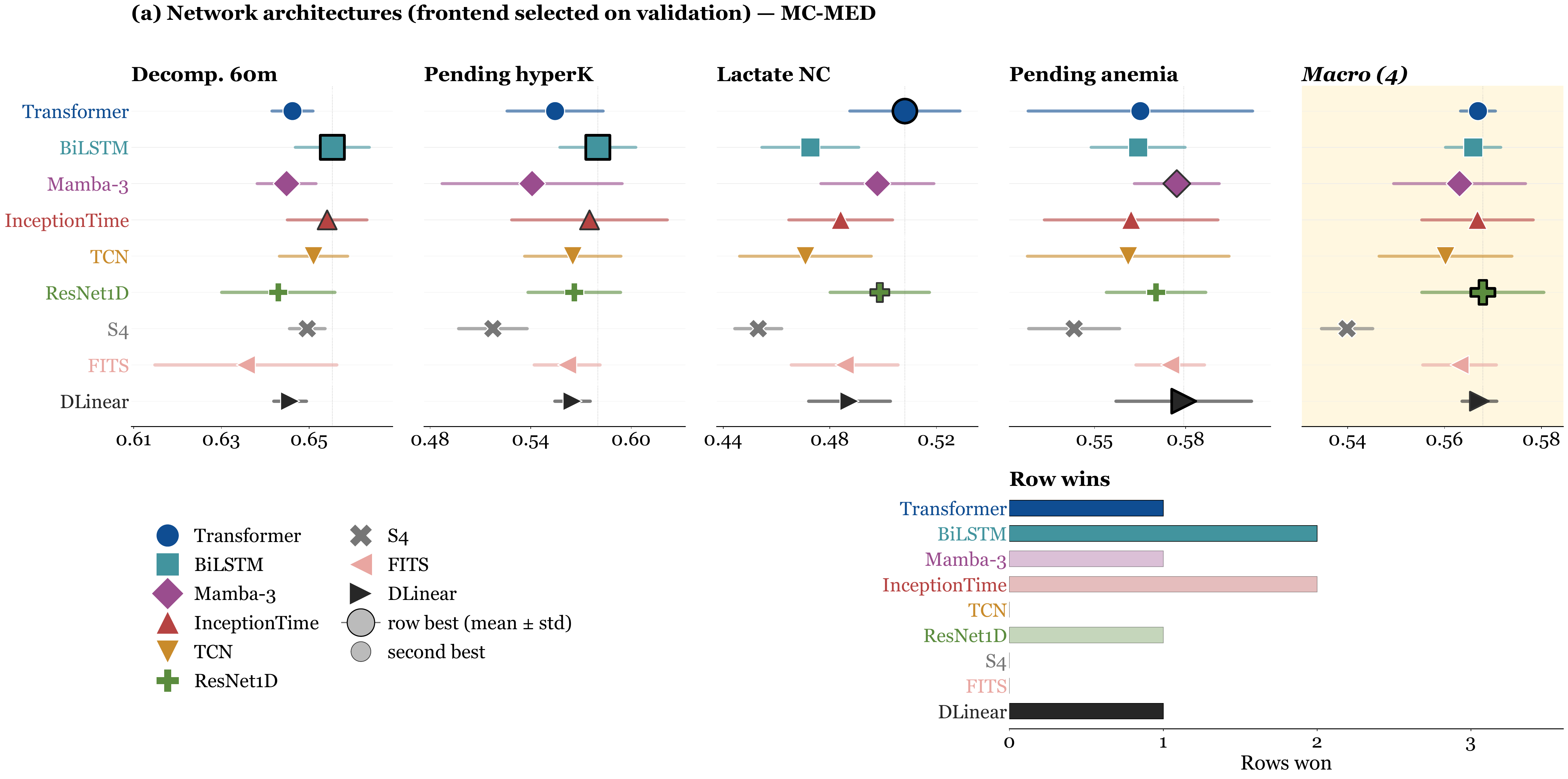}
\caption{\textbf{Architecture AUROC by task --- MC-MED.} Nine encoders with selected tokenizers on four ED endpoints and their macro, using the input and search grids in Appendix~\ref{app:architecture-tokenizer-protocol}. Numerical values are in Table~\ref{tab:main_auroc}.}
\label{fig:auroc-architectures-mcmed}
\end{figure*}

\begin{figure*}[!t]
\centering
\includegraphics[width=\linewidth]{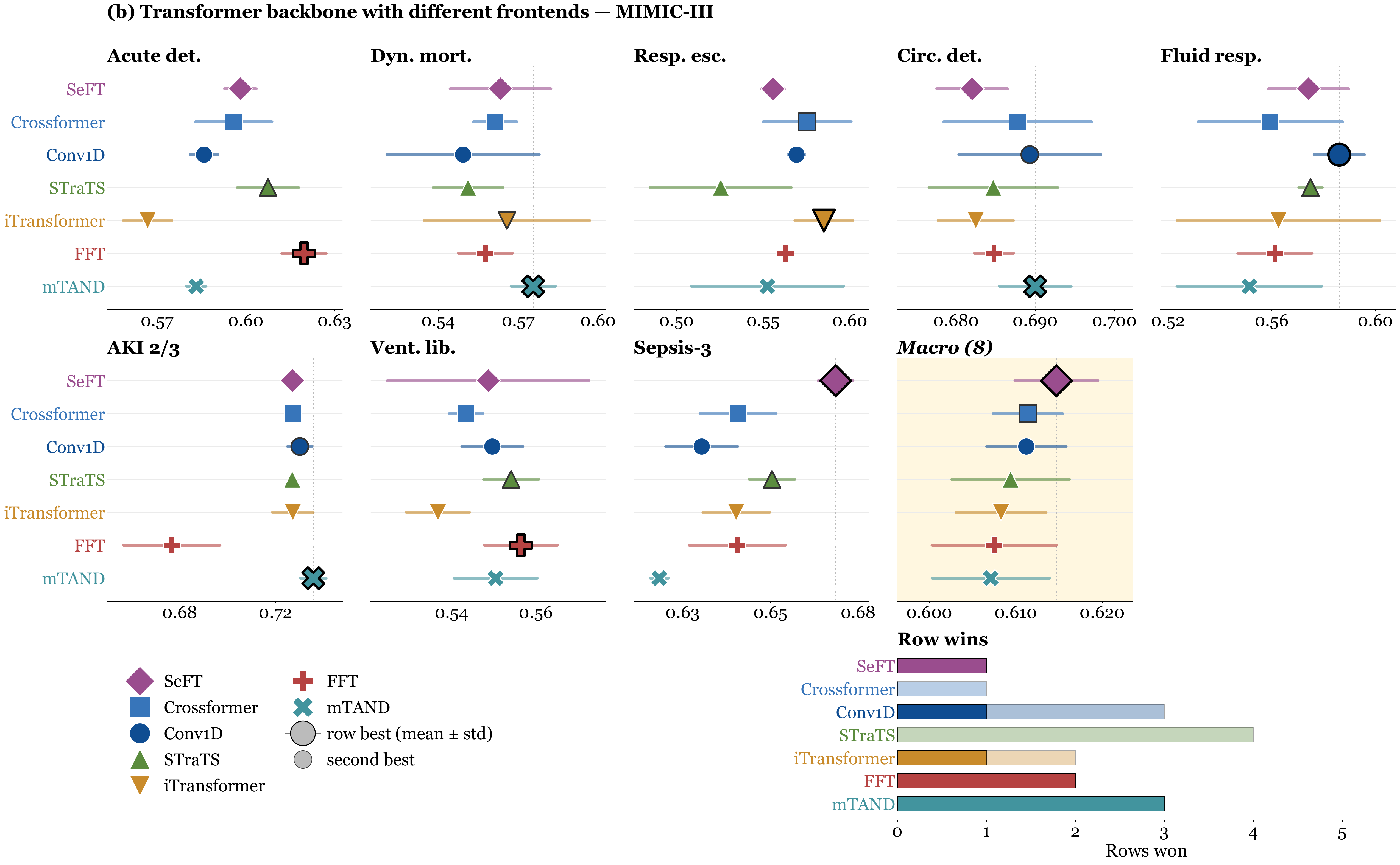}
\caption{\textbf{Transformer tokenizer AUROC by task --- MIMIC-III.} Seven tokenizers across eight ICU tasks; SeFT has the highest macro AUROC. Numerical values are in Table~\ref{tab:main_auroc}.}
\label{fig:auroc-tokenizers-mimic}
\end{figure*}

\begin{figure*}[!t]
\centering
\includegraphics[width=\linewidth]{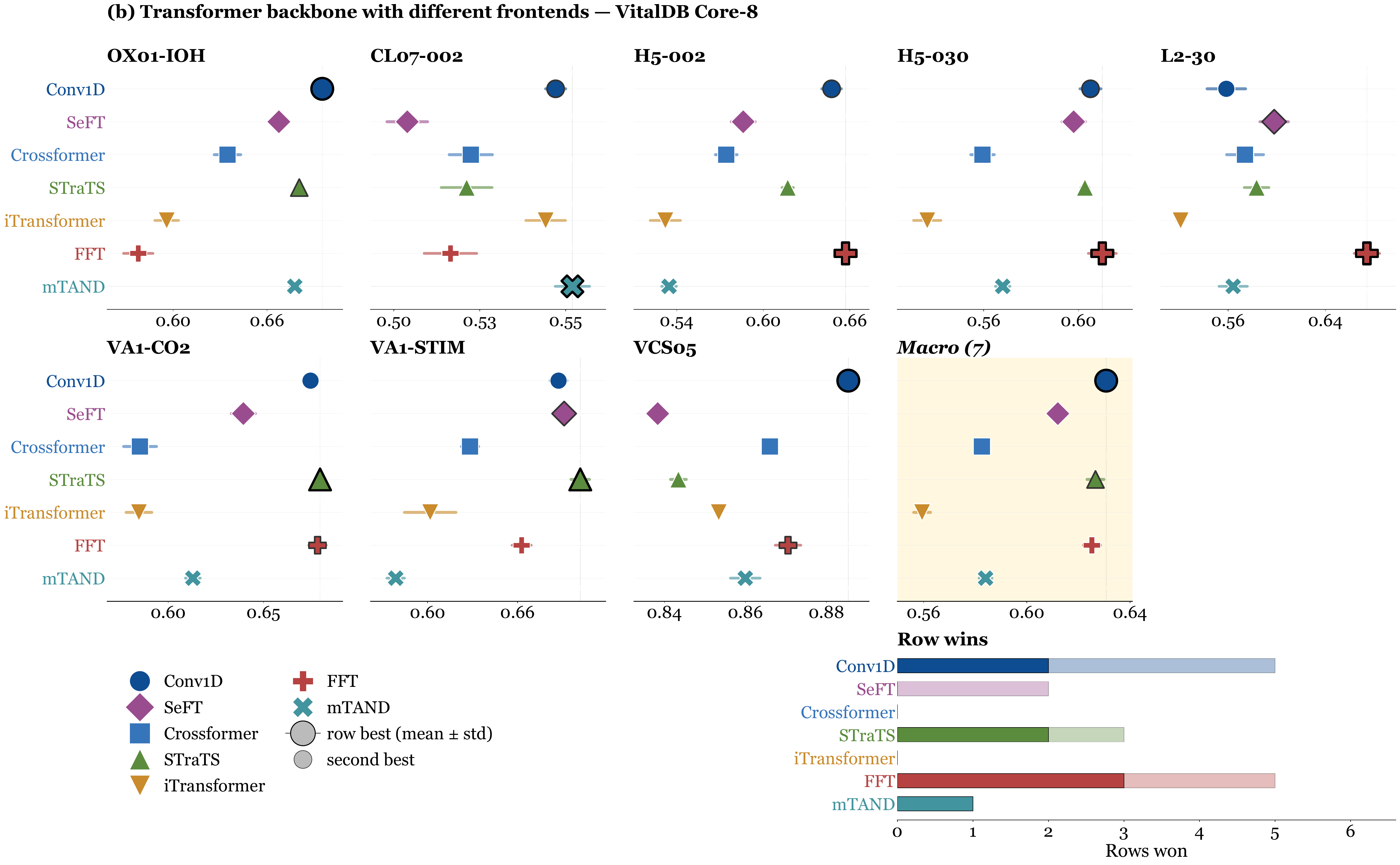}
\caption{\textbf{Transformer tokenizer AUROC by task --- VitalDB Core-8.} Seven tokenizers with the gap weighting and scoring of Figure~\ref{fig:auroc-architectures-vitaldb}. FFT's gas-discordance gain and hypotension loss largely offset in the macro; concurrent VCS05 is separate. Numerical values are in Table~\ref{tab:main_auroc}.}
\label{fig:auroc-tokenizers-vitaldb}
\end{figure*}

\begin{figure*}[!t]
\centering
\includegraphics[width=\linewidth]{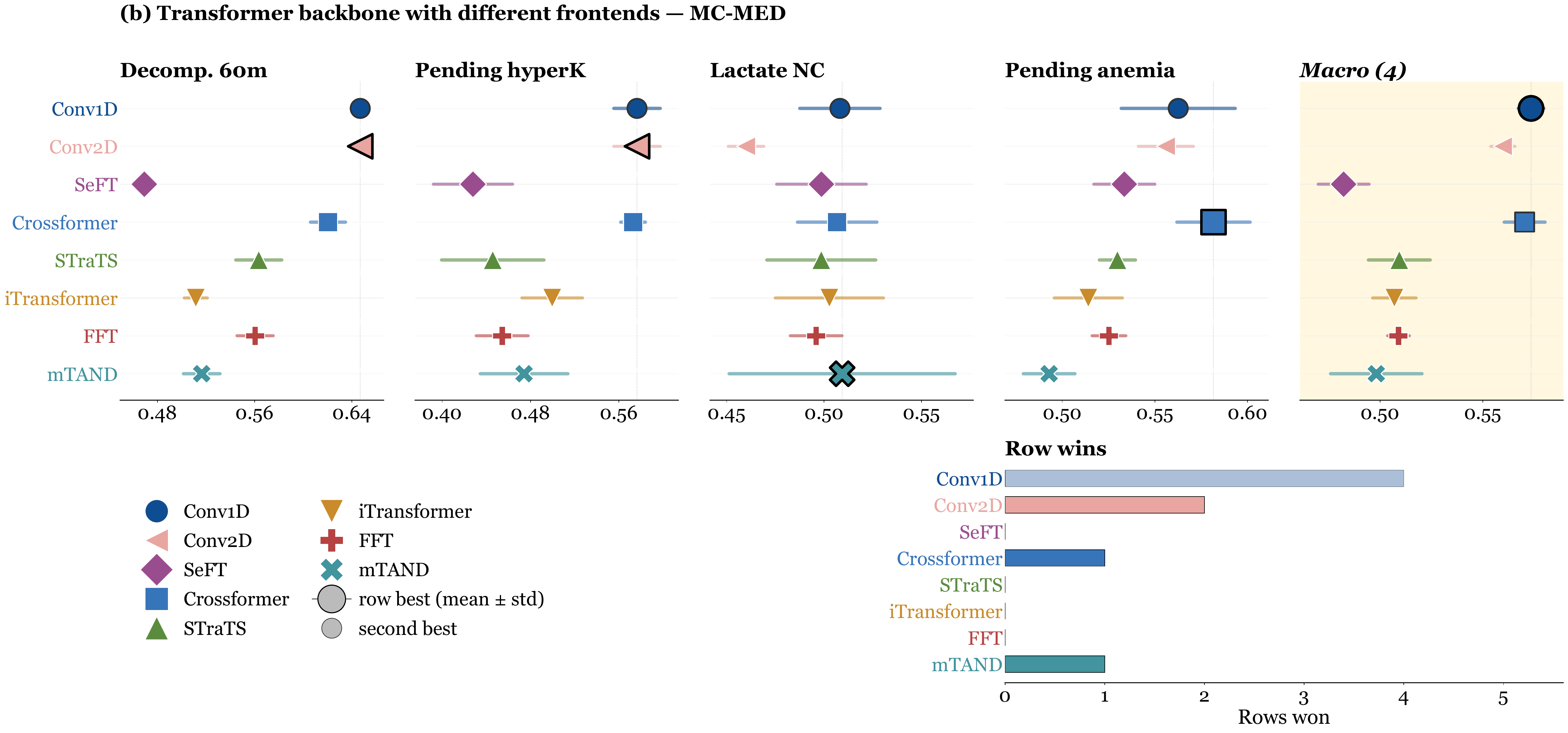}
\caption{\textbf{Transformer tokenizer AUROC by task --- MC-MED.} Eight tokenizers, reported as five-seed means and SDs. Conv1D leads the macro; Conv1D and Conv2D are nearly tied on decompensation and hyperkalemia, while Crossformer leads anemia. Numerical values are in Table~\ref{tab:main_auroc}.}
\label{fig:auroc-tokenizers-mcmed}
\end{figure*}

Figures~\ref{fig:auroc-architectures-mimic}, \ref{fig:auroc-architectures-vitaldb}, and~\ref{fig:auroc-architectures-mcmed} show architecture results; Figures~\ref{fig:auroc-tokenizers-mimic}, \ref{fig:auroc-tokenizers-vitaldb}, and~\ref{fig:auroc-tokenizers-mcmed} show Transformer tokenizer results.
They expand Table~\ref{tab:main_auroc} into individual task panels: markers are five-seed means, horizontal bars are $\pm1$ seed SD, and enlarged outlined markers identify the highest displayed means.
Axis ranges differ by task; shaded panels are macros, and row-win summaries count first and second places over non-macro tasks, including VitalDB's concurrent probe.

\subsection{Tokenizer--Encoder Test Contrasts}
\label{app:representation-test-interactions}

\begin{figure*}[!t]
\centering
\includegraphics[width=\linewidth]{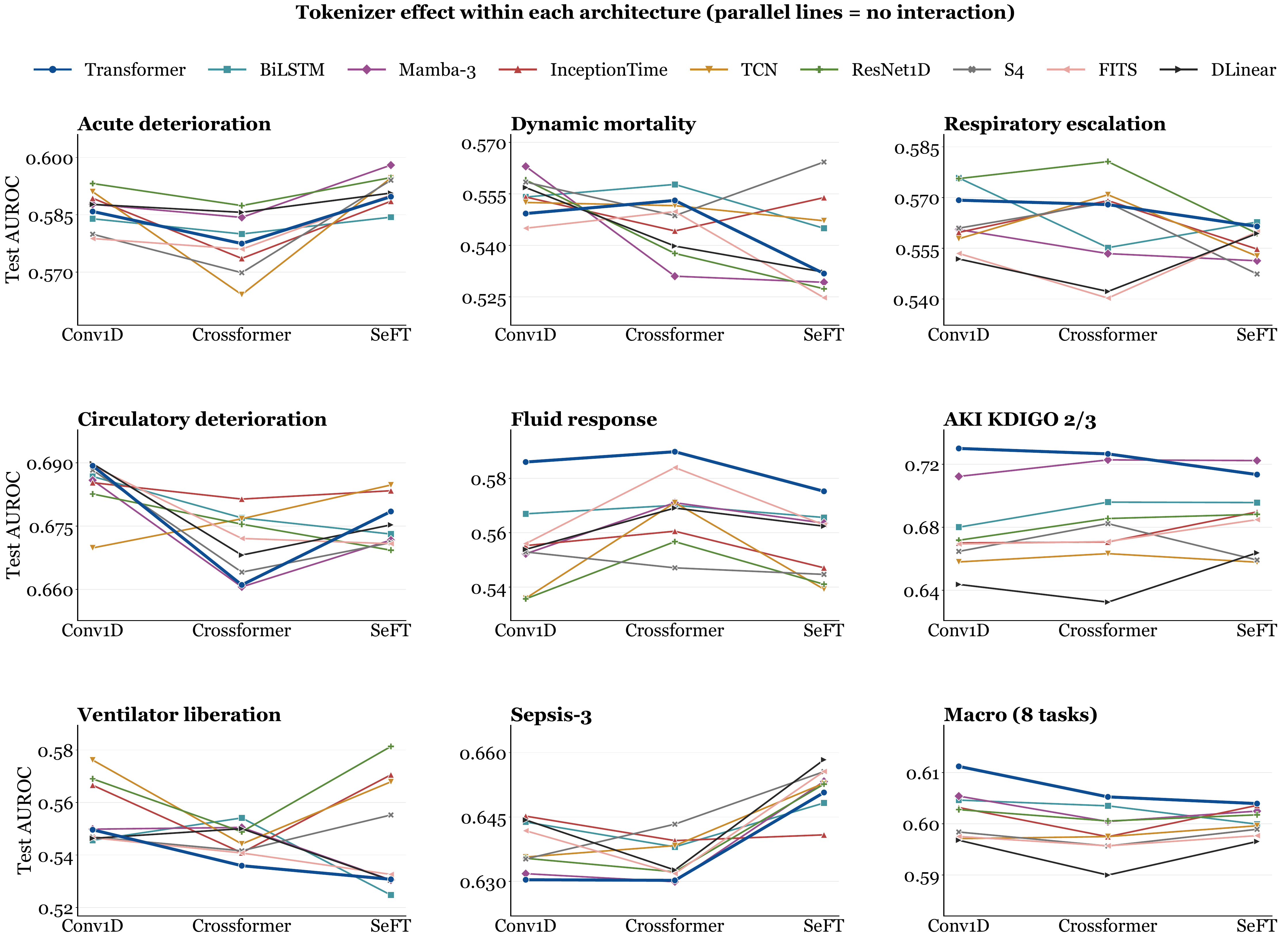}
\caption{\textbf{Task-level tokenizer--architecture interactions on MIMIC-III.} Test AUROC for nine encoders with Conv1D, Crossformer, and SeFT across eight tasks and their macro; each line follows one encoder, with nonparallel trends indicating encoder-dependent tokenizer effects and crossings indicating changes in encoder ranking. Mean task-level contrasts are listed in Table~\ref{tab:representation-test-interactions}.}
\label{fig:mimic-tokenizer-architecture-interactions}
\end{figure*}

Table~\ref{tab:representation-test-interactions} reports the AUROC differences shown in the rounded annotations of Figure~\ref{fig:representation-results}(c) for all three datasets.
For each encoder and task, we subtract the Conv1D test AUROC from the indicated tokenizer's test AUROC. We report these differences on the 0--1 AUROC scale, dividing the figure annotations by 100.
Each mean averages the nine encoder-specific contrasts. Agreement counts how many contrasts have the same sign as the unrounded mean.
MIMIC-III macros average eight tasks; VitalDB forecast macros average seven tasks after equal weighting of the two gaps, with concurrent VCS05 reported separately; MC-MED macros average four tasks.
MC-MED compares Conv2D and FFT with Conv1D; changes are rounded to three decimals, and agreement follows the sign of the unrounded mean.
Figure~\ref{fig:mimic-tokenizer-architecture-interactions} shows the absolute MIMIC-III scores behind the crossed-grid comparison, making encoder-rank changes visible within each task.
This crossed grid is separate from the fixed-Transformer tokenizer search in Table~\ref{tab:main_auroc}(b), whose reported cell means differ.

\noindent\textbf{MIMIC-III interaction patterns.}
\label{app:training-model-choices}
The mean macro changes relative to Conv1D are $-0.003$ AUROC for Crossformer and $-0.001$ for SeFT; eight and five encoders, respectively, share the mean's negative sign.
For fluid response, Crossformer improves AUROC by 0.014 on average, with eight of nine encoders improving, while SeFT improves it by 0.001, with five encoders improving.
For Sepsis-3, Crossformer's mean change is $-0.003$ AUROC, whereas SeFT's is $+0.014$ AUROC, with eight of nine encoders improving under SeFT.
The fixed-Transformer FFT comparison gives another reversal: acute-deterioration AUROC rises from 0.586 to 0.620, while AKI falls from 0.730 to 0.677.
The results favor a shortlist of Conv1D and SeFT for sepsis, and Conv1D and Crossformer for fluid response, where the alternative tokenizer's gain recurs across most encoders.

\noindent\textbf{VitalDB interaction patterns.}
The mean forecast-macro changes are $-0.015$ AUROC for Crossformer and $-0.010$ for SeFT, with negative changes in seven and nine encoders, respectively.
Both tokenizers lower hypotension AUROC across all nine encoders (mean changes of $-0.038$ and $-0.023$), yet improve blood-gas/ETCO$_2$-discordance AUROC by 0.032 and 0.039, with all nine encoders improving under each tokenizer.
For coupled hemodynamic response (VA1-STIM), SeFT improves AUROC by 0.008 in all nine encoders, whereas Crossformer has a mean change of $-0.009$ with only four encoders sharing the negative sign.
With a fixed Transformer, FFT similarly improves gas-discordance AUROC (0.674 versus 0.559) but lowers hypotension AUROC (0.578 versus 0.695).
Conv1D's repeated advantage for hypotension is consistent with the predictive value of local arterial-waveform changes reported by \citet{hatib2018machine}, whereas the gas-discordance target estimates an arterial-to-expired gas relation with different ventilation--perfusion determinants~\citep{nassar2016capnography}.
The concurrent VCS05 probe also favors Conv1D over Crossformer and SeFT across all nine encoders, with mean losses of 0.006 and 0.044, respectively.
Conv1D is better suited to the pressure and concurrent-state endpoints in this comparison; FFT, SeFT, and Crossformer are stronger candidates for gas discordance.

\noindent\textbf{MC-MED interaction patterns.}
FFT lowers mean AUROC for decompensation by 0.108, hyperkalemia by 0.076, and anemia by 0.041, with losses in all nine encoders for each endpoint.
Its lactate gain of 0.010 occurs in six encoders and does not offset these losses: macro AUROC falls by 0.054 in the nine-encoder average, with all nine declining.
Conv2D changes decompensation and hyperkalemia means by approximately 0.000 and 0.001, but lowers lactate nonclearance by 0.020 and anemia by 0.012, giving a macro change of $-0.008$ with eight encoders declining.
The FFT losses on hyperkalemia distinguish the clinical information in ECGs~\citep{galloway2019hyperkalemia} from the representation that best exposes it to the tested models: frequency projection is less effective than local patches for this endpoint.
Conv1D is the stronger shared tokenizer for this ED battery; Conv2D remains a close task-specific comparator for decompensation and hyperkalemia.

\noindent\textbf{Comparison scope.}
The alternative tokenizers differ across datasets: Crossformer and SeFT in the MIMIC-III and VitalDB interaction panels, and Conv2D and FFT on MC-MED.
The updated VitalDB representation table averages fitted-fold model metrics within each seed; the validation and probability-ensemble studies in Appendix~\ref{app:vitaldb-preprocessing} use separate estimators.

\subsection{Efficacy, Efficiency, and Safety of Representation Choices}
\label{app:ees-representation}
Figures~\ref{fig:ees-architectures-mimic}--\ref{fig:ees-architectures-mcmed} compare architectures, and Figures~\ref{fig:ees-tokenizers-mimic}--\ref{fig:ees-tokenizers-mcmed} compare Transformer tokenizers.
Appendix~\ref{app:evaluation-details} specifies metric definitions, model selection, and export differences.

\noindent\textbf{Discrimination, calibration, and selective risk.}
On MIMIC-III, TCN has higher architecture macro AUPRC than Transformer (0.098 versus 0.093) and lower ECE (0.138 versus 0.206), but higher AURC (0.319 versus 0.308).
On VitalDB, Transformer leads architecture AUPRC at 0.078, while BiLSTM has the lowest ECE at 0.036 versus Transformer's 0.038.
In the earlier MC-MED companion campaign, ResNet1D has the highest AUPRC (0.169), S4 the lowest ECE (0.245), and InceptionTime the lowest AURC (0.326).
The different AUPRC and ECE leaders agree with the distinction between discrimination and probability calibration emphasized by \citet{vancalster2019calibration}.
For AUPRC, TCN on MIMIC-III, Transformer on VitalDB, and ResNet1D on MC-MED are strong initial baselines, with the ECE and AURC leaders retained when calibrated risk or selective prediction is the primary objective.

\noindent\textbf{Spectral tokenization across endpoints and metrics.}
For MIMIC-III Transformer tokenizers, FFT improves AUPRC from 0.098 to 0.100, ECE from 0.178 to 0.128, and AURC from 0.330 to 0.271 relative to Conv1D.
In that earlier MC-MED campaign, FFT improves AUPRC (0.166 versus 0.148), but increases ECE (0.282 versus 0.255) and AURC (0.450 versus 0.326); these are separate configurations from the updated AUROC comparison in Table~\ref{tab:main_auroc}.
VitalDB's gas-discordance AUPRC rises from 0.101 with Conv1D to 0.138 with FFT, alongside the AUROC increase from 0.559 to 0.674.
On hypotension, however, AUROC falls from 0.695 to 0.578, leaving FFT below Conv1D in both macro AUROC and AUPRC (0.076 versus 0.078), despite its lower macro ECE (0.035 versus 0.038).

\noindent\textbf{Tokenizer cost and choice.}
MIMIC-III tokenizer training time ranges from 108.6 seconds for Conv1D to 1,084 seconds for mTAND, while peak memory increases from 3.395\% to 17.2\% of device capacity and AUPRC decreases from 0.098 to 0.095.
FFT's AUPRC, ECE, and AURC improvements in this comparison require 142.6 seconds and 3.404\% memory, substantially less than mTAND.
VitalDB tokenizer cost records cover Conv1D, SeFT, and Crossformer.
Conv1D and FFT offer a useful cost comparison on MIMIC-III; the updated VitalDB and MC-MED AUROC comparisons favor Conv1D for macro discrimination.

\eesfigure{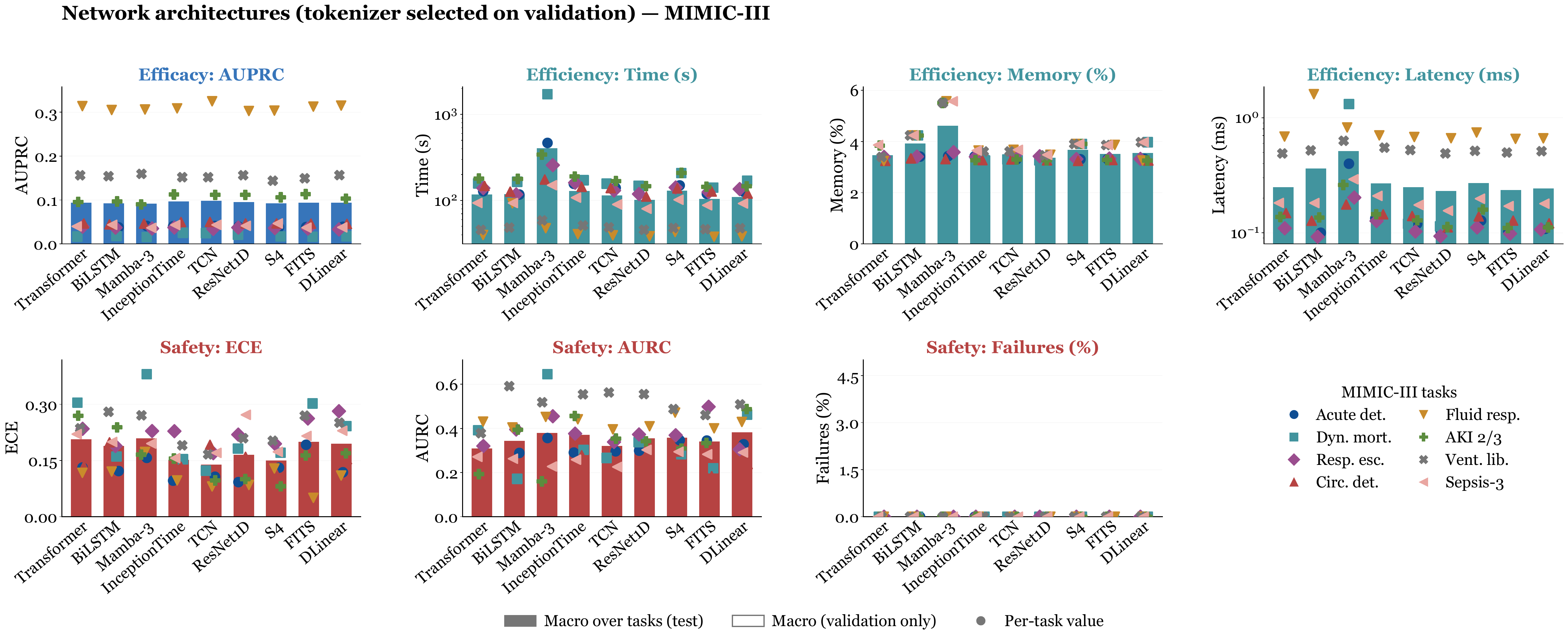}{fig:ees-architectures-mimic}{\textbf{Network architectures --- MIMIC-III.} Approximately 500k-parameter encoders with selected tokenizers. Panels show AUPRC, training time, memory, time per training sample, ECE, AURC, and recorded failures. Metric and mark definitions follow Appendix~\ref{app:evaluation-details}. Numerical values are in Tables~\ref{tab:ees-architectures-mimic} and~\ref{tab:ees-architectures-mimic-cost}.}

\eesfigure{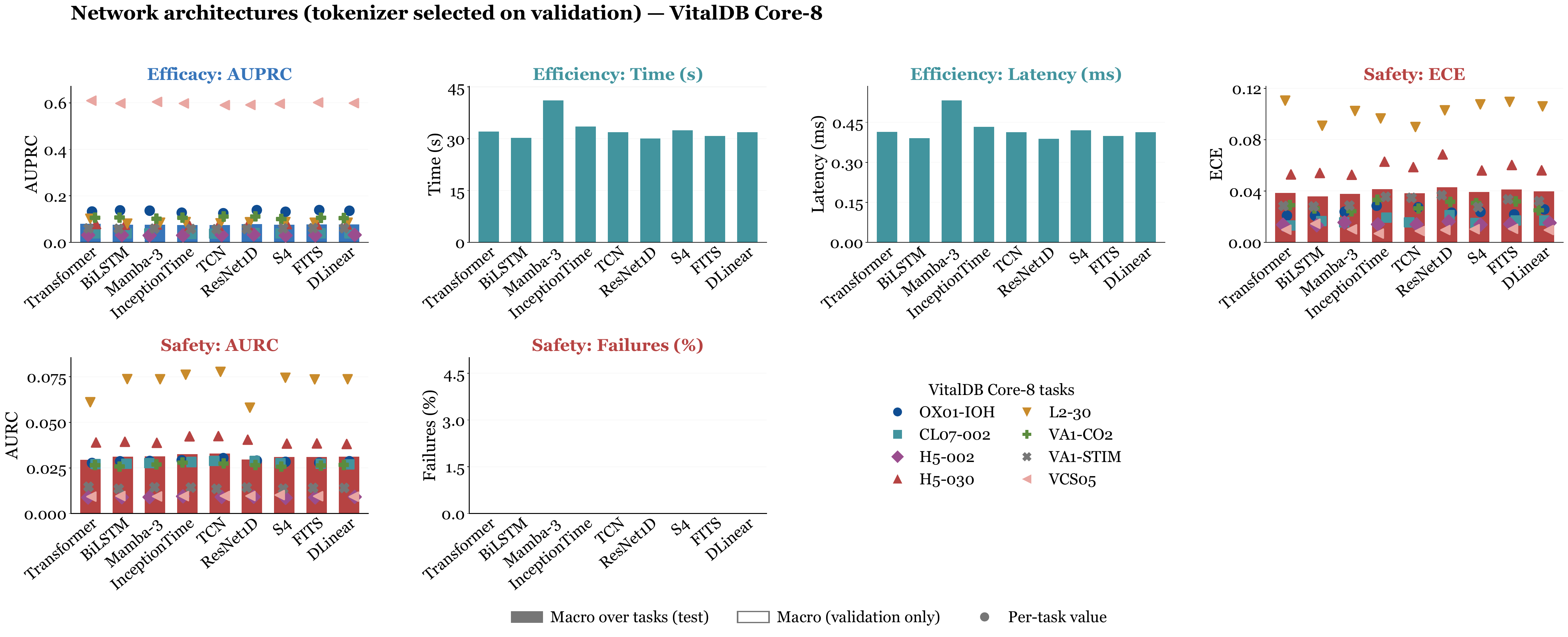}{fig:ees-architectures-vitaldb}{\textbf{Network architectures --- VitalDB Core-8.} Approximately 500k-parameter encoders with selected tokenizers. Forecast macros exclude concurrent VCS05. Cost panels use earlier-grid telemetry per validation cell. Panels show AUPRC, training time, time per training sample, ECE, AURC, and audited prediction-file failures. Metric and mark definitions follow Appendix~\ref{app:evaluation-details}. Numerical values are in Tables~\ref{tab:ees-architectures-vitaldb} and~\ref{tab:ees-architectures-vitaldb-cost}.}

\eesfigure{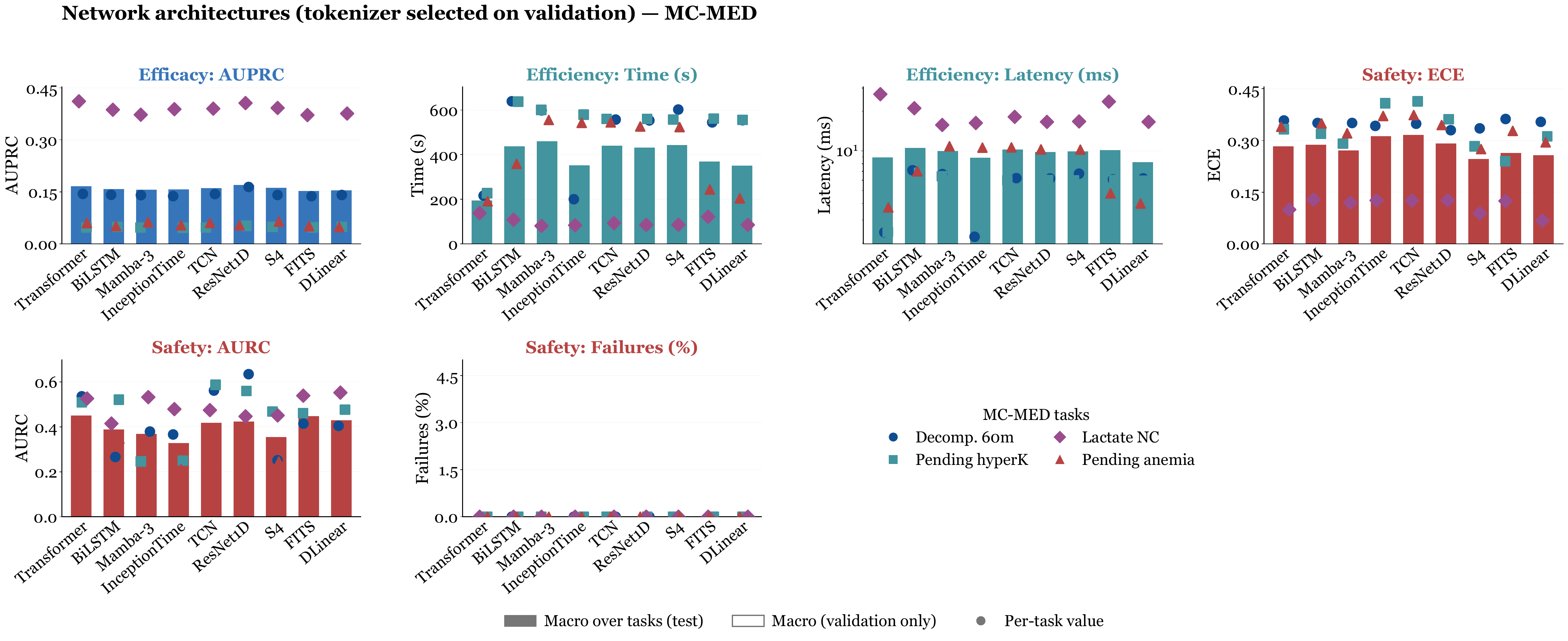}{fig:ees-architectures-mcmed}{\textbf{Network architectures --- MC-MED, earlier companion campaign.} Approximately 500k-parameter encoders with selected tokenizers, including FFT candidates. Panels show AUPRC, training time, time per training sample, ECE, AURC, and recorded failures. Metric and mark definitions follow Appendix~\ref{app:evaluation-details}. Numerical values are in Tables~\ref{tab:ees-architectures-mcmed} and~\ref{tab:ees-architectures-mcmed-cost}.}

\eesfigure{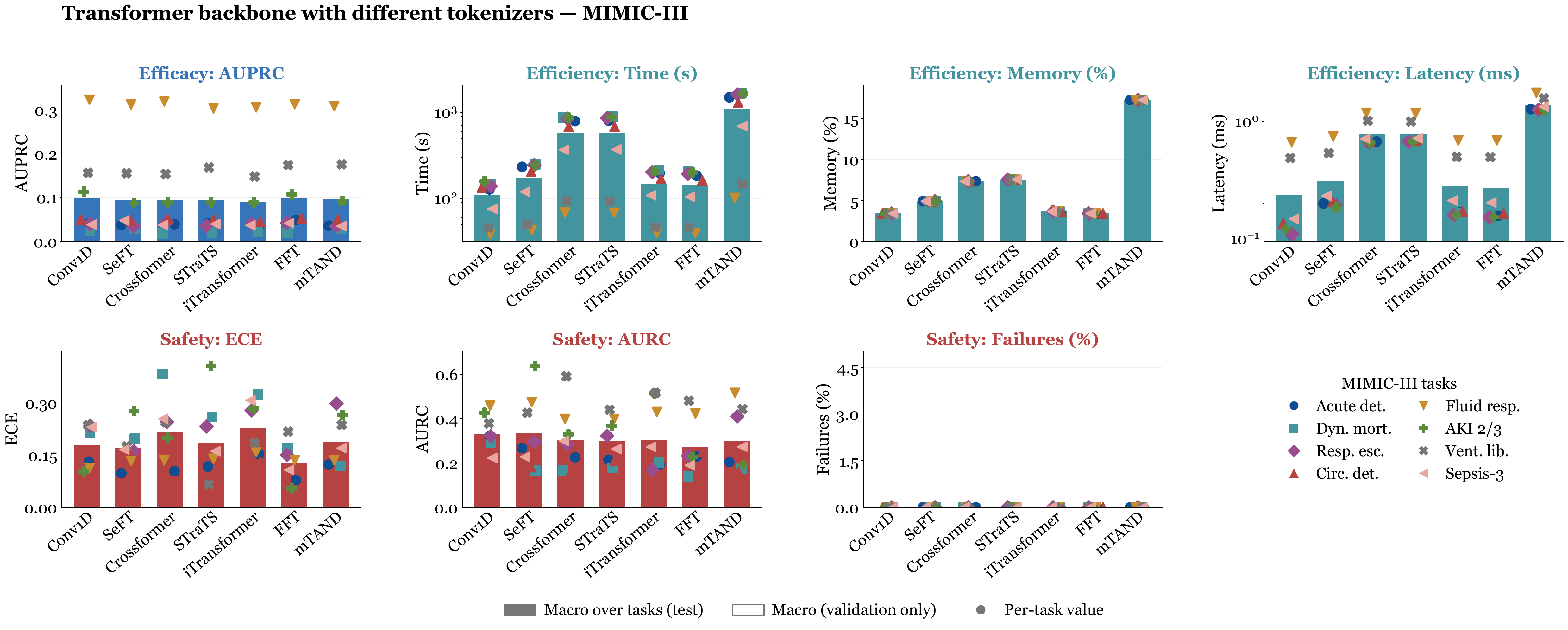}{fig:ees-tokenizers-mimic}{\textbf{Transformer tokenizers --- MIMIC-III.} Seven tokenizers with a fixed Transformer backbone. Panels show AUPRC, training time, memory, time per training sample, ECE, AURC, and recorded failures. Metric and mark definitions follow Appendix~\ref{app:evaluation-details}. Numerical values are in Tables~\ref{tab:ees-tokenizers-mimic} and~\ref{tab:ees-tokenizers-mimic-cost}.}

\eesfigure{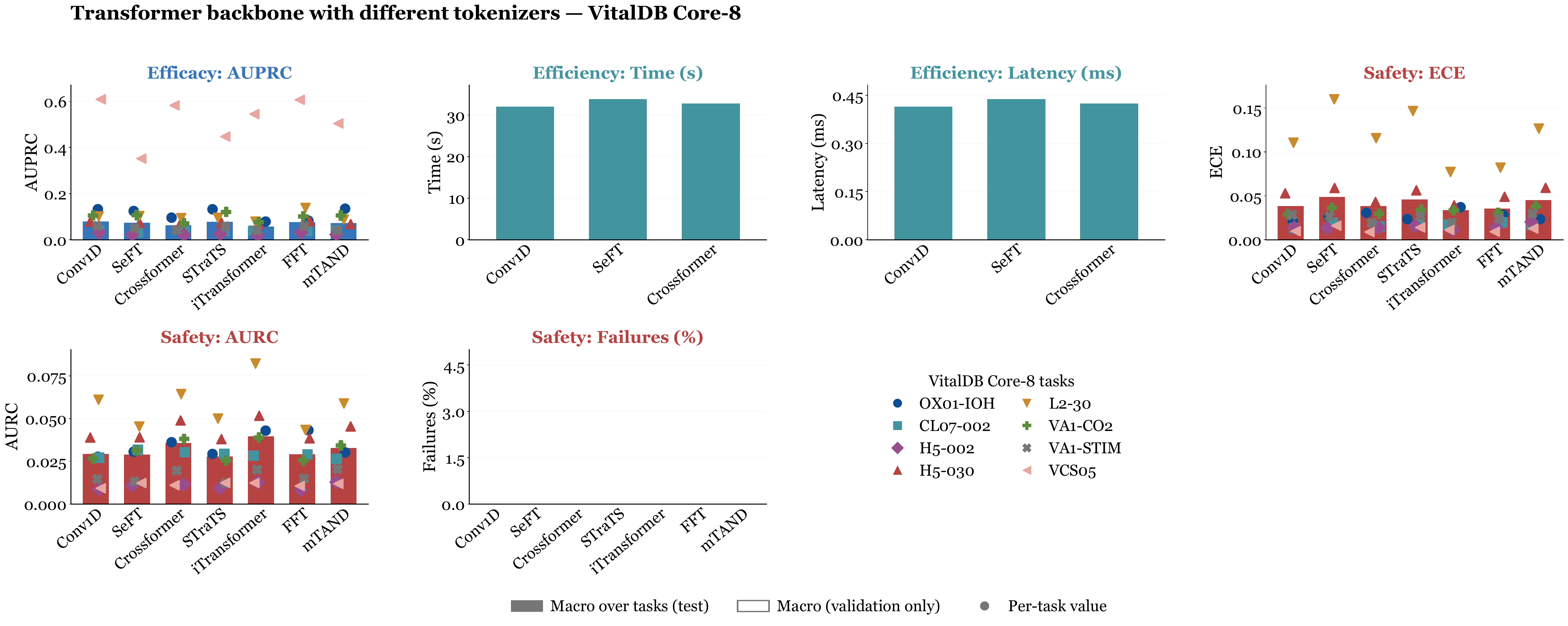}{fig:ees-tokenizers-vitaldb}{\textbf{Transformer tokenizers --- VitalDB Core-8.} Seven tokenizers with a fixed Transformer backbone. Forecast macros exclude concurrent VCS05. Cost panels use earlier-grid telemetry per validation cell. Panels show AUPRC, training time, time per training sample, ECE, AURC, and audited prediction-file failures. Metric and mark definitions follow Appendix~\ref{app:evaluation-details}. Numerical values are in Tables~\ref{tab:ees-tokenizers-vitaldb} and~\ref{tab:ees-tokenizers-vitaldb-cost}.}

\eesfigure{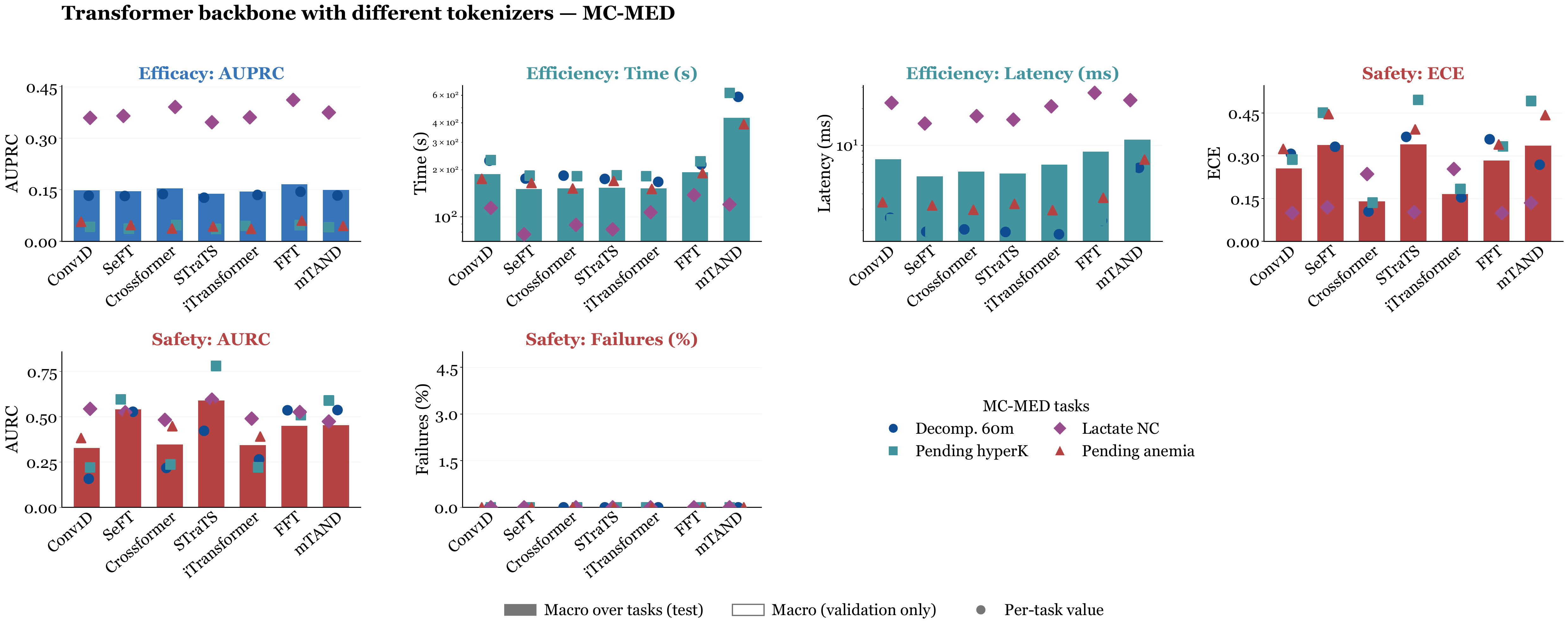}{fig:ees-tokenizers-mcmed}{\textbf{Transformer tokenizers --- MC-MED, earlier companion campaign.} Seven tokenizers with a fixed Transformer backbone. Panels show AUPRC, training time, time per training sample, ECE, AURC, and recorded failures. Metric and mark definitions follow Appendix~\ref{app:evaluation-details}. Numerical values are in Tables~\ref{tab:ees-tokenizers-mcmed} and~\ref{tab:ees-tokenizers-mcmed-cost}.}

\section{Temporal Comparison Details}
\label{app:temporal-comparisons}

\noindent\textbf{Comparison grids and evaluation status.}
Figure~\ref{fig:context-warning} compares two timing choices: how much waveform history is available before prediction, and how long the interval is between the input cutoff and the target window.
Table~\ref{tab:temporal-comparison-grids} records the displayed observation lengths and forecast-gap axis values.
All six panels in Figure~\ref{fig:context-warning} are labeled as test AUROC.
The VitalDB warning curve is distinct from the validation curve and the fixed-30-minute Transformer/Patch test results in Appendices~\ref{app:vitaldb-gaps} and~\ref{app:vitaldb-gap-test}. The scores, uncertainty intervals, and probability-quality metrics reported there apply only to those studies.
The MIMIC-III warning panel contains only dynamic mortality, AKI, and Sepsis-3; MC-MED includes all four endpoints with task-specific gaps.
The refreshed MIMIC-III duration comparison contains 10-, 20-, and 60-minute histories; the earlier five-minute result belongs to a separate campaign.
The duration curves average three encoder--tokenizer configurations on MIMIC-III and five on VitalDB; MC-MED uses ResNet1D with Conv2D. The VitalDB gap study uses Transformer with patch embedding.


\noindent\textbf{Observation length.}
Black lines in panel~(a) summarize eight MIMIC-III tasks, seven VitalDB forecast tasks, and four MC-MED tasks.
VitalDB weights its two core gaps equally within each forecast task; concurrent VCS05 is excluded from the forecast macro.
Numerical observations use the task-level exports underlying the updated figure.
MIMIC-III macro AUROC is 0.606 at both ten and 20 minutes and 0.609 at 60 minutes; respiratory escalation improves from 0.575 to 0.593 and acute deterioration from 0.646 to 0.656 across ten to 60 minutes.
Mortality stays near 0.517, while Sepsis-3 decreases from 0.659 to 0.652; the modest macro change therefore combines opposing endpoint trends.
VitalDB macro AUROC increases from 0.627 at five minutes to 0.636 at 60 minutes, with hypotension peaking at 30 minutes (0.729) and gas discordance at 60 minutes (0.594).
The hypotension gain aligns with prior evidence of predictive pre-event arterial-waveform patterns~\citep{hatib2018machine}, whereas oxygen desaturation decreases from 0.635 to 0.625 and hypothermia stays near 0.602.
MC-MED macro AUROC peaks at ten minutes (0.578 versus 0.567 at both five and 15 minutes).
From ten to 15 minutes, decompensation falls from 0.656 to 0.631, hyperkalemia from 0.576 to 0.528, and anemia from 0.577 to 0.530; lactate nonclearance instead rises from 0.502 to 0.577.
This lactate-specific gain is consistent with a serial-measurement endpoint shaped by evolving metabolism and treatment~\citep{vincent2016lactate}.
For AUROC, ten-minute histories are useful starting points for the MIMIC-III battery and the three non-lactate MC-MED tasks, while longer windows suit respiratory escalation, intraoperative hypotension and gas discordance, and MC-MED lactate nonclearance in these sweeps.

\noindent\textbf{Forecast gap.}
Panel~(b) shifts the target window beyond the same input cutoff, changing both the warning requirement and which future events qualify.
For MIMIC-III, mortality decreases from 0.545 at three hours to 0.539 at twelve hours and 0.537 at 24 hours; AKI instead peaks at six hours (0.697 versus 0.686 at three hours and 0.685 at twelve hours).
Sepsis-3 returns to 0.629 at six hours after falling to 0.619 at four hours.
AKI's intermediate peak is consistent with creatinine and urine-output criteria being assessed over time~\citep{kdigo2012aki}, linking its warning curve to outcome ascertainment as well as physiology.
For VitalDB, macro AUROC falls from 0.662 at five minutes to 0.621 at 15 minutes, then ranges from 0.619 to 0.632.
Coupled response rebounds from 0.603 at 45 minutes to 0.750 at 75 minutes, while gas discordance declines from 0.622 at 30 minutes to 0.555 at 75 minutes.
Hypotension falls from 0.791 at five minutes to 0.611 at 60 minutes before recovering to 0.650 at 75 minutes; its strongest discrimination at short gaps aligns with prior arterial-waveform forecasting results~\citep{hatib2018machine}.
In our clinical interpretation, the 60-minute hypothermia peak reflects evolving redistribution and warming~\citep{sessler2016thermoregulation}, and the coupled-response rebound reflects a later operative phase with changing surgical stimulation and anesthetic context~\citep{ledowski2019nociception}.
For MC-MED, decompensation declines from 0.647 at zero gap to 0.629 at ten minutes and 0.626 at 30 minutes.
Hyperkalemia is almost unchanged from zero to 15 minutes (0.566 and 0.567), while anemia improves from 0.556 to 0.566; both decline at 60 minutes (0.532 and 0.544).
Lactate nonclearance falls from 0.499 at 30 minutes to 0.489 at 60 minutes, then rises to 0.532 at 90 minutes.
The laboratory endpoints therefore have different preferred intervals, with the lactate rebound remaining close to chance.
These curves use fixed patient partitions but shifted outcome windows and gap-specific eligibility, so they describe prediction of different future intervals within the same partitioning scheme.
Model comparisons should fix the gap to the intended clinical intervention window and interpret later-gap gains alongside outcome timing and support.

\noindent\textbf{Temporal controls and interpretation.}
The observation-length comparisons in Figure~\ref{fig:context-warning}(a) use the same anchors and nested raw-waveform intervals $[T-W,T)$, holding the task, gap, horizon, channels, labels, and patient partitions fixed.
For MIMIC-III, the extracted history is $[T-6\,\mathrm{h},T)$; the five-minute and one-hour conditions take $[T-5\,\mathrm{min},T)$ and $[T-1\,\mathrm{h},T)$, respectively.
For warning-time comparisons, each patient retains the same training, validation, or test assignment across gaps, including the same fold assignment where folds are used.
The observation view and target horizon remain fixed, while the outcome interval shifts to $[T+G,T+G+H)$.
This fixes patient allocation across gaps; eligible anchors, outcome labels, and prevalence can still change with follow-up and task-specific exclusion rules.
The separate VitalDB validation extension reports this gap-specific support and prevalence with subject-bootstrap intervals conditional on fitted models (Appendix~\ref{app:vitaldb-gaps}).

\subsection{Efficacy, Efficiency, and Safety Across Temporal Settings}
\label{app:ees-temporal}
Figures~\ref{fig:ees-observation-length-mimic}--\ref{fig:ees-observation-length-mcmed} compare observation lengths, and Figures~\ref{fig:ees-forecast-gap-mimic}--\ref{fig:ees-forecast-gap-mcmed} compare forecast gaps.
These figures use the September 17 temporal collection corresponding to the AUROC studies in Figure~\ref{fig:context-warning}; observation histories and task-specific gap grids follow Appendix~\ref{app:temporal-comparisons}.
Predictive and confidence metrics report test results; VitalDB training costs retain earlier-grid telemetry, and MC-MED memory records sampled shared-device usage in MiB.

\noindent\textbf{Additional history and measured cost.}
MIMIC-III macro AUPRC changes from 0.1192 at ten minutes to 0.1177 at 20 and 0.1227 at 60, while 20 minutes gives the lowest macro ECE (0.1538) and AURC (0.3190).
Respiratory-escalation AUPRC rises from 0.0549 to 0.0725 across ten to 60 minutes, but its ECE increases from 0.2264 to 0.2787 and AURC from 0.3363 to 0.4530: extra waveform history improves discrimination without improving the confidence estimates used to prioritize review.
For the ICU battery, 20- and 60-minute views offer different starting points for confidence quality and precision--recall performance, respectively.

VitalDB macro AUPRC rises from 0.0762 at five minutes to 0.0810 at 30 and 0.0812 at 60; macro ECE decreases from 0.0411 to 0.0351 and AURC from 0.0300 to 0.0277 across five to 60 minutes.
Hypotension AUPRC increases from 0.1321 to 0.1597, consistent with predictive information in pre-event arterial-pressure patterns~\citep{hatib2018machine}, whereas oxygen-desaturation AUPRC decreases from 0.0295 to 0.0237.
Thirty- and 60-minute histories are useful hypotension candidates, while desaturation favors retaining the five-minute view.

MC-MED macro AUPRC increases from 0.1809 at five minutes to 0.1977 at ten and 0.2208 at 15, even though macro AUROC peaks at ten minutes.
Decompensation and lactate drive the ten-to-15-minute AUPRC gain (0.2763 to 0.3166 and 0.3966 to 0.4689); hyperkalemia and anemia both decline.
The lowest macro ECE occurs at ten minutes (0.0873), whereas five minutes has the lowest AURC (0.2385 versus 0.2661 at 15 minutes).
At batch size 128, mean training time per optimizer step increases from 14.20\,ms at five minutes to 18.80\,ms at ten and 25.27\,ms at 15, separating update cost from differences in the number of training windows and selected epochs.
Sampled shared-device memory is nonmonotonic and does not measure a model's isolated memory requirement.
Ten minutes is the stronger AUROC/calibration starting point, with 15 minutes as the AUPRC alternative for decompensation and lactate and five minutes retained for selective-risk comparisons.

\noindent\textbf{Earlier warning and endpoint differences.}
MIMIC-III AKI AUPRC decreases from 0.1509 at three hours to 0.1404 at six and 0.1202 at twelve, despite AUROC peaking at six hours.
Sepsis-3 AUPRC also falls from 0.0362 at two hours to 0.0301 at six, while ECE improves from 0.3108 to 0.2197.
Thus, neither the AKI AUROC peak nor the improved sepsis calibration identifies the gap with the strongest precision--recall performance.

VitalDB gap macro AUPRC decreases from 0.100 at five minutes to 0.071 at 15 minutes, then varies nonmonotonically and reaches 0.082 at 75 minutes.
Coupled hemodynamic response recovers from 0.047 at 45 minutes to 0.113 at 75 minutes, while oxygen-desaturation AUPRC ranges from 0.022 to 0.027 across 15--75 minutes.
We interpret the coupled-response rebound as reflecting a later operative period entering the outcome window, since stimulation and anesthetic context affect hemodynamic responses~\citep{ledowski2019nociception}.
Across the six gaps, macro ECE ranges from 0.030 to 0.035 and AURC from 0.027 to 0.030.
The 45-minute gap has the lowest ECE (0.030) but also the lowest AUPRC (0.068), consistent with calibration and discrimination measuring different properties~\citep{vancalster2019calibration}.
MC-MED decompensation AUPRC decreases from 0.2636 at zero gap to 0.1566 at 30 minutes, agreeing with its AUROC decline.
Anemia instead gains AUPRC from 0.0470 at zero gap to 0.1999 at 60 minutes, while AUROC falls from 0.556 to 0.544 and ECE increases from 0.0348 to 0.0776.
This later-window AUPRC gain accompanies weaker pairwise ranking and calibration; shifted outcome windows and prevalence must be considered when interpreting it.
MIMIC-III and MC-MED cover different task subsets at different gaps, so their figures show task markers without macro bars.
For waveform warning models, fix the intervention gap before comparing configurations, then assess precision--recall performance and confidence quality within that task and outcome window.

\eesfigure{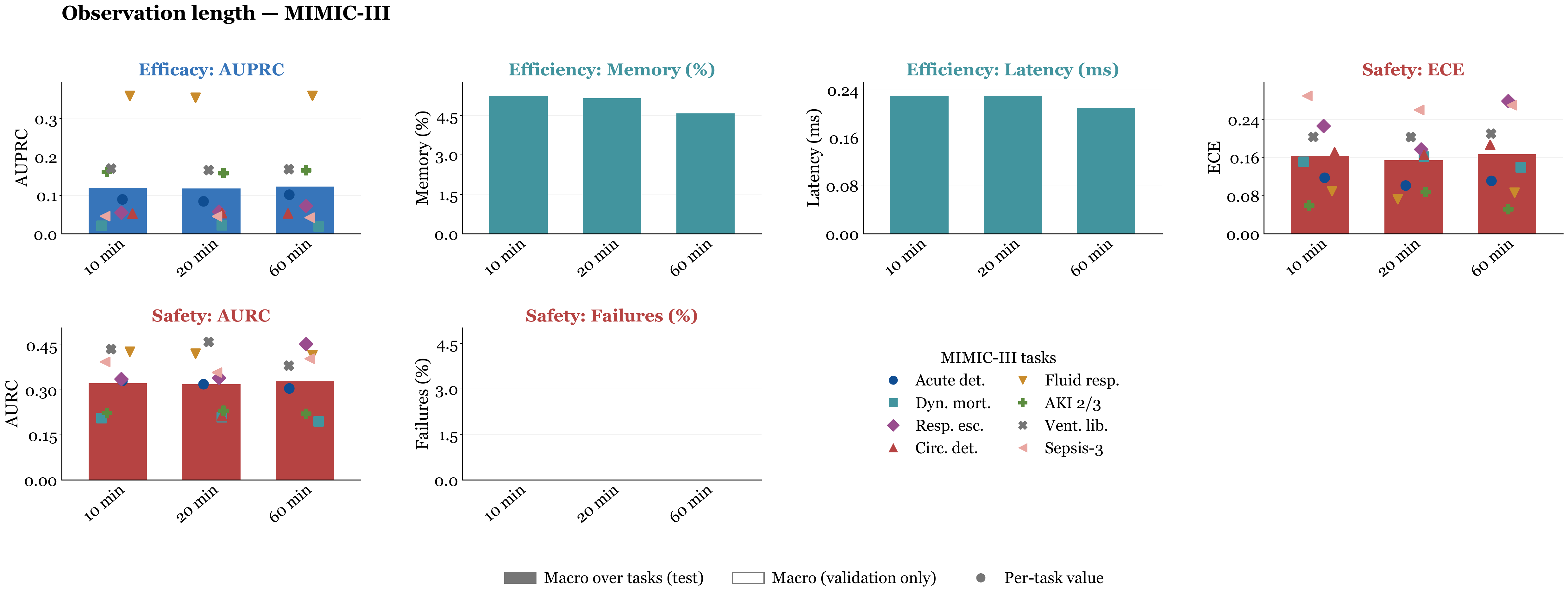}{fig:ees-observation-length-mimic}{\textbf{Observation length --- MIMIC-III.} Ten-, 20-, and 60-minute histories across eight tasks. Panels show AUPRC, peak accelerator memory (\%), time per training sample, ECE, AURC, and recorded failures. Metric and mark definitions follow Appendix~\ref{app:evaluation-details}. Numerical values are in Tables~\ref{tab:ees-observation-length-mimic} and~\ref{tab:ees-observation-length-mimic-cost}.}

\eesfigure{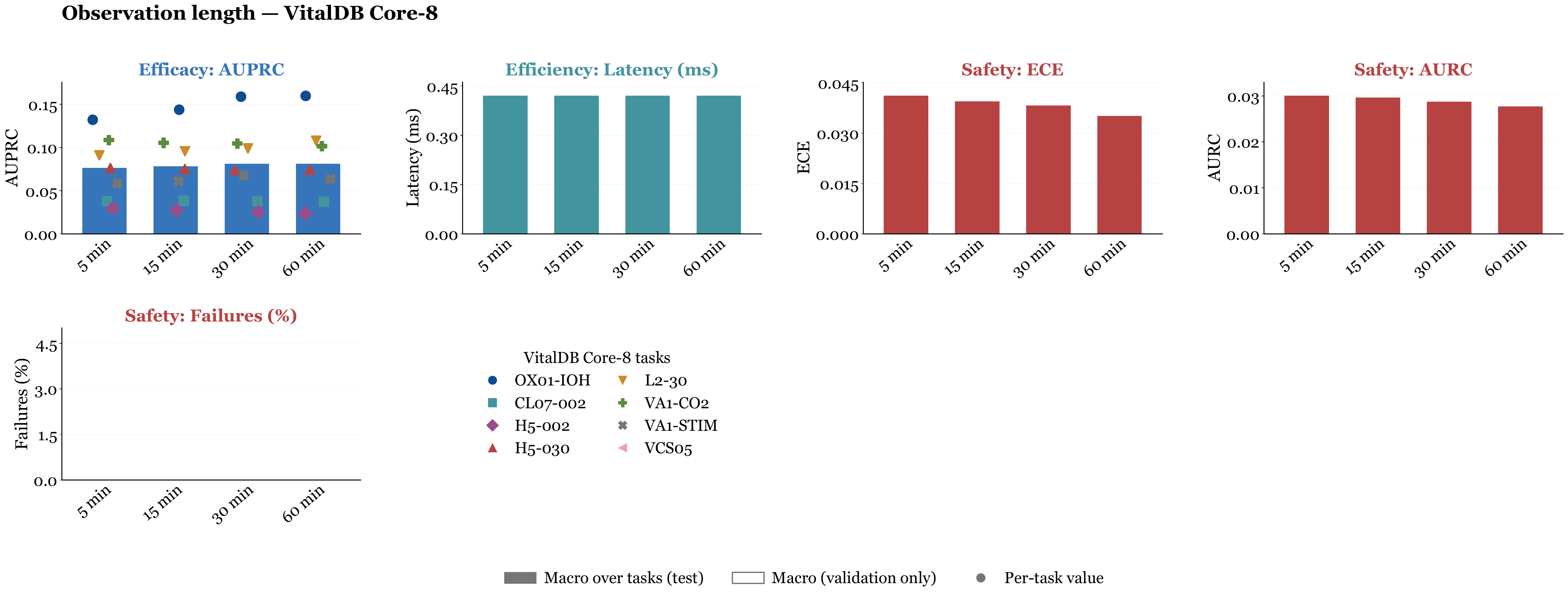}{fig:ees-observation-length-vitaldb}{\textbf{Observation length --- VitalDB Core-8.} Five-, 15-, 30-, and 60-minute histories; forecast macros exclude concurrent VCS05. Per-sample cost uses earlier-grid telemetry. Panels show AUPRC, time per training sample, ECE, AURC, and audited prediction-file failures. Metric and mark definitions follow Appendix~\ref{app:evaluation-details}. Numerical values are in Tables~\ref{tab:ees-observation-length-vitaldb} and~\ref{tab:ees-observation-length-vitaldb-cost}.}

\eesfigure{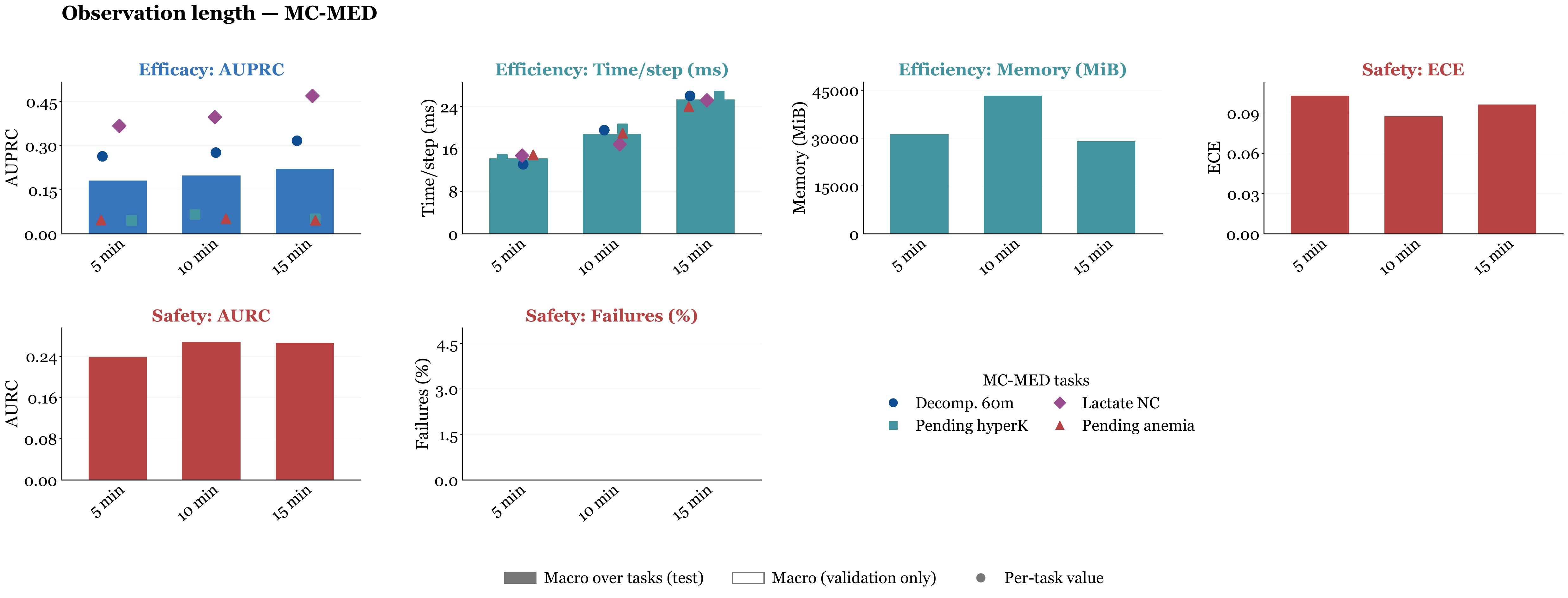}{fig:ees-observation-length-mcmed}{\textbf{Observation length --- MC-MED.} Five-, ten-, and 15-minute histories across four ED tasks. Panels show AUPRC, training time per optimizer step (ms; batch size 128), sampled shared-device memory (MiB), ECE, AURC, and training-attempt failures. Step times divide each fit's training seconds by its recorded update count before averaging. Memory is a device-level measurement rather than an isolated model peak. Metric and mark definitions follow Appendix~\ref{app:evaluation-details}. Numerical values are in Tables~\ref{tab:ees-observation-length-mcmed} and~\ref{tab:ees-observation-length-mcmed-cost}.}

\eesfigure{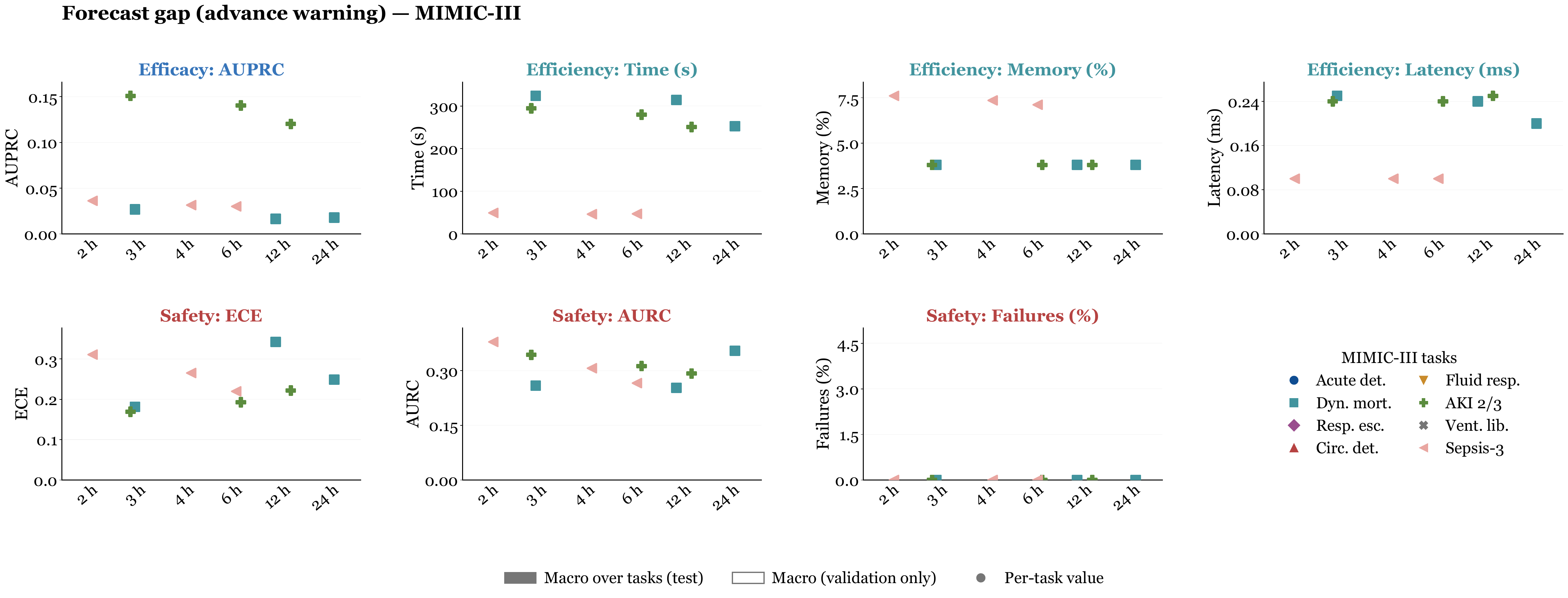}{fig:ees-forecast-gap-mimic}{\textbf{Forecast gap --- MIMIC-III.} Task-specific intervals for mortality (3/12/24\,h), AKI (3/6/12\,h), and Sepsis-3 (2/4/6\,h). Task coverage varies by gap, so macro bars are omitted. Panels show AUPRC, training time, peak accelerator memory (\%), time per training sample, ECE, AURC, and recorded failures. Metric and mark definitions follow Appendix~\ref{app:evaluation-details}. Numerical values are in Tables~\ref{tab:ees-forecast-gap-mimic} and~\ref{tab:ees-forecast-gap-mimic-cost}.}

\eesfigure{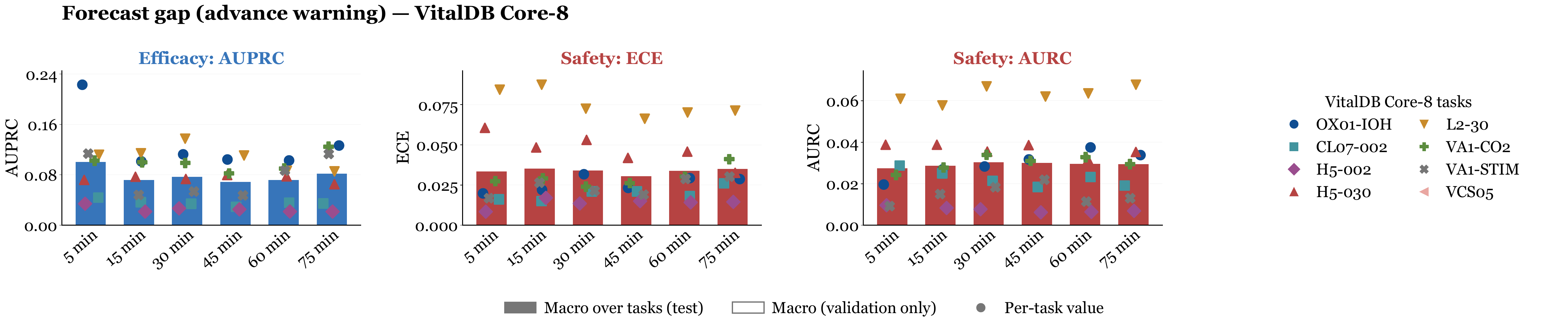}{fig:ees-forecast-gap-vitaldb}{\textbf{Forecast gap --- VitalDB Core-8.} Five- to 75-minute advance-warning intervals across seven forecasts, excluding concurrent VCS05. Panels show AUPRC, ECE, and AURC. Metric and mark definitions follow Appendix~\ref{app:evaluation-details}. Numerical values are in Table~\ref{tab:ees-forecast-gap-vitaldb}.}

\eesfigure{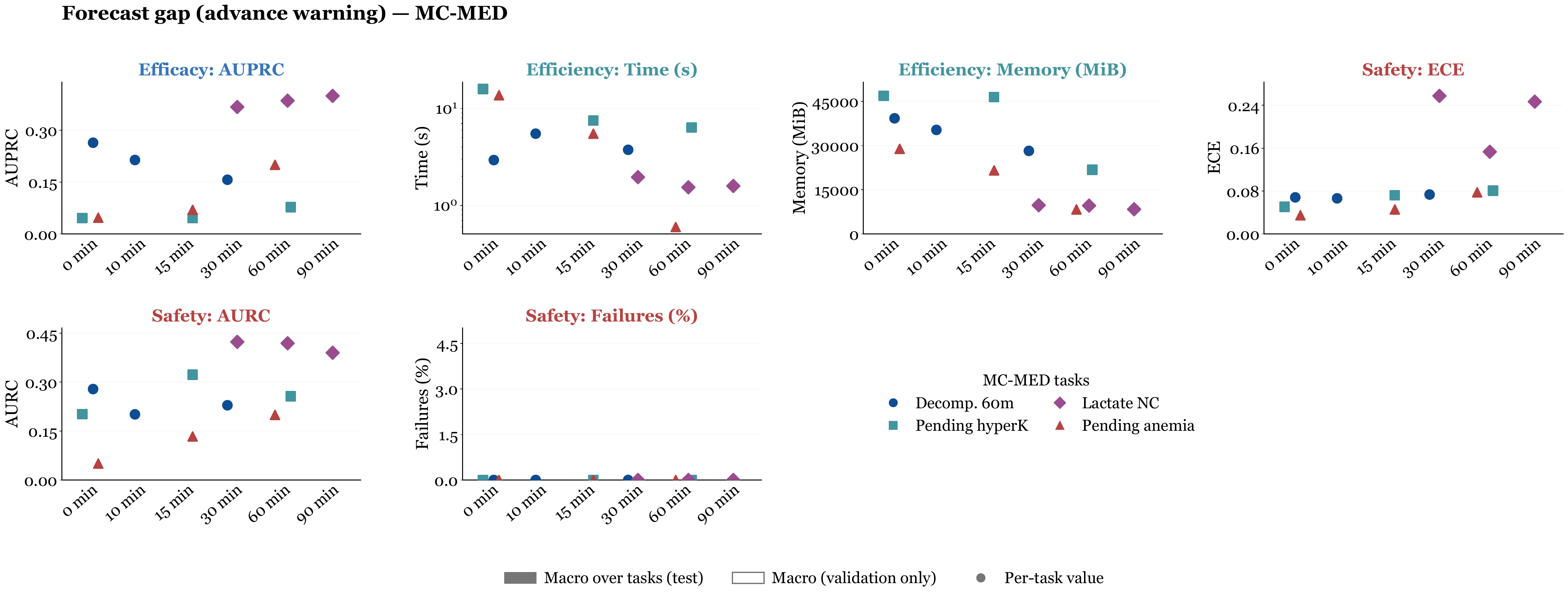}{fig:ees-forecast-gap-mcmed}{\textbf{Forecast gap --- MC-MED.} Task-specific advance-warning intervals for four ED endpoints. Task coverage varies by gap, so macro bars are omitted. Panels show AUPRC, training time, sampled shared-device memory (MiB), ECE, AURC, and training-attempt failures. Memory measures device usage rather than an isolated model peak. Metric and mark definitions follow Appendix~\ref{app:evaluation-details}. Numerical values are in Tables~\ref{tab:ees-forecast-gap-mcmed} and~\ref{tab:ees-forecast-gap-mcmed-cost}.}

\section{Information Availability Comparison Details}
\label{app:information-comparisons}

\noindent\textbf{References and evidence scope.}
Figure~\ref{fig:information-multimodal} reports AUROC differences relative to a separate reference for each experimental axis (Table~\ref{tab:information-comparison-grids}).
The top two rows represent MIMIC-III and VitalDB; the lower-right cells show MC-MED fusion and subject selection, and the two lower-left cells show transfer to MIMIC-IV.
Colored symbols identify tasks within each dataset, and bars summarize aggregate differences; the transfer cells use the distinct five-target key in Appendix~\ref{app:information-transfer}.
The task legends contain eight ICU tasks, seven VitalDB forecasting tasks plus the concurrent VCS05 probe, and four MC-MED tasks.
The concurrent probe remains separate from the forecast aggregate under the documented VitalDB protocol.
Comparisons in Section~\ref{sec:information-availability-draft} describe the plotted means; uncertainty definitions follow the corresponding detailed studies below.
The composite and task-level figures use the September 17 collection; MC-MED subject selection and fusion share the September 16 campaign with ECG-II and respiration inputs.
MIMIC-III sensor, missingness, and subset results average Transformer/Patch, Transformer/Crossformer, and selective-SSM/Patch; VitalDB averages Transformer/Patch, Transformer/Crossformer, and Mamba-3/Patch.
Fusion uses Transformer/Patch on MIMIC-III and MC-MED, and Mamba-3/Patch with 60-minute history on VitalDB.
The revised VitalDB plotted means are separate from the earlier validation summaries and completed fixed-recipe tests in Appendices~\ref{app:training-budget-test} and~\ref{app:vitaldb-fusion-test}.
Transfer uses held-out outer-role test predictions under the retrospective development protocol in Appendix~\ref{app:information-transfer}.


\subsection{Sensor Access and Missing Inputs}
\label{app:information-sensors}
\noindent\textbf{Channel selection.}
Figure~\ref{fig:information-multimodal}(a) compares channel-selection policies against each dataset's all-channel reference; Figures~\ref{fig:info-tasks-sensors-mimic} and~\ref{fig:info-tasks-sensors-vitaldb} give task-level AUROCs.
MC-MED channel selection and channel imputation are omitted because its input configuration has only two available waveform channels.
At ten channels, MIMIC-III clinical-priority and SQI selection reach macro AUROCs of 0.6062 and 0.6073 versus the all-channel 0.6056, while random selection falls to 0.5579.
These policies prioritize clinical relevance or signal reliability, whereas MI-based policies prioritize statistical association with the training label.
Every VitalDB selector lowers macro AUROC, with temporal MI giving the smallest loss and random selection the largest, particularly for hypotension.
At the displayed eight-channel budget, temporal MI improves gas-discordance AUROC from 0.5559 to 0.5676, the largest gain among selectors; SQI also improves it, to 0.5596.
Clinical-priority and SQI are the stronger MIMIC-III choices under a sensor budget; VitalDB favors all available channels, with temporal MI as its closest reduced-channel comparator.

\noindent\textbf{Missing-input handling.}
Panel~(b) compares eight methods with zero-mask plus $\Delta$ on MIMIC-III and nine methods with architecture-only handling on VitalDB; Figures~\ref{fig:info-tasks-missingness-mimic} and~\ref{fig:info-tasks-missingness-vitaldb} give task-level AUROCs.
Lead-agnostic SSL and GRU-D lead the MIMIC-III macro at 0.6086 and 0.6082 versus 0.6037 for the reference, but all eight alternatives improve acute deterioration and ventilator liberation while reducing AKI and fluid-response AUROC.
GRU-D models observation and missingness histories~\citep{che2018missing}; its mortality gain of 0.0195 contrasts with a fluid-response loss of 0.0128, favoring different handling for prognosis and the short response to an intervention.
On VitalDB, Raindrop approximately matches the architecture-only macro (both 0.6128); the other eight methods lower it by 0.0207--0.0314.
Gas discordance improves with eight methods and accelerating MAP decline with all nine, while oxygen desaturation declines with every method.
Raindrop is the exception for coupled hemodynamic response, improving AUROC by 0.0067 while the other eight methods lower it.
The separate Transformer/Patch companion favors native handling in macro AUPRC (0.078) over GRIN (0.076) and lead-agnostic SSL (0.077).
In the separate VitalDB ensemble test, native handling loses AUROC from 0.619 to 0.583 under 50\% channel dropout, identifying additional sensor loss as a failure mode despite its favorable natural-missingness ranking (Appendix~\ref{app:vitaldb-preprocessing}).
Lead-agnostic SSL and GRU-D are useful MIMIC-III macro candidates, retaining zero-mask plus $\Delta$ for AKI and fluid response; VitalDB favors native handling or Raindrop for macro AUROC, with other methods reserved for gas discordance and MAP decline.

\subsection{Fusion Gains and Their Interpretation}
\label{app:information-fusion}
Figure~\ref{fig:information-multimodal}(c) compares nine fusion methods on MIMIC-III, five on VitalDB, and four on MC-MED with waveform-only and EHR-only controls.
On MIMIC-III, all nine fusion methods improve macro AUROC over waveform-only (0.6062), but the strongest, bilinear fusion (0.6217), remains below EHR-only (0.6322).
EHR-only also exceeds every fusion method for AKI, mortality, sepsis, and ventilator liberation; its mortality AUROC is 0.5898 versus at most 0.5567 for fusion, consistent with prognostic information in clinical records and clinician actions~\citep{beaulieujones2021clinicians}.
Circulatory deterioration instead favors waveform-based prediction: bilinear fusion reaches 0.6901 versus waveform-only 0.6878 and EHR-only 0.6620, aligning with the direct hemodynamic information used for circulatory warning~\citep{hyland2020circulatory}.
VitalDB concatenation leads the macro at 0.7123, followed by weighted late (0.7106) and gated fusion (0.7093), all above EHR-only (0.6902).
Every VitalDB fusion method improves hypotension AUROC despite EHR-only underperforming waveforms, showing that a weak standalone modality can still add useful conditional information.
MC-MED late fusion leads macro AUROC (0.6093 versus waveform-only 0.5669 and EHR-only 0.6065), while gated and late fusion are closely ranked for anemia (0.6708 and 0.6706; Figure~\ref{fig:info-tasks-fusion-mcmed}).
EHR-only leads lactate nonclearance at 0.5360; late fusion is the only fusion method to improve this task over waveform-only, reaching 0.5203 versus 0.5082.
Lactate's preference for EHR information is consistent with the role of metabolic and treatment trajectories in lactate kinetics~\citep{vincent2016lactate}.
The separate matched VitalDB validation, test, and missing-branch controls are detailed in Appendix~\ref{app:vitaldb-fusion}.
Bilinear fusion is the leading MIMIC-III candidate for augmenting waveform models, concatenation, weighted late, and gated fusion form the VitalDB shortlist, and late fusion leads the MC-MED macro; retain EHR-only controls to determine which gains include an additional waveform contribution.

\subsection{Training-Data Budgets}
\label{app:information-subjects}
The comparison holds evaluation cohorts fixed and applies eight selectors, including random selection, under restricted training-data budgets.
MIMIC-III uses quotas of 10/20/30\% of training windows; VitalDB and MC-MED select whole subjects at 5/10/20\% and 10/20/50\%, respectively.
These dataset-specific fractions run from light to dark in Figures~\ref{fig:info-tasks-subjects-mimic}--\ref{fig:info-tasks-subjects-mcmed}; the composite's numeric shade legend illustrates MIMIC-III's fractions.
The earlier MC-MED efficacy/efficiency companion remains a separate validation sweep (Appendix~\ref{app:ees-information}).
Every reduced-data condition lowers MIMIC-III macro AUROC and VitalDB forecast-macro AUROC relative to the full training set.
At 30\% of MIMIC-III windows, random selection leads at 0.5898 versus the full-data 0.6056; NLDR reaches 0.5887 at 20\%, with no further improvement at 30\%.
Task-anchor selection worsens from 0.5044 to 0.4960 as its window quota increases from 10\% to 30\%, so additional transition-focused examples do not recover the performance of random sampling.
VitalDB deficits shrink as subsets grow: at 20\%, gradient matching and noise filtering lead at 0.5941 and 0.5933 versus the full-cohort 0.6125.
On MC-MED, most subsets lower test macro AUROC; at 50\%, diversity and training-dynamics selection reach 0.5673 and 0.5695 versus the full-cohort 0.5669.
Gradient matching and task-anchor selection remain below the MC-MED reference at every displayed budget.
Under data constraints, random selection at 30\% or NLDR at 20\% are MIMIC-III starting points, gradient matching and noise filtering at 20\% form the VitalDB shortlist, and training dynamics and diversity at 50\% are the MC-MED candidates.

\noindent\textbf{Completed VitalDB test: training-subject budgets.}\label{app:training-budget-test}
The later test release evaluates the frozen five-minute Transformer/Patch recipe with gradient matching at 5\%, 10\%, and 20\% of training subjects and a matched full-data comparator.
It averages 25 individual model metrics, giving equal weight to folds within each seed and then to the five seeds, without probability ensembling.
All subject budgets retain 600 training updates, and intervals use 2,000 shared whole-subject draws conditional on fitted models.
At 10\% and 20\%, concurrent BIS AUROC is 0.882 and 0.884 versus 0.883 with all subjects, whereas seven-task forecast AUROC is 0.544 and 0.578 versus 0.630.
The 20\%-minus-full forecast AUROC difference is $-0.0520$ (95\% subject-bootstrap CI $-0.0692$ to $-0.0340$).
The preserved concurrent-state AUROC and lower forecast AUROC indicate that a subset sufficient for estimating current anesthetic state is insufficient for learning future-event discrimination under the same recipe.
BIS AUPRC remains lower at 10\% and 20\% (0.584 and 0.592 versus 0.606), and both paired AUPRC intervals exclude zero in the negative direction.
The full cohort is the stronger forecasting choice, whereas 10--20\% gradient-matched subsets are viable for concurrent-state AUROC, where the observed loss is small.

\subsection{Task-Level Information Availability Results}
\label{app:information-task-results}

\begin{figure*}[!t]
\centering
\includegraphics[width=\linewidth]{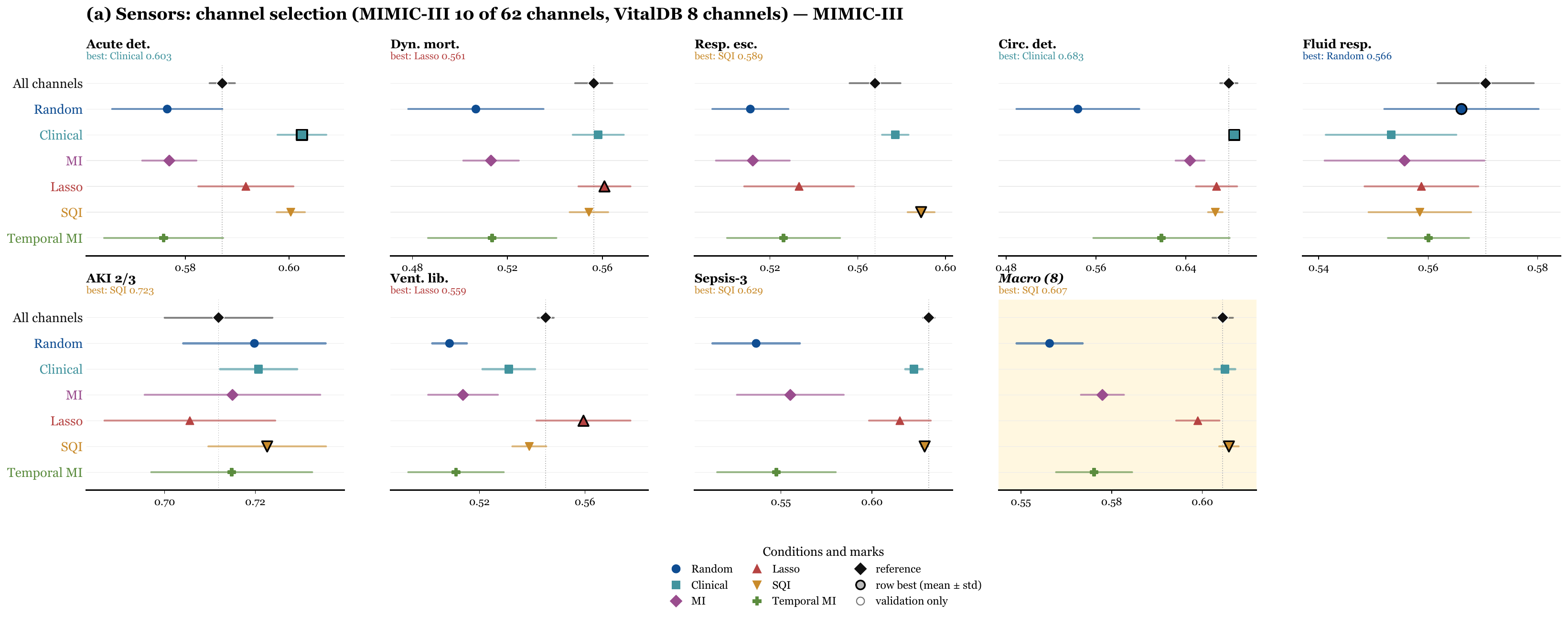}
\caption{\textbf{Channel selection: MIMIC-III.} Comparison of six policies selecting ten of 62 canonical channels with all channels across eight tasks and their macro. Mark and uncertainty definitions follow Appendix~\ref{app:information-task-results}. Numerical values are in Table~\ref{tab:tasks-channel-selection-mimic}.}
\label{fig:info-tasks-sensors-mimic}
\end{figure*}

\begin{figure*}[!t]
\centering
\includegraphics[width=\linewidth]{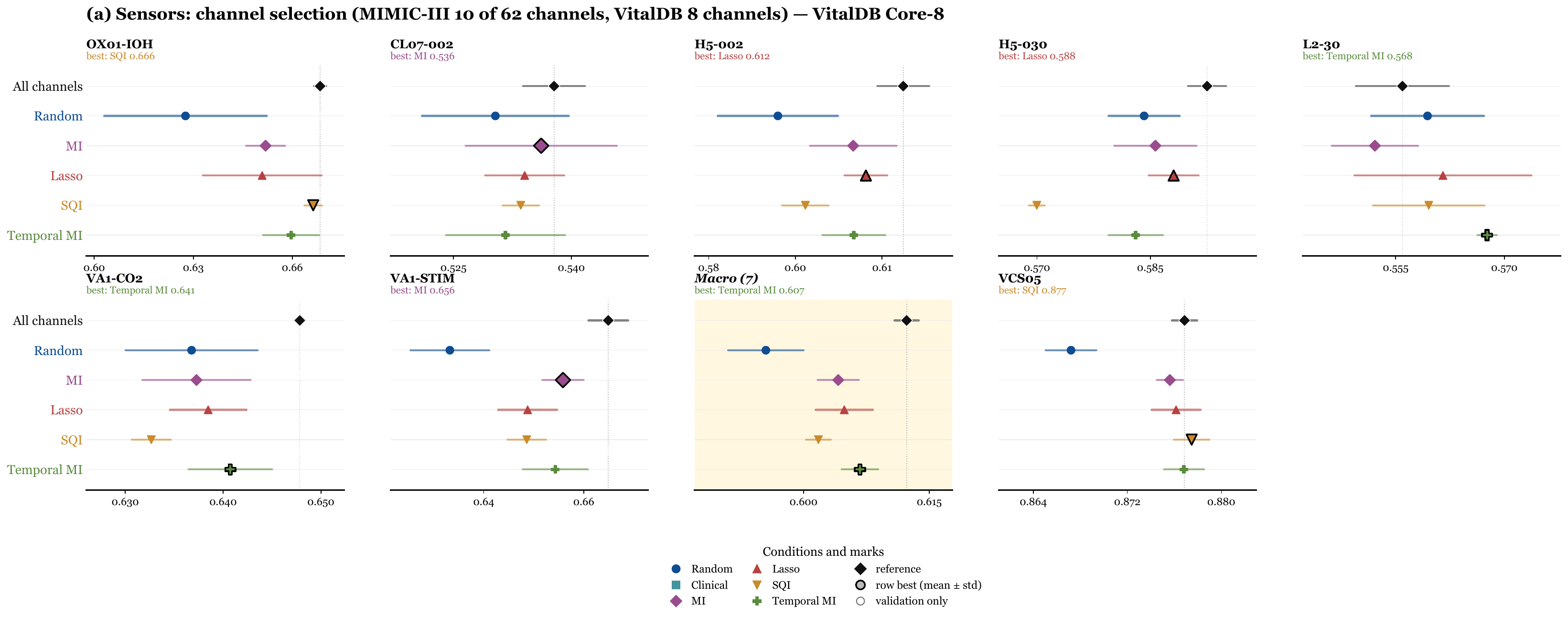}
\caption{\textbf{Channel selection: VitalDB.} Five eight-channel policies against all permitted channels. The macro excludes concurrent VCS05. This fixed-budget AUROC comparison differs from the pooled-budget efficacy/efficiency companion. Mark definitions follow Appendix~\ref{app:information-task-results}. Numerical values are in Table~\ref{tab:tasks-channel-selection-vitaldb}.}
\label{fig:info-tasks-sensors-vitaldb}
\end{figure*}

\begin{figure*}[!t]
\centering
\includegraphics[width=\linewidth]{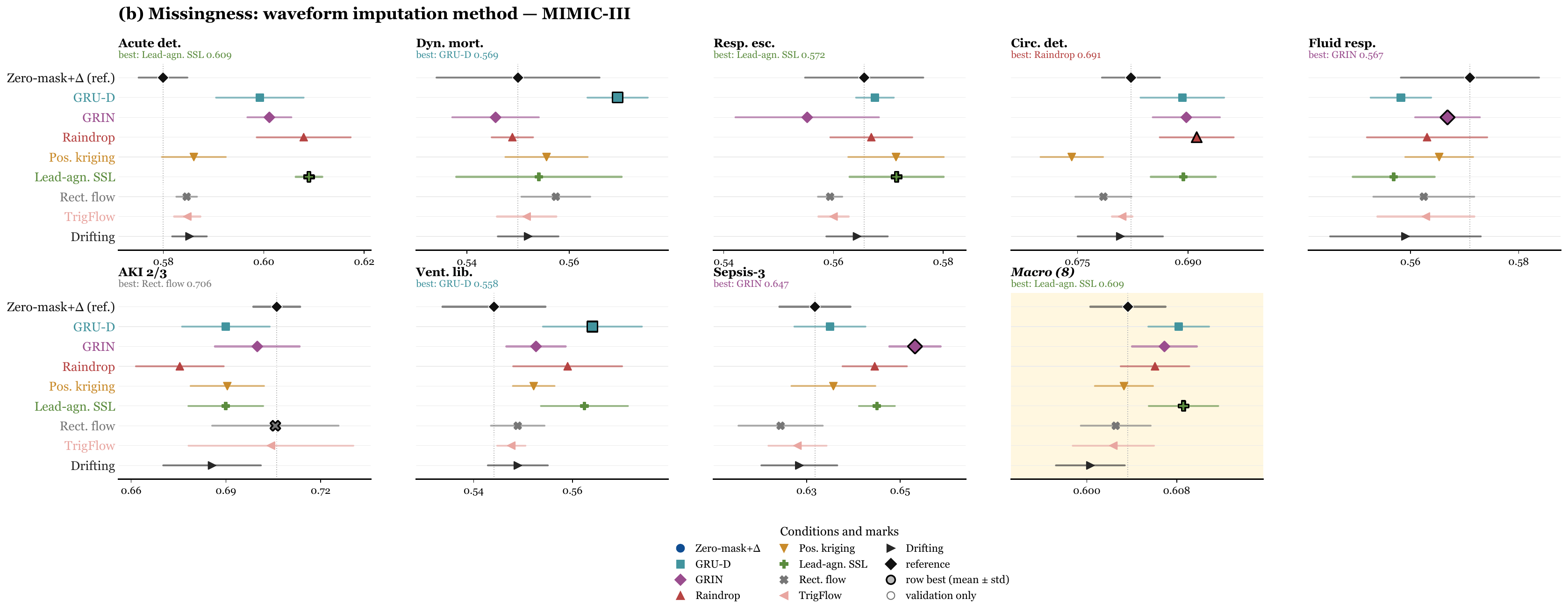}
\caption{\textbf{Missing-input handling: MIMIC-III.} Eight alternatives compared with zero-mask plus $\Delta$ across eight tasks and their macro. The zero-mask reference supplies missingness features without reconstructing channels. Mark and uncertainty definitions follow Appendix~\ref{app:information-task-results}. Numerical values are in Table~\ref{tab:tasks-missingness-mimic}.}
\label{fig:info-tasks-missingness-mimic}
\end{figure*}

\begin{figure*}[!t]
\centering
\includegraphics[width=\linewidth]{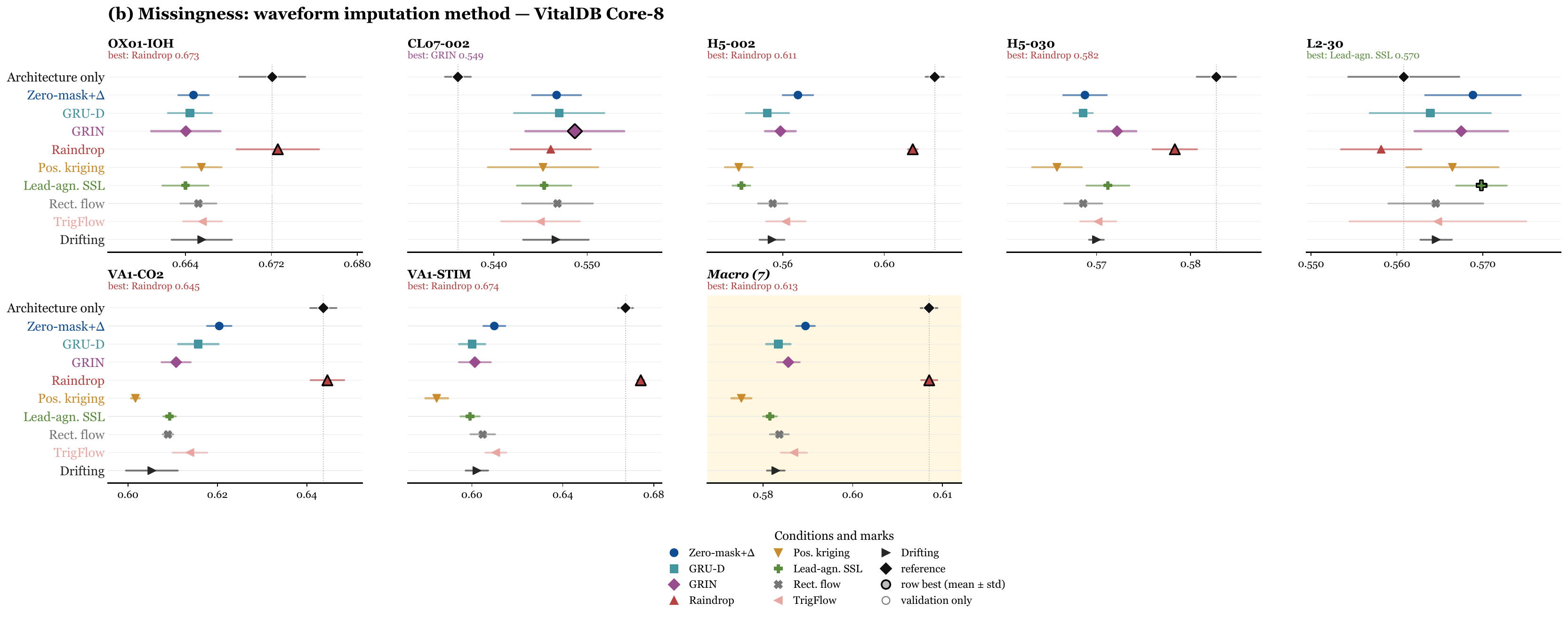}
\caption{\textbf{Missing-input handling: VitalDB.} Nine methods compared with architecture-only handling under natural channel availability, for seven forecasting tasks and their macro. Mark and uncertainty definitions follow Appendix~\ref{app:information-task-results}. Numerical values are in Table~\ref{tab:tasks-missingness-vitaldb}.}
\label{fig:info-tasks-missingness-vitaldb}
\end{figure*}

\begin{figure*}[!t]
\centering
\includegraphics[width=\linewidth]{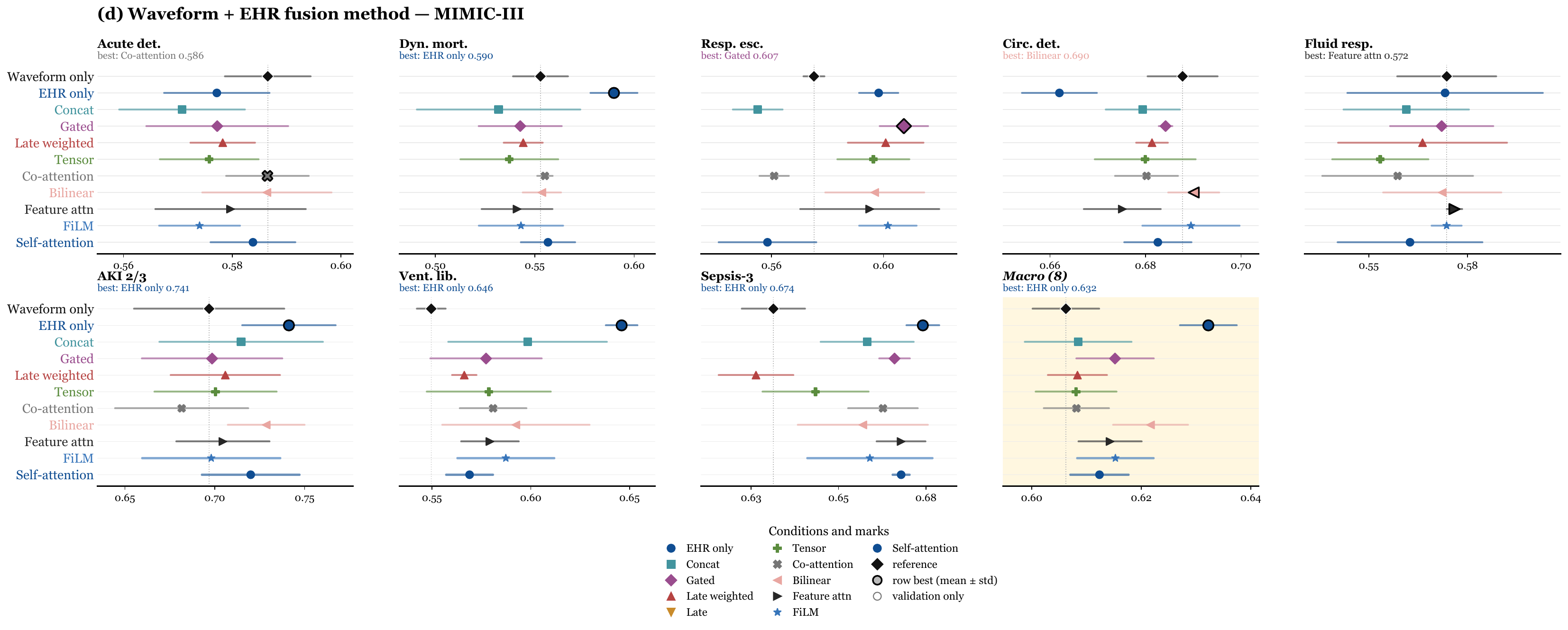}
\caption{\textbf{Waveform--EHR fusion: MIMIC-III.} Nine fusion methods and EHR-only prediction against the Transformer/Patch waveform-only reference across eight tasks and their macro. Mark and uncertainty definitions follow Appendix~\ref{app:information-task-results}. Numerical values are in Table~\ref{tab:tasks-ehr-fusion-mimic}.}
\label{fig:info-tasks-fusion-mimic}
\end{figure*}

\begin{figure*}[!t]
\centering
\includegraphics[width=\linewidth]{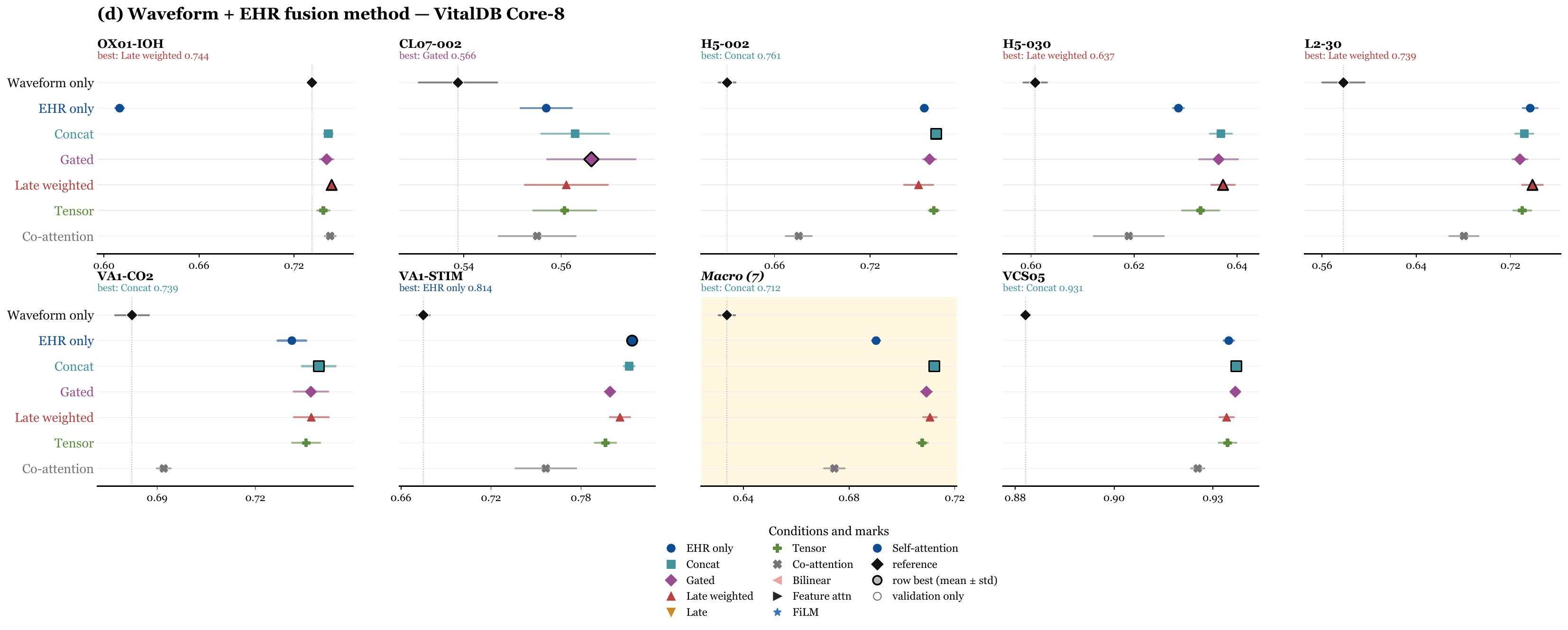}
\caption{\textbf{Waveform--EHR fusion: VitalDB.} Five fusion methods and EHR-only prediction against the Mamba-3/Patch waveform-only reference with 60-minute history. The forecast macro excludes concurrent VCS05. Mark and uncertainty definitions follow Appendix~\ref{app:information-task-results}. Numerical values are in Table~\ref{tab:tasks-ehr-fusion-vitaldb}.}
\label{fig:info-tasks-fusion-vitaldb}
\end{figure*}

\begin{figure*}[!t]
\centering
\includegraphics[width=\linewidth]{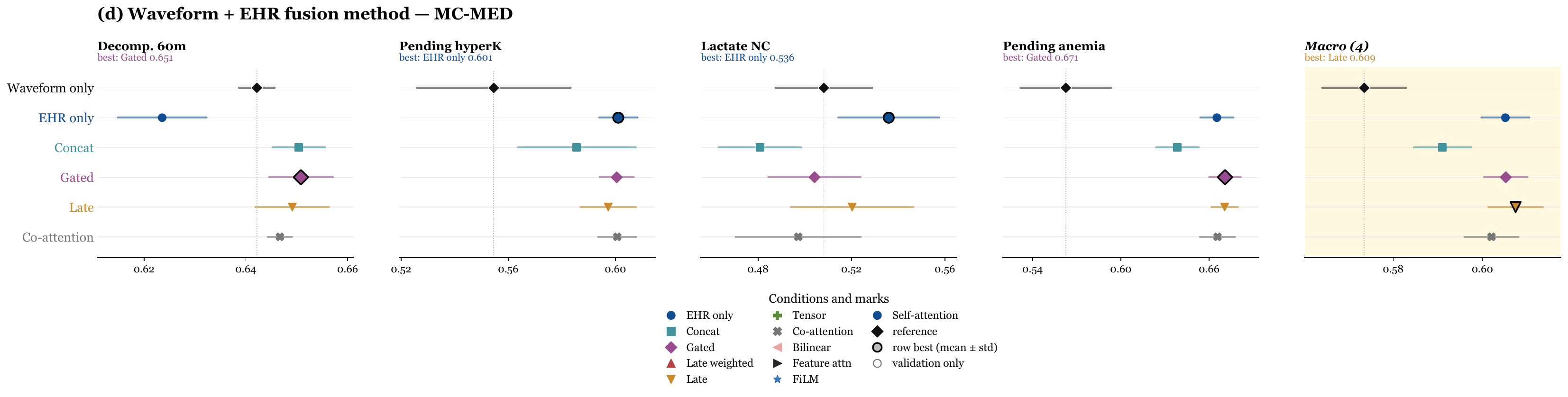}
\caption{\textbf{Waveform--EHR fusion: MC-MED.} Four fusion methods and EHR-only prediction against the Transformer/Patch waveform-only reference with ECG-II and respiration across four tasks. Five-seed test results match the main composite. Mark definitions follow Appendix~\ref{app:information-task-results}. Numerical values are in Table~\ref{tab:tasks-ehr-fusion-mcmed}.}
\label{fig:info-tasks-fusion-mcmed}
\end{figure*}

\begin{figure*}[!t]
\centering
\includegraphics[width=\linewidth]{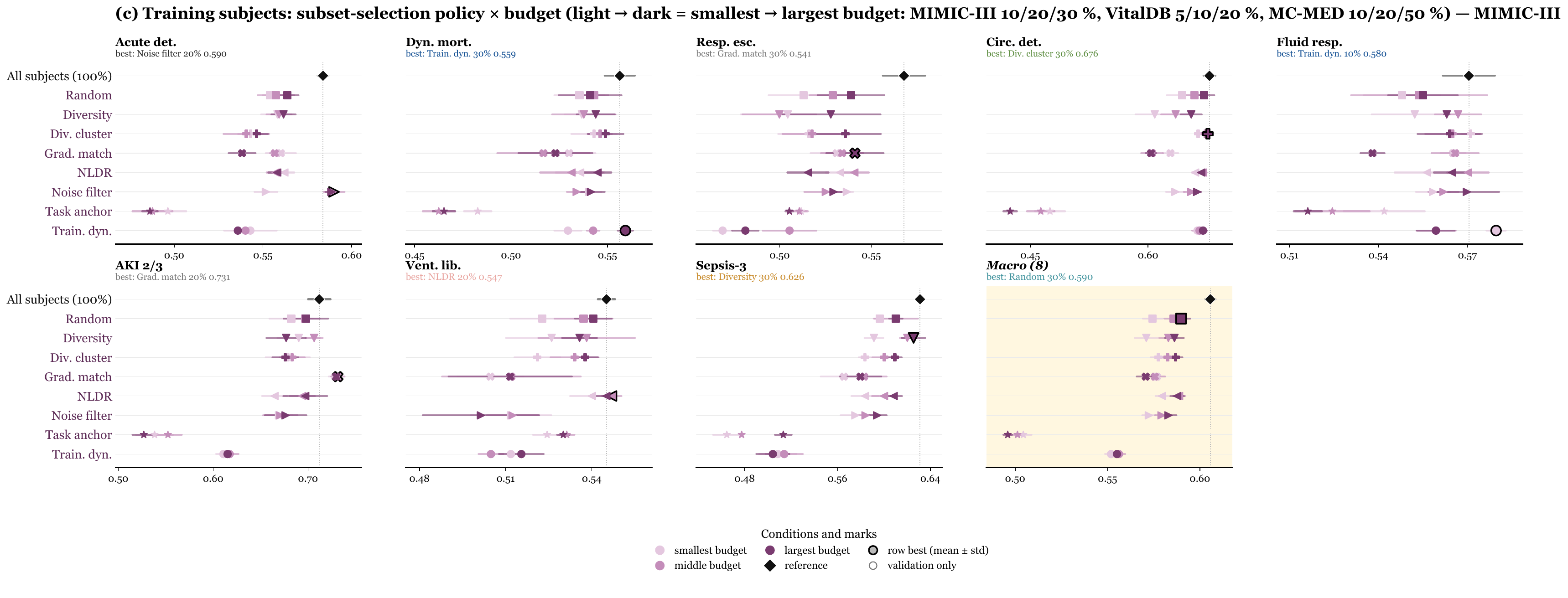}
\caption{\textbf{Training-window selection: MIMIC-III.} Eight policies, including Random, at 10\%, 20\%, and 30\% of training windows, ordered from light to dark, are compared with the full training set. Mark and uncertainty definitions follow Appendix~\ref{app:information-task-results}. Numerical values are in Table~\ref{tab:tasks-training-subjects-mimic}.}
\label{fig:info-tasks-subjects-mimic}
\end{figure*}

\begin{figure*}[!t]
\centering
\includegraphics[width=\linewidth]{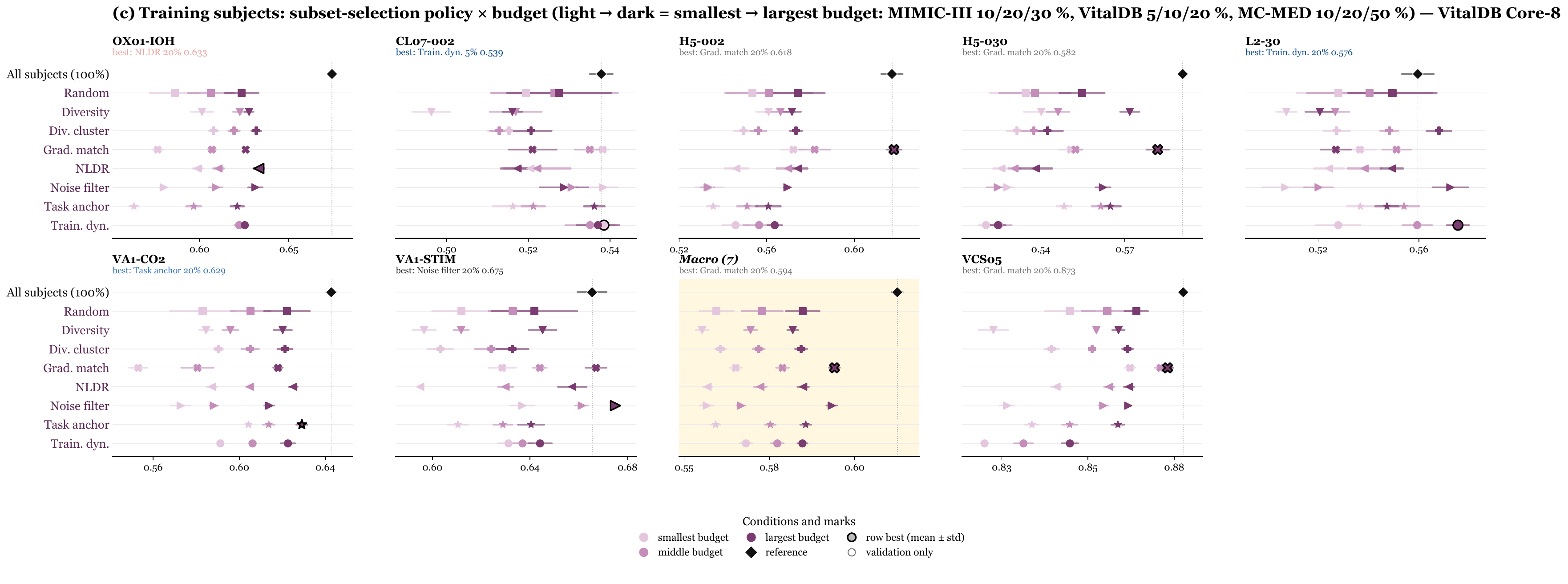}
\caption{\textbf{Training-subject selection: VitalDB.} Eight policies, including Random, at 5\%, 10\%, and 20\% of training subjects, ordered from light to dark, against the full cohort. VCS05 is separate from the forecast macro. Mark definitions follow Appendix~\ref{app:information-task-results}. Numerical values are in Table~\ref{tab:tasks-training-subjects-vitaldb}.}
\label{fig:info-tasks-subjects-vitaldb}
\end{figure*}

\begin{figure*}[!t]
\centering
\includegraphics[width=\linewidth]{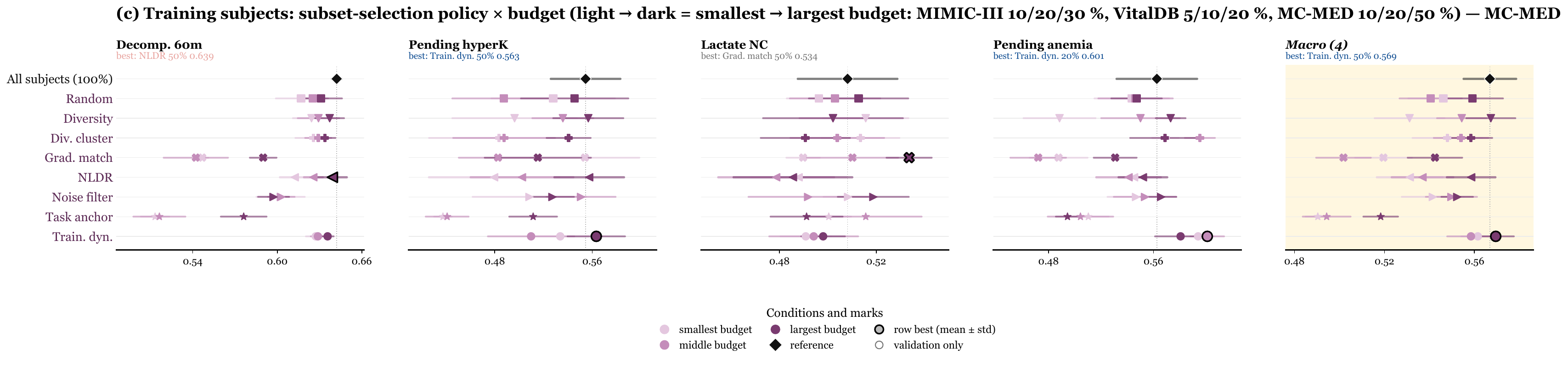}
\caption{\textbf{Training-subject selection: MC-MED.} Eight selectors, including Random, using 10\%, 20\%, and 50\% of training subjects, ordered from light to dark, compared with the full cohort. Results are five-seed test means from the September 16 campaign. Mark definitions follow Appendix~\ref{app:information-task-results}. Numerical values are in Table~\ref{tab:tasks-training-subjects-mcmed}.}
\label{fig:info-tasks-subjects-mcmed}
\end{figure*}

Figures~\ref{fig:info-tasks-sensors-mimic}--\ref{fig:info-tasks-subjects-mcmed} show absolute test AUROC for each task and its dataset macro under the four information-availability comparisons.
Markers show five-seed means and horizontal intervals show one seed SD; the macro interval propagates task SDs as $\sqrt{\sum_t s_t^2}/K$ and omits cross-task covariance.
Black diamonds and vertical dotted lines identify the reference; black outlines mark the highest-mean tested alternative, excluding the reference.
VitalDB macros average seven forecasts, with concurrent VCS05 displayed separately where available.
These figures and the main composite use the same September 17 collection; MC-MED inputs are ECG-II and respiration (Appendix~\ref{app:information-comparisons}).

\subsection{Transfer to MIMIC-IV}
\label{app:information-transfer}

\begin{figure*}[!t]
\centering
\includegraphics[width=\linewidth]{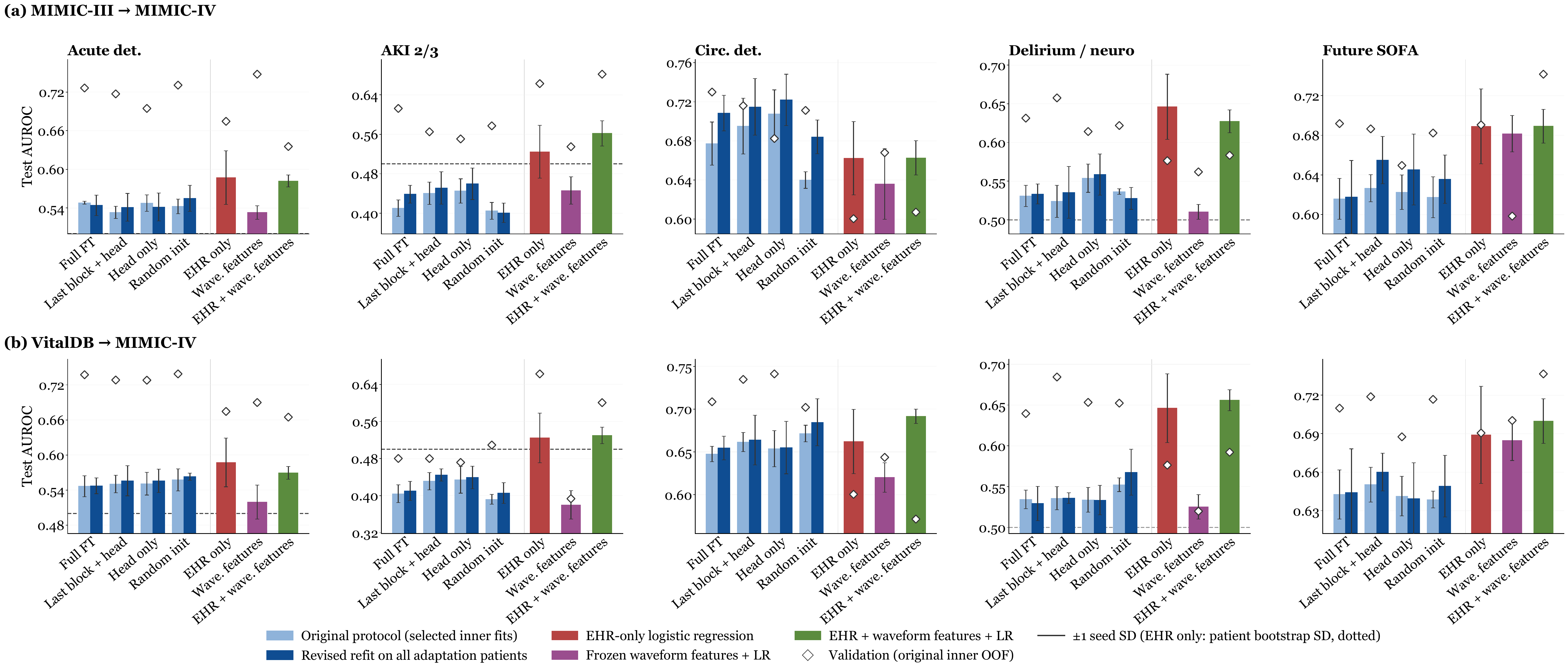}
\caption{\textbf{Detailed transfer to MIMIC-IV: AUROC.} Absolute patient-level test AUROC for five targets, using (a) MIMIC-III and (b) VitalDB source models. Light/dark blue bars show the original selected-inner-fit protocol and revised full-adaptation refits; red, purple, and green bars show EHR-only, frozen waveform-feature, and combined logistic-regression controls. Hollow diamonds are original inner-OOF validation scores. Error bars denote $\pm1$ seed SD, except EHR-only, which uses patient-bootstrap SD (dotted); these are not confidence intervals for paired gains. Inputs, patient roles, and selection follow this subsection. Numerical values are in Table~\ref{tab:transfer-auroc}.}
\label{fig:transfer-mimiciv-auroc}
\end{figure*}

\begin{figure*}[!t]
\centering
\includegraphics[width=\linewidth]{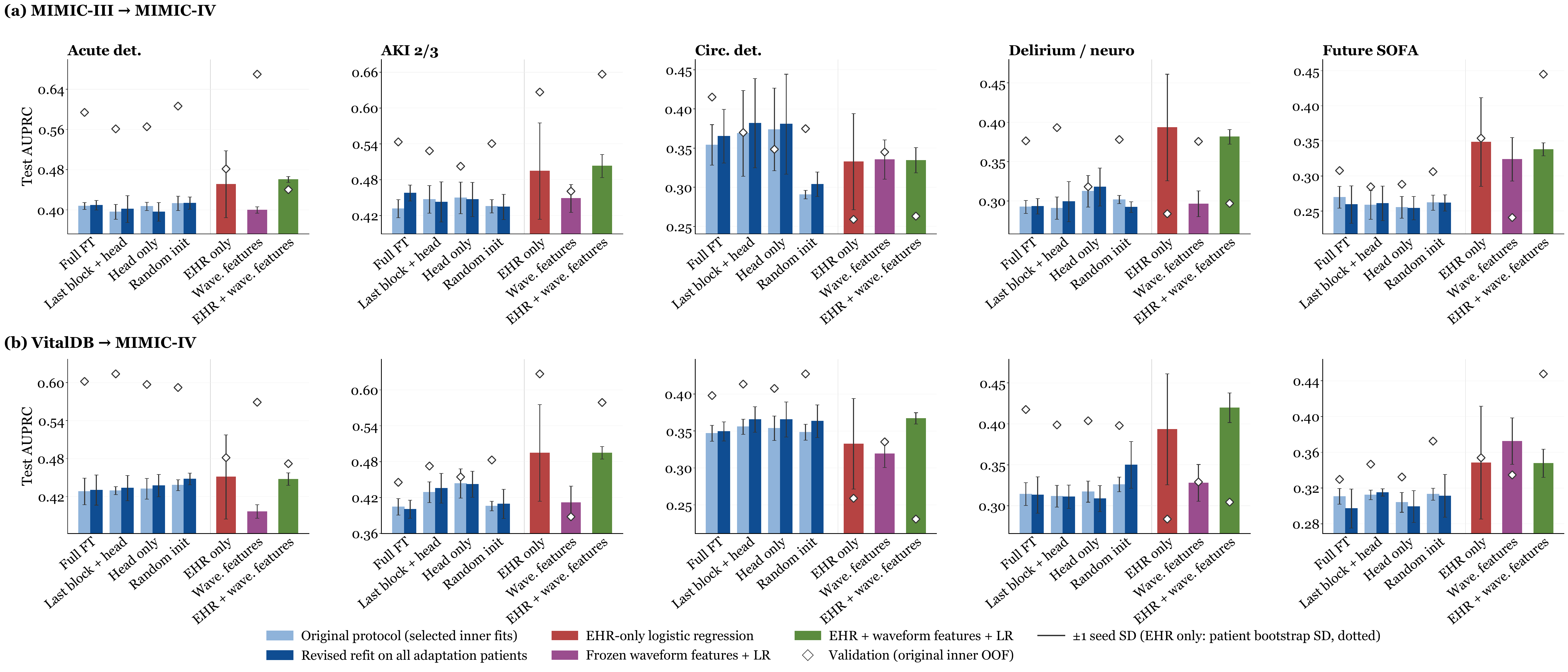}
\caption{\textbf{Detailed transfer to MIMIC-IV: AUPRC.} Patient-level test AUPRC for the same AUROC-selected configurations and targets as Figure~\ref{fig:transfer-mimiciv-auroc}, with (a) MIMIC-III and (b) VitalDB sources. Colors, validation diamonds, and SD bars use the same encoding; EHR-only bars use patient-bootstrap SD rather than seed variability. Compare methods within each target because prevalence differs across targets. Numerical values are in Table~\ref{tab:transfer-auprc}.}
\label{fig:transfer-mimiciv-auprc}
\end{figure*}

\noindent\textbf{Targets and patient roles.}
The two transfer cells in Figure~\ref{fig:information-multimodal}(e) evaluate five waveform-eligible MIMIC-IV targets (Table~\ref{tab:mimiciv-transfer-cohorts}). Figures~\ref{fig:transfer-mimiciv-auroc} and~\ref{fig:transfer-mimiciv-auprc} provide the detailed target-level results.
Each target is an operational forecasting label with its original temporal contract; these target cohorts form an additional adaptation study and are not added to the core 20-task battery or its window count.
Patients overlap across targets, so the table cannot be summed to obtain a unique-patient total.
Mortality has 157 waveform-eligible patients and eight positive patients but is not estimable under this protocol and is omitted.
The target cohorts were inspected in earlier AutoAdapt work, and MIMIC-III/MIMIC-IV patient overlap remains unverified; these transfer results are retrospective development comparisons.


Patients rotate through five outer roles, with approximately 20\% used for adaptation and 80\% for testing in each role.
Each patient enters adaptation once and testing four times; each adaptation role has four inner folds, except AKI role 4, which has three.
Both patient predictions and patient labels are the maxima over that patient's windows.
This patient-level endpoint and its repeated outer-role evaluation differ from the core battery's window-level scores.

\noindent\textbf{Source models and target inputs.}
Both source conditions use our supervised Transformer with the Conv1D tokenizer, approximately 500k parameters in total: 275k in the encoder/tokenizer and 225k in the task head.
For MIMIC-III, the acute-deterioration source model initializes the acute-deterioration, delirium/neurological, and future-SOFA targets; AKI and circulatory-deterioration source models initialize their corresponding targets.
The VitalDB source is the OX01-IOH model with a five-minute forecast gap.
Five source seeds are matched to five target-initialization seeds; the VitalDB checkpoints are also fold matched.
The source head and optimizer are discarded, and each target receives a fresh head.

MIMIC-IV inputs use raw waveform samples from the selected window, passed directly to the model tokenizer; NaN marks unobserved channels.
The MIMIC-III-source condition uses raw physical units on the canonical 62-channel layout.
The VitalDB-source condition pads this layout with four unobserved slots and uses median/IQR normalization fitted on adaptation patients only.
Each condition has a randomly initialized waveform reference with matching inputs and preprocessing.
Cross-source differences therefore combine initialization and preprocessing effects.

\noindent\textbf{Adaptation and model selection.}
The four waveform arms update the full model, the last Transformer block plus head, the head only, or a matched randomly initialized full model.
Each fit uses batch size 128, BF16, at most 600 updates, and learning rates $\{10^{-4},3\!\times\!10^{-4},10^{-3}\}$.
Checkpoint updates and learning rates are selected using inner-validation patient-max AUROC.
The revised protocol used in Figure~\ref{fig:information-multimodal}(e) refits once on all adaptation patients in each outer role, using the selected learning rate and an update budget equal to the lower median of the selected inner update counts.
That refit is evaluated on the role's test patients; it is not an ensemble of the inner-fold models.
The detailed figures also retain the original protocol's selected-inner-fit results for comparison.

\noindent\textbf{Feature-based controls.}
EHR only uses regularized logistic regression on 183 numeric features and one categorical feature summarizing the six hours before the index time, with observed-value indicators.
Waveform features use logistic regression on the source encoder's frozen 128-dimensional embedding, and EHR plus waveform features combines both inputs.
All controls use the same waveform-eligible patients as the waveform arms.
Regularization uses $C\in\{0.01,0.1,1,10\}$; imputation and scaling are fitted on adaptation patients only, and each patient receives equal total weight.
The selected classifier is refitted once on all adaptation patients and scored once on the role's test patients.

\noindent\textbf{Result encoding.}
In Figure~\ref{fig:information-multimodal}(e), bars average AUROC changes across the five targets relative to the revised random-initialization refit for the corresponding source condition; markers show individual target changes.
Blue circles denote acute deterioration, teal squares AKI, purple diamonds circulatory deterioration, red upward triangles delirium/neurological outcomes, and gold downward triangles future SOFA.
Test values average outer-role results over five matched seeds; EHR-only logistic regression is deterministic.
Figures~\ref{fig:transfer-mimiciv-auroc} and~\ref{fig:transfer-mimiciv-auprc} show absolute scores for each target, with MIMIC-III sources in row~(a) and VitalDB sources in row~(b).
Light and dark blue bars distinguish original selected inner fits from revised refits on all adaptation patients; the feature-based controls are shown separately.
Hollow diamonds report the original inner out-of-fold validation scores and should be distinguished from the test bars.
The AUPRC panel evaluates the same AUROC-selected configurations.

\noindent\textbf{Adaptation depth and clinical information.}
MIMIC-III head-only adaptation has the highest macro AUROC among the three tuning strategies, making preservation of the source encoder competitive with updating it.
Circulatory deterioration and AKI improve over random initialization under all three MIMIC-III adaptation strategies.
VitalDB initialization offers no consistent macro benefit, with partial adaptation helping AKI but circulatory deterioration and delirium remaining below random initialization.
For MIMIC-III circulatory deterioration, head-only adaptation improves AUROC while logistic regression on the same frozen encoder features lowers it, locating this difference in the readout and fitting procedure rather than encoder updates.
EHR-only and combined-feature controls give the largest macro gains, about 0.05 AUROC in the VitalDB-source condition, driven mainly by AKI and delirium.
Waveform adaptation's advantage over EHR-only for circulatory deterioration aligns with the relevance of hemodynamic measurements to circulatory failure~\citep{hyland2020circulatory}.
Conversely, EHR-based gains on AKI align with its dependence on renal measurements in the diagnostic criteria~\citep{kdigo2012aki}.
For MIMIC-III-source AKI, partial adaptation has higher AUROC than full fine-tuning, while full fine-tuning has the highest AUPRC among the adaptation arms.
MIMIC-III head-only adaptation is a useful waveform-transfer starting point, with partial and full fine-tuning added to the AKI shortlist according to whether AUROC or AUPRC is the development objective; EHR-only remains a control for the value added by waveform adaptation.

\subsection{Efficacy, Efficiency, and Safety Under Information Constraints}
\label{app:ees-information}
The companion figures report precision--recall performance, training cost, calibration, and selective risk under changes in available information.
They retain earlier experimental exports; the refreshed AUROC comparison uses different model averages and a zero-mask plus $\Delta$ reference for MIMIC-III missingness (Appendix~\ref{app:information-comparisons}).
Appendix~\ref{app:evaluation-details} specifies the definitions and measurement coverage.

\noindent\textbf{Sensor selection.}
Figures~\ref{fig:ees-channel-selection-mimic} and~\ref{fig:ees-channel-selection-vitaldb} compare the predictive and resource costs of channel reduction.
MIMIC-III clinical channel selection retains macro AUPRC 0.096 versus 0.098 with all channels, while peak memory decreases from 3.395\% to 2.353\% and training time from 108.6 to 100.1 seconds.
Its ECE increases from 0.178 to 0.228 and AURC from 0.330 to 0.382, making the memory saving less attractive when confidence quality is a priority.
Clinical selection is useful when MIMIC-III memory use is constrained; all channels provide the stronger reference for calibration and selective risk.

\noindent\textbf{Missing-input handling.}
Figures~\ref{fig:ees-missingness-mimic} and~\ref{fig:ees-missingness-vitaldb} assess whether added missing-input modules improve predictive quality enough to justify their cost.
MIMIC-III GRU-D improves AUPRC from 0.098 to 0.100 and ECE from 0.183 to 0.136, consistent with prior evidence for modeling informative missingness~\citep{che2018missing}.
Its AURC increases from 0.340 to 0.347 and training time from 109.0 to 369.3 seconds, so these benefits require more computation without improving selective risk.
On VitalDB, all nine missing-input methods have higher macro ECE and AURC than architecture-only handling; their AUPRCs span 0.073--0.078 versus the reference 0.078.
Native handling is a practical initial configuration; GRU-D offers better MIMIC-III calibration when its additional training cost is acceptable.

\noindent\textbf{Subject budgets and optimization schedules.}
Figures~\ref{fig:ees-training-subjects-mimic}--\ref{fig:ees-training-subjects-mcmed} compare reduced training cohorts.
At a 20\% budget, MIMIC-III diversity-cluster selection has AUPRC 0.095 and training time 35.0 seconds, versus 0.098 and 108.6 seconds for the full cohort; ECE rises from 0.178 to 0.213.
VitalDB reduced-cohort AUPRCs remain below the full-cohort value of 0.080.
At 20\%, noise filtering and diversity-cluster selection both round to AUPRC 0.072, but noise filtering has lower ECE (0.022 versus 0.044) and AURC (0.029 versus 0.033).
VitalDB's earlier training cells remain between 31.8 and 32.2 seconds under fixed update budgets; the smaller cohorts receive similar scheduled optimization time.
MC-MED subject-budget results and the full-cohort reference both summarize five seeds.
At a 20\% budget, diversity-cluster selection reduces MIMIC-III training time, while noise filtering is preferable to diversity clustering on VitalDB when confidence quality matters.

\noindent\textbf{Fusion across tasks and metrics.}
Figures~\ref{fig:ees-ehr-fusion-mimic}--\ref{fig:ees-ehr-fusion-mcmed} compare waveform--EHR fusion.
MIMIC-III fusion AUPRCs remain between 0.096 and 0.098, while co-attention reduces ECE from the waveform-only 0.183 to 0.136 and AURC from 0.340 to 0.310.
On VitalDB, gated and weighted late fusion have similar AUPRCs (0.1256 and 0.1255), but weighted late fusion has lower ECE (0.0251 versus 0.0306).
For hypotension in this companion study, weighted late fusion improves AUPRC over EHR alone (0.151 versus 0.093), yet remains below waveform-only (0.161).
MC-MED fusion now covers all four endpoints for AUPRC: late fusion records macro 0.210, EHR-only 0.209, co-attention 0.193, and waveform-only 0.179.
For pending hyperkalemia and anemia, EHR-only has higher AUPRC than every fusion method, whereas late fusion leads decompensation and gated fusion leads lactate nonclearance.
MC-MED fusion ECE and AURC cover decompensation alone: late fusion improves its AUPRC from 0.210 to 0.260 but increases AURC from 0.279 to 0.327, while co-attention has the lowest AURC at 0.178.
Weighted late and gated fusion are the leading VitalDB AUPRC candidates, with lower ECE favoring weighted late fusion; late fusion is the strongest MC-MED macro-AUPRC option for waveform augmentation, with EHR-only retained as a control.

\eesfigure{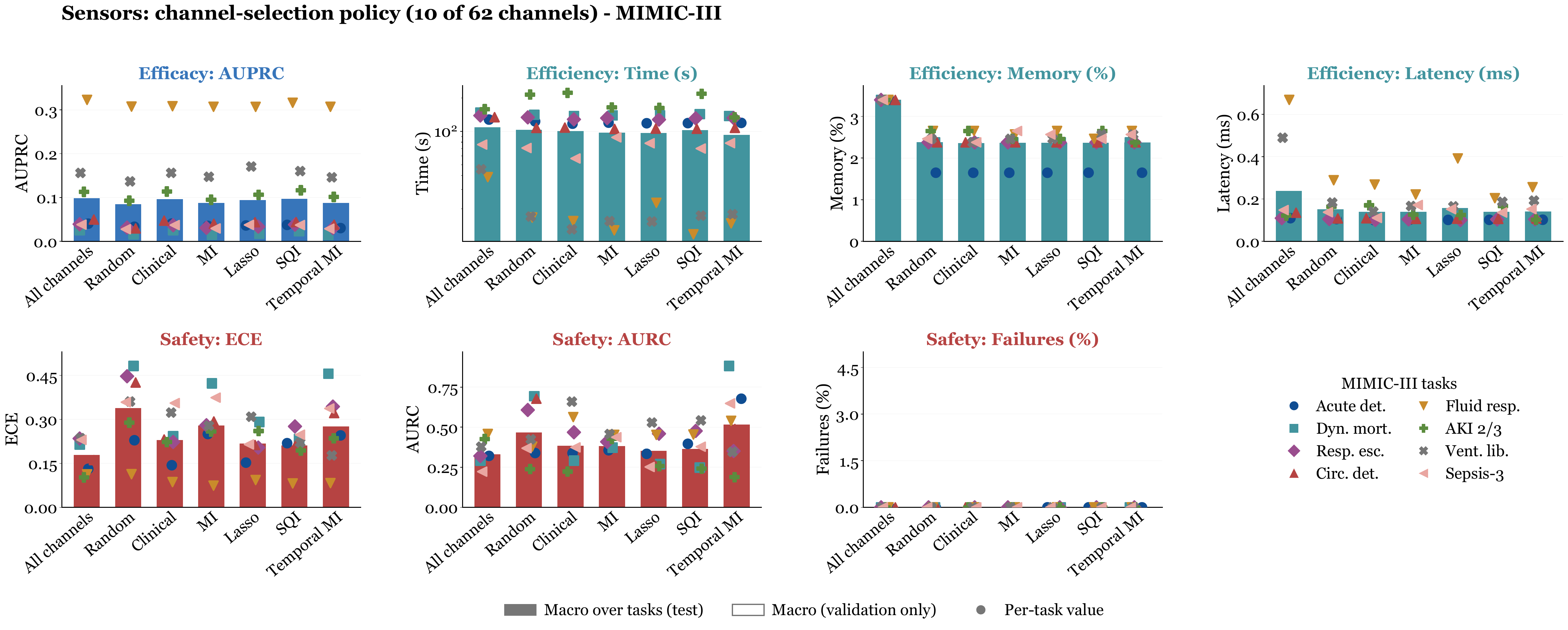}{fig:ees-channel-selection-mimic}{\textbf{Sensor selection --- MIMIC-III.} Ten selected channels out of 62 canonical slots against the all-channel reference. Panels show AUPRC, training time, memory, time per training sample, ECE, AURC, and recorded failures. Metric and mark definitions follow Appendix~\ref{app:evaluation-details}. Numerical values are in Tables~\ref{tab:ees-channel-selection-mimic} and~\ref{tab:ees-channel-selection-mimic-cost}.}

\eesfigure{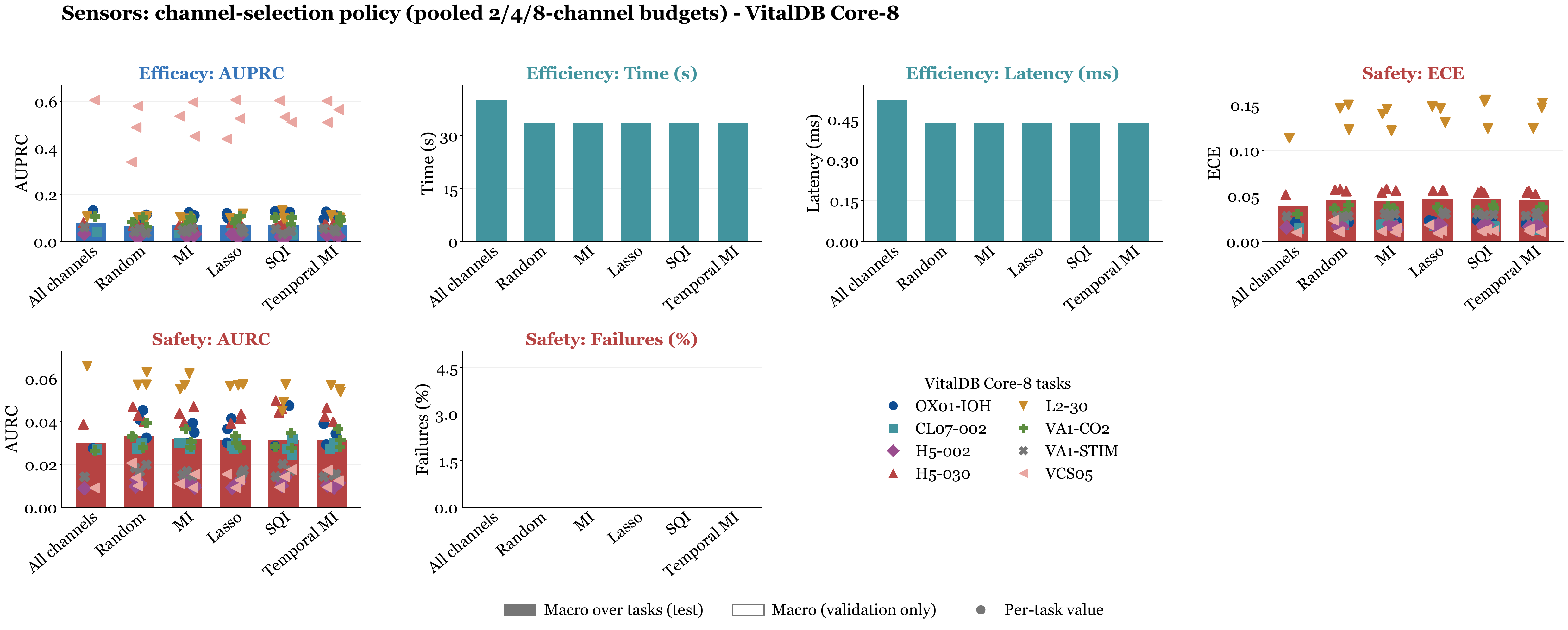}{fig:ees-channel-selection-vitaldb}{\textbf{Sensor selection --- VitalDB Core-8.} Policy means pool two-, four-, and eight-channel budgets against the all-channel reference. Forecast macros exclude concurrent VCS05. Cost panels use earlier-grid telemetry per validation cell. Panels show AUPRC, training time, time per training sample, ECE, AURC, and audited prediction-file failures. Metric and mark definitions follow Appendix~\ref{app:evaluation-details}. Numerical values are in Tables~\ref{tab:ees-channel-selection-vitaldb} and~\ref{tab:ees-channel-selection-vitaldb-cost}.}

\eesfigure{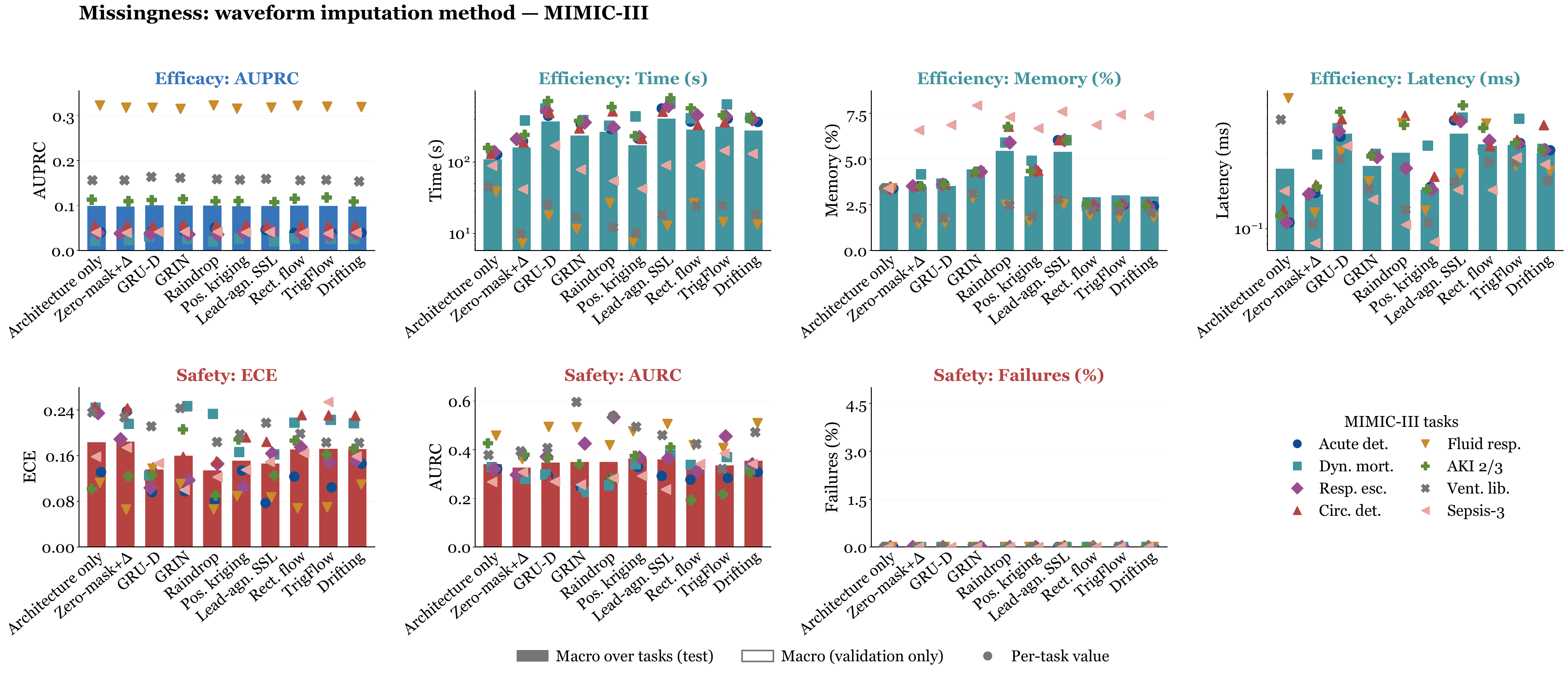}{fig:ees-missingness-mimic}{\textbf{Missing-input handling --- MIMIC-III.} Missing-input methods compared with the architecture-only reference. Panels show AUPRC, training time, memory, time per training sample, ECE, AURC, and recorded failures. Metric and mark definitions follow Appendix~\ref{app:evaluation-details}. Numerical values are in Tables~\ref{tab:ees-missingness-mimic} and~\ref{tab:ees-missingness-mimic-cost}.}

\eesfigure{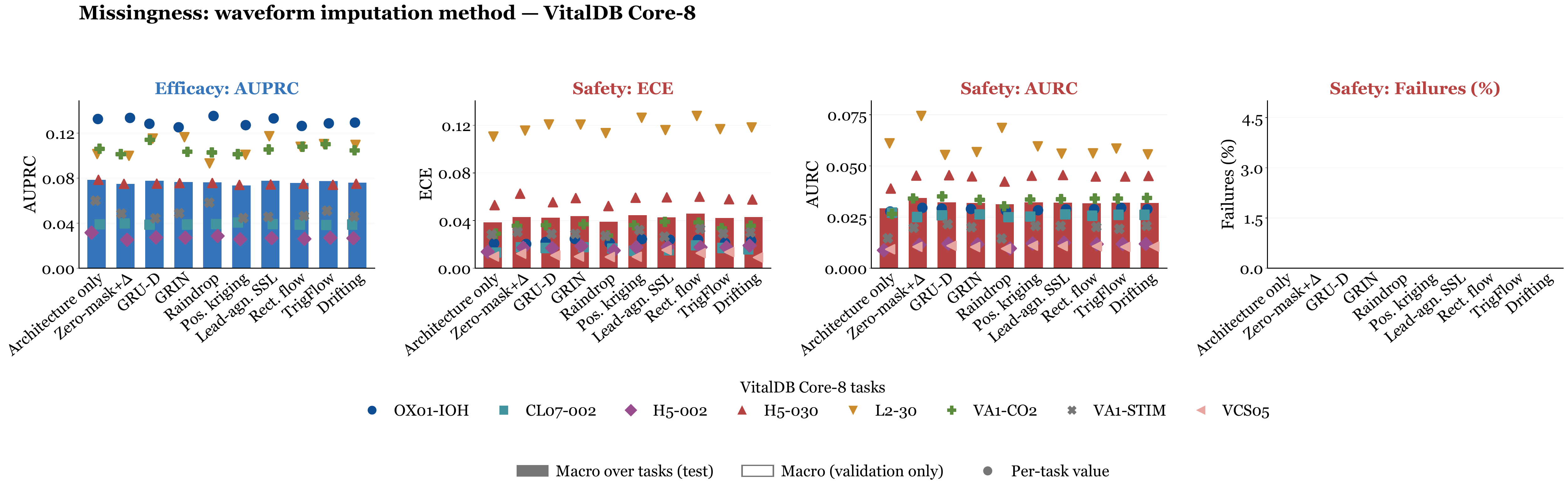}{fig:ees-missingness-vitaldb}{\textbf{Missing-input handling --- VitalDB Core-8.} Missing-input methods against the architecture-only reference. Forecast macros exclude concurrent VCS05. Panels show AUPRC, ECE, AURC, and audited prediction-file failures. Metric and mark definitions follow Appendix~\ref{app:evaluation-details}. Numerical values are in Tables~\ref{tab:ees-missingness-vitaldb} and~\ref{tab:ees-missingness-vitaldb-cost}.}

\eesfigure{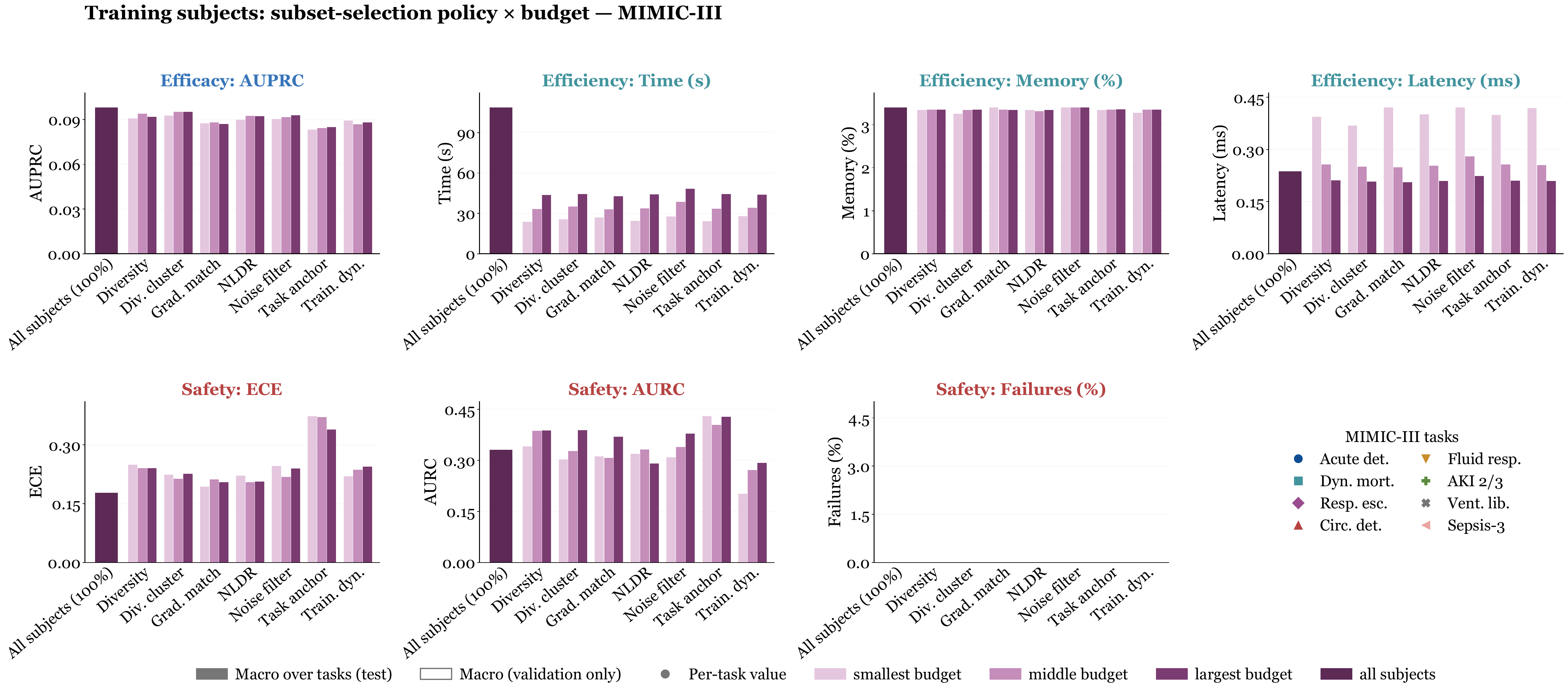}{fig:ees-training-subjects-mimic}{\textbf{Training-subject budgets --- MIMIC-III.} Grouped bars compare 10\%, 20\%, and 30\% subject budgets with the full cohort. Panels show AUPRC, training time, memory, time per training sample, ECE, AURC, and recorded failures. Metric and mark definitions follow Appendix~\ref{app:evaluation-details}. Numerical values are in Tables~\ref{tab:ees-training-subjects-mimic} and~\ref{tab:ees-training-subjects-mimic-cost}.}

\eesfigure{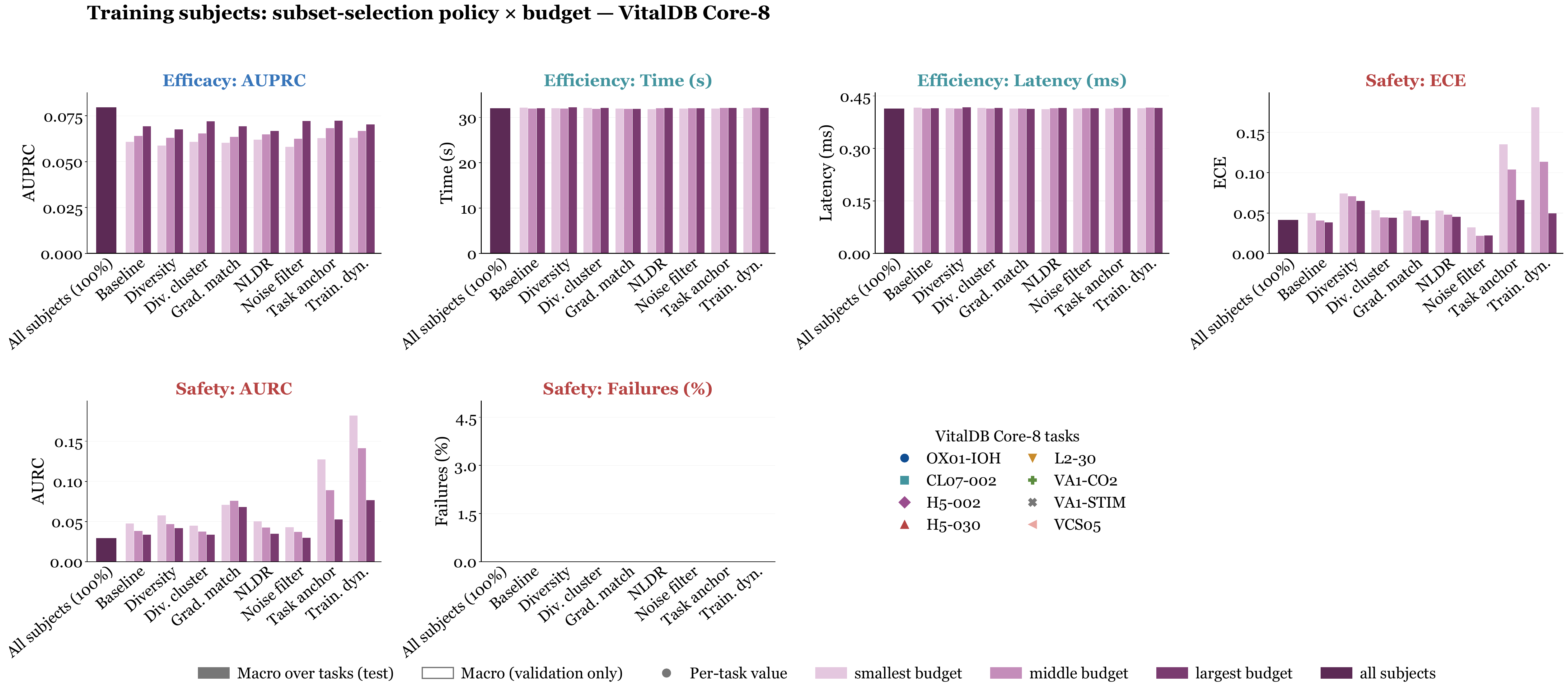}{fig:ees-training-subjects-vitaldb}{\textbf{Training-subject budgets --- VitalDB Core-8.} Grouped bars compare 5\%, 10\%, and 20\% subject budgets with the full cohort. Forecast macros exclude concurrent VCS05. Cost panels use earlier-grid telemetry per validation cell. Panels show AUPRC, training time, time per training sample, ECE, AURC, and audited prediction-file failures. Metric and mark definitions follow Appendix~\ref{app:evaluation-details}. Numerical values are in Tables~\ref{tab:ees-training-subjects-vitaldb} and~\ref{tab:ees-training-subjects-vitaldb-cost}.}

\eesfigure{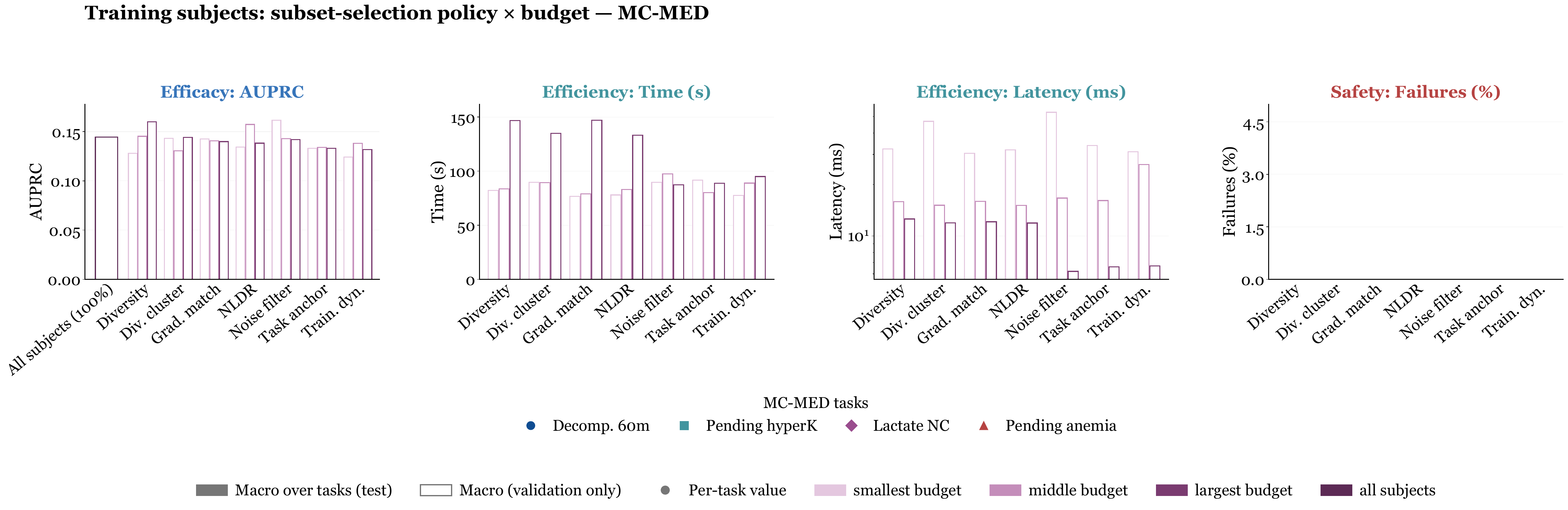}{fig:ees-training-subjects-mcmed}{\textbf{Training-subject budgets --- MC-MED.} Grouped bars compare 10\%, 20\%, and 50\% subject budgets with the full cohort. Hollow bars show validation results averaged over five seeds, including the full-cohort reference. Panels show AUPRC, training time, time per training sample, and recorded failures. Metric and mark definitions follow Appendix~\ref{app:evaluation-details}. Numerical values are in Tables~\ref{tab:ees-training-subjects-mcmed} and~\ref{tab:ees-training-subjects-mcmed-cost}.}

\eesfigure{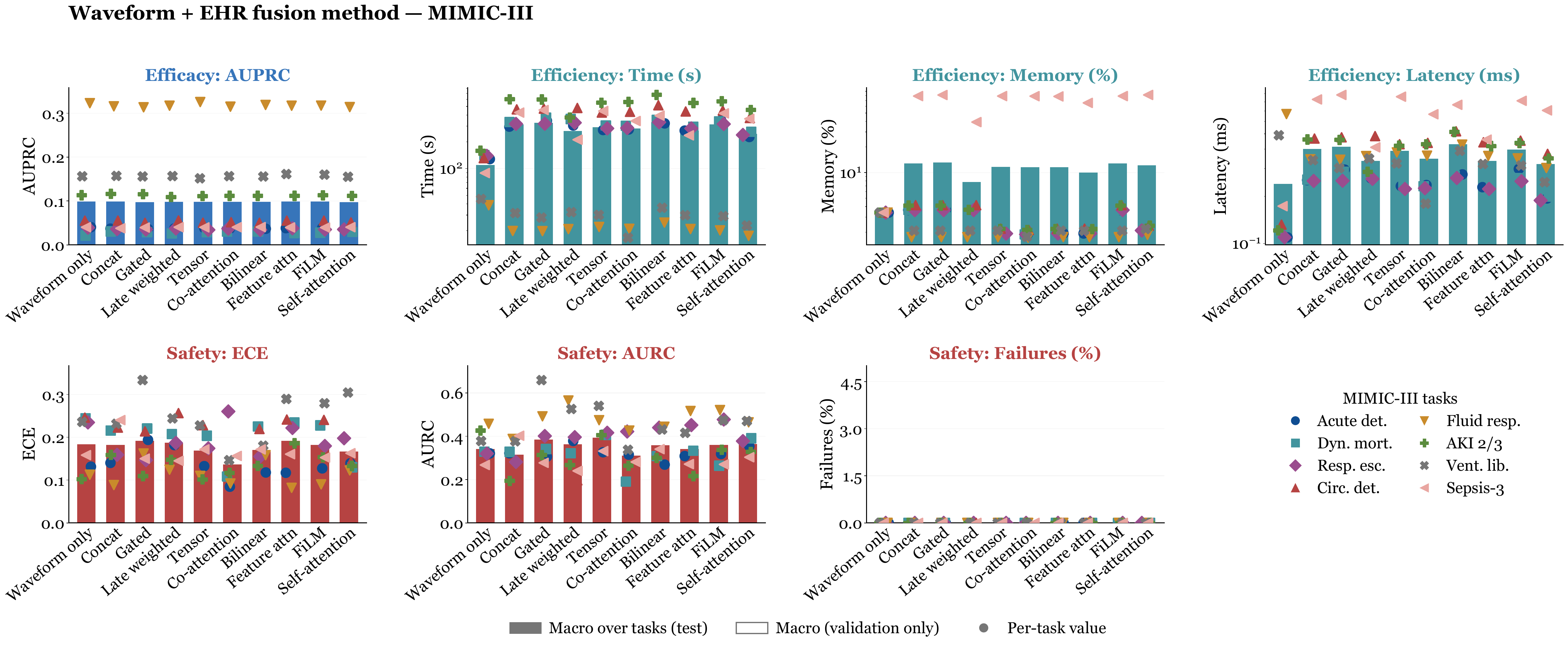}{fig:ees-ehr-fusion-mimic}{\textbf{Waveform--EHR fusion --- MIMIC-III.} Fusion methods against waveform-only prediction. Panels show AUPRC, training time, memory, time per training sample, ECE, AURC, and recorded failures. Metric and mark definitions follow Appendix~\ref{app:evaluation-details}. Numerical values are in Tables~\ref{tab:ees-ehr-fusion-mimic} and~\ref{tab:ees-ehr-fusion-mimic-cost}.}

\eesfigure{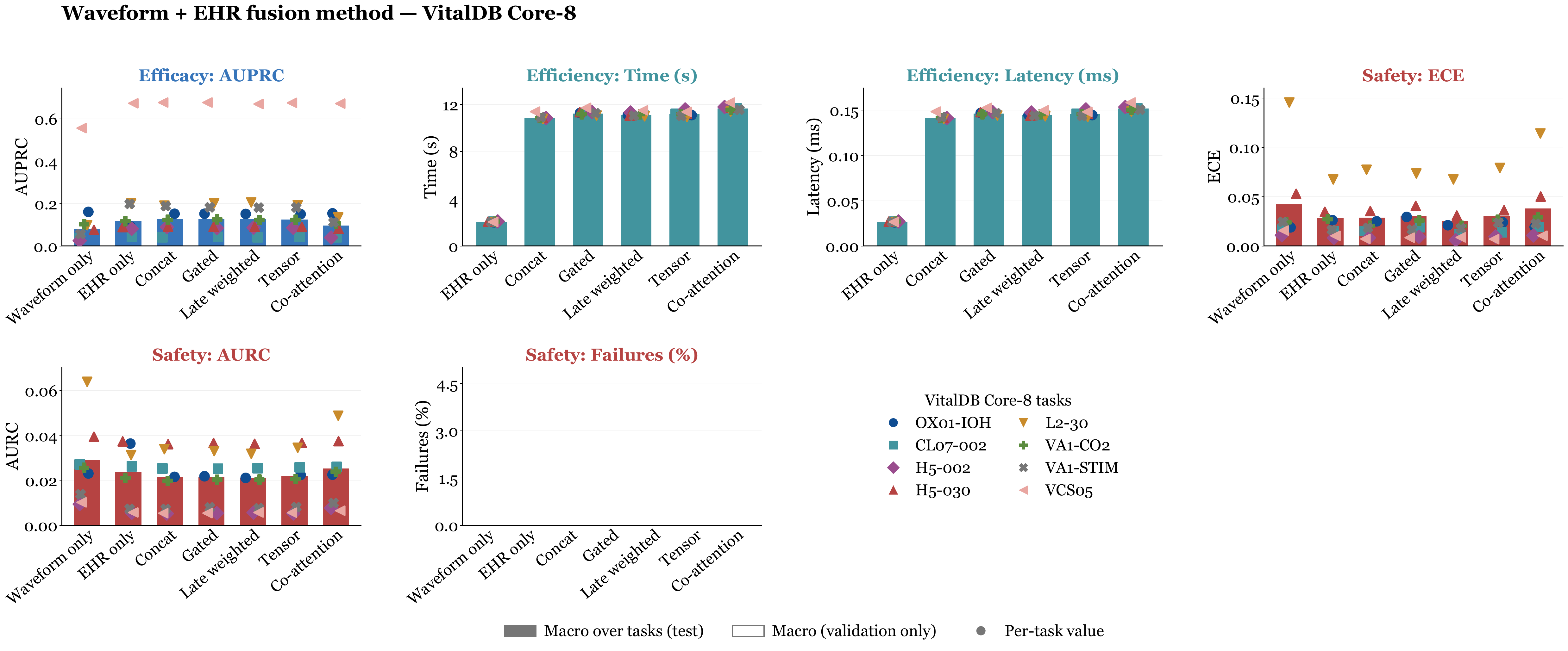}{fig:ees-ehr-fusion-vitaldb}{\textbf{Waveform--EHR fusion --- VitalDB Core-8.} Fusion methods against waveform-only and EHR-only references. Forecast macros exclude concurrent VCS05. Cost panels use earlier-grid telemetry per validation cell. Panels show AUPRC, training time, time per training sample, ECE, AURC, and audited prediction-file failures. Metric and mark definitions follow Appendix~\ref{app:evaluation-details}. Numerical values are in Tables~\ref{tab:ees-ehr-fusion-vitaldb} and~\ref{tab:ees-ehr-fusion-vitaldb-cost}.}

\eesfigure{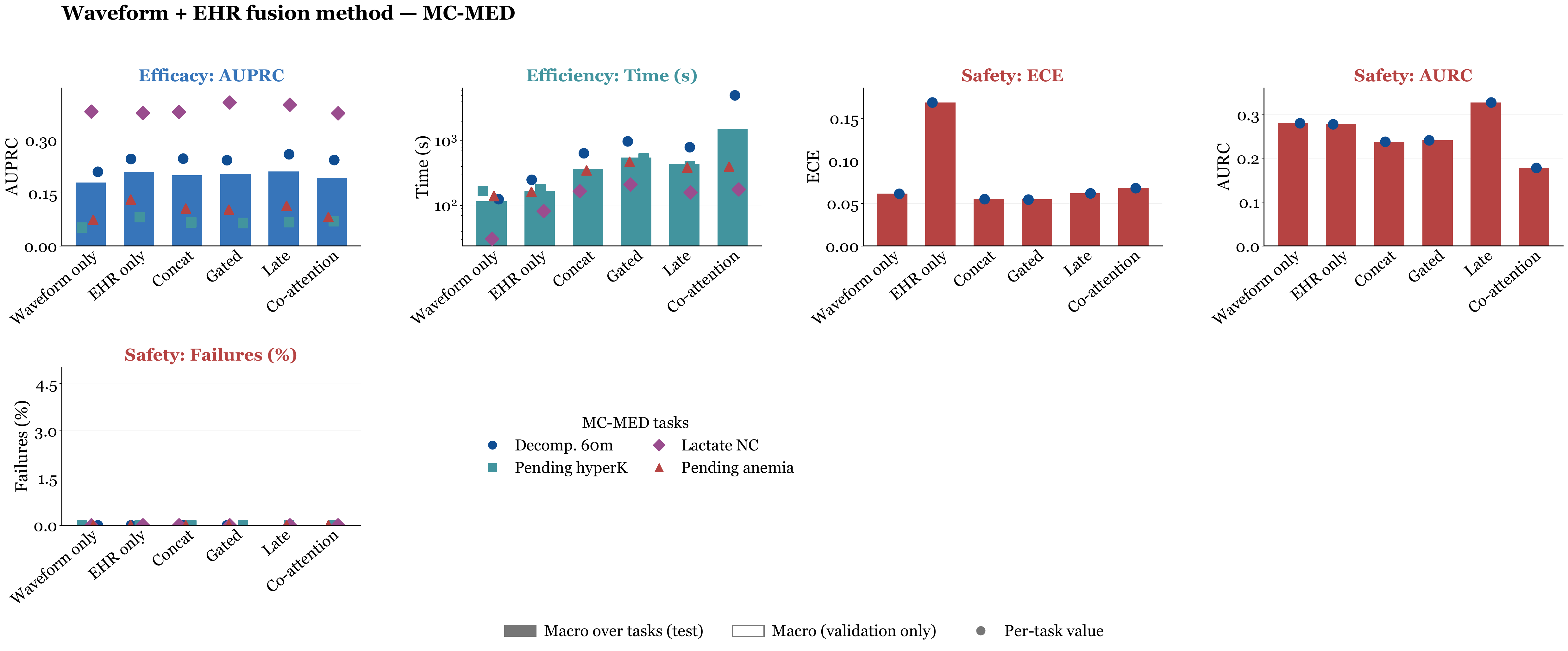}{fig:ees-ehr-fusion-mcmed}{\textbf{Waveform--EHR fusion --- MC-MED.} Fusion methods against waveform-only and EHR-only references. AUPRC, time, and failures cover four tasks; ECE and AURC cover decompensation only. Panels show AUPRC, training time, ECE, AURC, and recorded failures. Metric and mark definitions follow Appendix~\ref{app:evaluation-details}. Numerical values are in Tables~\ref{tab:ees-ehr-fusion-mcmed} and~\ref{tab:ees-ehr-fusion-mcmed-cost}.}

\section{Additional VitalDB Results}
\label{app:vitaldb-results}
\subsection{Architecture, Data, and Missingness}
\label{app:vitaldb-preprocessing}
\begin{figure*}[!t]
\centering
\includegraphics[width=\linewidth]{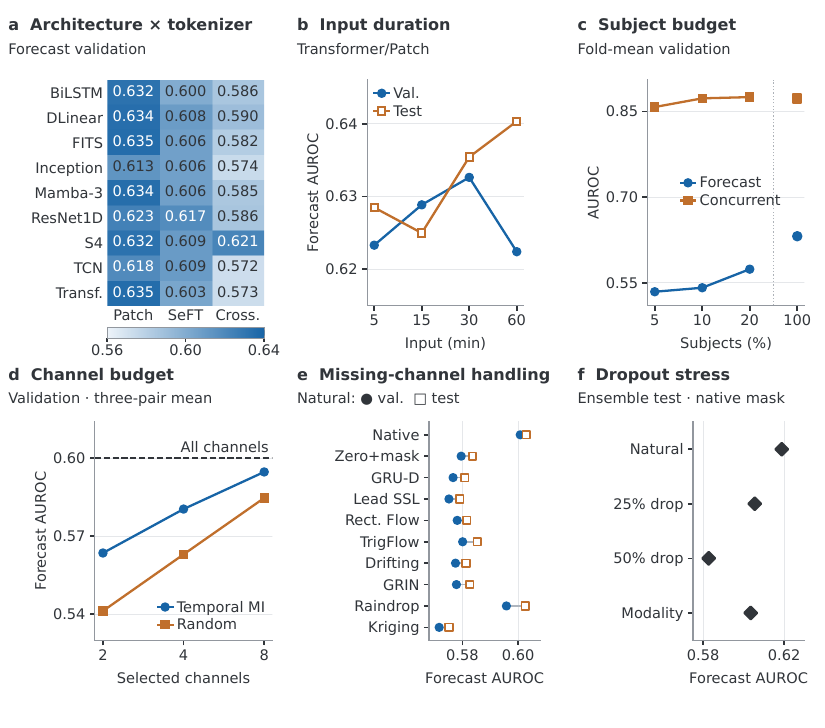}
\caption{\textbf{VitalDB preprocessing and missingness.}
All 27 architecture/tokenizer pairs and ten natural-missingness methods;
selected duration, subject, and channel contrasts. Validation pools held-fold predictions
within seed; test averages fixed-fold model metrics. Subject selection instead
uses fold-mean validation, including its full-data comparator; the later reduced-budget
test is reported separately in Appendix~\ref{app:training-budget-test}. The separate stress panel scores five-fold probability
ensembles on test. These estimators are not interchangeable. Channel and missingness panels average three fixed pairs. Architecture
validation follows recipe selection, not nested evaluation. Numerical values are in Table~\ref{tab:vitaldb-preprocessing-values}.}
\label{fig:vitaldb-preprocessing}
\end{figure*}

\noindent\textbf{Scope.}
The waveform-only grids comprise 27 architecture/tokenizer pairs, 20 duration
conditions, eight subject selectors, 48 channel conditions, and 30 missingness
conditions. Across five seeds and five subject folds, normalization uses training data only;
missing streams and source exclusions are preserved.
Forecast summaries give equal weight to the 5/15-minute gaps within each task, then
to the seven tasks. Concurrent BIS is reported separately, with its EEG and BIS sources
withheld.
Figure~\ref{fig:vitaldb-preprocessing} retains the earlier crossed grid and ensemble-dropout controls, which are not included in the current test-metric figures in Appendix~\ref{app:ees-information}.

\noindent\textbf{Representation and history.}
Patch gives the highest forecast-validation AUROC for every architecture, with Transformer and FITS tied at 0.635.
The encoder ordering nevertheless changes with the tokenizer: Transformer exceeds S4 with Patch (0.635 versus 0.632), while S4 exceeds Transformer with Crossformer (0.621 versus 0.573).
Extending S4/SeFT history from five to 60 minutes lowers concurrent BIS AUROC from 0.812 to 0.607, making this longer-history configuration unfavorable for current-state estimation.
Transformer/Patch and FITS/Patch provide a competitive initial pair from this grid, with the five-minute view sufficient for the concurrent-state comparison.

\noindent\textbf{Data and sensor constraints.}
Gradient matching at 10--20\% of subjects approaches full-data concurrent AUROC but still falls short in forecasting.
Eight temporally selected channels approach full-panel performance, while native masking gives the highest natural-missingness validation mean.
The native-masking ensemble's test AUROC falls from 0.619 to 0.583 under 50\% dropout.
Forecast development can start from native masking and the full cohort, then compare the eight-channel temporal selector under a sensor budget and dropout under additional sensor loss.


\subsection{Forecast Gaps from Five to 75 Minutes}
\label{app:vitaldb-gaps}
\begin{figure*}[!t]
\centering
\includegraphics[width=\linewidth]{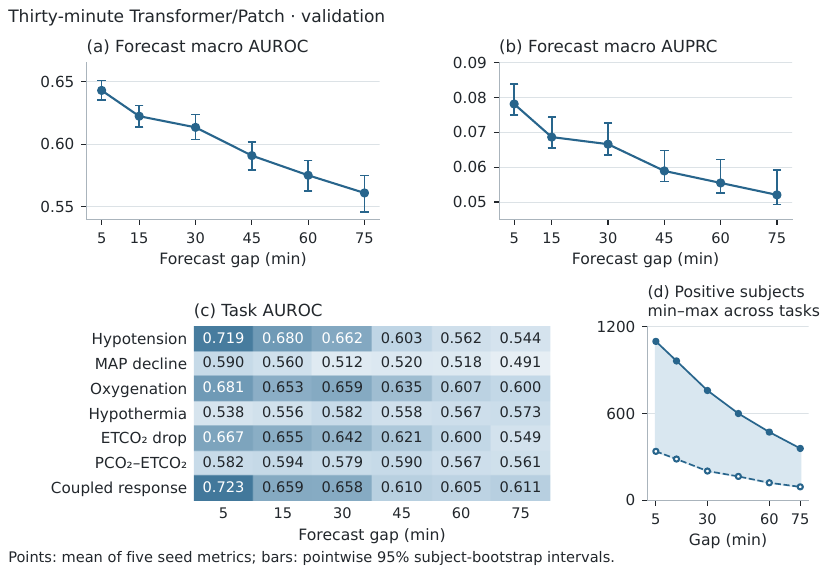}
\caption{\textbf{Longer-gap forecasting.}
Macro discrimination with pointwise 95\% subject-bootstrap intervals,
task AUROC, and the range of positive-window subject counts across tasks.
Support ranges are neither pooled counts nor matched-cohort sizes. A
positive-window subject has at least one positive episode and may also
contribute negative episodes. Eligibility and incidence definitions differ
across gaps; the plotted trend is descriptive. Numerical values and subject support are in Table~\ref{tab:vitaldb-gap-values}.}
\label{fig:vitaldb-gaps}
\end{figure*}

\noindent\textbf{Design.}
A fixed 30-minute Transformer/Patch recipe is evaluated at six gaps on seven
forecast tasks: 1,050 seed--fold fits, including 350 reused 5/15-minute
references and 700 new fits. Figure~\ref{fig:vitaldb-gaps} reports this validation study, separately from the main test curves in Figure~\ref{fig:context-warning}. Metrics pool
held-fold predictions within each seed, average five seeds, then weight tasks
equally. The concurrent probe is excluded.

\noindent\textbf{Results and probability quality.}
From five to 75 minutes, macro AUROC declines from 0.643
(95\% CI 0.635--0.651) to 0.561 (0.546--0.575), and AUPRC from
0.078 to 0.052. Brier score increases from 0.0522 to 0.0608,
log loss from 0.214 to 0.288, and ECE from 0.042 to 0.057.
Task responses differ: core hypothermia is nonmonotonic, while MAP-decline
AUROC at 75 minutes is 0.491 (0.446--0.535). This configuration is therefore unsuitable as a starting point for long-lead MAP-decline warning.
The overall lead-time loss agrees with earlier arterial-waveform hypotension forecasting~\citep{hatib2018machine}, while the nonmonotonic temperature curve shows that this pattern is endpoint dependent.

\noindent\textbf{Support and interpretation.}
At 75 minutes, tasks retain 710--2,251 subjects, including 94--359 with
positive windows. All 42 heads pass support checks; the smallest held fold
contains 130 subjects and 16 with positive windows. Intervals use 2,000 shared
whole-subject Poisson draws over the 4,311-subject union with gap-specific
eligibility masks. They condition on fitted models and omit retraining,
selection, and multiplicity uncertainty. Patient fold assignments remain fixed across gaps,
while follow-up support, event exclusions, and prevalence vary among the eligible anchors.
No temporal-order permutation control was run.

\noindent\textbf{Completed test: warning time.}\label{app:vitaldb-gap-test}
A later test readout of the fixed 30-minute Transformer/Patch recipe averages 25 individual fold/seed model metrics and seven tasks equally at each gap, separately from the validation estimator above.
Macro AUROC decreases from 0.659 (95\% CI 0.638--0.679) at five minutes to 0.558 (0.530--0.585) at 75 minutes; AUPRC decreases from 0.094 to 0.060 and ECE increases from 0.041 to 0.058.
The test readout uses fixed patient partitions, gap-specific eligible anchors, and 2,000 whole-subject draws conditional on fitted models; paired short--long-gap contrast intervals are not supplied.
Its numerical trajectory differs from the main-figure test curve, so these exact scores and intervals remain specific to this fixed-recipe readout.
This Transformer/Patch recipe provides a baseline for five- to 15-minute warning, with additional clinical context evaluated for longer-gap prediction where waveform discrimination is weaker.

\subsection{Waveform--EHR Fusion and Missing-Input Controls}
\label{app:vitaldb-fusion}
\begin{figure*}[!t]
\centering
\includegraphics[width=\linewidth]{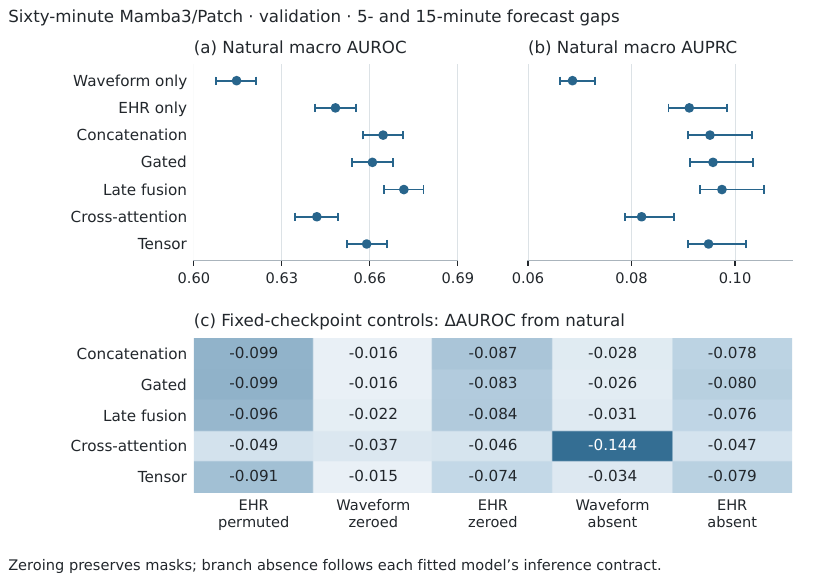}
\caption{\textbf{Fusion discrimination and sensitivity.}
Natural-input forecast means with 95\% subject-bootstrap intervals, and each
control's AUROC difference from its own natural-input model. EHR permutation
shuffles observed feature values within the held fold, preserving recipient
masks. Value zeroing retains masks; branch absence uses each architecture's
frozen missing-branch behavior. Controls are inference interventions, not
retrained unimodal baselines or causal feature importance. Numerical values are in Tables~\ref{tab:vitaldb-fusion-values} and~\ref{tab:vitaldb-fusion-controls}.}
\label{fig:vitaldb-fusion}
\end{figure*}

\noindent\textbf{Design.}
All seven conditions use identical episodes and task-specific exclusions.
Waveform arms use 60-minute Mamba-3/Patch inputs; EHR arms use pre-cutoff
information permitted by the task input rules. The conditions comprise two unimodal references and
five fusion methods. The validation battery has 15 heads
(seven tasks at 5/15-minute gaps and a separate concurrent probe), five folds,
and five seeds. New fits use 600 updates at batch size 128. Forecast aggregation follows
Appendix~\ref{app:vitaldb-preprocessing}.
Figure~\ref{fig:vitaldb-fusion} shows the original validation discrimination and inference-time controls; the current test AUPRC/ECE comparison appears in Figure~\ref{fig:ees-ehr-fusion-vitaldb}.

\noindent\textbf{Validation gains and probability quality.}
The stronger-unimodal envelope selects the better unimodal metric per head and bootstrap draw before aggregation.
Weighted late fusion reaches AUROC 0.6716 versus 0.6483 for EHR and 0.6146 for waveforms.
Its paired AUROC gain over the envelope is 0.0087 (95\% CI 0.0043--0.0127), whereas its AUPRC gain is 0.0016 (\(-0.0012\)--0.0043).
Brier score, log loss, and ECE worsen by 0.00116, 0.00447, and 0.00156, respectively, aligning with the established distinction between discrimination and probability quality~\citep{vancalster2019calibration}.
Concurrent late-fusion AUROC is 0.9069 versus EHR's 0.9060 (paired difference CI \(-0.0026\)--0.0046), leaving little observed benefit from fusion for this state-estimation task.
Weighted late fusion is the stronger candidate for augmenting waveform forecasts; the small concurrent-state gain over EHR-only does not establish a comparable benefit from the waveform branch for that endpoint.

\noindent\textbf{Robustness and uncertainty.}
All 1,875 frozen fusion checkpoints reproduce their natural-input predictions bitwise
before we apply the 9,375 controls. Every method/control pair reduces forecast
AUROC and AUPRC (all 50 unadjusted intervals exclude zero), with task-level
exceptions. Late-fusion EHR zeroing lowers Brier to 0.04478 despite worse
discrimination, whereas EHR branch absence raises it to 0.17294. Intervals
use 2,000 shared subject draws over 4,596 subjects, with the same conditional
uncertainty limits as the gap analysis.
Robustness evaluations should distinguish EHR-value disruption from complete branch absence, because retaining observation masks and removing the branch produce substantially different probability errors.

\noindent\textbf{Completed matched test comparison.}\label{app:vitaldb-fusion-test}
The subsequent test release verifies identical rows, subjects, and labels across waveform and EHR conditions.
It reports the mean of 25 individual model metrics, with equal gap weights within each forecast task and equal weights across the seven tasks; these scores are distinct from both pooled validation and probability-ensemble test estimates.
Weighted late fusion reaches macro AUROC 0.6937 and AUPRC 0.1166, versus 0.6723 and 0.1074 for EHR-only and 0.6247 and 0.0729 for waveform-only.
Relative to EHR-only, its paired AUROC gain is 0.0214 (95\% CI 0.0124--0.0307) and AUPRC gain is 0.0092 (0.0041--0.0155).
Relative to the headwise stronger-unimodal envelope, however, neither gain is conclusive: AUROC $+0.0053$ ($-0.0052$ to $0.0120$) and AUPRC $+0.0012$ ($-0.0060$ to $0.0071$).
ECE is approximately 0.0302 for late fusion versus 0.0306 for EHR-only, with an inconclusive paired difference; relative to the metric-specific envelope, ECE worsens by 0.00149 (0.00010--0.00417).
For blood-gas/ETCO$_2$ discordance, gap-averaged AUROC favors EHR-only over late fusion (0.730 versus 0.697).
The 2,000 shared whole-subject bootstrap draws condition on fitted models and are not adjusted for multiplicity.
Weighted late fusion is a useful waveform-augmentation candidate for this forecast battery, with calibration assessed alongside discrimination and EHR-only retained as a control, especially for gas discordance.

\section{Interpreting Absolute Waveform-Only Performance}
\label{app:absolute-performance}

\subsection{What the Reported AUROCs Establish}
To interpret why many tasks fall below AUROC 0.8--0.9, we examine the available clinical information, effective warning time, and outcome ascertainment.
Table~\ref{tab:absolute-performance-context} anchors the discussion in the numerical comparisons already reported in Table~\ref{tab:main_auroc}.
For example, the MIMIC-III Transformer architecture entry ranges from 0.532 for dynamic mortality to 0.727 for AKI stage~2/3; the corresponding VitalDB forecast entries range from 0.547 for accelerating MAP decline to 0.695 for hypotension.
The best MIMIC-III tokenizer macro, 0.615, comes from a different learning-rate search from the architecture panel, whose highest macro is 0.607 (Appendix~\ref{app:architecture-tokenizer-protocol}).


The concurrent BIS result demonstrates strong current-state discrimination with EEG and BIS-derived inputs excluded, whereas the forecasting tasks require anticipating a different endpoint after the input cutoff.
Within-task evidence comes from the separate VitalDB gap study, where macro validation AUROC falls from 0.643 at five minutes to 0.561 at 75 minutes as warning time and gap-specific eligibility change (Appendix~\ref{app:vitaldb-gaps}).

\subsection{Information, Timing, and Evaluation Constraints}

\noindent\textbf{Restricted information relative to clinical comparators.}
The waveform-only arm omits laboratory values, medication history, and other charted context that may directly inform organ dysfunction and treatment decisions.
For example, the circulatory-failure system of \citet{hyland2020circulatory} reports AUROC 0.94 using a broader clinical information set, while \citet{tomasev2019aki} predicts creatinine-defined AKI from longitudinal EHR data.
Within our VitalDB fusion validation study, EHR-only macro AUROC is 0.6483 versus 0.6146 for waveforms, using matched episodes; these values belong to that study's 60-minute waveform protocol (Appendix~\ref{app:vitaldb-fusion}).
Clinical actions themselves can carry predictive information: \citet{beaulieujones2021clinicians} obtained mortality AUROC 0.89 from administrative and clinician-initiated data.
High waveform-only AUROCs are also possible for particular endpoints and input resolutions, as illustrated by arterial-waveform hypotension prediction~\citep{hatib2018machine}.

\noindent\textbf{Warning requirements and input recency.}
Future-event tasks require information to remain useful beyond the waveform cutoff; concurrent BIS and pending laboratory-result tasks have distinct temporal interpretations.
The MIMIC-III headline baseline uses $[T-5\,\mathrm{min},T)$ from the extracted six-hour history, so its input ends at the prediction anchor.
The target interval $[T+G,T+G+H)$ begins $G$ after this cutoff; for dynamic mortality, the six-hour gap and 24-hour horizon place it 6--30 hours after the cutoff.
Observation-length comparisons extend the input backward from the same $T$, preserving the anchor and outcome window.

\noindent\textbf{Sparse and uneven sensor availability.}
Canonical channel counts describe tensor positions; observed sensor coverage varies by patient.
The MIMIC-III waveform audit covers 2,679 matched records from 2,315 subjects and reports ECG-II in 97.4\% of records, plethysmography in 66.9\%, arterial pressure in 49.6\%, and central venous pressure in 17.8\%.
These are record-level statistics for the audited subset, not per-window coverage rates for the full benchmark.
Consequently, some examples lack the pressure or respiratory signals most closely related to their endpoint, and the available channels can change over time.
Explicit masks preserve information about monitoring decisions and illness severity carried by missingness~\citep{che2018missing}.
The channel and imputation comparisons show that recovery of predictive value is task-dependent (Appendix~\ref{app:information-sensors}).

\noindent\textbf{Input representation and optimization budget.}
The models receive raw waveform samples: a five-minute MIMIC-III channel at 125\,Hz contains 37,500 samples before the selected tokenizer constructs its representation.
The approximately 500k-parameter models and schedules of ten epochs on MIMIC-III or 600 updates on VitalDB define a bounded comparison budget.
The 87,825 training runs distribute this budget across tasks, conditions, hyperparameters, seeds, and folds.

\noindent\textbf{The evaluated population and negative examples.}
Evaluation retains the anchors meeting each task's eligibility and outcome-ascertainment rules, with patient-disjoint partitions and no additional selection of only easy negative examples.
Task-specific stability criteria restrict evaluation to patients meeting the stated pre-anchor conditions.
Selecting positive and negative segments using different blood-pressure ranges can exaggerate apparent discrimination; \citet{enevoldsen2022selection} identified this concern in the development and validation design of the Hypotension Prediction Index.
Likewise, record-wise validation can overestimate generalization to new patients~\citep{saeb2017usecase}.

\noindent\textbf{Outcome support and uncertainty near chance.}
Low prevalence alone does not mathematically reduce population AUROC when the class-conditional score distributions are unchanged: AUROC measures the probability of correctly ordering a positive and a negative example~\citep{hanley1982roc}.
A small number of independent positive patients can nevertheless limit learning, widen uncertainty, and make learning-rate or checkpoint selection unstable.
In MC-MED, validation contains only 23 positive lactate windows, 33 positive anemia windows, and 48 positive hyperkalemia windows (Table~\ref{tab:cohorts}); repeated windows elsewhere in the benchmark do not create independent patient-level evidence.
Sample-size adequacy depends on the outcome frequency, expected predictive signal, and model complexity, rather than a universal events-per-parameter rule~\citep{riley2020sample}.
AUPRC and threshold-specific sensitivity and positive predictive value help interpret prediction under rare outcomes~\citep{saito2015precision}.
Calibration separately assesses whether predicted risks agree with observed event frequencies~\citep{vancalster2019calibration}.

\noindent\textbf{Recorded endpoints and healthcare processes.}
Laboratory ascertainment and interventions such as ventilation or vasopressor initiation reflect both patient physiology and clinical decisions.
Their recorded times can differ from the onset of the underlying physiological change, while treatment after the input cutoff can alter the eventual outcome.
The influence of ordering and recording processes on EHR-derived prediction targets is well documented~\citep{agniel2018biases}.
Outcome-coverage requirements therefore determine which anchors can receive an observed negative label.

\subsection{Why Difficulty Differs Across Clinical Tasks}

\noindent\textbf{MIMIC-III: downstream outcomes and intervention responses.}
Dynamic mortality concerns a delayed trajectory influenced by organ dysfunction, treatment, and comorbidity beyond a short waveform fragment; the composite acute-deterioration endpoint combines several such pathways.
Respiratory escalation and ventilator liberation additionally depend on clinical assessments that are incompletely represented by monitor waveforms.
For extubation, a meta-analysis identifies respiratory, circulatory, neurological, and secretion-management factors, supporting a broader assessment than a single physiological index~\citep{torrini2021extubation}.
The fluid-response task predicts a documented MAP improvement, which is a limited surrogate for a change in cardiac output: changes in MAP did not reliably track cardiac-index responses in the septic-shock study of \citet{pierrakos2012pressure}.
Circulatory deterioration is more directly related to available hemodynamic signals, consistent with its relatively stronger discrimination, while AKI classification uses serum creatinine or urine-output criteria~\citep{kdigo2012aki}, whose trajectories are excluded from the waveform-only input.
Sepsis identification also depends on the operational definition of infection and organ dysfunction; different retrospective definitions produce different cohorts and outcome incidences~\citep{johnson2018sepsis}.

\noindent\textbf{VitalDB: observable states, future transitions, and external influences.}
Hypotension is directly related to arterial-pressure dynamics, whereas predicting accelerating MAP decline requires anticipating a trajectory change rather than identifying an already low pressure.
Oxygen-desaturation prediction can draw on preoperative patient characteristics and intraoperative measurements, including ventilation parameters~\citep{lundberg2018hypoxaemia}.
Core hypothermia is influenced by anesthetic effects, redistribution of body heat, and warming~\citep{sessler2016thermoregulation}.
A short waveform input may not fully represent these influences.
The blood-gas--end-tidal CO$_2$ discordance endpoint also requires a qualifying future blood-gas measurement; capnography is informative about ventilation and perfusion but has important limits as a surrogate for arterial CO$_2$~\citep{nassar2016capnography}.
Abrupt ETCO$_2$ falls may combine cardiopulmonary changes with equipment or ventilation changes, whose antecedents need not be present in the waveform history.
Coupled hemodynamic responses can depend on subsequent stimulation or treatment, and their HR/MAP definition does not identify nociception specifically~\citep{ledowski2019nociception}.
The BIS probe measures a concurrent state and is reported separately from future-event forecasting.

\noindent\textbf{MC-MED: restricted cohorts and laboratory information.}
Decompensation is evaluated among initially stable patients and combines several future vital-sign abnormalities into one endpoint.
The related ED study of \citet{sundrani2023decompensation} predicts the abnormalities separately over 90 minutes and combines monitoring features with triage information, so its AUROCs are not directly comparable with our 60-minute waveform-only composite.
The pending hyperkalemia and anemia tasks estimate the values of tests ordered before $T$; a result arriving after $T$ need not represent a newly developing physiological abnormality.
Large ECG screening studies also differ in scale and acquisition: \citet{galloway2019hyperkalemia} trained on approximately 1.6 million ECGs and evaluated two- and four-lead inputs, while \citet{kwon2020anaemia} studied anemia screening using standard ECG records.
Our corresponding training sets contain 360 and 180 positive windows, respectively (Table~\ref{tab:cohorts}).
Lactate nonclearance depends on metabolic and treatment trajectories over hours~\citep{vincent2016lactate}; its training cohort contains only 518 patients and its validation set 85 windows.


\section{Recommendations for Waveform Model Development}
\label{app:model-development-recommendations}

Architecture shortlists refer to the selected encoder--tokenizer configurations in Table~\ref{tab:main_auroc}(a); tokenizer shortlists refer to its fixed-Transformer comparisons in Table~\ref{tab:main_auroc}(b).
History and input choices come from separate sweeps; their preferred settings have not been jointly evaluated.
Recommendations concern waveform models and their augmentation; EHR-only models serve as controls for the additional value of waveform information.

\subsection{ICU Tasks: MIMIC-III}
\label{app:recommendations-mimic}

\noindent\textbf{Sepsis-3 and acute deterioration.}
With a fixed Transformer, SeFT is the first alternative to Conv1D for Sepsis-3: it raises AUROC from 0.630 to 0.669, and its gain recurs in eight of nine encoders in the interaction study.
For acute deterioration, the corresponding shortlist is FFT and STraTS, with AUROCs of 0.620 and 0.607 versus Conv1D's 0.586.
If one tokenizer must serve the ICU battery, Conv1D is a practical shared baseline (Appendix~\ref{app:representation-test-interactions}).

\noindent\textbf{Circulatory deterioration and AKI stage 2/3.}
For waveform-only encoder searches, TCN and InceptionTime are strong circulatory-deterioration candidates, while Transformer and Mamba-3 lead the AKI comparison.
With a fixed Transformer, Conv1D and mTAND form a useful shortlist for both endpoints; FFT is less attractive for AKI, reducing AUROC from 0.730 to 0.677.
Under a training-time or memory constraint, Conv1D is the better initial choice: the MIMIC-III tokenizer sweep records approximately tenfold higher training time and fivefold higher peak memory for mTAND, with only a small task-AUROC advantage on these two endpoints (Appendix~\ref{app:ees-representation}).
The AKI gap sweep peaks at six hours for AUROC but at three hours for AUPRC; use the required clinical warning interval to fix the comparison, because outcome timing and eligibility also change (Appendices~\ref{app:temporal-comparisons} and~\ref{app:ees-temporal}).
With EHR access, bilinear fusion is a useful waveform-augmentation candidate for both tasks, but only circulatory deterioration exceeds the EHR-only control in this comparison (Appendix~\ref{app:information-fusion}).

\noindent\textbf{Dynamic mortality.}
With waveform-only inputs, BiLSTM and InceptionTime are suitable architecture candidates, while mTAND and iTransformer lead the fixed-Transformer tokenizer comparison.
Ten minutes is the preferred initial history among the tested views: extending to 20 or 60 minutes does not improve mortality AUROC, which remains near 0.517.
For missing-input handling, GRU-D improves mortality AUROC by 0.0195 over zero-mask plus $\Delta$; adding EHR through the tested fusion methods yields at most 0.557, below the EHR-only control of 0.590 (Appendix~\ref{app:information-sensors}).

\noindent\textbf{Respiratory escalation and ventilator liberation.}
For respiratory escalation, the selected BiLSTM, ResNet1D, and Transformer configurations form a close architecture shortlist; when the Transformer is fixed, iTransformer and Crossformer are the leading tokenizer alternatives.
For ventilator liberation, BiLSTM, Mamba-3, and Transformer are reasonable architecture candidates, while FFT and STraTS lead the tokenizer comparison by small absolute margins.
For respiratory escalation, the 60-minute view improves AUROC and AUPRC, while the 20-minute view has lower ECE and AURC and is worth retaining when confidence quality guides clinical review (Appendices~\ref{app:temporal-comparisons} and~\ref{app:ees-temporal}).

\noindent\textbf{Documented fluid response.}
Under a fixed Transformer, Conv1D is the preferred tokenizer, while FITS is a useful candidate when the architecture can also change.
For a broader encoder--tokenizer search, include Crossformer: it improves fluid-response AUROC in eight of nine encoders, despite lowering the fixed-Transformer score.
This interaction makes fluid response a case for joint encoder--tokenizer comparison, rather than selecting the tokenizer from its performance on one backbone (Appendix~\ref{app:representation-test-interactions}).

\subsection{Intraoperative Tasks: VitalDB}
\label{app:recommendations-vitaldb}

\noindent\textbf{Hypotension.}
With waveform-only inputs, Conv1D is the strongest fixed-Transformer tokenizer; Transformer, S4, and FITS provide a closely ranked architecture shortlist.
When more pre-cutoff history is available, 30- and 60-minute inputs are useful candidates because longer history improves both hypotension AUROC and AUPRC, while increasing the forecast gap weakens discrimination (Appendices~\ref{app:temporal-comparisons} and~\ref{app:ees-temporal}).
If EHR inputs are available, compare concatenation, weighted late, and gated fusion with both unimodal references: these methods improve AUROC, but weighted late fusion has lower hypotension AUPRC than waveform-only prediction.
For a battery-level AUPRC and calibration objective, weighted late fusion is preferable to gated fusion when their AUPRCs are similar, because its ECE is lower (0.0251 versus 0.0306; Appendix~\ref{app:ees-information}).

\noindent\textbf{Gas discordance and accelerating MAP decline.}
With a fixed Transformer, FFT leads the gas-discordance tokenizer comparison, with SeFT as the next candidate; MAP decline favors mTAND and Conv1D.
When inputs are incomplete, these endpoints justify targeted comparisons of missing-input methods: eight of nine methods improve gas discordance and all nine improve MAP decline, without establishing one common best imputer.
For gas discordance with EHR access, weighted late fusion is a candidate for augmenting the waveform model, evaluated against EHR-only to measure the waveform branch's additional value (Appendices~\ref{app:information-sensors} and~\ref{app:vitaldb-fusion-test}).

\noindent\textbf{Oxygen desaturation and coupled hemodynamic response.}
For fixed-Transformer development, FFT and Conv1D are the leading desaturation tokenizers, whereas STraTS and SeFT lead coupled response.
Under natural channel missingness, native handling leads desaturation; for coupled response, Raindrop is the only added method that improves AUROC, by 0.0067 (Appendix~\ref{app:information-sensors}).
If desaturation AUPRC is the objective, keep short histories in the comparison, because its AUPRC falls as history increases even while the battery macro improves (Appendix~\ref{app:ees-temporal}).
For longer warning intervals, reevaluate coupled response at the required gap, since its rebound accompanies a shifted outcome window (Appendix~\ref{app:temporal-comparisons}).

\noindent\textbf{Hypothermia and abrupt ETCO$_2$ fall.}
Under an architecture search, DLinear is a strong hypothermia candidate and BiLSTM leads abrupt ETCO$_2$ fall.
With the Transformer fixed, compare FFT and Conv1D for hypothermia, and STraTS, FFT, and Conv1D for ETCO$_2$ fall.
For hypothermia, retain the five-minute history because longer views do not improve AUROC in the reported sweep (Appendix~\ref{app:temporal-comparisons}).

\noindent\textbf{Concurrent light-BIS state.}
For waveform-only state estimation, Transformer/Conv1D with the five-minute view is a strong starting configuration.
Under a restricted subject budget and an AUROC objective, 10--20\% gradient-matched subsets nearly preserve the full-data score; if AUPRC is the objective, the full cohort remains preferable because both subsets lose precision--recall performance (Appendix~\ref{app:training-budget-test}).
The separate fusion validation study finds little concurrent-state AUROC gain over EHR-only, so its results do not favor adding fusion complexity to improve the waveform branch's contribution (Appendix~\ref{app:vitaldb-fusion}).

\subsection{Emergency-Department Tasks: MC-MED}
\label{app:recommendations-mcmed}

\noindent\textbf{Physiological decompensation.}
For waveform-only development, BiLSTM and InceptionTime are the leading architecture candidates (0.649 and 0.648), while Conv1D and Conv2D are nearly tied with a fixed Transformer (both 0.647 at the displayed precision).
FFT is unfavorable for this endpoint, lowering AUROC in all nine encoders in the interaction study.
For an AUROC objective, ten minutes is preferable to 15 minutes for decompensation (0.656 versus 0.631) and also gives the highest four-task macro.
For an AUPRC objective, retain the 15-minute waveform view, which instead improves this task from 0.2763 to 0.3166 (Appendix~\ref{app:ees-temporal}).
If EHR is available and AUPRC is the objective, start with late fusion because it leads decompensation. Its worse AURC makes co-attention worth retaining when selective prediction is the objective (Appendix~\ref{app:ees-information}).

\noindent\textbf{Pending hyperkalemia and severe anemia.}
For waveform-only hyperkalemia, BiLSTM and InceptionTime form the architecture shortlist; Conv1D, Conv2D, and Crossformer lead the fixed-Transformer comparison, while FFT lowers AUROC across all nine encoders.
For anemia, compare DLinear, Mamba-3, and FITS when choosing the encoder, and Crossformer with Conv1D when the Transformer is fixed; Crossformer reaches 0.582 versus Conv1D's 0.563.
If longer recordings are available, ten minutes is the stronger history for both hyperkalemia and anemia; extending to 15 minutes reduces their AUROCs by 0.048 and 0.046, respectively, and also lowers both AUPRCs (Appendices~\ref{app:temporal-comparisons} and~\ref{app:ees-temporal}).
For waveform augmentation, gated and late fusion are the leading anemia-AUROC candidates; the companion study's higher EHR-only AUPRC for both laboratory tasks makes that control necessary when assessing the added waveform contribution (Appendices~\ref{app:information-fusion} and~\ref{app:ees-information}).

\noindent\textbf{Lactate nonclearance.}
Transformer and ResNet1D lead the architecture comparison (0.508 and 0.499); with the Transformer fixed, mTAND, Conv1D, and Crossformer score 0.509, 0.508, and 0.507.
These configurations form a near-chance shortlist; FFT's small mean gain across encoders does not translate into the best fixed-Transformer result.
The 15-minute view is a useful candidate when history can be extended, because both AUROC and AUPRC improve for this endpoint in the duration sweep (Appendix~\ref{app:ees-temporal}).
With EHR access, late fusion is the only tested fusion method to improve lactate AUROC over waveform-only, while gated fusion leads task AUPRC in the companion study; retain EHR-only as a control because the AUROC improvement does not exceed it (Appendices~\ref{app:information-fusion} and~\ref{app:ees-information}).

\subsection{Shared Resource Constraints and Transfer}
\label{app:recommendations-resources-transfer}

\noindent\textbf{Reduced sensors and training data.}
For MIMIC-III under a sensor budget, clinical-priority and SQI selection are stronger initial choices than random or generic statistical selection; clinical selection also reduces memory, at the cost of worse calibration and selective risk.
For VitalDB, retain all available channels when possible, and use temporal MI as the first reduced-channel comparator.
Under a MIMIC-III window budget, random selection at 30\% and NLDR at 20\% are useful starting points; for subject reduction, gradient matching or noise filtering at 20\% of VitalDB subjects and training dynamics or diversity at 50\% of MC-MED subjects form the shortlists.
Full training sets remain stronger for MIMIC-III macro and VitalDB forecast-macro AUROC (Appendix~\ref{app:information-subjects}).
If the operating condition includes additional sensor loss, evaluate that condition separately: natural-missingness performance does not predict the observed AUROC decline under 50\% sensor dropout (Appendix~\ref{app:information-sensors}).

\noindent\textbf{Adapting to MIMIC-IV.}
In the reported comparison of circulatory deterioration with MIMIC-III source weights, head-only adaptation outperforms both random initialization and EHR-only, making it a strong initial choice.
For AKI, include partial adaptation when prioritizing AUROC and full fine-tuning when prioritizing AUPRC; with EHR access, retain EHR-only and combined-feature controls for both AKI and delirium, where they account for much of the gain.
For acute-deterioration and future-SOFA targets, compare head-only waveform adaptation with matched random initialization, using EHR-only to assess whether the waveform models add information beyond clinical context.
Under VitalDB initialization, retain the matched randomly initialized model as a primary candidate because source pretraining gives no consistent macro advantage in this adaptation study (Appendix~\ref{app:information-transfer}).

\section{Numerical Tables and Figure Data}
\label{app:numerical-tables}
\extrafloats{512}
\maxdeadcycles=1000
\makeatletter
\setlength{\@fptop}{0pt}
\setlength{\@fpsep}{12pt}
\setlength{\@fpbot}{0pt plus 1fil}
\makeatother
\raggedbottom
This section collects the cohort, protocol, and numerical result tables. Updated summaries retain the precision supplied in the latest numerical exports; task-level tables retain the data used to draw the corresponding figures. Earlier companion experiments are identified separately, so values from different model searches or scoring estimators are not combined.
A dash denotes an unavailable measurement or an unevaluated combination, not zero. Metric definitions and uncertainty conventions follow Appendix~\ref{app:evaluation-details}.

These tables give the task markers and shared telemetry from each EES figure. Forecast macros exclude concurrent BIS; a condition-level telemetry value is reported once as shared telemetry. Earlier representation and information-availability companions retain their original measurements, whereas the temporal tables use the refreshed collection.

\begin{table}[!t]
\centering
\small
\setlength{\tabcolsep}{4pt}
\renewcommand{\arraystretch}{1.12}
\caption{Short labels and full names of the 20 clinical tasks.}\label{tab:task-name-key}
\begin{tabular}{@{}p{3.6cm}p{9.5cm}@{}}
\toprule
Short label in results & Full clinical task name \\
\midrule
\multicolumn{2}{@{}l}{\textbf{MIMIC-III}} \\
Acute det. /\newline Acute deterior. & Acute deterioration / failure to rescue \\
Dyn. mort. /\newline Dyn. mortality & Dynamic in-hospital mortality \\
Resp. esc. /\newline Resp. escalation & Respiratory deterioration and escalation \\
Circ. det. /\newline Circ. deterior. & Circulatory deterioration and shock \\
Fluid resp. /\newline Fluid response & Rapid documented-fluid response (MAP-response proxy) \\
AKI 2/3 /\newline AKI KDIGO 2/3 & Incident Kidney Disease: Improving Global Outcomes (KDIGO) stage 2/3 acute kidney injury \\
Vent. lib. /\newline Vent. liberation & Ventilator course and liberation \\
Sepsis-3 & Sepsis-3 onset and shock trajectory \\
\midrule
\multicolumn{2}{@{}l}{\textbf{VitalDB Core-8}} \\
OX01-IOH & Sustained intraoperative hypotension \\
CL07-002 & Accelerating mean arterial pressure (MAP) decline \\
H5-002 & Incident oxygen desaturation \\
H5-030 & Incident core hypothermia \\
L2-30 & Blood-gas PCO$_2$--end-tidal CO$_2$ discordance \\
VA1-CO2 /\newline VA1-CO2-DROP & Abrupt end-tidal CO$_2$ (ETCO$_2$) fall \\
VA1-STIM /\newline VA1-STIM-RESPONSE & Coupled hemodynamic response \\
VCS05 & Concurrent light bispectral-index (BIS) state \\
\midrule
\multicolumn{2}{@{}l}{\textbf{MC-MED}} \\
Decomp. 60m & Physiological decompensation within 60 minutes \\
Pending hyperK & Pending hyperkalemia \\
Lactate NC & Lactate nonclearance \\
Pending anemia & Pending severe anemia \\
\bottomrule
\end{tabular}
\end{table}
\begin{table}[!t]
\centering
\small
\setlength{\tabcolsep}{7pt}
\caption{Conservative model-fit counts implied by the reported grids, including specified learning-rate searches, seeds, and folds, with shared conditions counted once.}
\label{tab:experiment-count}
\begin{tabular}{@{}lrrrr@{}}
\toprule
Comparison & MIMIC-III & VitalDB & MC-MED & Total \\
\midrule
Representation & 4,240 & 34,875 & 2,460 & 41,575 \\
Additional histories & 120 & 5,625 & 40 & 5,785 \\
Additional forecast gaps & -- & 700 & -- & 700 \\
Channel selection & 240 & 16,875 & 120 & 17,235 \\
Missing-input handling & 360 & 10,125 & 180 & 10,665 \\
Subject selection & 840 & 7,875 & 420 & 9,135 \\
Fusion and EHR-only models & 360 & 2,250 & 120 & 2,730 \\
\midrule
Total & 6,160 & 78,325 & 3,340 & 87,825 \\
\bottomrule
\end{tabular}
\end{table}
\begin{table}[!t]
\centering
\small
\caption{Official VitalDB waveform streams used by the waveform-only benchmark.
Canonical IDs denote positions in the fixed 66-slot representation; they are not a count of observed channels.
Native rates are nominal, while preprocessing uses the rate recorded in each source file.}
\label{tab:vitaldb-channels}
\begin{tabular}{rlll}
\toprule
ID & Canonical name & VitalDB source track & Nominal Hz \\
\midrule
0  & ECG-II       & \texttt{SNUADC/ECG\_II}    & 500 \\
59 & ECG-V5       & \texttt{SNUADC/ECG\_V5}    & 500 \\
2  & PLETH        & \texttt{SNUADC/PLETH}       & 500 \\
13 & ART          & \texttt{SNUADC/ART}         & 500 \\
8  & CVP          & \texttt{SNUADC/CVP}         & 500 \\
42 & Femoral AP   & \texttt{SNUADC/FEM}         & 500 \\
28 & CO2          & \texttt{Primus/CO2}         & 62.5 \\
62 & Airway pressure & \texttt{Primus/AWP}      & 62.5 \\
63 & EEG1         & \texttt{BIS/EEG1\_WAV}      & 128 \\
64 & EEG2         & \texttt{BIS/EEG2\_WAV}      & 128 \\
4  & CardioQ ABP  & \texttt{CardioQ/ABP}        & 180 \\
65 & CardioQ flow & \texttt{CardioQ/FLOW}       & 180 \\
\bottomrule
\end{tabular}
\end{table}
\begin{table}[!t]
\centering
\scriptsize
\setlength{\tabcolsep}{4pt}
\renewcommand{\arraystretch}{1.12}
\caption{Task-level cohort support across development and test partitions.
MIMIC-III and MC-MED cells give unique patients / clinical episodes / task-labelled windows; episodes are ICU stays and ED visits, respectively.
VitalDB cells give patients / windows because surgical-case counts were not supplied.
Parentheses give positive-window counts where reported.}
\label{tab:cohorts}
\label{tab:task-support}
\label{tab:vitaldb-current-support}
\label{tab:mcmed-task-support}
\resizebox{\textwidth}{!}{%
\begin{tabular}{@{}lccc@{}}
\toprule
\multicolumn{4}{@{}l}{\textbf{MIMIC-III: patient-disjoint partitions}} \\
Task & Train & Validation & Test \\
\midrule
Acute deterioration & 5,151 / 5,761 / 116,929 & 633 / 703 / 14,399 & 644 / 739 / 15,257 \\
Dynamic mortality & 5,327 / 5,953 / 130,504 & 651 / 722 / 15,215 & 894 / 1,052 / 81,359 \\
Respiratory escalation & 4,393 / 4,863 / 127,005 & 556 / 598 / 15,509 & 547 / 622 / 16,966 \\
Circulatory deterioration & 4,788 / 5,320 / 99,417 & 595 / 649 / 12,031 & 877 / 1,020 / 64,619 \\
Fluid response & 1,740 / 1,839 / 5,787 & 209 / 217 / 696 & 247 / 256 / 854 \\
AKI KDIGO 2/3 & 3,844 / 4,126 / 130,944 & 466 / 490 / 16,117 & 485 / 514 / 15,288 \\
Ventilator liberation & 2,064 / 2,188 / 9,302 & 239 / 252 / 1,120 & 246 / 262 / 1,239 \\
Sepsis-3 & 3,896 / 4,190 / 51,482 & 480 / 510 / 6,635 & 729 / 816 / 31,246 \\
\midrule
\textbf{Union / task totals} & 6,787 / 7,753 / 671,370 & 851 / 964 / 81,722 & 897 / 1,055 / 226,828 \\
\addlinespace[5pt]
\multicolumn{4}{@{}l}{\textbf{VitalDB Core-8: development uses five outer whole-subject folds}} \\
Task & \multicolumn{2}{c}{Development} & Test \\
\midrule
Intraoperative hypotension & \multicolumn{2}{c}{2,443 / 27,855 (1,252)} & 355 / 3,442 (189) \\
Accelerating MAP decline & \multicolumn{2}{c}{2,447 / 28,903 (918)} & 365 / 3,927 (123) \\
Oxygen desaturation & \multicolumn{2}{c}{4,175 / 61,293 (849)} & 608 / 8,395 (121) \\
Core hypothermia & \multicolumn{2}{c}{3,437 / 40,665 (2,164)} & 514 / 5,736 (314) \\
$\mathrm{PCO_2}$--$\mathrm{ETCO_2}$ discordance & \multicolumn{2}{c}{1,297 / 9,338 (782)} & 169 / 1,078 (65) \\
Abrupt $\mathrm{ETCO_2}$ fall & \multicolumn{2}{c}{3,580 / 32,602 (1,629)} & 500 / 4,409 (204) \\
Coupled hemodynamic response & \multicolumn{2}{c}{2,238 / 23,265 (634)} & 333 / 3,240 (84) \\
Concurrent light BIS state & \multicolumn{2}{c}{4,221 / 112,425 (5,231)} & 705 / 19,094 (890) \\
\midrule
\textbf{Union / task totals} & \multicolumn{2}{c}{4,596 / 336,346 (13,459)} & 750 / 49,321 (1,990) \\
\addlinespace[5pt]
\multicolumn{4}{@{}l}{\textbf{MC-MED: episodes are ED visits; one window per visit and task}} \\
Task & Train & Validation & Test \\
\midrule
\shortstack[l]{Cardiorespiratory decompensation\\(60 min)} & 7,875 / 9,016 / 9,016 (1,106) & 1,321 / 1,390 / 1,390 (167) & 2,100 / 2,288 / 2,288 (289) \\
Pending hyperkalemia & 8,171 / 9,422 / 9,422 (360) & 1,440 / 1,529 / 1,529 (48) & 2,139 / 2,364 / 2,364 (90) \\
Lactate non-clearance & 518 / 540 / 540 (145) & 84 / 85 / 85 (23) & 187 / 196 / 196 (70) \\
Pending severe anemia & 4,715 / 5,130 / 5,130 (180) & 896 / 929 / 929 (33) & 1,388 / 1,490 / 1,490 (54) \\
\midrule
\textbf{Union / task totals} & 11,051 / 13,216 / 24,108 (1,791) & 1,891 / 2,044 / 3,933 (271) & -- / -- / 6,338 (503) \\
\bottomrule
\end{tabular}%
}
\par\smallskip
\begin{minipage}{\textwidth}
\scriptsize
Union patients and clinical episodes are deduplicated across tasks within each partition; windows and positive windows are summed over task-labelled rows.
The same stay or visit can therefore contribute to several tasks.
VitalDB development is counted once, not once as training and again as validation.
Its forecast-window counts include the reported 5- and 15-minute gap heads; concurrent BIS is a separate task.
-- denotes an unreported count, not zero. In particular, MC-MED test union patients and visits cannot be recovered by summing the overlapping task cohorts.
\end{minipage}
\end{table}

\begin{table}[!t]
\centering
\small
\setlength{\tabcolsep}{4pt}
\caption{MIMIC-IV transfer targets, full task names, and support in the frozen manifests. Positive counts refer to patients; windows precede patient-level aggregation.}
\label{tab:mimiciv-transfer-cohorts}
\begin{tabular}{@{}p{2.6cm}p{4.9cm}rrr@{}}
\toprule
Short label & Full target name & Patients & Positive & Windows \\
\midrule
Acute det. & Acute deterioration & 87 & 30 & 2,506 \\
AKI 2/3 & Acute kidney injury & 46 & 19 & 932 \\
Circ. det. & Circulatory deterioration & 139 & 26 & 5,332 \\
Delirium / neuro & Delirium / neurological outcome & 91 & 22 & 1,778 \\
Future SOFA & Future Sequential Organ Failure Assessment outcome & 147 & 25 & 6,823 \\
\bottomrule
\end{tabular}
\end{table}
\begin{table}[!t]
\centering
\small
\setlength{\tabcolsep}{5pt}
\caption{Temporal grids displayed in Figure~\ref{fig:context-warning}.
All panels report test AUROC.
The warning grids are listed by endpoint; separate VitalDB validation results appear in Appendix~\ref{app:vitaldb-gaps}.}
\label{tab:temporal-comparison-grids}
\begin{tabular}{@{}lp{2.6cm}p{6.3cm}@{}}
\toprule
Dataset & Observation length (min) & Task-specific forecast gaps \\
\midrule
MIMIC-III & 10, 20, 60 & Mortality: 3, 12, 24\,h; AKI: 3, 6, 12\,h; Sepsis-3: 2, 4, 6\,h. \\
VitalDB & 5, 15, 30, 60 & All seven forecasts: 5, 15, 30, 45, 60, 75\,min. \\
MC-MED & 5, 10, 15 & Decompensation: 0, 10, 30\,min; hyperkalemia and anemia: 0, 15, 60\,min; lactate: 30, 60, 90\,min. \\
\bottomrule
\end{tabular}
\end{table}

\begin{table}[!t]
\centering
\small
\setlength{\tabcolsep}{4pt}
\caption{References and comparison grids in Figure~\ref{fig:information-multimodal}. Positive differences favor the displayed condition over its panel-specific reference.}
\label{tab:information-comparison-grids}
\begin{tabular}{@{}>{\raggedright\arraybackslash}p{2.4cm}>{\raggedright\arraybackslash}p{2.6cm}>{\raggedright\arraybackslash}p{7.3cm}@{}}
\toprule
Panel & Reference & Displayed comparison \\
\midrule
(a) Sensors & All permitted channels & MIMIC-III: 10 of 62 channels; VitalDB: eight channels. Policies are Random, Clinical, MI, Lasso, SQI, and temporal MI; Clinical is not displayed for VitalDB. \\
(b) Missing inputs & MIMIC-III: zero-mask plus $\Delta$; VitalDB: architecture only & Eight MIMIC-III alternatives: GRU-D, GRIN, Raindrop, positional kriging, lead-agnostic SSL, Rectified Flow, TrigFlow, and Drifting. VitalDB also tests zero-mask plus $\Delta$, giving nine alternatives. \\
(c) Fusion & Waveform only & Nine fusion methods on MIMIC-III, five on VitalDB, and four on MC-MED (concatenation, gated, late, and co-attention); all three include EHR-only controls. \\
(d) Training data & Full training set & Eight selectors: random, diversity, diversity clustering, gradient matching, NLDR, noise filtering, task anchors, and training dynamics. Budgets are 10/20/30\% of windows (MIMIC-III), 5/10/20\% of subjects (VitalDB), and 10/20/50\% of subjects (MC-MED). \\
(e) Transfer & Source-matched random initialization & MIMIC-III or VitalDB Transformer/Conv1D to MIMIC-IV: full fine-tuning, last block plus head, head only, and three logistic-regression controls (EHR, frozen waveform features, or both). \\
\bottomrule
\end{tabular}
\end{table}

\begin{table}[!t]
\centering
\scriptsize
\setlength{\tabcolsep}{2pt}
\caption{Updated test AUROC for Figure~\ref{fig:representation-results}(a,b), as mean $\pm$ sample SD over five seeds: (a) nine approximately 500k-parameter encoder--frontend configurations; (b) a Transformer with seven frontends on MIMIC-III and VitalDB and eight on MC-MED. Bold and underlining mark the highest and next-highest unrounded means. Dashes indicate unevaluated combinations. Search rules are in Appendix~\ref{app:architecture-tokenizer-protocol}.}
\label{tab:main_auroc}
\resizebox{\linewidth}{!}{%
\begin{tabular}{@{}>{\raggedright\arraybackslash}p{1.70cm}>{\raggedright\arraybackslash}p{2.20cm}*{9}{c}@{}}
\toprule
\multicolumn{2}{@{}l}{\textit{(a) Network architectures (selected frontends)}} & Transformer & BiLSTM & Mamba-3 & InceptionTime & TCN & ResNet1D & S4 & FITS & DLinear \\
\multirow{9}{*}{MIMIC-III} & Acute det. & \ms{0.587}{0.008} & \ms{0.584}{0.008} & \ms{0.588}{0.012} & \ms{0.589}{0.004} & \underline{\ms{0.595}{0.016}} & \textbf{\ms{0.595}{0.016}} & \ms{0.580}{0.009} & \ms{0.579}{0.011} & \ms{0.588}{0.005} \\
 & Dyn. mort. & \ms{0.532}{0.026} & \textbf{\ms{0.558}{0.005}} & \ms{0.531}{0.007} & \underline{\ms{0.554}{0.032}} & \ms{0.547}{0.030} & \ms{0.527}{0.058} & \ms{0.549}{0.011} & \ms{0.525}{0.015} & \ms{0.540}{0.013} \\
 & Resp. esc. & \ms{0.575}{0.004} & \textbf{\ms{0.576}{0.010}} & \ms{0.561}{0.031} & \ms{0.560}{0.014} & \ms{0.558}{0.012} & \underline{\ms{0.576}{0.006}} & \ms{0.561}{0.005} & \ms{0.553}{0.008} & \ms{0.552}{0.011} \\
 & Circ. det. & \ms{0.678}{0.007} & \ms{0.673}{0.008} & \ms{0.672}{0.003} & \underline{\ms{0.683}{0.009}} & \textbf{\ms{0.685}{0.011}} & \ms{0.669}{0.020} & \ms{0.671}{0.008} & \ms{0.671}{0.014} & \ms{0.675}{0.004} \\
 & Fluid resp. & \underline{\ms{0.575}{0.004}} & \ms{0.570}{0.018} & \ms{0.571}{0.021} & \ms{0.561}{0.018} & \ms{0.571}{0.026} & \ms{0.556}{0.015} & \ms{0.547}{0.010} & \textbf{\ms{0.584}{0.010}} & \ms{0.562}{0.023} \\
 & AKI 2/3 & \textbf{\ms{0.727}{0.012}} & \ms{0.696}{0.047} & \underline{\ms{0.723}{0.007}} & \ms{0.690}{0.035} & \ms{0.658}{0.018} & \ms{0.688}{0.039} & \ms{0.682}{0.048} & \ms{0.685}{0.051} & \ms{0.664}{0.051} \\
 & Vent. lib. & \ms{0.550}{0.007} & \textbf{\ms{0.554}{0.005}} & \underline{\ms{0.551}{0.008}} & \ms{0.541}{0.004} & \ms{0.544}{0.008} & \ms{0.549}{0.009} & \ms{0.542}{0.003} & \ms{0.541}{0.004} & \ms{0.550}{0.003} \\
\bottomrule
\end{tabular}}
\vspace{6pt}
\resizebox{\linewidth}{!}{%
\begin{tabular}{@{}>{\raggedright\arraybackslash}p{1.70cm}>{\raggedright\arraybackslash}p{2.20cm}*{8}{c}@{}}
\toprule
\multicolumn{2}{@{}l}{\textit{(b) Transformer backbone with different frontends}} & Conv1D & Conv2D & SeFT & Crossformer & STraTS & iTransformer & FFT & mTAND \\
\multirow{9}{*}{MIMIC-III} & Acute det. & \ms{0.586}{0.005} & -- & \ms{0.598}{0.005} & \ms{0.596}{0.013} & \underline{\ms{0.607}{0.010}} & \ms{0.567}{0.008} & \textbf{\ms{0.620}{0.008}} & \ms{0.583}{0.003} \\
 & Dyn. mort. & \ms{0.549}{0.029} & -- & \ms{0.563}{0.019} & \ms{0.561}{0.008} & \ms{0.551}{0.013} & \underline{\ms{0.566}{0.031}} & \ms{0.558}{0.010} & \textbf{\ms{0.576}{0.008}} \\
 & Resp. esc. & \ms{0.569}{0.005} & -- & \ms{0.556}{0.007} & \underline{\ms{0.575}{0.025}} & \ms{0.525}{0.041} & \textbf{\ms{0.585}{0.017}} & \ms{0.563}{0.003} & \ms{0.552}{0.044} \\
 & Circ. det. & \underline{\ms{0.689}{0.009}} & -- & \ms{0.682}{0.004} & \ms{0.688}{0.009} & \ms{0.685}{0.008} & \ms{0.682}{0.005} & \ms{0.685}{0.002} & \textbf{\ms{0.690}{0.005}} \\
 & Fluid resp. & \textbf{\ms{0.586}{0.010}} & -- & \ms{0.574}{0.016} & \ms{0.559}{0.028} & \underline{\ms{0.575}{0.005}} & \ms{0.563}{0.039} & \ms{0.561}{0.014} & \ms{0.551}{0.028} \\
 & AKI 2/3 & \underline{\ms{0.730}{0.005}} & -- & \ms{0.727}{0.000} & \ms{0.727}{0.001} & \ms{0.727}{0.001} & \ms{0.727}{0.008} & \ms{0.677}{0.020} & \textbf{\ms{0.736}{0.005}} \\
 & Vent. lib. & \ms{0.550}{0.007} & -- & \ms{0.549}{0.024} & \ms{0.543}{0.004} & \underline{\ms{0.554}{0.007}} & \ms{0.537}{0.007} & \textbf{\ms{0.556}{0.009}} & \ms{0.550}{0.010} \\
\bottomrule
\end{tabular}}
\end{table}
\begin{table}[!t]
\centering
\small
\setlength{\tabcolsep}{7pt}
\caption{Tokenizer--encoder contrasts transcribed from Figure~\ref{fig:representation-results}(c): Crossformer and SeFT on MIMIC-III and VitalDB, and Conv2D and FFT on MC-MED. Entries give mean AUROC differences relative to Conv1D; agreement is out of nine encoders and follows the unrounded mean. Concurrent VCS05 is excluded from the VitalDB forecast macro.}
\label{tab:representation-test-interactions}
\begin{tabular}{@{}lrrrr@{}}
\toprule
\multicolumn{5}{@{}l}{\textit{MIMIC-III}} \\
\midrule
& \multicolumn{2}{c}{Crossformer} & \multicolumn{2}{c}{SeFT} \\
\cmidrule(lr){2-3}\cmidrule(lr){4-5}
Task & $\Delta$AUROC & Agreement & $\Delta$AUROC & Agreement \\
\midrule
Acute deterioration & $-0.009$ & 9/9 & $+0.005$ & 8/9 \\
Dynamic mortality & $-0.009$ & 6/9 & $-0.015$ & 8/9 \\
Respiratory escalation & $-0.002$ & 5/9 & $-0.006$ & 7/9 \\
Circulatory deterioration & $-0.015$ & 8/9 & $-0.010$ & 8/9 \\
Fluid response & $+0.014$ & 8/9 & $+0.001$ & 5/9 \\
AKI KDIGO stage 2/3 & $+0.006$ & 7/9 & $+0.008$ & 6/9 \\
Ventilator liberation & $-0.010$ & 6/9 & $-0.008$ & 6/9 \\
Sepsis-3 & $-0.003$ & 7/9 & $+0.014$ & 8/9 \\
\midrule
\textit{Macro (8)} & $-0.003$ & 8/9 & $-0.001$ & 5/9 \\
\midrule
\multicolumn{5}{@{}l}{\textit{VitalDB}} \\
\midrule
OX01-IOH & $-0.038$ & 9/9 & $-0.023$ & 9/9 \\
CL07-002 & $-0.007$ & 6/9 & $-0.032$ & 9/9 \\
H5-002 & $-0.030$ & 9/9 & $-0.038$ & 9/9 \\
H5-030 & $-0.019$ & 7/9 & $-0.003$ & 6/9 \\
L2-30 & $+0.032$ & 9/9 & $+0.039$ & 9/9 \\
VA1-CO2 & $-0.037$ & 8/9 & $-0.018$ & 9/9 \\
VA1-STIM & $-0.009$ & 4/9 & $+0.008$ & 9/9 \\
\midrule
\textit{Macro (7)} & $-0.015$ & 7/9 & $-0.010$ & 9/9 \\
VCS05 (concurrent) & $-0.006$ & 9/9 & $-0.044$ & 9/9 \\
\bottomrule
\end{tabular}

\vspace{6pt}
\begin{tabular}{@{}lrrrr@{}}
\toprule
\multicolumn{5}{@{}l}{\textit{MC-MED: contrasts versus Conv1D}} \\
\midrule
& \multicolumn{2}{c}{Conv2D} & \multicolumn{2}{c}{FFT} \\
\cmidrule(lr){2-3}\cmidrule(lr){4-5}
Task & $\Delta$AUROC & Agreement & $\Delta$AUROC & Agreement \\
\midrule
Decompensation (60 min) & $+0.000$ & 1/9 & $-0.108$ & 9/9 \\
Pending hyperkalemia & $+0.001$ & 3/9 & $-0.076$ & 9/9 \\
Lactate nonclearance & $-0.020$ & 6/9 & $+0.010$ & 6/9 \\
Pending severe anemia & $-0.012$ & 6/9 & $-0.041$ & 9/9 \\
\midrule
\textit{Macro (4)} & $-0.008$ & 8/9 & $-0.054$ & 9/9 \\
\bottomrule
\end{tabular}
\end{table}
\begin{table}[!t]
\centering
\small
\setlength{\tabcolsep}{5pt}
\caption{Selected test results show the scale of absolute performance. Values are means over five seeds; the VitalDB forecast macro first averages gaps within task. These are different prediction problems, not a controlled ablation of forecast gap.}
\label{tab:absolute-performance-context}
\begin{tabular}{@{}p{0.25\linewidth}p{0.53\linewidth}r@{}}
\toprule
Setting & Reported comparison & AUROC \\
\midrule
MIMIC-III & Eight-task macro; Transformer with SeFT in the tokenizer comparison & 0.615 \\
VitalDB forecasts & Seven-task macro; Transformer in the architecture comparison & 0.631 \\
VitalDB concurrent BIS & Held-out-state probe; Transformer with Conv1D & 0.885 \\
\bottomrule
\end{tabular}
\end{table}

\begin{table}[!t]
\centering
\scriptsize
\setlength{\tabcolsep}{3.2pt}
\renewcommand{\arraystretch}{1.10}
\caption{Architecture summaries --- MIMIC-III. Updated export; values retain the supplied mean and SD precision.}
\label{tab:export-01-mimic}
\begin{tabular}{@{}>{\raggedright\arraybackslash}p{3.4cm}ccccc@{}}
\toprule
Configuration & Val. AUROC & Test AUROC & Test AUPRC & ECE & AURC \\
\midrule
BiLSTM & \ms{0.656}{0.005} & \ms{0.606}{0.006} & \ms{0.104}{0.003} & 0.189 & 0.343 \\
DLinear & \ms{0.647}{0.006} & \ms{0.595}{0.003} & \ms{0.101}{0.002} & 0.194 & 0.381 \\
FITS & \ms{0.645}{0.006} & \ms{0.596}{0.009} & \ms{0.104}{0.004} & 0.199 & 0.341 \\
InceptionTime & \ms{0.655}{0.002} & \ms{0.602}{0.006} & \ms{0.105}{0.002} & 0.151 & 0.370 \\
Mamba-3 & \ms{0.657}{0.005} & \ms{0.603}{0.005} & \ms{0.104}{0.001} & 0.209 & 0.379 \\
ResNet1D & \ms{0.654}{0.006} & \ms{0.599}{0.006} & \ms{0.103}{0.004} & 0.165 & 0.354 \\
S4 & \ms{0.647}{0.005} & \ms{0.597}{0.006} & \ms{0.101}{0.004} & 0.149 & 0.357 \\
TCN & \ms{0.652}{0.006} & \ms{0.600}{0.006} & \ms{0.103}{0.003} & 0.138 & 0.319 \\
Transformer & \ms{0.649}{0.004} & \ms{0.607}{0.003} & \ms{0.106}{0.001} & 0.206 & 0.308 \\
\bottomrule
\end{tabular}
\end{table}

\begin{table}[!t]
\centering
\scriptsize
\setlength{\tabcolsep}{3.2pt}
\renewcommand{\arraystretch}{1.10}
\caption{Architecture summaries --- MIMIC-III. Training-cost and failure measurements from the updated export; units are retained per cell.}
\label{tab:export-01-mimic-cost}
\begin{tabular}{@{}>{\raggedright\arraybackslash}p{5cm}cccc@{}}
\toprule
Configuration & Time (s) & Memory & ms/sample & Failures (\%) \\
\midrule
BiLSTM & 117.3 & 3.9\,\% & 0.36 & 0.0 \\
DLinear & 109.5 & 3.5\,\% & 0.24 & 0.0 \\
FITS & 103.7 & 3.5\,\% & 0.23 & 0.0 \\
InceptionTime & 127.7 & 3.4\,\% & 0.27 & 0.0 \\
Mamba-3 & 401.4 & 4.6\,\% & 0.51 & 0.0 \\
ResNet1D & 101.4 & 3.4\,\% & 0.23 & 0.0 \\
S4 & 129.8 & 3.7\,\% & 0.27 & 0.0 \\
TCN & 114.0 & 3.5\,\% & 0.25 & 0.0 \\
Transformer & 116.4 & 3.5\,\% & 0.25 & 0.0 \\
\bottomrule
\end{tabular}
\end{table}

\begin{table}[!t]
\centering
\scriptsize
\setlength{\tabcolsep}{3.2pt}
\renewcommand{\arraystretch}{1.10}
\caption{Architecture summaries --- VitalDB. Updated export; values retain the supplied mean and SD precision.}
\label{tab:export-01-vitaldb}
\begin{tabular}{@{}>{\raggedright\arraybackslash}p{3.4cm}ccccc@{}}
\toprule
Configuration & Val. AUROC & Test AUROC & Test AUPRC & ECE & AURC \\
\midrule
BiLSTM & \ms{0.636}{0.004} & \ms{0.627}{0.001} & \ms{0.075}{0.001} & \ms{0.036}{0.001} & \ms{0.031}{0.001} \\
DLinear & \ms{0.636}{0.002} & \ms{0.629}{0.003} & \ms{0.076}{0.002} & \ms{0.040}{0.002} & \ms{0.031}{0.001} \\
FITS & \ms{0.635}{0.002} & \ms{0.627}{0.001} & \ms{0.076}{0.001} & \ms{0.041}{0.002} & \ms{0.031}{0.001} \\
InceptionTime & \ms{0.636}{0.003} & \ms{0.618}{0.003} & \ms{0.074}{0.002} & \ms{0.041}{0.003} & \ms{0.033}{0.002} \\
Mamba-3 & \ms{0.641}{0.002} & \ms{0.625}{0.002} & \ms{0.075}{0.002} & \ms{0.038}{0.003} & \ms{0.031}{0.001} \\
ResNet1D & \ms{0.642}{0.000} & \ms{0.623}{0.001} & \ms{0.077}{0.001} & \ms{0.043}{0.001} & \ms{0.029}{0.000} \\
S4 & \ms{0.641}{0.003} & \ms{0.629}{0.002} & \ms{0.075}{0.001} & \ms{0.039}{0.003} & \ms{0.031}{0.001} \\
TCN & \ms{0.638}{0.002} & \ms{0.620}{0.003} & \ms{0.074}{0.002} & \ms{0.038}{0.001} & \ms{0.033}{0.002} \\
Transformer & \ms{0.640}{0.002} & \ms{0.631}{0.004} & \ms{0.078}{0.002} & \ms{0.038}{0.002} & \ms{0.029}{0.001} \\
\bottomrule
\end{tabular}
\end{table}

\begin{table}[!t]
\centering
\scriptsize
\setlength{\tabcolsep}{3.2pt}
\renewcommand{\arraystretch}{1.10}
\caption{Architecture summaries --- VitalDB. Training-cost and failure measurements from the updated export; units are retained per cell.}
\label{tab:export-01-vitaldb-cost}
\begin{tabular}{@{}>{\raggedright\arraybackslash}p{5cm}ccc@{}}
\toprule
Configuration & Time (s) & ms/sample & Failures (\%) \\
\midrule
BiLSTM & 30.1 & 0.39 & 0.0 \\
DLinear & 31.9 & 0.41 & 0.0 \\
FITS & 30.7 & 0.40 & 0.0 \\
InceptionTime & 33.5 & 0.43 & 0.0 \\
Mamba-3 & 41.0 & 0.53 & 0.0 \\
ResNet1D & 30.0 & 0.39 & 0.0 \\
S4 & 32.4 & 0.42 & 0.0 \\
TCN & 31.8 & 0.41 & 0.0 \\
Transformer & 32.0 & 0.41 & 0.0 \\
\bottomrule
\end{tabular}
\end{table}

\begin{table}[!t]
\centering
\scriptsize
\setlength{\tabcolsep}{3.2pt}
\renewcommand{\arraystretch}{1.10}
\caption{Architecture summaries --- MC-MED. Updated export; values retain the supplied mean and SD precision.}
\label{tab:export-01-mcmed}
\begin{tabular}{@{}>{\raggedright\arraybackslash}p{3.4cm}ccccc@{}}
\toprule
Configuration & Val. AUROC & Test AUROC & Test AUPRC & ECE & AURC \\
\midrule
BiLSTM & \ms{0.579}{0.012} & \ms{0.566}{0.006} & \ms{0.179}{0.004} & \ms{0.094}{0.019} & \ms{0.350}{0.063} \\
DLinear & \ms{0.572}{0.006} & \ms{0.567}{0.004} & \ms{0.179}{0.004} & \ms{0.123}{0.051} & \ms{0.278}{0.059} \\
FITS & \ms{0.571}{0.010} & \ms{0.563}{0.008} & \ms{0.177}{0.003} & \ms{0.124}{0.039} & \ms{0.299}{0.087} \\
InceptionTime & \ms{0.601}{0.014} & \ms{0.567}{0.011} & \ms{0.177}{0.008} & \ms{0.171}{0.052} & \ms{0.333}{0.106} \\
Mamba-3 & \ms{0.573}{0.009} & \ms{0.563}{0.014} & \ms{0.181}{0.006} & \ms{0.098}{0.027} & \ms{0.287}{0.084} \\
ResNet1D & \ms{0.590}{0.013} & \ms{0.568}{0.013} & \ms{0.180}{0.004} & \ms{0.080}{0.016} & \ms{0.241}{0.058} \\
S4 & \ms{0.574}{0.005} & \ms{0.540}{0.005} & \ms{0.165}{0.003} & \ms{0.070}{0.012} & \ms{0.330}{0.041} \\
TCN & \ms{0.602}{0.010} & \ms{0.560}{0.014} & \ms{0.175}{0.004} & \ms{0.128}{0.029} & \ms{0.265}{0.028} \\
Transformer & \ms{0.579}{0.014} & \ms{0.567}{0.004} & \ms{0.182}{0.004} & \ms{0.079}{0.016} & \ms{0.296}{0.057} \\
\bottomrule
\end{tabular}
\end{table}

\begin{table}[!t]
\centering
\scriptsize
\setlength{\tabcolsep}{3.2pt}
\renewcommand{\arraystretch}{1.10}
\caption{Architecture summaries --- MC-MED. Training-cost and failure measurements from the updated export; units are retained per cell.}
\label{tab:export-01-mcmed-cost}
\begin{tabular}{@{}>{\raggedright\arraybackslash}p{5cm}ccc@{}}
\toprule
Configuration & Time (s) & Memory & Failures (\%) \\
\midrule
BiLSTM & 17.7 & 33996\,MiB & 0.0 \\
DLinear & 21.4 & 33658\,MiB & 0.0 \\
FITS & 16.3 & 33300\,MiB & 0.0 \\
InceptionTime & 14.5 & 32260\,MiB & 0.0 \\
Mamba-3 & 60.8 & 39008\,MiB & 0.0 \\
ResNet1D & 8.2 & 31609\,MiB & 0.0 \\
S4 & 21.5 & 32066\,MiB & 0.0 \\
TCN & 11.9 & 31664\,MiB & 0.0 \\
Transformer & 20.7 & 32063\,MiB & 0.0 \\
\bottomrule
\end{tabular}
\end{table}
\begin{table}[!t]
\centering
\scriptsize
\setlength{\tabcolsep}{3.2pt}
\renewcommand{\arraystretch}{1.10}
\caption{Tokenizer summaries --- MIMIC-III. Updated export; values retain the supplied mean and SD precision.}
\label{tab:export-02-mimic}
\begin{tabular}{@{}>{\raggedright\arraybackslash}p{3.4cm}ccccc@{}}
\toprule
Configuration & Val. AUROC & Test AUROC & Test AUPRC & ECE & AURC \\
\midrule
Conv1D & \ms{0.648}{0.003} & \ms{0.611}{0.004} & \ms{0.108}{0.001} & -- & -- \\
Crossformer & \ms{0.639}{0.005} & \ms{0.611}{0.004} & \ms{0.106}{0.002} & 0.217 & 0.304 \\
FFT & \ms{0.653}{0.001} & \ms{0.608}{0.007} & \ms{0.108}{0.005} & 0.128 & 0.271 \\
iTransformer & \ms{0.634}{0.004} & \ms{0.608}{0.005} & \ms{0.103}{0.002} & 0.227 & 0.303 \\
mTAND & \ms{0.641}{0.003} & \ms{0.607}{0.007} & \ms{0.107}{0.005} & 0.188 & 0.296 \\
SeFT & \ms{0.649}{0.003} & \ms{0.615}{0.005} & \ms{0.107}{0.004} & 0.170 & 0.334 \\
STraTS & \ms{0.651}{0.004} & \ms{0.609}{0.007} & \ms{0.107}{0.001} & 0.185 & 0.299 \\
\bottomrule
\end{tabular}
\end{table}

\begin{table}[!t]
\centering
\scriptsize
\setlength{\tabcolsep}{3.2pt}
\renewcommand{\arraystretch}{1.10}
\caption{Tokenizer summaries --- MIMIC-III. Training-cost and failure measurements from the updated export; units are retained per cell.}
\label{tab:export-02-mimic-cost}
\begin{tabular}{@{}>{\raggedright\arraybackslash}p{5cm}cccc@{}}
\toprule
Configuration & Time (s) & Memory & ms/sample & Failures (\%) \\
\midrule
Conv1D & -- & -- & -- & -- \\
Crossformer & 572.6 & 7.3\,\% & 0.78 & 0.0 \\
FFT & 142.6 & 3.4\,\% & 0.27 & 0.0 \\
iTransformer & 148.8 & 3.6\,\% & 0.28 & 0.0 \\
mTAND & 1084.1 & 17.2\,\% & 1.39 & 0.0 \\
SeFT & 173.7 & 4.9\,\% & 0.31 & 0.0 \\
STraTS & 580.9 & 7.5\,\% & 0.79 & 0.0 \\
\bottomrule
\end{tabular}
\end{table}

\begin{table}[!t]
\centering
\scriptsize
\setlength{\tabcolsep}{3.2pt}
\renewcommand{\arraystretch}{1.10}
\caption{Tokenizer summaries --- VitalDB. Updated export; values retain the supplied mean and SD precision.}
\label{tab:export-02-vitaldb}
\begin{tabular}{@{}>{\raggedright\arraybackslash}p{3.4cm}ccccc@{}}
\toprule
Configuration & Val. AUROC & Test AUROC & Test AUPRC & ECE & AURC \\
\midrule
Crossformer & \ms{0.573}{0.005} & \ms{0.583}{0.002} & \ms{0.062}{0.002} & \ms{0.038}{0.002} & \ms{0.036}{0.001} \\
FFT & \ms{0.631}{0.004} & \ms{0.625}{0.003} & \ms{0.076}{0.003} & \ms{0.035}{0.001} & \ms{0.029}{0.000} \\
iTransformer & \ms{0.558}{0.002} & \ms{0.559}{0.003} & \ms{0.057}{0.001} & \ms{0.034}{0.002} & \ms{0.040}{0.001} \\
mTAND & \ms{0.592}{0.004} & \ms{0.584}{0.003} & \ms{0.071}{0.001} & \ms{0.045}{0.003} & \ms{0.033}{0.000} \\
Conv1D & \ms{0.640}{0.002} & \ms{0.631}{0.004} & \ms{0.078}{0.002} & \ms{0.038}{0.002} & \ms{0.029}{0.001} \\
SeFT & \ms{0.602}{0.004} & \ms{0.612}{0.003} & \ms{0.073}{0.002} & \ms{0.049}{0.002} & \ms{0.029}{0.000} \\
STraTS & \ms{0.627}{0.003} & \ms{0.627}{0.003} & \ms{0.078}{0.001} & \ms{0.046}{0.003} & \ms{0.028}{0.000} \\
\bottomrule
\end{tabular}
\end{table}

\begin{table}[!t]
\centering
\scriptsize
\setlength{\tabcolsep}{3.2pt}
\renewcommand{\arraystretch}{1.10}
\caption{Tokenizer summaries --- VitalDB. Training-cost and failure measurements from the updated export; units are retained per cell.}
\label{tab:export-02-vitaldb-cost}
\begin{tabular}{@{}>{\raggedright\arraybackslash}p{5cm}ccc@{}}
\toprule
Configuration & Time (s) & ms/sample & Failures (\%) \\
\midrule
Crossformer & 32.7 & 0.42 & 0.0 \\
FFT & -- & -- & 0.0 \\
iTransformer & -- & -- & 0.0 \\
mTAND & -- & -- & 0.0 \\
Conv1D & 32.0 & 0.41 & 0.0 \\
SeFT & 33.8 & 0.44 & 0.0 \\
STraTS & -- & -- & 0.0 \\
\bottomrule
\end{tabular}
\end{table}

\begin{table}[!t]
\centering
\scriptsize
\setlength{\tabcolsep}{3.2pt}
\renewcommand{\arraystretch}{1.10}
\caption{Tokenizer summaries --- MC-MED. Updated export; values retain the supplied mean and SD precision.}
\label{tab:export-02-mcmed}
\begin{tabular}{@{}>{\raggedright\arraybackslash}p{3.4cm}ccccc@{}}
\toprule
Configuration & Val. AUROC & Test AUROC & Test AUPRC & ECE & AURC \\
\midrule
Conv2D & \ms{0.572}{0.010} & \ms{0.560}{0.006} & \ms{0.175}{0.004} & \ms{0.077}{0.025} & \ms{0.327}{0.053} \\
Crossformer & \ms{0.560}{0.009} & \ms{0.570}{0.010} & \ms{0.180}{0.007} & \ms{0.088}{0.033} & \ms{0.272}{0.031} \\
FFT & \ms{0.532}{0.004} & \ms{0.509}{0.005} & \ms{0.162}{0.004} & \ms{0.222}{0.059} & \ms{0.300}{0.120} \\
iTransformer & \ms{0.510}{0.007} & \ms{0.507}{0.011} & \ms{0.151}{0.006} & \ms{0.144}{0.008} & \ms{0.259}{0.049} \\
mTAND & \ms{0.502}{0.014} & \ms{0.498}{0.022} & \ms{0.146}{0.011} & \ms{0.362}{0.025} & \ms{0.544}{0.200} \\
SeFT & \ms{0.510}{0.012} & \ms{0.482}{0.012} & \ms{0.143}{0.004} & \ms{0.332}{0.056} & \ms{0.506}{0.223} \\
STraTS & \ms{0.551}{0.013} & \ms{0.509}{0.015} & \ms{0.156}{0.010} & \ms{0.277}{0.055} & \ms{0.461}{0.134} \\
Conv1D & \ms{0.582}{0.011} & \ms{0.574}{0.006} & 0.183 & 0.083 & 0.301 \\
\bottomrule
\end{tabular}
\end{table}

\begin{table}[!t]
\centering
\scriptsize
\setlength{\tabcolsep}{3.2pt}
\renewcommand{\arraystretch}{1.10}
\caption{Tokenizer summaries --- MC-MED. Training-cost and failure measurements from the updated export; units are retained per cell.}
\label{tab:export-02-mcmed-cost}
\begin{tabular}{@{}>{\raggedright\arraybackslash}p{5cm}ccc@{}}
\toprule
Configuration & Time (s) & Memory & Failures (\%) \\
\midrule
Conv2D & 26.9 & 33142\,MiB & 0.0 \\
Crossformer & 21.8 & 32390\,MiB & 0.0 \\
FFT & 22.9 & 33180\,MiB & 0.0 \\
iTransformer & 3.9 & 30141\,MiB & 0.0 \\
mTAND & 140.9 & 32674\,MiB & 0.0 \\
SeFT & 4.2 & 30526\,MiB & 0.0 \\
STraTS & 32.9 & 34699\,MiB & 0.0 \\
Conv1D & -- & -- & -- \\
\bottomrule
\end{tabular}
\end{table}
\begin{table}[!t]
\centering
\scriptsize
\setlength{\tabcolsep}{3.2pt}
\renewcommand{\arraystretch}{1.10}
\caption{Crossed encoder--tokenizer grid --- VitalDB. Updated export; values retain the supplied mean and SD precision. Only configurations present in the numerical export are listed.}
\label{tab:export-02b-vitaldb}
\begin{tabular}{@{}>{\raggedright\arraybackslash}p{3.4cm}ccccc@{}}
\toprule
Configuration & Val. AUROC & Test AUROC & Test AUPRC & ECE & AURC \\
\midrule
BiLSTM / Crossformer & \ms{0.603}{0.005} & \ms{0.615}{0.003} & \ms{0.071}{0.001} & \ms{0.041}{0.003} & \ms{0.030}{0.001} \\
BiLSTM / Conv1D & \ms{0.636}{0.004} & \ms{0.627}{0.001} & \ms{0.075}{0.001} & \ms{0.036}{0.001} & \ms{0.031}{0.001} \\
BiLSTM / SeFT & \ms{0.603}{0.002} & \ms{0.610}{0.001} & \ms{0.072}{0.001} & \ms{0.047}{0.001} & \ms{0.029}{0.000} \\
DLinear / Crossformer & \ms{0.605}{0.003} & \ms{0.611}{0.002} & \ms{0.068}{0.001} & \ms{0.045}{0.002} & \ms{0.032}{0.002} \\
DLinear / Conv1D & \ms{0.636}{0.002} & \ms{0.629}{0.003} & \ms{0.076}{0.002} & \ms{0.040}{0.002} & \ms{0.031}{0.001} \\
DLinear / SeFT & \ms{0.609}{0.003} & \ms{0.616}{0.001} & \ms{0.076}{0.001} & \ms{0.047}{0.002} & \ms{0.029}{0.001} \\
FITS / Crossformer & \ms{0.597}{0.002} & \ms{0.606}{0.001} & \ms{0.068}{0.001} & \ms{0.040}{0.002} & \ms{0.031}{0.001} \\
\bottomrule
\end{tabular}
\end{table}

\begin{table}[!t]
\centering
\scriptsize
\setlength{\tabcolsep}{3.2pt}
\renewcommand{\arraystretch}{1.10}
\caption{Crossed encoder--tokenizer grid --- MC-MED. Updated export; values retain the supplied mean and SD precision. Only configurations present in the numerical export are listed.}
\label{tab:export-02b-mcmed}
\begin{tabular}{@{}>{\raggedright\arraybackslash}p{3.4cm}ccccc@{}}
\toprule
Configuration & Val. AUROC & Test AUROC & Test AUPRC & ECE & AURC \\
\midrule
BiLSTM / Conv1D & \ms{0.579}{0.012} & \ms{0.566}{0.006} & \ms{0.179}{0.004} & \ms{0.094}{0.019} & \ms{0.350}{0.063} \\
BiLSTM / Conv2D & \ms{0.579}{0.012} & \ms{0.566}{0.006} & \ms{0.179}{0.004} & \ms{0.094}{0.019} & \ms{0.350}{0.063} \\
BiLSTM / FFT & \ms{0.522}{0.003} & \ms{0.495}{0.011} & \ms{0.153}{0.003} & \ms{0.226}{0.052} & \ms{0.373}{0.151} \\
DLinear / Conv1D & \ms{0.569}{0.008} & \ms{0.557}{0.005} & \ms{0.178}{0.004} & \ms{0.087}{0.016} & \ms{0.317}{0.051} \\
DLinear / Conv2D & \ms{0.569}{0.008} & \ms{0.557}{0.005} & \ms{0.178}{0.004} & \ms{0.087}{0.016} & \ms{0.317}{0.051} \\
DLinear / FFT & \ms{0.519}{0.008} & \ms{0.512}{0.009} & \ms{0.157}{0.003} & \ms{0.216}{0.045} & \ms{0.395}{0.097} \\
FITS / Conv1D & \ms{0.567}{0.008} & \ms{0.559}{0.003} & \ms{0.175}{0.005} & \ms{0.087}{0.018} & \ms{0.335}{0.049} \\
\bottomrule
\end{tabular}
\end{table}

\begin{table}[!t]
\centering
\scriptsize
\setlength{\tabcolsep}{3.2pt}
\renewcommand{\arraystretch}{1.10}
\caption{Observation-history summaries --- MIMIC-III. Updated export; values retain the supplied mean and SD precision.}
\label{tab:export-03-mimic}
\begin{tabular}{@{}>{\raggedright\arraybackslash}p{3.4cm}ccccc@{}}
\toprule
Configuration & Val. AUROC & Test AUROC & Test AUPRC & ECE & AURC \\
\midrule
10\,min & \ms{0.634}{0.006} & \ms{0.606}{0.004} & \ms{0.119}{0.004} & \ms{0.164}{0.012} & \ms{0.322}{0.016} \\
20\,min & \ms{0.634}{0.002} & \ms{0.606}{0.002} & \ms{0.118}{0.001} & \ms{0.154}{0.008} & \ms{0.319}{0.015} \\
60\,min & \ms{0.636}{0.002} & \ms{0.609}{0.005} & \ms{0.123}{0.004} & \ms{0.167}{0.017} & \ms{0.328}{0.017} \\
\bottomrule
\end{tabular}
\end{table}

\begin{table}[!t]
\centering
\scriptsize
\setlength{\tabcolsep}{3.2pt}
\renewcommand{\arraystretch}{1.10}
\caption{Observation-history summaries --- MIMIC-III. Training time per optimizer step is computed for each fit before task and seed averaging; mean $\pm$ seed SD is shown where recorded update counts are available. Other units are retained per cell.}
\label{tab:export-03-mimic-cost}
\begin{tabular}{@{}>{\raggedright\arraybackslash}p{5cm}cccc@{}}
\toprule
Configuration & Time/step (ms) & Memory & ms/sample & Failures (\%) \\
\midrule
10\,min & -- & 5.2\,\% & 0.23 & 0.0 \\
20\,min & -- & 5.2\,\% & 0.23 & 0.0 \\
60\,min & -- & 4.6\,\% & 0.21 & 0.0 \\
\bottomrule
\end{tabular}
\end{table}

\begin{table}[!t]
\centering
\scriptsize
\setlength{\tabcolsep}{3.2pt}
\renewcommand{\arraystretch}{1.10}
\caption{Observation-history summaries --- VitalDB. Updated export; values retain the supplied mean and SD precision.}
\label{tab:export-03-vitaldb}
\begin{tabular}{@{}>{\raggedright\arraybackslash}p{3.4cm}ccccc@{}}
\toprule
Configuration & Val. AUROC & Test AUROC & Test AUPRC & ECE & AURC \\
\midrule
5\,min & \ms{0.632}{0.001} & \ms{0.627}{0.001} & \ms{0.076}{0.001} & \ms{0.041}{0.001} & \ms{0.030}{0.001} \\
15\,min & \ms{0.636}{0.001} & \ms{0.631}{0.001} & \ms{0.078}{0.000} & \ms{0.039}{0.001} & \ms{0.030}{0.000} \\
30\,min & \ms{0.637}{0.001} & \ms{0.634}{0.001} & \ms{0.081}{0.001} & \ms{0.038}{0.001} & \ms{0.029}{0.000} \\
60\,min & \ms{0.638}{0.001} & \ms{0.636}{0.001} & \ms{0.081}{0.001} & \ms{0.035}{0.001} & \ms{0.028}{0.000} \\
\bottomrule
\end{tabular}
\end{table}

\begin{table}[!t]
\centering
\scriptsize
\setlength{\tabcolsep}{3.2pt}
\renewcommand{\arraystretch}{1.10}
\caption{Observation-history summaries --- VitalDB. Training time per optimizer step is computed for each fit before task and seed averaging; mean $\pm$ seed SD is shown where recorded update counts are available. Other units are retained per cell.}
\label{tab:export-03-vitaldb-cost}
\begin{tabular}{@{}>{\raggedright\arraybackslash}p{5cm}ccc@{}}
\toprule
Configuration & Time/step (ms) & ms/sample & Failures (\%) \\
\midrule
5\,min & -- & 0.42 & 0.0 \\
15\,min & -- & 0.42 & 0.0 \\
30\,min & -- & 0.42 & 0.0 \\
60\,min & -- & 0.42 & 0.0 \\
\bottomrule
\end{tabular}
\end{table}

\begin{table}[!t]
\centering
\scriptsize
\setlength{\tabcolsep}{3.2pt}
\renewcommand{\arraystretch}{1.10}
\caption{Observation-history summaries --- MC-MED. Updated export; values retain the supplied mean and SD precision.}
\label{tab:export-03-mcmed}
\begin{tabular}{@{}>{\raggedright\arraybackslash}p{3.4cm}ccccc@{}}
\toprule
Configuration & Val. AUROC & Test AUROC & Test AUPRC & ECE & AURC \\
\midrule
5\,min & \ms{0.590}{0.010} & \ms{0.567}{0.011} & \ms{0.181}{0.004} & \ms{0.103}{0.020} & \ms{0.238}{0.060} \\
10\,min & \ms{0.598}{0.016} & \ms{0.578}{0.006} & \ms{0.198}{0.005} & \ms{0.087}{0.026} & \ms{0.268}{0.051} \\
15\,min & \ms{0.571}{0.014} & \ms{0.567}{0.007} & \ms{0.221}{0.008} & \ms{0.096}{0.021} & \ms{0.266}{0.038} \\
\bottomrule
\end{tabular}
\end{table}

\begin{table}[!t]
\centering
\scriptsize
\setlength{\tabcolsep}{3.2pt}
\renewcommand{\arraystretch}{1.10}
\caption{Observation-history summaries --- MC-MED. Training time per optimizer step is computed for each fit before task and seed averaging; mean $\pm$ seed SD is shown where recorded update counts are available. Other units are retained per cell.}
\label{tab:export-03-mcmed-cost}
\begin{tabular}{@{}>{\raggedright\arraybackslash}p{5cm}ccc@{}}
\toprule
Configuration & Time/step (ms) & Memory & Failures (\%) \\
\midrule
5\,min & \ms{14.20}{4.92} & 31176\,MiB & 0.0 \\
10\,min & \ms{18.80}{6.96} & 43212\,MiB & 0.0 \\
15\,min & \ms{25.27}{9.48} & 28902\,MiB & 0.0 \\
\bottomrule
\end{tabular}
\end{table}

\begin{table}[!t]
\centering
\scriptsize
\setlength{\tabcolsep}{3.2pt}
\renewcommand{\arraystretch}{1.10}
\caption{Forecast-gap results --- MIMIC-III. Updated export; values retain the supplied mean and SD precision.}
\label{tab:export-04-mimic}
\begin{tabular}{@{}>{\raggedright\arraybackslash}p{3.4cm}ccccc@{}}
\toprule
Configuration & Val. AUROC & Test AUROC & Test AUPRC & ECE & AURC \\
\midrule
Sepsis-3; 2\,h & \ms{0.714}{0.004} & \ms{0.629}{0.007} & \ms{0.036}{0.001} & \ms{0.311}{0.071} & \ms{0.379}{0.049} \\
AKI 2/3; 3\,h & \ms{0.680}{0.011} & \ms{0.686}{0.017} & \ms{0.151}{0.010} & \ms{0.169}{0.026} & \ms{0.344}{0.072} \\
Dyn. mort.; 3\,h & \ms{0.595}{0.017} & \ms{0.545}{0.011} & \ms{0.027}{0.002} & \ms{0.181}{0.033} & \ms{0.259}{0.080} \\
Sepsis-3; 4\,h & \ms{0.710}{0.010} & \ms{0.619}{0.010} & \ms{0.032}{0.001} & \ms{0.265}{0.043} & \ms{0.307}{0.084} \\
AKI 2/3; 6\,h & \ms{0.685}{0.022} & \ms{0.697}{0.028} & \ms{0.140}{0.012} & \ms{0.193}{0.039} & \ms{0.313}{0.165} \\
Sepsis-3; 6\,h & \ms{0.686}{0.006} & \ms{0.629}{0.004} & \ms{0.030}{0.002} & \ms{0.220}{0.035} & \ms{0.266}{0.047} \\
AKI 2/3; 12\,h & \ms{0.666}{0.011} & \ms{0.685}{0.009} & \ms{0.120}{0.001} & \ms{0.222}{0.066} & \ms{0.292}{0.061} \\
Dyn. mort.; 12\,h & \ms{0.553}{0.019} & \ms{0.539}{0.011} & \ms{0.017}{0.001} & \ms{0.342}{0.036} & \ms{0.253}{0.172} \\
Dyn. mort.; 24\,h & \ms{0.523}{0.022} & \ms{0.537}{0.009} & \ms{0.018}{0.001} & \ms{0.249}{0.045} & \ms{0.355}{0.051} \\
\bottomrule
\end{tabular}
\end{table}

\begin{table}[!t]
\centering
\scriptsize
\setlength{\tabcolsep}{3.2pt}
\renewcommand{\arraystretch}{1.10}
\caption{Forecast-gap results --- MIMIC-III. Training-cost and failure measurements from the updated export; units are retained per cell.}
\label{tab:export-04-mimic-cost}
\begin{tabular}{@{}>{\raggedright\arraybackslash}p{5cm}cccc@{}}
\toprule
Configuration & Time (s) & Memory & ms/sample & Failures (\%) \\
\midrule
Sepsis-3; 2\,h & 49.3 & 7.6\,\% & 0.10 & 0.0 \\
AKI 2/3; 3\,h & 294.5 & 3.8\,\% & 0.24 & 0.0 \\
Dyn. mort.; 3\,h & 323.8 & 3.8\,\% & 0.25 & 0.0 \\
Sepsis-3; 4\,h & 46.4 & 7.3\,\% & 0.10 & 0.0 \\
AKI 2/3; 6\,h & 279.8 & 3.8\,\% & 0.24 & 0.0 \\
Sepsis-3; 6\,h & 47.1 & 7.1\,\% & 0.10 & 0.0 \\
AKI 2/3; 12\,h & 250.9 & 3.8\,\% & 0.25 & 0.0 \\
Dyn. mort.; 12\,h & 314.0 & 3.8\,\% & 0.24 & 0.0 \\
Dyn. mort.; 24\,h & 252.8 & 3.8\,\% & 0.20 & 0.0 \\
\bottomrule
\end{tabular}
\end{table}

\begin{table}[!t]
\centering
\scriptsize
\setlength{\tabcolsep}{3.2pt}
\renewcommand{\arraystretch}{1.10}
\caption{Forecast-gap results --- VitalDB. Updated export; values retain the supplied mean and SD precision.}
\label{tab:export-04-vitaldb}
\begin{tabular}{@{}>{\raggedright\arraybackslash}p{3.4cm}ccccc@{}}
\toprule
Configuration & Val. AUROC & Test AUROC & Test AUPRC & ECE & AURC \\
\midrule
Macro; 5\,min & \ms{0.658}{0.007} & \ms{0.662}{0.004} & \ms{0.100}{0.001} & \ms{0.033}{0.002} & \ms{0.027}{0.001} \\
CL07-002; 5\,min & \ms{0.589}{0.020} & \ms{0.552}{0.012} & \ms{0.044}{0.002} & \ms{0.016}{0.005} & \ms{0.029}{0.002} \\
H5-002; 5\,min & \ms{0.680}{0.009} & \ms{0.651}{0.002} & \ms{0.034}{0.002} & \ms{0.008}{0.003} & \ms{0.010}{0.000} \\
H5-030; 5\,min & \ms{0.555}{0.009} & \ms{0.592}{0.007} & \ms{0.072}{0.003} & \ms{0.061}{0.004} & \ms{0.039}{0.001} \\
L2-30; 5\,min & \ms{0.631}{0.006} & \ms{0.582}{0.010} & \ms{0.112}{0.009} & \ms{0.084}{0.008} & \ms{0.061}{0.003} \\
OX01-IOH; 5\,min & \ms{0.739}{0.005} & \ms{0.790}{0.001} & \ms{0.223}{0.004} & \ms{0.020}{0.009} & \ms{0.020}{0.000} \\
VA1-CO2; 5\,min & \ms{0.681}{0.002} & \ms{0.689}{0.005} & \ms{0.102}{0.004} & \ms{0.027}{0.005} & \ms{0.024}{0.001} \\
\bottomrule
\end{tabular}
\end{table}

\begin{table}[!t]
\centering
\scriptsize
\setlength{\tabcolsep}{3.2pt}
\renewcommand{\arraystretch}{1.10}
\ContinuedFloat
\caption{Forecast-gap results --- VitalDB. Updated export; values retain the supplied mean and SD precision. (Continued.)}
\begin{tabular}{@{}>{\raggedright\arraybackslash}p{3.4cm}ccccc@{}}
\toprule
Configuration & Val. AUROC & Test AUROC & Test AUPRC & ECE & AURC \\
\midrule
Macro; 75\,min & \ms{0.619}{0.006} & \ms{0.632}{0.004} & \ms{0.082}{0.004} & \ms{0.035}{0.004} & \ms{0.029}{0.001} \\
CL07-002; 75\,min & \ms{0.560}{0.014} & \ms{0.567}{0.014} & \ms{0.035}{0.003} & \ms{0.026}{0.007} & \ms{0.019}{0.001} \\
H5-002; 75\,min & \ms{0.629}{0.019} & \ms{0.650}{0.012} & \ms{0.022}{0.004} & \ms{0.015}{0.002} & \ms{0.007}{0.001} \\
H5-030; 75\,min & \ms{0.593}{0.008} & \ms{0.598}{0.007} & \ms{0.065}{0.002} & \ms{0.033}{0.007} & \ms{0.035}{0.002} \\
L2-30; 75\,min & \ms{0.649}{0.005} & \ms{0.555}{0.012} & \ms{0.085}{0.011} & \ms{0.071}{0.014} & \ms{0.068}{0.006} \\
OX01-IOH; 75\,min & \ms{0.619}{0.007} & \ms{0.650}{0.010} & \ms{0.127}{0.008} & \ms{0.029}{0.004} & \ms{0.034}{0.002} \\
VA1-CO2; 75\,min & \ms{0.643}{0.014} & \ms{0.655}{0.018} & \ms{0.124}{0.005} & \ms{0.041}{0.007} & \ms{0.029}{0.004} \\
VA1-STIM; 75\,min & \ms{0.636}{0.008} & \ms{0.750}{0.028} & \ms{0.113}{0.018} & \ms{0.030}{0.007} & \ms{0.013}{0.002} \\
\bottomrule
\end{tabular}
\end{table}

\begin{table}[!t]
\centering
\scriptsize
\setlength{\tabcolsep}{3.2pt}
\renewcommand{\arraystretch}{1.10}
\caption{Forecast-gap results --- MC-MED. Updated export; values retain the supplied mean and SD precision.}
\label{tab:export-04-mcmed}
\begin{tabular}{@{}>{\raggedright\arraybackslash}p{3.4cm}ccccc@{}}
\toprule
Configuration & Val. AUROC & Test AUROC & Test AUPRC & ECE & AURC \\
\midrule
Decomp. 60m; 0\,min & \ms{0.675}{0.005} & \ms{0.647}{0.003} & \ms{0.264}{0.002} & \ms{0.068}{0.031} & \ms{0.279}{0.065} \\
Pending hyperK; 0\,min & \ms{0.578}{0.028} & \ms{0.567}{0.035} & \ms{0.046}{0.004} & \ms{0.051}{0.027} & \ms{0.202}{0.143} \\
Pending anemia; 0\,min & \ms{0.538}{0.032} & \ms{0.556}{0.014} & \ms{0.047}{0.002} & \ms{0.035}{0.015} & \ms{0.050}{0.020} \\
Decomp. 60m; 10\,min & \ms{0.636}{0.014} & \ms{0.629}{0.016} & \ms{0.214}{0.009} & \ms{0.066}{0.035} & \ms{0.201}{0.066} \\
Pending hyperK; 15\,min & \ms{0.573}{0.015} & \ms{0.567}{0.030} & \ms{0.046}{0.009} & \ms{0.072}{0.039} & \ms{0.323}{0.126} \\
Pending anemia; 15\,min & \ms{0.573}{0.047} & \ms{0.566}{0.016} & \ms{0.070}{0.003} & \ms{0.046}{0.022} & \ms{0.133}{0.063} \\
Decomp. 60m; 30\,min & \ms{0.664}{0.019} & \ms{0.626}{0.017} & \ms{0.157}{0.010} & \ms{0.073}{0.031} & \ms{0.229}{0.088} \\
Lactate NC; 30\,min & \ms{0.567}{0.022} & \ms{0.499}{0.020} & \ms{0.367}{0.015} & \ms{0.258}{0.028} & \ms{0.423}{0.067} \\
Lactate NC; 60\,min & \ms{0.534}{0.023} & \ms{0.489}{0.016} & \ms{0.385}{0.011} & \ms{0.153}{0.034} & \ms{0.419}{0.046} \\
Pending hyperK; 60\,min & \ms{0.555}{0.023} & \ms{0.532}{0.017} & \ms{0.077}{0.013} & \ms{0.081}{0.068} & \ms{0.257}{0.173} \\
Pending anemia; 60\,min & \ms{0.544}{0.053} & \ms{0.544}{0.057} & \ms{0.200}{0.031} & \ms{0.078}{0.018} & \ms{0.200}{0.086} \\
Lactate NC; 90\,min & \ms{0.577}{0.039} & \ms{0.532}{0.014} & \ms{0.399}{0.017} & \ms{0.247}{0.037} & \ms{0.390}{0.047} \\
\bottomrule
\end{tabular}
\end{table}

\begin{table}[!t]
\centering
\scriptsize
\setlength{\tabcolsep}{3.2pt}
\renewcommand{\arraystretch}{1.10}
\caption{Forecast-gap results --- MC-MED. Training-cost and failure measurements from the updated export; units are retained per cell.}
\label{tab:export-04-mcmed-cost}
\begin{tabular}{@{}>{\raggedright\arraybackslash}p{5cm}ccc@{}}
\toprule
Configuration & Time (s) & Memory & Failures (\%) \\
\midrule
Decomp. 60m; 0\,min & 2.9 & 39238\,MiB & 0.0 \\
Pending hyperK; 0\,min & 15.9 & 46850\,MiB & 0.0 \\
Pending anemia; 0\,min & 13.7 & 28876\,MiB & 0.0 \\
Decomp. 60m; 10\,min & 5.5 & 35301\,MiB & 0.0 \\
Pending hyperK; 15\,min & 7.5 & 46461\,MiB & 0.0 \\
Pending anemia; 15\,min & 5.5 & 21562\,MiB & 0.0 \\
Decomp. 60m; 30\,min & 3.8 & 28188\,MiB & 0.0 \\
Lactate NC; 30\,min & 2.0 & 9741\,MiB & 0.0 \\
Lactate NC; 60\,min & 1.5 & 9606\,MiB & 0.0 \\
Pending hyperK; 60\,min & 6.4 & 21782\,MiB & 0.0 \\
Pending anemia; 60\,min & 0.6 & 8324\,MiB & 0.0 \\
Lactate NC; 90\,min & 1.6 & 8381\,MiB & 0.0 \\
\bottomrule
\end{tabular}
\end{table}
\begin{table}[!t]
\centering
\scriptsize
\setlength{\tabcolsep}{3.2pt}
\renewcommand{\arraystretch}{1.10}
\caption{Temporal-order control --- MC-MED. Updated export; values retain the supplied mean and SD precision.}
\label{tab:export-06b-mcmed}
\begin{tabular}{@{}>{\raggedright\arraybackslash}p{3.4cm}ccc@{}}
\toprule
Configuration & Val. AUROC & Test AUROC & Test AUPRC \\
\midrule
10-s block shuffle & \ms{0.576}{0.011} & \ms{0.563}{0.009} & \ms{0.179}{0.004} \\
\bottomrule
\end{tabular}
\end{table}

\begin{table}[!t]
\centering
\scriptsize
\setlength{\tabcolsep}{3.2pt}
\renewcommand{\arraystretch}{1.10}
\caption{Observation length --- MIMIC-III. Task-level test AUROC (mean $\pm$ seed SD) underlying the AUROC figures. Dashes mark combinations absent from that plotted grid.}
\label{tab:tasks-observation-length-mimic}
\begin{tabular}{@{}>{\raggedright\arraybackslash}p{4.0cm}cccc@{}}
\toprule
\multicolumn{5}{@{}l}{\textit{Acute det. / AKI 2/3 / Circ. det. / Fluid resp.}}\\*
\midrule
Condition / budget & Acute det. & AKI 2/3 & Circ. det. & Fluid resp. \\
\midrule
10 min & \ms{0.646}{0.009} & \ms{0.678}{0.009} & \ms{0.657}{0.007} & \ms{0.565}{0.010} \\
20 min & \ms{0.654}{0.015} & \ms{0.671}{0.007} & \ms{0.656}{0.004} & \ms{0.560}{0.008} \\
60 min & \ms{0.656}{0.009} & \ms{0.680}{0.018} & \ms{0.662}{0.004} & \ms{0.567}{0.004} \\
\addlinespace
\multicolumn{5}{@{}l}{\textit{Dyn. mort. / Resp. esc. / Sepsis-3 / Vent. lib.}}\\*
\midrule
Condition / budget & Dyn. mort. & Resp. esc. & Sepsis-3 & Vent. lib. \\
\midrule
10 min & \ms{0.517}{0.006} & \ms{0.575}{0.034} & \ms{0.659}{0.010} & \ms{0.550}{0.006} \\
20 min & \ms{0.516}{0.006} & \ms{0.585}{0.022} & \ms{0.659}{0.004} & \ms{0.547}{0.007} \\
60 min & \ms{0.517}{0.012} & \ms{0.593}{0.019} & \ms{0.652}{0.005} & \ms{0.548}{0.004} \\
\addlinespace
\bottomrule
\end{tabular}
\end{table}
\begin{table}[!t]
\centering
\scriptsize
\setlength{\tabcolsep}{3.2pt}
\renewcommand{\arraystretch}{1.10}
\caption{Observation length --- VitalDB. Task-level test AUROC (mean $\pm$ seed SD) underlying the AUROC figures. Dashes mark combinations absent from that plotted grid.}
\label{tab:tasks-observation-length-vitaldb}
\begin{tabular}{@{}>{\raggedright\arraybackslash}p{4.0cm}cccc@{}}
\toprule
\multicolumn{5}{@{}l}{\textit{CL07-002 / H5-002 / H5-030 / L2-30}}\\*
\midrule
Condition / budget & CL07-002 & H5-002 & H5-030 & L2-30 \\
\midrule
5 min & \ms{0.537}{0.001} & \ms{0.635}{0.003} & \ms{0.602}{0.001} & \ms{0.560}{0.004} \\
15 min & \ms{0.533}{0.002} & \ms{0.635}{0.002} & \ms{0.602}{0.002} & \ms{0.576}{0.004} \\
30 min & \ms{0.532}{0.002} & \ms{0.626}{0.002} & \ms{0.602}{0.001} & \ms{0.579}{0.002} \\
60 min & \ms{0.536}{0.003} & \ms{0.625}{0.002} & \ms{0.601}{0.002} & \ms{0.594}{0.002} \\
\addlinespace
\multicolumn{5}{@{}l}{\textit{OX01-IOH / VA1-CO2 / VA1-STIM}}\\*
\midrule
Condition / budget & OX01-IOH & VA1-CO2 & VA1-STIM &  \\
\midrule
5 min & \ms{0.689}{0.003} & \ms{0.676}{0.002} & \ms{0.687}{0.002} &  \\
15 min & \ms{0.712}{0.002} & \ms{0.674}{0.003} & \ms{0.686}{0.003} &  \\
30 min & \ms{0.729}{0.001} & \ms{0.677}{0.001} & \ms{0.696}{0.002} &  \\
60 min & \ms{0.727}{0.004} & \ms{0.678}{0.001} & \ms{0.692}{0.003} &  \\
\addlinespace
\bottomrule
\end{tabular}
\end{table}
\begin{table}[!t]
\centering
\scriptsize
\setlength{\tabcolsep}{3.2pt}
\renewcommand{\arraystretch}{1.10}
\caption{Observation length --- MC-MED. Task-level test AUROC (mean $\pm$ seed SD) underlying the AUROC figures. Dashes mark combinations absent from that plotted grid.}
\label{tab:tasks-observation-length-mcmed}
\begin{tabular}{@{}>{\raggedright\arraybackslash}p{4.0cm}cccc@{}}
\toprule
\multicolumn{5}{@{}l}{\textit{Decomp. 60m / Lactate NC / Pending hyperK / Pending anemia}}\\*
\midrule
Condition / budget & Decomp. 60m & Lactate NC & Pending hyperK & Pending anemia \\
\midrule
5 min & \ms{0.647}{0.003} & \ms{0.499}{0.020} & \ms{0.566}{0.035} & \ms{0.556}{0.014} \\
10 min & \ms{0.656}{0.009} & \ms{0.502}{0.013} & \ms{0.576}{0.027} & \ms{0.577}{0.022} \\
15 min & \ms{0.631}{0.014} & \ms{0.577}{0.022} & \ms{0.528}{0.057} & \ms{0.530}{0.050} \\
\addlinespace
\bottomrule
\end{tabular}
\end{table}
\begin{table}[!t]
\centering
\scriptsize
\setlength{\tabcolsep}{3.2pt}
\renewcommand{\arraystretch}{1.10}
\caption{Forecast gap --- MIMIC-III. Task-level test AUROC (mean $\pm$ seed SD) underlying the AUROC figures. Dashes mark combinations absent from that plotted grid.}
\label{tab:tasks-forecast-gap-mimic}
\begin{tabular}{@{}>{\raggedright\arraybackslash}p{4.0cm}cccc@{}}
\toprule
\multicolumn{5}{@{}l}{\textit{AKI 2/3 / Dyn. mort. / Sepsis-3}}\\*
\midrule
Condition / budget & AKI 2/3 & Dyn. mort. & Sepsis-3 &  \\
\midrule
2 h & -- & -- & \ms{0.629}{0.007} &  \\
3 h & \ms{0.686}{0.017} & \ms{0.545}{0.011} & -- &  \\
4 h & -- & -- & \ms{0.619}{0.010} &  \\
6 h & \ms{0.697}{0.028} & -- & \ms{0.629}{0.004} &  \\
12 h & \ms{0.685}{0.009} & \ms{0.539}{0.011} & -- &  \\
24 h & -- & \ms{0.537}{0.009} & -- &  \\
\addlinespace
\bottomrule
\end{tabular}
\end{table}
\begin{table}[!t]
\centering
\scriptsize
\setlength{\tabcolsep}{3.2pt}
\renewcommand{\arraystretch}{1.10}
\caption{Forecast gap --- VitalDB. Task-level test AUROC (mean $\pm$ seed SD) underlying the AUROC figures. Dashes mark combinations absent from that plotted grid.}
\label{tab:tasks-forecast-gap-vitaldb}
\begin{tabular}{@{}>{\raggedright\arraybackslash}p{4.0cm}cccc@{}}
\toprule
\multicolumn{5}{@{}l}{\textit{CL07-002 / H5-002 / H5-030 / L2-30}}\\*
\midrule
Condition / budget & CL07-002 & H5-002 & H5-030 & L2-30 \\
\midrule
5 min & \ms{0.552}{0.012} & \ms{0.651}{0.002} & \ms{0.592}{0.007} & \ms{0.582}{0.010} \\
15 min & \ms{0.552}{0.016} & \ms{0.626}{0.012} & \ms{0.607}{0.005} & \ms{0.577}{0.009} \\
30 min & \ms{0.548}{0.009} & \ms{0.650}{0.010} & \ms{0.615}{0.006} & \ms{0.622}{0.002} \\
45 min & \ms{0.559}{0.016} & \ms{0.670}{0.010} & \ms{0.606}{0.010} & \ms{0.607}{0.007} \\
60 min & \ms{0.535}{0.006} & \ms{0.652}{0.010} & \ms{0.631}{0.009} & \ms{0.589}{0.006} \\
75 min & \ms{0.567}{0.014} & \ms{0.650}{0.012} & \ms{0.598}{0.007} & \ms{0.555}{0.012} \\
\addlinespace
\multicolumn{5}{@{}l}{\textit{OX01-IOH / VA1-CO2 / VA1-STIM}}\\*
\midrule
Condition / budget & OX01-IOH & VA1-CO2 & VA1-STIM &  \\
\midrule
5 min & \ms{0.791}{0.001} & \ms{0.689}{0.005} & \ms{0.775}{0.006} &  \\
15 min & \ms{0.682}{0.003} & \ms{0.664}{0.007} & \ms{0.641}{0.013} &  \\
30 min & \ms{0.660}{0.008} & \ms{0.640}{0.011} & \ms{0.630}{0.008} &  \\
45 min & \ms{0.653}{0.006} & \ms{0.633}{0.002} & \ms{0.603}{0.009} &  \\
60 min & \ms{0.611}{0.008} & \ms{0.609}{0.014} & \ms{0.743}{0.014} &  \\
75 min & \ms{0.650}{0.010} & \ms{0.655}{0.018} & \ms{0.750}{0.028} &  \\
\addlinespace
\bottomrule
\end{tabular}
\end{table}
\begin{table}[!t]
\centering
\scriptsize
\setlength{\tabcolsep}{3.2pt}
\renewcommand{\arraystretch}{1.10}
\caption{Forecast gap --- MC-MED. Task-level test AUROC (mean $\pm$ seed SD) underlying the AUROC figures. Dashes mark combinations absent from that plotted grid.}
\label{tab:tasks-forecast-gap-mcmed}
\begin{tabular}{@{}>{\raggedright\arraybackslash}p{4.0cm}cccc@{}}
\toprule
\multicolumn{5}{@{}l}{\textit{Decomp. 60m / Lactate NC / Pending hyperK / Pending anemia}}\\*
\midrule
Condition / budget & Decomp. 60m & Lactate NC & Pending hyperK & Pending anemia \\
\midrule
0 min & \ms{0.647}{0.003} & -- & \ms{0.566}{0.035} & \ms{0.556}{0.014} \\
10 min & \ms{0.629}{0.016} & -- & -- & -- \\
15 min & -- & -- & \ms{0.567}{0.031} & \ms{0.566}{0.016} \\
30 min & \ms{0.626}{0.017} & \ms{0.499}{0.020} & -- & -- \\
60 min & -- & \ms{0.489}{0.016} & \ms{0.532}{0.017} & \ms{0.544}{0.057} \\
90 min & -- & \ms{0.532}{0.014} & -- & -- \\
\addlinespace
\bottomrule
\end{tabular}
\end{table}

\input{sections/tables/export_05_subject_selection}
\begin{table}[!t]
\centering
\scriptsize
\setlength{\tabcolsep}{3.2pt}
\renewcommand{\arraystretch}{1.10}
\caption{Channel-budget summaries --- MIMIC-III. Updated export; values retain the supplied mean and SD precision.}
\label{tab:export-06-mimic}
\begin{tabular}{@{}>{\raggedright\arraybackslash}p{3.4cm}ccccc@{}}
\toprule
Configuration & Val. AUROC & Test AUROC & Test AUPRC & ECE & AURC \\
\midrule
All channels (62) & \ms{0.635}{0.001} & \ms{0.606}{0.002} & \ms{0.107}{0.001} & -- & -- \\
Lasso (10) & \ms{0.625}{0.006} & \ms{0.599}{0.004} & \ms{0.105}{0.001} & \ms{0.204}{0.029} & \ms{0.347}{0.032} \\
Mutual information (10) & \ms{0.598}{0.007} & \ms{0.573}{0.005} & \ms{0.098}{0.002} & \ms{0.283}{0.032} & \ms{0.357}{0.077} \\
Random (10) & \ms{0.579}{0.008} & \ms{0.558}{0.008} & \ms{0.097}{0.001} & \ms{0.334}{0.033} & \ms{0.451}{0.037} \\
Clinical (10) & \ms{0.633}{0.002} & \ms{0.606}{0.004} & \ms{0.104}{0.002} & \ms{0.202}{0.021} & \ms{0.343}{0.031} \\
Waveform SQI (10) & \ms{0.635}{0.004} & \ms{0.607}{0.003} & \ms{0.106}{0.002} & \ms{0.197}{0.009} & \ms{0.359}{0.031} \\
Temporal MI (10) & \ms{0.595}{0.010} & \ms{0.570}{0.008} & \ms{0.099}{0.003} & \ms{0.285}{0.033} & \ms{0.444}{0.059} \\
\bottomrule
\end{tabular}
\end{table}

\begin{table}[!t]
\centering
\scriptsize
\setlength{\tabcolsep}{3.2pt}
\renewcommand{\arraystretch}{1.10}
\caption{Channel-budget summaries --- MIMIC-III. Training-cost and failure measurements from the updated export; units are retained per cell.}
\label{tab:export-06-mimic-cost}
\begin{tabular}{@{}>{\raggedright\arraybackslash}p{5cm}cccc@{}}
\toprule
Configuration & Time (s) & Memory & ms/sample & Failures (\%) \\
\midrule
All channels (62) & 108.6 & 3.4\,\% & 0.24 & 0.0 \\
Lasso (10) & 96.4 & 2.4\,\% & 0.16 & 0.0 \\
Mutual information (10) & 97.2 & 2.4\,\% & 0.14 & 0.0 \\
Random (10) & 103.2 & 2.4\,\% & 0.15 & 0.0 \\
Clinical (10) & 100.1 & 2.4\,\% & 0.14 & 0.0 \\
Waveform SQI (10) & 102.1 & 2.4\,\% & 0.14 & 0.0 \\
Temporal MI (10) & 92.8 & 2.4\,\% & 0.14 & 0.0 \\
\bottomrule
\end{tabular}
\end{table}

\begin{table}[!t]
\centering
\scriptsize
\setlength{\tabcolsep}{3.2pt}
\renewcommand{\arraystretch}{1.10}
\caption{Channel-budget summaries --- VitalDB. Updated export; values retain the supplied mean and SD precision.}
\label{tab:export-06-vitaldb}
\begin{tabular}{@{}>{\raggedright\arraybackslash}p{3.4cm}ccccc@{}}
\toprule
Configuration & Val. AUROC & Test AUROC & Test AUPRC & ECE & AURC \\
\midrule
All channels (12) & \ms{0.617}{0.001} & \ms{0.612}{0.001} & \ms{0.072}{0.001} & \ms{0.039}{0.001} & \ms{0.032}{0.001} \\
Lasso (2) & \ms{0.571}{0.005} & \ms{0.565}{0.006} & \ms{0.058}{0.001} & \ms{0.044}{0.002} & \ms{0.034}{0.000} \\
Lasso (4) & \ms{0.591}{0.002} & \ms{0.587}{0.005} & \ms{0.066}{0.001} & \ms{0.045}{0.001} & \ms{0.033}{0.001} \\
Lasso (8) & \ms{0.610}{0.002} & \ms{0.605}{0.003} & \ms{0.071}{0.001} & \ms{0.042}{0.001} & \ms{0.032}{0.001} \\
Mutual information (2) & \ms{0.569}{0.006} & \ms{0.563}{0.006} & \ms{0.058}{0.002} & \ms{0.044}{0.001} & \ms{0.035}{0.001} \\
Mutual information (4) & \ms{0.591}{0.004} & \ms{0.584}{0.003} & \ms{0.066}{0.002} & \ms{0.044}{0.001} & \ms{0.033}{0.001} \\
Mutual information (8) & \ms{0.611}{0.001} & \ms{0.604}{0.002} & \ms{0.070}{0.001} & \ms{0.041}{0.003} & \ms{0.032}{0.001} \\
\bottomrule
\end{tabular}
\end{table}

\begin{table}[!t]
\centering
\scriptsize
\setlength{\tabcolsep}{3.2pt}
\renewcommand{\arraystretch}{1.10}
\caption{Channel-budget summaries --- VitalDB. Training-cost and failure measurements from the updated export; units are retained per cell.}
\label{tab:export-06-vitaldb-cost}
\begin{tabular}{@{}>{\raggedright\arraybackslash}p{5cm}ccc@{}}
\toprule
Configuration & Time (s) & ms/sample & Failures (\%) \\
\midrule
All channels (12) & -- & -- & -- \\
Lasso (2) & 33.3 & 0.43 & 0.0 \\
Lasso (4) & 33.4 & 0.43 & 0.0 \\
Lasso (8) & 33.6 & 0.44 & 0.0 \\
Mutual information (2) & 33.5 & 0.43 & 0.0 \\
Mutual information (4) & 33.6 & 0.44 & 0.0 \\
Mutual information (8) & 33.5 & 0.43 & 0.0 \\
Random (2) & 33.3 & 0.43 & 0.0 \\
Random (4) & 33.8 & 0.44 & 0.0 \\
Random (8) & 33.2 & 0.43 & 0.0 \\
Waveform SQI (2) & 33.4 & 0.43 & 0.0 \\
Waveform SQI (4) & 33.7 & 0.44 & 0.0 \\
Waveform SQI (8) & 33.4 & 0.43 & 0.0 \\
Temporal MI (2) & 33.6 & 0.44 & 0.0 \\
Temporal MI (4) & 33.5 & 0.43 & 0.0 \\
Temporal MI (8) & 33.3 & 0.43 & 0.0 \\
\bottomrule
\end{tabular}
\end{table}
\input{sections/tables/export_07_missingness}
\input{sections/tables/export_08_ehr_fusion_main}
\input{sections/tables/export_09_ehr_fusion_expanded_mimic}
\begin{table}[!t]
\centering
\scriptsize
\setlength{\tabcolsep}{3.2pt}
\renewcommand{\arraystretch}{1.10}
\caption{Channel selection --- MIMIC-III. Task-level test AUROC (mean $\pm$ seed SD) underlying the AUROC figures. Dashes mark combinations absent from that plotted grid.}
\label{tab:tasks-channel-selection-mimic}
\begin{tabular}{@{}>{\raggedright\arraybackslash}p{4.0cm}cccc@{}}
\toprule
\multicolumn{5}{@{}l}{\textit{Acute det. / AKI 2/3 / Circ. det. / Fluid resp.}}\\*
\midrule
Condition / budget & Acute det. & AKI 2/3 & Circ. det. & Fluid resp. \\
\midrule
All channels (62) & \ms{0.584}{0.003} & \ms{0.712}{0.012} & \ms{0.678}{0.008} & \ms{0.570}{0.009} \\
Lasso (10) & \ms{0.590}{0.011} & \ms{0.706}{0.019} & \ms{0.667}{0.018} & \ms{0.559}{0.010} \\
MI (10) & \ms{0.571}{0.007} & \ms{0.715}{0.019} & \ms{0.644}{0.013} & \ms{0.556}{0.015} \\
Random (10) & \ms{0.571}{0.013} & \ms{0.720}{0.016} & \ms{0.544}{0.055} & \ms{0.566}{0.014} \\
Clinical (10) & \ms{0.603}{0.006} & \ms{0.721}{0.009} & \ms{0.683}{0.004} & \ms{0.553}{0.012} \\
SQI (10) & \ms{0.600}{0.003} & \ms{0.723}{0.013} & \ms{0.666}{0.006} & \ms{0.558}{0.009} \\
Temporal MI (10) & \ms{0.570}{0.014} & \ms{0.715}{0.018} & \ms{0.618}{0.061} & \ms{0.560}{0.007} \\
\addlinespace
\multicolumn{5}{@{}l}{\textit{Dyn. mort. / Resp. esc. / Sepsis-3 / Vent. lib.}}\\*
\midrule
Condition / budget & Dyn. mort. & Resp. esc. & Sepsis-3 & Vent. lib. \\
\midrule
All channels (62) & \ms{0.556}{0.008} & \ms{0.568}{0.012} & \ms{0.631}{0.003} & \ms{0.545}{0.003} \\
Lasso (10) & \ms{0.561}{0.011} & \ms{0.533}{0.025} & \ms{0.615}{0.017} & \ms{0.559}{0.018} \\
MI (10) & \ms{0.513}{0.012} & \ms{0.512}{0.017} & \ms{0.555}{0.029} & \ms{0.514}{0.013} \\
Random (10) & \ms{0.507}{0.028} & \ms{0.511}{0.017} & \ms{0.536}{0.024} & \ms{0.509}{0.007} \\
Clinical (10) & \ms{0.558}{0.011} & \ms{0.577}{0.006} & \ms{0.623}{0.005} & \ms{0.531}{0.010} \\
SQI (10) & \ms{0.554}{0.008} & \ms{0.589}{0.006} & \ms{0.629}{0.002} & \ms{0.539}{0.006} \\
Temporal MI (10) & \ms{0.514}{0.027} & \ms{0.526}{0.026} & \ms{0.548}{0.033} & \ms{0.511}{0.018} \\
\addlinespace
\bottomrule
\end{tabular}
\end{table}
\input{sections/tables/tasks_channel_selection_vitaldb}
\input{sections/tables/tasks_missingness_mimic}
\input{sections/tables/tasks_missingness_vitaldb}
\input{sections/tables/tasks_ehr_fusion_mimic}
\input{sections/tables/tasks_ehr_fusion_vitaldb}
\input{sections/tables/tasks_ehr_fusion_mcmed}
\input{sections/tables/tasks_training_subjects_mimic}
\input{sections/tables/tasks_training_subjects_vitaldb}
\input{sections/tables/tasks_training_subjects_mcmed}

\input{sections/tables/export_10_transfer_mimiciv}
\input{sections/tables/transfer_auroc}
\input{sections/tables/transfer_auprc}

\input{sections/tables/ees_architectures_mimic}
\input{sections/tables/ees_architectures_vitaldb}
\input{sections/tables/ees_architectures_mcmed}
\input{sections/tables/ees_tokenizers_mimic}
\input{sections/tables/ees_tokenizers_vitaldb}
\input{sections/tables/ees_tokenizers_mcmed}
\input{sections/tables/ees_observation_length_mimic}
\input{sections/tables/ees_observation_length_vitaldb}
\input{sections/tables/ees_observation_length_mcmed}
\input{sections/tables/ees_forecast_gap_mimic}
\input{sections/tables/ees_forecast_gap_vitaldb}
\input{sections/tables/ees_forecast_gap_mcmed}
\input{sections/tables/ees_channel_selection_mimic}
\input{sections/tables/ees_channel_selection_vitaldb}
\input{sections/tables/ees_missingness_mimic}
\input{sections/tables/ees_missingness_vitaldb}
\input{sections/tables/ees_training_subjects_mimic}
\input{sections/tables/ees_training_subjects_vitaldb}
\input{sections/tables/ees_training_subjects_mcmed}
\input{sections/tables/ees_ehr_fusion_mimic}
\input{sections/tables/ees_ehr_fusion_vitaldb}
\input{sections/tables/ees_ehr_fusion_mcmed}

\input{sections/tables/legacy_vitaldb_preprocessing}
\input{sections/tables/legacy_vitaldb_gap}
\input{sections/tables/legacy_vitaldb_fusion}

\end{document}